\pdfoutput=1
\documentclass{article}

\PassOptionsToPackage{numbers, compress}{natbib}

\def\status{final} 
\usepackage[\status]{neurips_2024}

\usepackage[\status]{neurips_2024}
\usepackage{titletoc}

\usepackage[utf8]{inputenc} 
\usepackage[T1]{fontenc}    
\usepackage{hyperref}       
\usepackage{url}            
\usepackage{booktabs}       
\usepackage{amsfonts}       
\usepackage{nicefrac}       
\usepackage{microtype}      
\usepackage{xcolor}         

\usepackage{booktabs}       
\usepackage{amsfonts}       
\usepackage{nicefrac}       
\usepackage{microtype}      

\makeatletter%
\def\input@path{{latex-utilities}}%
\makeatother%
\usepackage{amsmath}%
\usepackage{amssymb}%
\usepackage{bm}%
\usepackage{mathtools}%

\makeatletter
\def\assertPackageLoaded#1{
  \@ifpackageloaded{#1}{}{\PackageError{fdtexutils}{Please load the #1 package}{}}
}
\makeatother
\assertPackageLoaded{amsmath}
\assertPackageLoaded{amssymb}
\assertPackageLoaded{bm}
\assertPackageLoaded{mathtools}

\def\vzero{{\bm{0}}}
\def\vone{{\bm{1}}}

\def\vdelta{{\bm{\delta}}}

\def\va{{\bm{a}}}
\def\vb{{\bm{b}}}

\def\vd{{\bm{d}}}

\def\vf{{\bm{f}}}
\def\vg{{\bm{g}}}

\def\vt{{\bm{t}}}

\def\vw{{\bm{w}}}
\def\vx{{\bm{x}}}
\def\vy{{\bm{y}}}

\def\evb{{b}}

\def\evy{{y}}

\def\mA{{\bm{A}}}
\def\mB{{\bm{B}}}
\def\mC{{\bm{C}}}

\def\mG{{\bm{G}}}

\def\mI{{\bm{I}}}
\def\mJ{{\bm{J}}}
\def\mK{{\bm{K}}}

\def\mS{{\bm{S}}}

\def\mW{{\bm{W}}}
\def\mX{{\bm{X}}}
\def\mY{{\bm{Y}}}

\def\mGamma{{\bm{\Gamma}}}

\def\mOmega{{\bm{\Omega}}}

\DeclareMathAlphabet{\mathsfit}{\encodingdefault}{\sfdefault}{m}{sl}
\SetMathAlphabet{\mathsfit}{bold}{\encodingdefault}{\sfdefault}{bx}{n}
\newcommand{\tens}[1]{\bm{\mathsfit{#1}}}
\def\tA{{\tens{A}}}
\def\tB{{\tens{B}}}
\def\tC{{\tens{C}}}
\def\tD{{\tens{D}}}

\def\tG{{\tens{G}}}

\def\tJ{{\tens{J}}}

\def\tP{{\tens{P}}}
\def\tPi{\bm{\mathsf{\Pi}}}

\def\tS{{\tens{S}}}

\def\tV{{\tens{V}}}
\def\tW{{\tens{W}}}
\def\tX{{\tens{X}}}
\def\tY{{\tens{Y}}}

\def\gX{{\mathcal{X}}}
\def\gY{{\mathcal{Y}}}

\def\sN{{\mathbb{N}}}

\def\sR{{\mathbb{R}}}

\DeclareSymbolFont{bbold}{U}{bbold}{m}{n}
\DeclareSymbolFontAlphabet{\mathbbold}{bbold}

\def\emA{{A}}

\newcommand{\etens}[1]{\mathsfit{#1}}

\def\etP{{\etens{P}}}
\def\etPi{\mathsf{\Pi}}

\def\etV{{\etens{V}}}
\def\etW{{\etens{W}}}
\def\etX{{\etens{X}}}
\def\etY{{\etens{Y}}}

\DeclareMathOperator{\diag}{diag}

\DeclarePairedDelimiterX{\KLdivx}[2]{(}{)}{%
  #1\;\delimsize\|\;#2%
}

\usepackage{stmaryrd} 

\newtheorem{transformation}{Transformation} 

\makeatletter
\def\assertPackageLoaded#1{
  \@ifpackageloaded{#1}{}{\PackageError{fdtexutils}{Please load the #1 package}{}}
}
\makeatother
\assertPackageLoaded{xcolor}

\definecolor{VectorBlack}{RGB}{34, 34, 34}
\definecolor{VectorGray}{RGB}{239, 238, 237}

\definecolor{VectorBlue}{RGB}{59, 69, 227}
\definecolor{VectorPink}{RGB}{253, 8, 238}
\definecolor{VectorOrange}{RGB}{250, 173, 26}
\definecolor{VectorTeal}{RGB}{82, 199, 222}

\colorlet{maincolor}{VectorBlue}
\colorlet{secondcolor}{VectorPink}
\colorlet{thirdcolor}{VectorOrange}
\hypersetup{%
  colorlinks,
  citecolor = maincolor,%
  linkcolor = maincolor,%
  urlcolor = secondcolor,%
}%
\usepackage{url}%
\usepackage{cleveref} 
\crefname{section}{\S\!\!}{\S\!\!}
\crefname{appendix}{\S\!\!}{\S\!\!}
\crefname{transformation}{transformation}{transformations} 

\usepackage{blindtext}%
\usepackage{enumitem} 
\usepackage{graphicx}%
\usepackage{subcaption}%
\usepackage{csvsimple} 
\usepackage{siunitx} 
\usepackage{wrapfig} 
\usepackage{multirow}%
\usepackage{tabularx} 
\usepackage{pdflscape} 
\usepackage{tikz} 
\usetikzlibrary{calc}
\usetikzlibrary{arrows.meta} 
\usetikzlibrary{backgrounds,scopes} 
\usepackage{pgfplots}

\pgfkeys{/pgfplots/mybasic/.style={
    every axis plot/.append style={line width = 1.5pt},
    tick pos = left,
    ylabel near ticks,
    xlabel near ticks,
    xtick align = outside,
    ytick align = outside,
    legend cell align = left,
    legend columns = 1,
    legend pos = north east,
    legend style = {
      fill opacity = 0.7,
      text opacity = 1,
      font = \scriptsize,
    },
    xticklabel style = {font = \scriptsize},
    xlabel style = {font = \scriptsize},
    axis line style = {black},
    yticklabel style = {font = \scriptsize},
    ylabel style = {font = \scriptsize},
    title style = {font = \scriptsize},
    grid = major,
  }
}

\pgfkeys{/pgfplots/boxplotbasicstyle/.style={
    mybasic,
    grid=none,
    xmajorgrids,
    width = \linewidth,
    height = 0.6\linewidth,
    every axis plot post/.append style={
      mark size=1.5, mark options={solid, line width = 1pt}
    },
    execute at begin picture={\colorlet{color0}{maincolor}},
    execute at begin picture={\colorlet{color1}{secondcolor}},
    execute at begin picture={\colorlet{color2}{maincolor!30!white}},
    execute at begin picture={\colorlet{color3}{secondcolor!30!white}},
  }
}
\pgfkeys{/pgfplots/tacoboxplotbasicstyle/.style={
    boxplotbasicstyle,
    execute at begin picture={\colorlet{color0}{thirdcolor}},
    execute at begin picture={\colorlet{color1}{thirdcolor!30!white}},
  }
}

\usepackage{algorithm}
\usepackage{algpseudocode}

\usepackage{listings}
\usepackage{xpatch}

\makeatletter
\def\assertPackageLoaded#1{
  \@ifpackageloaded{#1}{}{\PackageError{fdtexutils}{Please load the #1 package}{}}
}
\makeatother
\assertPackageLoaded{listings}
\assertPackageLoaded{xpatch}

\makeatletter
\let\old@lstKV@SwitchCases\lstKV@SwitchCases
\def\lstKV@SwitchCases#1#2#3{}
\makeatother
\usepackage{lstlinebgrd}
\makeatletter
\let\lstKV@SwitchCases\old@lstKV@SwitchCases

\lst@Key{numbers}{none}{%
	\def\lst@PlaceNumber{\lst@linebgrd}%
	\lstKV@SwitchCases{#1}%
	{none:\\%
		left:\def\lst@PlaceNumber{\llap{\normalfont
				\lst@numberstyle{\thelstnumber}\kern\lst@numbersep}\lst@linebgrd}\\%
		right:\def\lst@PlaceNumber{\rlap{\normalfont
				\kern\linewidth \kern\lst@numbersep
				\lst@numberstyle{\thelstnumber}}\lst@linebgrd}%
	}{\PackageError{Listings}{Numbers #1 unknown}\@ehc}}
\makeatother

\makeatletter
\xpatchcmd\lst@linebgrd{\color{-.}}{\lst@bkgcolor}{}{\fail}
\makeatother

\lstdefinestyle{vector_institute}{
  backgroundcolor=\color{VectorGray!50},
  commentstyle=\bfseries\color{VectorBlue},
  keywordstyle=\bfseries\color{VectorBlack},
  numberstyle=\tiny\color{VectorBlack!50},
  stringstyle=\bfseries\color{VectorPink},
  basicstyle=\ttfamily\tiny,
  breakatwhitespace=false,
  breaklines=true,
  captionpos=t,
  keepspaces=true,
  showspaces=false,
  showstringspaces=false,
  showtabs=false,
  tabsize=2,
}
\newcommand{\papertitle}{%
  Convolutions and More as Einsum: A Tensor Network Perspective with Advances for Second-Order Methods%
}%
\title{\papertitle}

\author{%
  Felix Dangel \\
  Vector Institute \\
  Toronto, Canada\\
  \texttt{fdangel@vectorinstitute.ai} \\
}

\begin{document}

\maketitle

\begin{abstract}
  Despite their simple intuition, convolutions are more tedious to analyze than dense layers, which complicates the transfer of theoretical and algorithmic ideas to convolutions.
  We simplify convolutions by viewing them as tensor networks (TNs) that allow reasoning about the underlying tensor multiplications by drawing diagrams, manipulating them to perform function transformations like differentiation, and efficiently evaluating them with \texttt{einsum}.
  To demonstrate their simplicity and expressiveness, we derive diagrams of various autodiff operations and popular curvature approximations with full hyper-parameter support, batching, channel groups, and generalization to any convolution dimension.
  Further, we provide convolution-specific transformations based on the connectivity pattern which allow to simplify diagrams before evaluation.
  Finally, we probe performance.
  Our TN implementation accelerates a recently-proposed KFAC variant up to 4.5\,x while removing the standard implementation's memory overhead, and enables new hardware-efficient tensor dropout for approximate backpropagation.
\end{abstract}

\section{Introduction}\label{sec:introduction}
Convolutional neural networks~\citep[CNNs,][]{lecun1989backpropagation} mark a milestone in the development of deep learning architectures as their `sliding window' approach represents an important inductive bias for vision tasks.
Their intuition is simple to explain with graphical illustrations \citep[e.g.\,][]{dumoulin2016guide}.
Yet, convolutions are more challenging to analyze than dense layers in multi-layer perceptrons (MLPs) or transformers~\citep{vaswani2017attention}.
One reason is that they are hard to express in matrix notation and---even in index notation---compact expressions that are convenient to work with only exist for special hyper-parameters~\citep[e.g.][]{grosse2016kroneckerfactored,arora2019exact}.
Many hyper-parameters (stride, padding, \dots) and additional features like channel groups~\citep{krizhevsky2012imagenet} introduce even more complexity that is inherited by related routines, e.g.\,for autodiff.
We observe a delay of analytic and algorithmic developments between MLPs vs.\,CNNs, e.g.
\begin{itemize}[itemindent=-20pt]
\item Approximate Hessian diagonal: \citeyear{becker1989improving} vs.\,\citeyear{elsayed2024revisiting}

\item Hessian rank: \citeyear{singh2021analytic} vs.\,\citeyear{singh2023hessian}

\item Gradient descent learning dynamics: \citeyear{saxe2014exact} vs.\,\citeyear{pinson2023linear}

\item Neural tangent kernel (NTK): \defcitealias{jacot2020neural}{2018}\citetalias{jacot2020neural} vs.\,\citeyear{arora2019exact}

\item Kronecker-factored quasi-Newton methods: \citeyear{goldfarb2021practical} vs.\,\citeyear{ren2022kroneckerfactored}

\item Kronecker-factored curvature (KFAC, KFRA, KFLR): (\citeyear{martens2015optimizing}, \citeyear{botev2017practical}, \citeyear{botev2017practical}) vs.\,(\citeyear{grosse2016kroneckerfactored}, \citeyear{dangel2020backpack}, \citeyear{dangel2020backpack})

\end{itemize}
The software support for less standard routines some of these methods require also reflects this gap.
Some functions only support special dimensions~\cite{dangel2021unfoldnd}.
Others use less efficient workarounds (\Cref{subsec:evaluation}) or are not provided at all (\Cref{subsec:app:kfac-transpose-convolution}).
And they are hard to modify as the code is either closed-source~\cite{chetlur2014cudnn} or written in a low-level language.
This complicates the advance of existing, and the exploration of new, algorithmic ideas for convolutions.

\newsavebox\imcolbox
\begin{lrbox}{\imcolbox}
\makeatletter%
\ifcsname c@listlen\endcsname%
\else \newcounter{listlen}\fi%
\makeatother%
\def\setListlen#1{%
  \setcounter{listlen}{0}%
  \foreach \x in #1{%
    \stepcounter{listlen}%
  }%
}%
\tikzset{%
  tensor/.style={%
    rectangle,%
    minimum width=4ex,%
    minimum height=3.5ex,%
    inner sep=0.3ex,%
    rounded corners,%
    fill=maincolor!30!white,%
    draw=black, thick%
  },%
  index/.style={%
    fill=white,%
    inner sep=0.3ex%
  },%
}%
\def\tensornode#1#2#3#4#5#6{%
  \node[tensor] (#1) {#2};%
  %
  \setListlen{#3}%
  \foreach[count=\n] \name in #3%
  \draw ($ (#1.north east)!{\n/(\the\value{listlen}+1)}!(#1.south east) $) to
  ++(1.5ex, 0) coordinate (#1-\name);%
  %
  \setListlen{#4}%
  \foreach[count=\n] \name in #4%
  \draw ($ (#1.north west)!{\n/(\the\value{listlen}+1)}!(#1.north east) $) to
  ++(0, 1.5ex) coordinate (#1-\name);%
  %
  \setListlen{#5}%
  \foreach[count=\n] \name in #5%
  \draw ($ (#1.north west)!{\n/(\the\value{listlen}+1)}!(#1.south west) $) to
  ++(-1.5ex, 0) coordinate (#1-\name);%
  %
  \setListlen{#6}%
  \foreach[count=\n] \name in #6%
  \draw ($ (#1.south west)!{\n/(\the\value{listlen}+1)}!(#1.south east) $) to
  ++(0, -1.5ex) coordinate (#1-\name);%
}%
\def\P#1{%
  \tensornode{P#1}%
  {$\tPi^{(#1)}$}%
  {{k#1}}
  {{o#1}}
  {{i#1}}
  {{}}
}%
\def\X{%
  \tensornode{X}%
  {$\tX$}%
  {{cin}}
  {{i1, i2}}
  {{}}
  {{}}
}%
\def\W{%
  \tensornode{W}%
  {$\tW$}%
  {{cout}}
  {{k2, k1}}
  {{cin}}
  {{}}
}%
\def\V{%
  \tensornode{V}%
  {$\tV^{(\tY)}$}%
  {{cout}}
  {{o1, o2}}
  {{}}
  {{}}
}%
\def\contract#1#2#3#4#5{%
  (#1-#3) edge[#5] node[index] {#4} (#2-#3)%
}%

  \def\Xunfold#1{%
    \tensornode{Xunfold#1}%
    {$\llbracket\tX\rrbracket$}%
    {{cin, k1, k2}}
    {{o1, o2}}
    {{}}
    {{}}
  }%
  \resizebox{0.215\linewidth}{!}{%
    \begin{tikzpicture}[ultra thick,%
      index/.append style={font={\scriptsize}},%
      ]%
      \matrix [row sep=3ex,column sep=3ex,%
      ampersand replacement=\&, 
      ]{%
        \Xunfold{1} \& \& \Xunfold{2}\\
      };
      \draw %
      (Xunfold2-o1) to ++(0, 0.2) -| (Xunfold1-o1) node [index, midway] {$o_1$}
      \contract{Xunfold1}{Xunfold2}{o2}{$o_2$}{out=5, in=175}%
      (Xunfold1-cin) node[index, right, yshift=0.5ex, fill opacity=0, text opacity=1] {$c_{\text{in}}$}%
      (Xunfold1-k1) node[index, right, fill opacity=0, text opacity=1] {$k_1$}%
      (Xunfold1-k2) node[index, right, yshift=-0.5ex, fill opacity=0, text opacity=1] {$k_2$}%
      (Xunfold2-cin) node[index, right, yshift=1.0ex, fill opacity=0, text opacity=1] {$c_{\text{in}}'$}%
      (Xunfold2-k1) node[index, right, fill opacity=0, text opacity=1] {$k_1'$}%
      (Xunfold2-k2) node[index, right, yshift=-1.0ex, fill opacity=0, text opacity=1] {$k_2'$}%
      ;%
    \end{tikzpicture}%
  }
\end{lrbox}

\newsavebox\tnbox
\begin{lrbox}{\tnbox}
  \resizebox{0.215\linewidth}{!}{%

    \def\P#1#2{%
      \tensornode{P#1-#2}%
      {$\tPi^{(#1)}$}%
      {{k#1}}
      {{o#1}}
      {{i#1}}
      {{}}
    }%
    \def\X#1{%
      \tensornode{X#1}%
      {$\tX$}%
      {{cin}}
      {{i1, i2}}
      {{}}
      {{}}
    }%
    \begin{tikzpicture}[ultra thick]%
      \matrix[row sep=1.5ex,column sep=0ex,%
      ampersand replacement=\&, 
      ]{
        \& \P{1}{1} \& \& \P{1}{2} \\
        \& \P{2}{1} \& \& \P{2}{2} \\
        \X{1} \& \& \begin{scope}[xshift=6ex] \X{2} \end{scope} \& \\
      };%
      \draw %
      \contract{X1}{P1-1}{i1}{$i_1$}{out=90, in=180}%
      \contract{X1}{P2-1}{i2}{$i_2$}{out=90, in=180}%
      \contract{X2}{P2-2}{i2}{$i'_2$}{out=90, in=180};%
      \begin{scope}[index/.append style={yshift=-3.5ex, xshift=-0.5ex}]
        \draw \contract{X2}{P1-2}{i1}{$i'_1$}{out=90, in=180};%
      \end{scope}
      \begin{scope}[opacity=1, index/.append style={text opacity=1}]
        \draw%
        \contract{P1-1}{P1-2}{o1}{$o_1$}{out=0, in=180}%
        \contract{P2-1}{P2-2}{o2}{$o_2$}{out=0, in=180}%
        (P1-1-k1) node[index, right] {$k_1$}%
        (P1-2-k1) node[index, right] {$k'_1$}%
        (P2-1-k2) node[index, right] {$k_2$}%
        (P2-2-k2) node[index, right] {$k'_2$}%
        (X1-cin) node[index, right] {$c_\text{in}$}%
        (X2-cin) node[index, right] {$c'_\text{in}$};%
      \end{scope}
    \end{tikzpicture}
  }
\end{lrbox}

\newsavebox\tnsimplifiedbox
\begin{lrbox}{\tnsimplifiedbox}
  \resizebox{0.215\linewidth}{!}{%

    \def\P#1#2{%
      \tensornode{P#1-#2}%
      {$\tPi^{(#1)}$}%
      {{k#1}}
      {{o#1}}
      {{i#1}}
      {{}}
    }%
    \def\X#1{%
      \tensornode{X#1}%
      {$\tX$}%
      {{cin}}
      {{i1, i2}}
      {{}}
      {{}}
    }%
    \begin{tikzpicture}[ultra thick]%
      \matrix[row sep=1.5ex,column sep=0ex,%
      ampersand replacement=\&, 
      ]{
        \&
        \begin{scope}[tensor/.append style={opacity=0}]
          \P{1}{1}
        \end{scope}
        \& \&
        \begin{scope}[tensor/.append style={opacity=0}]
          \P{1}{2}
        \end{scope}
        \\
        \&
        \begin{scope}[tensor/.append style={opacity=0}]
          \P{2}{1}
        \end{scope}
        \& \&
        \begin{scope}[tensor/.append style={opacity=0}]
          \P{2}{2}
        \end{scope}
        \\
        \X{1} \& \& \begin{scope}[xshift=6ex] \X{2} \end{scope} \& \\
      };%
      \coordinate (P1-1-i1-inner) at ($(P1-1-i1)+(1.3ex, 0)$);%
      \coordinate (P1-1-o1-inner) at ($(P1-1-o1)+(0, -1.3ex)$);%
      \coordinate (P1-1-k1-inner) at ($(P1-1-k1)+(-1.3ex, 0)$);%
      \coordinate (P1-2-i1-inner) at ($(P1-2-i1)+(1.3ex, 0)$);%
      \coordinate (P1-2-o1-inner) at ($(P1-2-o1)+(0, -1.3ex)$);%
      \coordinate (P1-2-k1-inner) at ($(P1-2-k1)+(-1.3ex, 0)$);%
      \coordinate (P2-1-i2-inner) at ($(P2-1-i2)+(1.3ex, 0)$);%
      \coordinate (P2-1-o2-inner) at ($(P2-1-o2)+(0, -1.3ex)$);%
      \coordinate (P2-1-k2-inner) at ($(P2-1-k2)+(-1.3ex, 0)$);%
      \coordinate (P2-2-i2-inner) at ($(P2-2-i2)+(1.3ex, 0)$);%
      \coordinate (P2-2-o2-inner) at ($(P2-2-o2)+(0, -1.3ex)$);%
      \coordinate (P2-2-k2-inner) at ($(P2-2-k2)+(-1.3ex, 0)$);%
      \draw[arrows={Triangle}-] (P1-1-i1-inner) to ++(0.01ex,0);%
      \draw [out=30, in=270] (P1-1-i1-inner) to (P1-1-o1-inner);%
      \draw [out=-30, in=180] (P1-1-i1-inner) to (P1-1-k1-inner);%
      \draw[arrows={Triangle}-] (P2-1-i2-inner) to ++(0.01ex,0);%
      \draw [out=30, in=270] (P2-1-i2-inner) to (P2-1-o2-inner);%
      \draw [out=-30, in=180] (P2-1-i2-inner) to (P2-1-k2-inner);%
      \draw[arrows={Triangle}-] (P1-2-i1-inner) to ++(0.01ex,0);%
      \draw [out=30, in=270] (P1-2-i1-inner) to (P1-2-o1-inner);%
      \draw [out=-30, in=180] (P1-2-i1-inner) to (P1-2-k1-inner);%
      \draw[arrows={Triangle}-] (P2-2-i2-inner) to ++(0.01ex,0);%
      \draw [out=30, in=270] (P2-2-i2-inner) to (P2-2-o2-inner);%
      \draw [out=-30, in=180] (P2-2-i2-inner) to (P2-2-k2-inner);%
      \draw %
      \contract{X1}{P1-1}{i1}{$i_1$}{out=90, in=180}%
      \contract{X1}{P2-1}{i2}{$i_2$}{out=90, in=180}%
      \contract{X2}{P2-2}{i2}{$i'_2$}{out=90, in=180};%
      \begin{scope}[index/.append style={yshift=-3.5ex, xshift=-0.5ex}]
        \draw \contract{X2}{P1-2}{i1}{$i'_1$}{out=90, in=180};%
      \end{scope}
      \draw%
      \contract{P1-1}{P1-2}{o1}{$o_1$}{out=0, in=180}%
      \contract{P2-1}{P2-2}{o2}{$o_2$}{out=0, in=180}%
      (P1-1-k1) node[index, right] {$k_1$}%
      (P1-2-k1) node[index, right] {$k'_1$}%
      (P2-1-k2) node[index, right] {$k_2$}%
      (P2-2-k2) node[index, right] {$k'_2$}%
      (X1-cin) node[index, right] {$c_\text{in}$}%
      (X2-cin) node[index, right] {$c'_\text{in}$};%
    \end{tikzpicture}
  }
\end{lrbox}

\begin{figure*}[t]
  \centering
  \begin{minipage}[t]{\linewidth}
\begin{lstlisting}[%
    language=Python,
    linebackgroundcolor={%
      \ifnum \value{lstnumber}=7\color{black!10!white}\fi%
      \ifnum \value{lstnumber}=8\color{black!10!white}\fi%
      \ifnum \value{lstnumber}=9\color{black!10!white}\fi%
      \ifnum \value{lstnumber}=10\color{black!10!white}\fi%
      %
      \ifnum \value{lstnumber}=12\color{maincolor!10!white}\fi%
      \ifnum \value{lstnumber}=13\color{maincolor!10!white}\fi%
      \ifnum \value{lstnumber}=14\color{maincolor!10!white}\fi%
      \ifnum \value{lstnumber}=15\color{maincolor!10!white}\fi%
      \ifnum \value{lstnumber}=16\color{maincolor!10!white}\fi%
      \ifnum \value{lstnumber}=17\color{maincolor!10!white}\fi%
      \ifnum \value{lstnumber}=18\color{maincolor!10!white}\fi%
      %
      \ifnum \value{lstnumber}=20\color{secondcolor!10!white}\fi%
      \ifnum \value{lstnumber}=21\color{secondcolor!10!white}\fi%
      \ifnum \value{lstnumber}=22\color{secondcolor!10!white}\fi%
      \ifnum \value{lstnumber}=23\color{secondcolor!10!white}\fi%
      \ifnum \value{lstnumber}=24\color{secondcolor!10!white}\fi%
    }%
    ]
from einconv import conv_index_pattern; from einops import einsum; import torch

I, K, D, P, S = ...  # convolution hyper-parameters (2-tuples)
X = torch.rand(C_in, I[0], I[1])  # input to convolution
kfc_shape = (C_in * K[0] * K[1], C_in * K[0] * K[1]) # final shape

def kfc_im2col():
  """Compute Kronecker factor via im2col."""
  X_unf = torch.nn.functional.unfold(X, K, D, P, S)
  return einsum(X_unf, X_unf, "i out, j out -> i j")

def kfc_tn():
  """Compute Kronecker factor via its tensor network."""
  Pi1 = conv_index_pattern(I[0], K[0], S[0], P[0], D[0])
  Pi2 = conv_index_pattern(I[1], K[1], S[1], P[1], D[1])
  return einsum(X, Pi1, Pi2, X, Pi1, Pi2,
     "c_in i1 i2, i1 o1 k1, i2 o2 k2, c_in_ i1_ i2_, i1_ o1_ k1_ i2_ o2_ k2_"
     + " -> c_in k1 k2 c_in_ k1_ k2_").reshape(kfc_shape)

def kfc_simplified_tn(): # for dense convolutions
  """Compute Kronecker factor via its simplified tensor network."""
  X = X.reshape(C_in, I[0] // K[0], K[0], I[1] // K[1], K[1])
  return einsum(X, X, "c_in o1 k1 o2 k2, c_in_ o1_ k1_ o2_ k2_"
     + "-> c_in k1 k2 c_in_ k1_ k2_").reshape(kfc_shape)
\end{lstlisting}
  \end{minipage}
  \caption{Many convolution-related routines can be expressed as TNs and evaluated with \texttt{einsum}.
    We illustrate this for the input-based factor of KFAC for convolutions~\citep[KFC,][]{grosse2016kroneckerfactored}, whose standard implementation (\emph{top}) requires unfolding the input (high memory).
    The TN (\emph{middle}) enables internal optimizations inside \texttt{einsum} (e.g.\,with contraction path optimizers like \texttt{opt\_einsum}~\cite{smith2018opteinsum}).
    (\emph{Bottom}) In many cases, the TN further simplifies due to structures in the index pattern, which reduces cost.}\label{fig:visual-abstract-code}
  \begin{tikzpicture}[overlay, remember picture, ultra thick]%
    \node (im2col) [ultra thick, rectangle, fill=white, rounded corners, draw=black!10!white, anchor = north east]%
    at ($(current page.north east)+(-3.85,-3)$)%
    {\usebox\imcolbox};
    \node [anchor=south east, inner sep=0pt] at (im2col.north east) {
    };
    \node (tn-label) [anchor=north east, inner sep=0pt, maincolor] at (im2col.south east) {
    };
    \node (tn) [ultra thick, rectangle, fill=white, rounded corners, draw=maincolor!10!white, anchor = north east]%
    at (tn-label.south east)%
    {\usebox\tnbox};
    \node (tn-simplified-label) [anchor=north east, inner sep=0pt, secondcolor] at (tn.south east) {
    };
    \node (tn-simplified) [ultra thick, rectangle, fill=white, rounded corners, draw=secondcolor!10!white, anchor = north east]%
    at (tn-simplified-label.south east)%
    {\usebox\tnsimplifiedbox};
    ;%
  \end{tikzpicture}
  \vspace*{-3ex}
\end{figure*}


Here, we seek to reduce this complexity gap by viewing convolutions as tensor networks~\citep[TNs,][]{penrose1971applications,biamonte2017tensor,bridgeman2017hand} which express the underlying tensor multiplications as diagrams.
These diagrams are simpler to parse than mathematical equations and can seamlessly be (i) manipulated to take derivatives, add batching, or extract sub-tensors, (ii) merged with other diagrams, and (iii) evaluated with \texttt{einsum}.
This yields simple, modifiable implementations that benefit from automated under-the-hood-optimizations for efficient TN contraction developed by the quantum simulation community~\citep[e.g.][]{smith2018opteinsum,gray2021hyper,zhang2020parallel,nvidia2023cuquantum}, like finding a high-quality contraction order or distributing computations:
\begin{enumerate}
\item We use the TN format of convolution from~\citet{hayashi2019einconv} to derive diagrams and \texttt{einsum} formulas for autodiff and less standard routines for curvature approximations with support for all hyper-parameters, batching, groups, and any dimension (\Cref{tab:einsum-expressions}).

\item We present transformations based on the convolution's connectivity pattern to re-wire and symbolically simplify TNs before evaluation (example in \Cref{fig:visual-abstract-code}).

\item We compare default and TN implementations, demonstrating optimal peak memory reduction and run time improvements up to 4.5\,x for a recent KFAC variant, and showcase their flexibility to impose hardware-efficient dropout for randomized backpropagation.
\end{enumerate}
Our work not only provides simpler perspectives and implementations that facilitate the exploration of algorithmic ideas for convolutions, but also directly advances second-order methods like KFAC: It enables more frequent pre-conditioner updates, using larger batches without going out of memory, and extending KFAC to transpose convolution.
These improvements are important for second-order optimization and other applications like Laplace approximations~\cite{daxberger2021laplace} and influence functions~\cite{grosse2023studying}.


\section{Preliminaries}\label{sec:background}
We briefly review 2d convolution (\Cref{subsec:convolution}), tensor multiplication and \texttt{einsum} (\Cref{subsec:tensor-multiplication}), then introduce the graphical TN notation and apply it to convolution (\Cref{subsec:tn-convolution}).
Bold lower-case ($\va$), upper-case ($\mA$), and upper-case sans-serif ($\tA$) symbols indicate vectors, matrices, and tensors.
Entries follow the same convention but use regular font weight; $[\cdot]$ denotes slicing (\,$[\mA]_{i,j} = \emA_{i,j}$).
Parenthesized indices mean reshapes, e.g.\,$[\va]_{(i,j)} = [\mA]_{i,j} $ with $\va$ the flattened matrix $\mA$.

\subsection{Convolution}\label{subsec:convolution}
2d convolutions process channels of 2d signals $\tX \in \sR^{C_{\text{in}} \times I_1 \times I_2}$ with $C_{\text{in}}$ channels of spatial dimensions\footnote{ We prefer $I_1, I_2$ over the more common choice $H, W$ to simplify the generalization to higher dimensions.
} $I_1, I_2$ by sliding a collection of $C_{\text{out}}$ filter banks, arranged in a kernel $\tW \in \sR^{C_{\text{out}} \times C_{\text{in}} \times K_1 \times K_2}$ with kernel size $K_1, K_2$, over the input.
The sliding operation depends on various hyper-parameters~\citep[padding, stride, \dots, see][]{dumoulin2016guide}.
At each step, the filters are contracted with the overlapping area, yielding the channel values of a pixel in the output $\tY \in \sR^{C_{\text{out}} \times O_1 \times O_2}$ with spatial dimensions $O_1, O_2$.
Optionally, a bias from $\vb \in \sR^{C_{\text{out}}}$ is added per channel.

One way to implement convolution is via matrix multiplication~\citep{chellapilla2006high}, similar to fully-connected layers.
First, one extracts the overlapping patches from the input for each output, then flattens and column-stacks them into a matrix $\llbracket \tX \rrbracket \in \sR^{C_{\text{in}} K_1 K_2 \times O_1 O_2}$, called the \emph{unfolded input} (or \texttt{im2col}).
Multiplying a matrix view $\mW \in \sR^{C_{\text{out}} \times C_{\text{in}} K_1 K_2}$ of the kernel onto the unfolded input then yields a matrix view $\mY$ of $\tY$ (the vector of ones, $\vone_{O_1 O_2}$, copies the bias for each channel),
\begin{equation}\label{eq:convolution-with-unfolded-input}
  \mY
  =
  \mW
  \llbracket \tX \rrbracket
  +
  \vb\, \vone^{\top}_{O_1 O_2}
  \in
  \sR^{C_{\text{out}} \times O_1 O_2}\,.
\end{equation}
We can also view convolution as an affine map of the flattened input $\vx \in \sR^{C_{\text{in}} I_1 I_2}$ into a vector view $\vy$ of $\tY$ with a Toeplitz-structured matrix $\mA(\tW) \in \sR^{C_{\text{out}} O_1 O_2 \times C_{\text{in}} I_1 I_2}$,
\begin{equation}\label{eq:convolution-with-unfolded-kernel}
  \vy = \mA(\tW) \vx + \vb \otimes \vone_{O_1 O_2} \in \sR^{C_{\text{out}} O_1 O_2}\,.
\end{equation}
This perspective is uncommon in code, but used in theoretical works~\citep[e.g.][]{singh2023hessian} as it highlights the similarity between convolutions and dense layers.

\begin{figure*}[t]
  \centering
  \begin{subfigure}[b]{0.312\linewidth}
    \centering
    \input{fig/tensor_networks/commands.tex}
    \resizebox{!}{3.6cm}{%
      \begin{tikzpicture}[ultra thick]%
        \matrix[row sep=3ex,%
        column sep=0ex,%
        ampersand replacement=\&%
        ]{
          \& \P{1} \& \\
          \& \P{2} \& \\
          \X \& \& \W \\
        };%
        \draw %
        \contract{X}{P1}{i1}{$i_1$}{out=90, in=180}%
        \contract{X}{P2}{i2}{$i_2$}{out=90, in=180}%
        \contract{P1}{W}{k1}{$k_1$}{out=0, in=90}%
        \contract{P2}{W}{k2}{$k_2$}{out=0, in=90}%
        \contract{X}{W}{cin}{$c_{\text{in}}$}{}%
        (W-cout) node[index, right] {$c_{\text{out}}$}%
        (P1-o1) node[index, above] {$o_1$}%
        (P2-o2) node[index, above] {$o_2$};%
        \begin{scope}[tensor/.append style={fill=secondcolor!30!white},%
          xshift=13ex, yshift=7.5ex]%
          \tensornode{Y}%
          {$\tY$}%
          {{cout}}
          {{o1, o2}}
          {{}}
          {{}};
          \draw%
          (Y.west) node [left, xshift=-0.5ex] {$=$}%
          (Y-o1) node [index, above, xshift=-0.5ex] {$o_1$}%
          (Y-o2) node [index, above, xshift=0.5ex] {$o_2$}%
          (Y-cout) node [index, right] {$c_{\text{out}}$}%
          ;%
        \end{scope}
      \end{tikzpicture}
    }
    \caption{Convolution}\label{subfig:visual-abstract-convolution}
  \end{subfigure}
  \hfill
  \begin{subfigure}[b]{0.3\linewidth}
    \centering
    \input{fig/tensor_networks/commands.tex}
    \resizebox{!}{3.6cm}{%
      \begin{tikzpicture}[%
        ultra thick,%
        ]
        \matrix[row sep=3ex,column sep=0ex, ampersand replacement=\&]{
          \& \P{1} \\
          \& \P{2} \\
          \X \& \\
        };%
        \draw %
        \contract{X}{P1}{i1}{$i_1$}{out=90, in=180}%
        \contract{X}{P2}{i2}{$i_2$}{out=90, in=180};%
        \coordinate (o1o2) at ($(P1-o1)+(-6ex,3ex)$);%
        \draw [out=150, in=285] (P1-o1) to (o1o2);%
        \draw [out=150, in=255] (P2-o2) to (o1o2);%
        \draw[arrows=-{Triangle}] (o1o2) to ++(0,1ex);%
        \draw (o1o2) to ++(0, 2ex) node [index, above] {$(o_1, o_2)$};%
        \coordinate (cink1k2) at ($(P2-k2)+(7ex,0)$);%
        \draw [out=330, in=150, looseness=1.75] (P1-k1) to (cink1k2);%
        \draw (P2-k2) to (cink1k2);%
        \draw [out=30, in=210, looseness=1.1] (X-cin) to (cink1k2);%
        \draw[arrows=-{Triangle}] (cink1k2) to ++(1ex,0);%
        \draw (cink1k2) to ++(2ex, 0) node [index, right] {$(c_{\text{in}}, k_1, k_2)$};%
        \draw %
        (X-cin) node[index, right] {$c_{\text{in}}$}%
        (P1-k1) node[index, right] {$k_1$}%
        (P2-k2) node[index, right] {$k_2$}%
        (P1-o1) node[index, above] {$o_1$}%
        (P2-o2) node[index, above] {$o_2$};%
        \begin{scope}[tensor/.append style={fill=secondcolor!30!white},%
          xshift=12.5ex, yshift=12.5ex]
          \tensornode{X}%
          {$\llbracket \tX \rrbracket$}%
          {{cink1k2}}
          {{o1o2}}
          {{}}
          {{}};
          \node[right, index] at (X-cink1k2) {$(c_{\text{in}}, k_1, k_2)$};%
          \node[above, index] at (X-o1o2) {$(o_1, o_2)$};%
          \draw (X.west) node [left, xshift=-0.5ex] {$=$};%
        \end{scope}
      \end{tikzpicture}
    }
    \caption{Input unfolding}\label{subfig:visual-abstract-unfolded-input}
  \end{subfigure}
  \hfill
  \begin{subfigure}[b]{0.336\linewidth}
    \centering
    \input{fig/tensor_networks/commands.tex}
    \resizebox{!}{3.6cm}{%
      \begin{tikzpicture}[%
        ultra thick,%
        ]
        \matrix[row sep=3ex,column sep=0ex, ampersand replacement=\&]{
          \P{1} \\
          \P{2} \\
          \& \W \\
        };%
        \draw %
        \contract{W}{P1}{k1}{$k_1$}{out=90, in=0}%
        \contract{W}{P2}{k2}{$k_2$}{out=90, in=0};%
        \coordinate (couto1o2) at ($(P1-o1)+(6ex,3ex)$);%
        \draw [out=30, in=240] (P1-o1) to (couto1o2);%
        \draw [out=30, in=270] (P2-o2) to (couto1o2);%
        \draw [out=30, in=300] (W-cout) to (couto1o2);%
        \draw[arrows=-{Triangle}] (couto1o2) to ++(0,1ex);%
        \draw (couto1o2) to ++(0, 2ex) node [index, above] {$(c_{\text{out}}, o_1, o_2)$};%
        \coordinate (cini1i2) at ($(P2-i2)+(-7ex,0)$);%
        \draw [out=210, in=30, looseness=1.75] (P1-i1) to (cini1i2);%
        \draw (P2-i2) to (cini1i2);%
        \draw [out=150, in=330, looseness=1.1] (W-cin) to (cini1i2);%
        \draw[arrows=-{Triangle}] (cini1i2) to ++(-1ex,0);%
        \draw (cini1i2) to ++(-2ex, 0) node [index, left] {$(c_{\text{in}}, i_1, i_2)$};%
        \draw %
        (W-cin) node[index, left] {$c_{\text{in}}$}%
        (W-cout) node[index, right] {$c_{\text{out}}$}%
        (P1-i1) node[index, left] {$i_1$}%
        (P2-i2) node[index, left] {$i_2$}%
        (P1-o1) node[index, above] {$o_1$}%
        (P2-o2) node[index, above] {$o_2$}%
        ;%
        \begin{scope}[tensor/.append style={fill=secondcolor!30!white},%
          xshift=-12.5ex, yshift=12.5ex]
          \tensornode{A}%
          {$\mA(\tW)$}%
          {{}}
          {{couto1o2}}
          {{cini1i2}}
          {{}};
          \node[left, index] at (A-cini1i2) {$(c_{\text{in}}, i_1, i_2)$};%
          \node[above, index] at (A-couto1o2) {$(c_{\text{out}}, o_1, o_2)$};%
          \draw (A.east) node [right, xshift=0.5ex] {$=$};%
        \end{scope}
      \end{tikzpicture}
    }
    \caption{Kernel unfolding}\label{subfig:visual-abstract-unfolded-kernel}
  \end{subfigure}
  \caption{TNs of (\subref{subfig:visual-abstract-convolution}) 2d convolution and (\subref{subfig:visual-abstract-unfolded-input},\subref{subfig:visual-abstract-unfolded-kernel}) connections to its matrix multiplication view.
    The connectivity along each dimension is explicit via an index pattern tensor $\tPi$.
  }\label{fig:visual-abstract}
\end{figure*}

\subsection{Tensor Multiplication}\label{subsec:tensor-multiplication}
Tensor multiplication unifies outer (Kronecker), element-wise (Hadamard), and inner products and uses the input-output index relation to infer the multiplication type.
We start with the binary case, then generalize to more inputs: Consider $\tA, \tB, \tC$ whose index names are described by the index tuples $S_1, S_2, S_3$ where $S_3 \subseteq ( S_1 \cup S_2)$ (converting tuples to sets if needed).
Any product of $\tA$ and $\tB$ can be described by the multiplication operator $*_{(S_1, S_2, S_3)}$ with
\begin{align}\label{eq:tensor-multiplication}
  \textstyle
  \tC = *_{(S_1, S_2, S_3)}(\tA, \tB)
  \quad\Leftrightarrow\quad
  [\tC]_{S_3}
  =
  \sum_{(S_1 \cup S_2) \setminus S_3}
  [\tA]_{S_1}
  [\tB]_{S_2}
\end{align}
summing over indices that are not present in the output.
E.g., for two matrices $\mA, \mB$, their product is $\mA \mB = *_{((i,j), (j, k), (i, k))}(\mA, \mB)$ (see \Cref{subsec:app:example:matrix-multiplication}), their Hadamard product $\mA \odot \mB = *_{((i,j), (i, j), (i, j))}(\mA, \mB)$, and their Kronecker product $\mA \otimes \mB = *_{((i,j), (k, l), ((i,k),(j,l)))} (\mA, \mB)$.
Libraries support this functionality via \texttt{einsum}, which takes a string encoding of $S_1, S_2, S_3$, followed by $\tA, \tB$.
It also accepts longer sequences $\tA_1, \dots, \tA_N$ with index tuples $S_1, S_2, \dots, S_N$ and output index tuple $S_{N+1}$,
\begin{align}\label{eq:n-ary-tensor-multiplication}
  \tA_{N+1}
  =
  *_{(S_1, \dots, S_N, S_{N+1})}(\tA_1, \dots, \tA_N)
  \,\Leftrightarrow\,
  [\tA_{N+1}]_{S_{N+1}}
  \!\!&=\!
        \sum_{\left( \bigcup_{n=1}^{N}S_n \right) \setminus S_{N+1}}
        \!\!
        \left(
        \prod_{n=1}^N
        [\tA_n]_{S_n}
        \!
        \right).
\end{align}
Binary and $N$-ary tensor multiplication are commutative: We can simultaneously permute operands and their index tuples without changing the result,
\begin{align*}
  *_{(S_1, S_2, S_3)}(\tA, \tB)
  &=
    *_{(S_2, S_1, S_3)}(\tB, \tA)\,,
    \quad
    *_{(.,S_i,.,S_j,.)}(., \tA_i, ., \tA_j, .)
  &=
    *_{(., S_j, ., S_i, .)}(., \tA_j, ., \tA_i, .)
\end{align*}
They are also associative, i.e.\,we can multiply operands in any order.
However, the notation becomes involved as it requires additional set arithmetic to detect summable indices (see \Cref{subsec:associativity-example} for an example).

\begin{table*}[t]
  \centering%
  \tikzset{%
    einsum/.style={%
      inner sep=0pt,%
      font=\ttfamily\bfseries\color{secondcolor},%
      align=left,%
      scale=0.7,%
    }%
  }%
  \caption{Contraction expressions of operations related to 2d convolution.
    They include batching and channel groups, which are standard features in implementations.
    We describe each operation by a tuple of input tensors and a contraction string that uses the \texttt{einops} library's syntax~\citep{rogozhnikov2022einops} which can express index (un-)grouping.
    Some quantities are only correct up to a scalar factor which is suppressed for brevity.
    See \Cref{sec:app:visual-tour} for visualizations and \Cref{tab:app:einsum-expressions-extensive} for more operations.}\label{tab:einsum-expressions}
  \begin{small}
    \begin{tabular}{lll}
      \toprule
      \textbf{Operation}
      & \textbf{Operands}
      & \textbf{Contraction string (\texttt{einops}}~\citep{rogozhnikov2022einops} \textbf{convention)}
      \\ \midrule
      Conv.\,(no bias)
      & $\tX, \tPi^{(1)}, \tPi^{(2)}, \tW$
      & \tikz[baseline=-0.5ex]{%
        \node[einsum]{%
        "n (g c\_in) i1 i2, i1 o1 k1, i2 o2 k2, (g c\_out) c\_in k1 k2 \\
      \ -> n (g c\_out) o1 o2"%
      }}%
      \\
      Unf.\,input (\texttt{im2col})
      & $\tX, \tPi^{(1)}, \tPi^{(2)}$
      & \tikz[baseline=-0.5ex]{%
        \node[einsum]{%
        "n c\_in i1 i2, i1 o1 k1, i2 o2 k2 -> n (c\_in k1 k2) (o1 o2)"%
      }}\!\!\!%
      \\
      Unf.\,kernel (Toeplitz)\!\!\!
      & $\tPi^{(1)}, \tPi^{(2)}, \tW$
      & \tikz[baseline=-0.5ex]{%
        \node[einsum]{%
        "i1 o1 k1, i2 o2 k2, c\_out c\_in k1 k2\\
      \ -> (c\_out o1 o2) (c\_in i1 i2)"%
      }}%
      \\
      \midrule
      Weight VJP
      & $\tX, \tPi^{(1)}, \tPi^{(2)}, \tV^{(\tY)}$
      & \tikz[baseline=-0.5ex]{%
        \node[einsum]{%
        "n (g c\_in) i1 i2, i1 o1 k1, i2 o2 k2, n (g c\_out) o1 o2 \\
      \ -> (g c\_out) c\_in k1 k2"%
      }}%
      \\
      Input VJP (tr.\,conv.)
      & $\tW, \tPi^{(1)}, \tPi^{(2)}, \tV^{(\tY)}$
      & \tikz[baseline=-0.5ex]{%
        \node[einsum]{%
        "(g c\_out) c\_in k1 k2, i1 o1 k1, i2 o2 k2, n (g c\_out) o1 o2 \\
      \ -> n (g c\_in) i1 i2"%
      }}\!\!\!\!%
      \\
      \midrule
      KFC/KFAC-expand
      & $\tX, \tPi^{(1)}, \tPi^{(2)}, \tX, \tPi^{(1)}, \tPi^{(2)}$\!\!\!\!\!
      & \tikz[baseline=-0.5ex]{%
        \node[einsum]{%
        "n (g c\_in) i1 i2, i1 o1 k1, i2 o2 k2, n (g c\_in\_) i1 i2,\\
      \ i1 o1 k1\_, i2 o2 k2\_ -> g (c\_in k1 k2) (c\_in\_ k1\_ k2\_)"%
      }}%
      \\
      KFAC-reduce
      & $\tX, \tPi^{(1)}, \tPi^{(2)}, \tX, \tPi^{(1)}, \tPi^{(2)}$\!\!\!\!\!
      & \tikz[baseline=-0.5ex]{%
        \node[einsum]{%
        "n (g c\_in) i1 i2, i1 o1 k1, i2 o2 k2, n (g c\_in\_) i1 i2,\\
      \ i1 o1\_ k1\_, i2 o2\_ k2\_ -> g (c k1 k2) (c\_ k1\_ k2\_)"%
      }}%
      \\
      \bottomrule
    \end{tabular}
  \end{small}
\end{table*}

\subsection{Tensor Networks \& Convolution}\label{subsec:tn-convolution}
A simpler way to understand tensor multiplications is via diagrams developed by
e.g.\,\citet{penrose1971applications}. Rank-$K$ tensors are represented by nodes
with $K$ legs labelled by the index's name\footnote{We use identical shapes for
  all tensors. Leg orientation does not assign properties like
  co-/contra-variance.}.
\resizebox{!}{1.7ex}{%
  \tikz[baseline=-1.2ex,%
  index/.append style={fill=white, fill opacity = 1},%
  very thick,%
  ]{%
    \tensornode{a}%
    {$\phantom{\mA}\!\!\!\!\!\va$}%
    {{i}}
    {{}}
    {{}}
    {{}};
    \node[right, index] at (a-i) {$i$};%
  }%
} %
denotes a vector $\va$,
\resizebox{!}{1.7ex}{%
  \tikz[baseline=-1.2ex,%
  index/.append style={fill=white, fill opacity = 1},%
  very thick,%
  ]{%
    \tensornode{B}%
    {$\mB$}%
    {{j}}
    {{}}
    {{i}}
    {{}};
    \node[left, index] at (B-i) {$i$};%
    \node[right, index] at (B-j) {$j$};%
  }%
} %
a matrix $\mB$, and
\resizebox{!}{1.7ex}{%
  \tikz[baseline=-1.2ex,%
  index/.append style={fill=white, fill opacity = 1},%
  very thick,%
  ]{%
    \tensornode{C}%
    {$\tC$}%
    {{j, k}}
    {{}}
    {{i}}
    {{}};
    \node[left, index] at (C-i) {$i$};%
    \node[right, index] at (C-j) {$j$};%
    \node[right, index, fill opacity=0, text opacity=1] at (C-k) {$k$};%
  }%
} %
a rank-3 tensor $\tC$. A Kronecker delta $[\vdelta]_{i,j} = \delta_{i,j}$
is simply a line,
\resizebox{!}{1.7ex}{%
  \tikz[baseline=-1.2ex,%
  index/.append style={fill=white, fill opacity = 1},%
  very thick,%
  ]{%
    \tensornode{Delta}%
    {$\phantom{\mA}\!\!\!\!\!\vdelta$}%
    {{i}}
    {{}}
    {{j}}
    {{}};
    \node[right, index] (i) at (Delta-i) {$i$};%
    \node[left, index] at (Delta-j) {$j$};%
    \node[scale=1.5, anchor=west] (eq2) at (i.east) {$=$};%
    \begin{scope}[xshift=15ex]
      \tensornode{I}%
      {$\mI$}%
      {{i}}
      {{}}
      {{j}}
      {{}};
      \node[right, index] (i2) at (I-i) {$i$};%
      \node[left, index] at (I-j) {$j$};%
    \end{scope}
    \node[scale=1.5, anchor=west] (eq) at (i2.east) {$=$};%
    \draw (eq.east) node [left, index] {$j$} to ++(4.5ex, 0) node [right, index] {$i$} ;%
  }%
}. %
Multiplications are indicated by connections between legs. For inner
multiplication, we join the legs of the involved indices, e.g.\,the
matrix multiplication diagram is %
\resizebox{!}{1.7ex}{%
  \tikz[baseline=-1.2ex,%
  index/.append style={fill=white, fill opacity = 1},%
  very thick,%
  ]{%
    \tensornode{AB}%
    {$\mA \mB$}%
    {{k}}
    {{}}
    {{i}}
    {{}};
    \node[left, index] at (AB-i) {$i$};%
    \node[right, index] at (AB-k) {$k$};%
  }%
} $=$ \resizebox{!}{1.7ex}{%
  \tikz[baseline=-1.2ex,%
  index/.append style={fill=white, fill opacity = 1},%
  very thick,%
  ]{%
    \tensornode{A}%
    {$\mA$}%
    {{j}}
    {{}}
    {{i}}
    {{}};
    \begin{scope}[xshift=8ex]
      \tensornode{B}%
      {$\mB$}%
      {{k}}
      {{}}
      {{j}}
      {{}};
    \end{scope}
    \node[left, index] at (A-i) {$i$};%
    \node[right, index] at (B-k) {$k$};%
    \draw \contract{A}{B}{j}{$j$}{};%
  }}. %
Element-wise multiplication is similar, but with a leg sticking out.
The Hadamard and Kronecker product diagrams are
\begin{align}\label{eq:hadamard-and-kronecker}
  \scalebox{0.66}{%
  \tikz[baseline=-0.6ex, very thick]{%
  \tensornode{AB}%
  {$\mA \odot \mB$}%
  {{j}}
  {{}}
  {{i}}
  {{}};
  \node[left, index] at (AB-i) {$i$};%
  \node[right, index] at (AB-j) {$j$};%
  }%
  \ =
  \tikz[baseline=-0.6ex, very thick]{%
  \matrix[%
  row sep=1ex,column sep=2ex,%
  ampersand replacement=\&%
  ]{%
  \tensornode{A}%
  {$\mA$}%
  {{j}}
  {{}}
  {{i}}
  {{}};
  \\
  \tensornode{B}%
  {$\mB$}%
  {{j}}
  {{}}
  {{i}}
  {{}};
  \\
  };
  \draw (A-i) to (B-i);
  \draw ($(A-i)!0.5!(B-i)$) to ++(-2ex,0) node [left, index] {$i$};
  \draw (A-j) to (B-j);
  \draw ($(A-j)!0.5!(B-j)$) to ++(2ex,0) node [right, index] {$j$};
  }},
  \qquad
  \scalebox{0.66}{%
  \tikz[baseline=-0.6ex, very thick]{%
  \tensornode{AB}%
  {$\mA \otimes \mB$}%
  {{jl}}
  {{}}
  {{ik}}
  {{}};
  \node[left, index] at (AB-ik) {$(i,k)$};%
  \node[right, index] at (AB-jl) {$(j,l)$};%
  }%
  \ =
  \tikz[baseline=-0.6ex, very thick]{%
  \matrix[%
  row sep=1ex,column sep=2ex,%
  ampersand replacement=\&%
  ]{%
  \tensornode{A}%
  {$\mA$}%
  {{j}}
  {{}}
  {{i}}
  {{}};
  \\
  \tensornode{B}%
  {$\mB$}%
  {{l}}
  {{}}
  {{k}}
  {{}};
  \\
  };
  \coordinate (ik) at ($(A)!0.5!(B)+(-7ex,0)$);%
  \draw (A-i) to[out=180, in=30] (ik);%
  \draw[out=180, in=330] (B-k) to (ik);%
  \draw[arrows=-{Triangle[reversed]}] (ik) to ++(0.1ex,0);%
  \draw (ik) to ++(-2ex,0) node [index, left] {$(i,k)$};%
  \coordinate (jl) at ($(A)!0.5!(B)+(7ex,0)$);%
  \draw[out=0, in=150] (A-j) to (jl);%
  \draw[out=0, in=210] (B-l) to (jl);%
  \draw[arrows=-{Triangle[reversed]}] (jl) to ++(-0.1ex,0);%
  \draw (jl) to ++(2ex,0) node [index, right] {$(j,l)$};%
  \node[left, index] at (A-i) {$i$};%
  \node[left, index] at (B-k) {$k$};%
  \node[right, index] at (A-j) {$j$};%
  \node[right, index] at (B-l) {$l$};%
  }}.
\end{align}
Note that the outer tensor product is a rank-4 tensor and must be reshaped (indicated by black triangles\footnote{Reshape can be seen as multiplication with a one-hot tensor, but we decided to use a separate symbol to emphasize it merely serves for re-interpretation and does not cause much computation.})
into a matrix.
This syntax allows for extracting and embedding tensors along diagonals; e.g.\,taking a matrix diagonal, %
\resizebox{!}{1.7ex}{%
  \tikz[baseline=-1.2ex,%
  index/.append style={fill=white, fill opacity = 1},%
  very thick,%
  ]{%
    \tensornode{A}%
    {$\diag(\mA)$}%
    {{i}}
    {{}}
    {{}}
    {{}};
    \node[right, index] at (A-i) {$i$};%
  }%
}%
$=$ \resizebox{!}{1.7ex}{%
  \tikz[baseline=-1.2ex,%
  index/.append style={fill=white, fill opacity = 1},%
  very thick,%
  ]{%
    \tensornode{A}%
    {$\mA$}%
    {{i}}
    {{}}
    {{j}}
    {{}};
    \draw[] (A-j) to ++(0,2.5ex) -| (A-i) to ++(2ex,0) node[right, index] {$i$};%
  }%
}%
, or forming a diagonal matrix, %
\resizebox{!}{1.7ex}{%
  \tikz[baseline=-1.2ex,%
  index/.append style={fill=white, fill opacity = 1},%
  very thick,%
  ]{%
    \tensornode{A}%
    {$\diag(\va)$}%
    {{i}}
    {{}}
    {{j}}
    {{}};
    \node[right, index] at (A-i) {$i$};%
    \node[left, index] at (A-j) {$i$};%
  }%
}%
$=$ \resizebox{!}{1.7ex}{%
  \tikz[baseline=-1.2ex,%
  index/.append style={fill=white, fill opacity = 1},%
  very thick,%
  ]{%
    \tensornode{A}%
    {$\va$}%
    {{i}}
    {{}}
    {{}}
    {{}};
    \draw[] (A-i) to ++(2ex,0) node[right, index] {$i$};%
    \draw[] (A-i) to ++(0,2.5ex) to ++(-7ex, 0) to ++(0,-2.5ex) to ++(-2ex,0) node[left, index] {$i$};%
  }%
}%
; and generalizes to larger diagonal blocks (\Cref{sec:app:visual-tour}).
In the following, we stick to the simplest case to avoid the more advanced syntax.
However, it shows the expressive power of TNs and is required to support common
features of convolutions like channel groups (known as separable convolutions).

\paragraph{Application to convolution:} We define a binary tensor $\tP \in \{0,
1\}^{I_1 \times O_1 \times K_1 \times I_2 \times O_2 \times K_2}$ which
represents the connectivity pattern between input, output, and kernel.
$\etP_{i_1, o_1, k_1, i_2, o_2, k_2}$ is 1 if input locations $(i_1, i_2)$
overlap with kernel positions $(k_1, k_2)$ when computing output locations
$(o_1, o_2)$ and 0 otherwise. The spatial couplings are independent along each
dimension, hence $\tP$ decomposes into $ \etP_{i_1, o_1, k_1, i_2, o_2, k_2} =
\smash{\etPi^{(1)}_{i_1, o_1, k_1}} \smash{\etPi^{(2)}_{i_2, o_2, k_2}}$ where the index
pattern tensor $\tPi^{(j)} \in \{0, 1\}^{I_j \times O_j \times K_j}$ encodes the
connectivity along dimension $j$. With that, one obtains
\begin{equation*}
  \etY_{c_{\text{out}}, o_1, o_2}
  =
  \evb_{c_{\text{out}}}
  +
  \sum_{c_{\text{in}}=1}^{C_{\text{in}}}
  \sum_{i_1,i_2=1}^{I_1, I_2}
  \sum_{k_1,k_2=1}^{K_1, K_2}
  \etX_{c_{\text{in}}, i_1, i_2 }
  \etPi^{(1)}_{i_1, o_1, k_1}
  \etPi^{(2)}_{i_2, o_2, k_2}
  \etW_{c_{\text{out}}, c_{\text{in}}, k_1, k_2}
\end{equation*}
Without bias, this translates into the diagram in \Cref{subfig:visual-abstract-convolution}.

\section{TNs for Convolution Operations}\label{sec:operations}
We now demonstrate the elegance of TNs for computing derivatives (\Cref{subsec:differentiation}), autodiff operations (\Cref{subsec:autodiff-conv-transpose}), and approximate second-order information (\Cref{subsec:kfac}) by graphical manipulation.
For simplicity, we exclude batching (\texttt{vmap}-ing like in JAX~\citep{bradbury2018jax}) and channel groups, and provide the diagrams with full support in \Cref{sec:app:visual-tour}.
\Cref{tab:einsum-expressions} summarizes our derivations (with batching and groups).
As a warm-up, we identify the unfolded input and kernel from the
matrix-multiplication view (\Cref{eq:convolution-with-unfolded-input,eq:convolution-with-unfolded-kernel}).
They follow by contracting the index patterns with either the input or kernel
(\Cref{subfig:visual-abstract-unfolded-input,subfig:visual-abstract-unfolded-kernel}),
\begin{align*}
  \textstyle
  \left[
  \llbracket \tX \rrbracket
  \right]_{(c_{\text{in}}, k_1, k_2), (o_1, o_1)}
  &=
    \sum_{i_1,i_2}
    \etX_{c_{\text{in}}, i_1, i_2}
    \etPi^{(1)}_{i_1, o_1, k_1}
    \etPi^{(2)}_{i_2, o_2, k_2}\,,
  \\
  \textstyle
  \left[
  \mA(\tW)
  \right]_{(c_{\text{out}}, o_1, o_2), (c_{\text{in}}, i_1, i_2)}
  &=
    \sum_{k_1, k_2}
    \etPi^{(1)}_{i_1, o_1, k_1}
    \etPi^{(2)}_{i_2, o_2, k_2}
    \etW_{c_{\text{out}}, c_{\text{in}}, k_1, k_2}\,.
\end{align*}

\subsection{Tensor Network Differentiation}\label{subsec:differentiation}
Derivatives play a crucial role in theoretical and practical ML.
First, we show that differentiating a TN diagram amounts to a simple graphical manipulation.
Then, we derive the Jacobians of convolution.
Consider an arbitrary TN represented by the tensor multiplication from \Cref{eq:n-ary-tensor-multiplication}.
The Jacobian tensor $\smash{[\tJ_{\tA_j}\tA_{N+1}]_{S_{N+1}, S'_j}} = \smash{\nicefrac{\partial [\tA_{N+1}]_{S_{N+1}}}{\partial [\tA_j]_{S_j'}}}$ w.r.t.\,an input $\tA_j$ collects all partial derivatives and is addressed through indices $S_{n+1} \times S'_j$ with $S_j'$ an independent copy of $S_j$.
Assume that $\tA_j$ only enters once in the tensor multiplication.
Then, taking the derivative of \Cref{eq:n-ary-tensor-multiplication} w.r.t.\,$[\tA_j]_{S_j'}$ simply replaces the tensor by a Kronecker delta $\delta_{S_j, S_j'}$,
\begin{align}
  \label{eq:tensor-network-derivative}
  \frac{\partial [\tA_{N+1}]_{S_{N+1}}}{
  \partial [\tA_j]_{S'_j}
  }
  =
  \sum_{\left( \bigcup_{n=1}^{N}\! S_n\right) \setminus S_{n+1}}
  [\tA_1]_{S_1}
  \cdots
  [\tA_{j-1}]_{S_{j-1}}
  \prod_{i \in S_j}
  \delta_{i, i'}
  [\tA_{j+1}]_{S_{j+1}}
  \cdots
  [\tA_N]_{S_N}
\end{align}
If an index $i \in S_j$ is summed, $i \not\in S_{n+1}$, we can sum the Kronecker delta $\delta_{i,i'}$, effectively replacing all occurrences of $i$ by $i'$.
If instead $i$ is part of the output index, $i \in S_{n+1}$, the Kronecker delta remains part of the Jacobian and imposes structure.
\Cref{subfig:example-tensor-network-derivative-delta} illustrates this process in diagrams for differentiating a convolution w.r.t.\,its kernel.
\Cref{eq:tensor-network-derivative} amounts to cutting out the argument of differentiation and assigning new indices to the resulting open legs.
For the weight Jacobian $\tJ_{\tW}\tY$, this introduces structure: If we re-interpret the two sub-diagrams in \Cref{subfig:example-tensor-network-derivative-delta} as matrices, compare with the Kronecker diagram from \Cref{eq:hadamard-and-kronecker} and use \Cref{subfig:visual-abstract-unfolded-input}, we find $\llbracket \tX \rrbracket^{\top} \otimes \mI_{C_{\text{out}}}$ for the Jacobian's matrix view~\citep[e.g.][]{dangel2020modular}.
\Cref{subfig:example-input-jacobian} shows the input Jacobian $\tJ_{\tX}\tY$ which is a tensor view of $\mA(\tW)$, as expected from the matrix-vector perspective of \Cref{eq:convolution-with-unfolded-kernel}.

Differentiating a TN is more convenient than using matrix calculus~\citep{magnus1999matrix} as it amounts to a simple graphical manipulation, does not rely on a flattening convention, and therefore preserves the full index structure.
The resulting TN can still be translated back to matrix language, if desired.
It also simplifies the computation of higher-order derivatives (e.g.\ $\nicefrac{\partial^2\tY}{\partial \tW \partial \tX}$), since differentiation yields another TN and can thus be repeated.
If a tensor occurs more than once in a TN, the product rule applies and the derivative is a sum of TNs with one occurrence removed.

\begin{figure}[t]
  \centering
  \begin{subfigure}[b]{0.3\linewidth}
    \centering
    \input{fig/tensor_networks/commands.tex}
    \resizebox{!}{18ex}{%
      \begin{tikzpicture}[ultra thick]%
        \matrix[row sep=3ex,column sep=0ex, ampersand replacement=\&]{%
          \& \P{1} \& \\
          \& \P{2} \& \\
          \X \& \& \begin{scope}[opacity=0, tensor/.append style={inner sep = 3ex}]\W\end{scope}
          \& \tensornode{I}%
          {$\mI$}%
          {{cout}}
          {{}}
          {{coutprime}}
          {{}}
          \\
        };%
        \draw %
        \contract{X}{P1}{i1}{$i_1$}{out=90, in=180}%
        \contract{X}{P2}{i2}{$i_2$}{out=90, in=180}%
        \contract{P1}{W}{k1}{$k_1$}{out=0, in=90}%
        \contract{P2}{W}{k2}{$k_2$}{out=0, in=90}%
        \contract{X}{W}{cin}{$c_{\text{in}}$}{}%
        (I-coutprime) node[index, left] {$c_{\text{out}}'$}%
        (P1-o1) node[index, above] {$o_1$}%
        (P2-o2) node[index, above] {$o_2$};%
        \draw %
        (W-cin) node [index, right] {$c_{\text{in}}'$}%
        (W-k2) node [index, below, xshift=-0.35ex] {$k_2'$}%
        (W-k1) node [index, below, xshift=0.35ex] {$k_1'$}%
        (I-cout) node [index, right] {$c_{\text{out}}$}%
        ;%
      \end{tikzpicture}
    }
    \vspace{-1ex}
    \caption{Differentiation/weight Jac.}\label{subfig:example-tensor-network-derivative-delta}
  \end{subfigure}
  \hspace{-0.75ex}
  \begin{subfigure}[b]{0.19\linewidth}
    \centering
    \resizebox{!}{17ex}{%
\input{fig/tensor_networks/commands.tex}
\begin{tikzpicture}[ultra thick]%
  \matrix[row sep=3ex,column sep=0ex]{
    \P{1} & \\
    \P{2} & \\
    &\W \\
  };
  \draw %
  \contract{P1}{W}{k1}{$k_1$}{out=0, in=90}%
  \contract{P2}{W}{k2}{$k_2$}{out=0, in=90}%
  (W-cin) node[index, left] {$c'_{\text{in}}$}%
  (W-cout) node[index, right] {$c_{\text{out}}$}%
  (P1-i1) node[index, left] {$i'_1$}%
  (P2-i2) node[index, left] {$i'_2$}%
  (P1-o1) node[index, above] {$o_1$}%
  (P2-o2) node[index, above] {$o_2$};%
\end{tikzpicture}
%
    }
    \caption{Input Jac.}\label{subfig:example-input-jacobian}
  \end{subfigure}
  \begin{subfigure}[b]{0.217\linewidth}
    \centering
    \resizebox*{!}{17ex}{%
\input{fig/tensor_networks/commands.tex}
\begin{tikzpicture}[ultra thick, opacity=0.3, index/.append style={fill opacity=1, text opacity=0.3}]%
  \matrix[row sep=3ex,column sep=0ex]{
    &\P{1} &\\
    &\P{2} & \\
    \X & & \begin{scope}[opacity=1, xshift=4ex] \V \end{scope} \\
  };%
  \draw %
  \contract{X}{P1}{i1}{$i_1$}{out=90, in=180}%
  \contract{X}{P2}{i2}{$i_2$}{out=90, in=180}%
  (X-cin) node[index, right] {$c'_{\text{in}}$}%
  (P1-k1) node[index, right] {$k'_1$}%
  (P2-k2) node[index, right] {$k'_2$};%
  \begin{scope}[opacity=1, index/.append style={text opacity=1}]
    \draw %
    (V-o2) |- node[index, xshift=-3.5ex] {$o_1$} (P1-o1)%
    (V-o1) |- node[index, xshift=-2ex] {$o_2$} (P2-o2)%
    (V-cout) node[index, right] {$c'_{\text{out}}$};%
  \end{scope}
\end{tikzpicture}
    }
    \caption{Weight VJP}
    \label{subfig:weight-vjp}
  \end{subfigure}
  \begin{subfigure}[b]{0.277\linewidth}
    \centering
    \resizebox*{!}{17ex}{%
\input{fig/tensor_networks/commands.tex}
\begin{tikzpicture}[ultra thick, opacity=0.3, index/.append style={fill opacity=1, text opacity=0.3}]%
  \matrix[row sep=3ex,column sep=0ex]{
    &\P{1} &\\
    &\P{2} & \\
    \begin{scope}[opacity=1]
      \V
    \end{scope}
    & & \W \\
  };
  \draw %
  \contract{P1}{W}{k1}{$k_1$}{out=0, in=90}%
  \contract{P2}{W}{k2}{$k_2$}{out=0, in=90}%
  (W-cin) node[index, left] {$c'_{\text{in}}$}%
  (P1-i1) node[index, left] {$i'_1$}%
  (P2-i2) node[index, left] {$i'_2$};%
  \begin{scope}[opacity=1, index/.append style={text opacity=1}]
    \draw %
    (V-o1) |- node[index, xshift=3.5ex] {$o_1$} (P1-o1)%
    (V-o2) |- node[index, xshift=2ex] {$o_2$} (P2-o2)%
    (V-cout) -- ++(0, 3ex) node[index, xshift=3ex] {$c_{\text{out}}$} -| (W-cout);%
  \end{scope}
\end{tikzpicture}
    }
    \caption{Input VJP/tr.\,conv.}
    \label{subfig:input-vjp}
  \end{subfigure}

  \caption{TN differentiation as graphical manipulation.
    (\subref{subfig:example-tensor-network-derivative-delta}) Differentiating convolution w.r.t.\,$\tW$ is cutting it out of the diagram and yields the weight Jacobian.
    (\subref{subfig:example-input-jacobian}) Same procedure applied to the Jacobian w.r.t.\,$\tX$.
    (\subref{subfig:weight-vjp}) VJP for the weight and (\subref{subfig:input-vjp}) input Jacobian (transpose convolution).
    Jacobians are shaded, only their contraction with $\smash{\tV^{(\tY)}}$ is highlighted.
  }\label{fig:example-tensor-network-derivative}
\end{figure}

\subsection{Autodiff \& Connections to Transpose Convolution}\label{subsec:autodiff-conv-transpose}
Although Jacobians are useful, crucial routines for autodiff are vector-Jacobian and Jacobian-vector products (VJPs, JVPs).
Both are simple to realize with TNs due to access to full Jacobians.
VJPs are used in backpropagation to pull back a tensor $\smash{\tV^{(\tY)}}\in \sR^{C_{\text{out}} \times O_1 \times O_2}$ from the output to the input or weight space.
The VJP results $\smash{\tV^{(\tX)}}\in \sR^{C_{\text{in}} \times I_1 \times I_2}$ and $\smash{\tV^{(\tW)}} \in \sR^{C_{\text{out}}\times C_{\text{in}} \times K_1 \times K_2}$ are
\begin{align*}
  \etV^{(\tX)}_{c'_{\text{in}}, i'_1, i'_2}
  =
  \sum_{c_{\text{out}}, o_1, o_2}
  \etV_{c_{\text{out}}, o_1, o_2}^{(\tY)}
  \frac{\partial \etY_{c_{\text{out}}, o_1, o_2}}{
  \partial \etX_{c'_{\text{in}}, i'_1, i'_2}
  }\,,\qquad
  \etV^{(\tW)}_{c'_{\text{out}}, c'_{\text{in}}, k'_1, k'_2}
  =
  \sum_{c_{\text{out}}, o_1, o_2}
  \etV_{c_{\text{out}}, o_1, o_2}^{(\tY)}
  \frac{\partial \etY_{c_{\text{out}}, o_1, o_2}}{
  \partial \etW_{c'_{\text{out}}, c'_{\text{in}}, k'_1, k'_2}
  }\,.
\end{align*}
Both are simply new TNs constructed from contracting the vector with the respective Jacobian, see \Cref{subfig:weight-vjp,subfig:input-vjp} (VJPs are analogous).
The input VJP is often used to define transpose convolution~\citep{dumoulin2016guide}.
In the matrix-multiplication perspective (\Cref{eq:convolution-with-unfolded-kernel}), this operation is defined relative to a convolution with kernel $\tW$ by multiplication with $\mA(\tW)^{\top}$, i.e.\,using the same connectivity pattern but mapping from the convolution's output to input space.
The TN in \Cref{subfig:input-vjp} makes this sharing explicit and cleanly defines transpose convolution.\footnote{Standalone implementations of transpose convolution require another parameter to unambiguously reconstruct the convolution's input dimension (see \Cref{sec:app:implementation-details} for how to compute $\tPi$ in this case).}

\subsection{Kronecker-factored Approximate Curvature}\label{subsec:kfac}
The Jacobian diagrams allow us to construct the TNs of second-order information like the Fisher/generalized Gauss-Newton (GGN) matrix and sub-tensors like its diagonal (\Cref{sec:app:second-order}).
Here, we focus on the popular Kronecker-factored approximation of the GGN~\citep{martens2015optimizing,grosse2016kroneckerfactored,eschenhagen2023kroneckerfactored,martens2018kroneckerfactored} whose input-based Kronecker factor relies on the unfolded input $\llbracket \tX \rrbracket$ which requires large memory.
State-of-the-art libraries that provide access to KFAC~\citep{dangel2020backpack,osawa2023asdl} also use this approach.
Using TNs, we can often avoid expanding $\llbracket \tX \rrbracket$ explicitly and save memory.
Here, we describe the existing KFAC approximations and their TNs (see \Cref{subsec:evaluation} for their run time evaluation).

\textbf{KFC (KFAC-expand):} \citet{grosse2016kroneckerfactored} introduce a
Kronecker approximation for the kernel's GGN, $\mG \approx \mOmega \otimes
\mGamma$ where $\mGamma \in \sR^{C_{\text{out}} \times C_{\text{out}}}$ and the
input-based factor $\mOmega = \llbracket \tX \rrbracket \llbracket \tX
\rrbracket^{\top} \in \sR^{C_{\text{in}} K_1 K_2 \times C_{\text{in}} K_1 K_2}$
(\Cref{subfig:kfc-factor}), the unfolded input's self-inner product (averaged
over a batch).

\textbf{KFAC-reduce:} \citet{eschenhagen2023kroneckerfactored} generalized KFAC to graph neural networks and transformers based on the concept of weight sharing, also present in convolutions.
They identify two approximations: KFAC-expand and KFAC-reduce.
The former corresponds to KFC~\citep{grosse2016kroneckerfactored}.
The latter shows similar performance in downstream tasks, but is cheaper to compute.
It relies on the column-averaged unfolded input, i.e.\,the average over all patches sharing the same weights.
KFAC-reduce approximates $\mG \approx \hat{\mOmega} \otimes \hat{\mGamma}$ with $\hat{\mGamma} \in \sR^{C_{\text{out}}\times C_{\text{out}}}$ and $\hat{\mOmega} = \nicefrac{1}{(O_1 O_2)^2} \vone_{O_1 O_2}^{\top} \llbracket \tX \rrbracket (\vone_{O_1 O_2}^{\top} \llbracket \tX \rrbracket)^{\top} \in \sR^{C_{\text{in}} K_1 K_2 \times C_{\text{in}} K_1 K_2}$ (\Cref{subfig:kfac-reduce-factor}; averaged over a batch).
For convolutions, this is arguably a `more natural' approximation as it becomes exact in certain limits~\cite{eschenhagen2023kroneckerfactored}, in contrast to the expand approximation.

\begin{figure}[t]
  \centering
  \begin{subfigure}[b]{0.308\linewidth}
    \centering
    \resizebox{\linewidth}{!}{%
\input{fig/tensor_networks/commands.tex}
\def\P#1#2{%
  \tensornode{P#1-#2}%
  {$\tPi^{(#1)}$}%
  {{k#1}}
  {{o#1}}
  {{i#1}}
  {{}}
}%
\def\X#1{%
  \tensornode{X#1}%
  {$\tX$}%
  {{cin}}
  {{i1, i2}}
  {{}}
  {{}}
}%
\begin{tikzpicture}[ultra thick, opacity=0.3, index/.append style={fill opacity=1, text opacity=0.3}]%
  \matrix[row sep=3ex,column sep=0ex]{
    & \P{1}{1} & & \P{1}{2} \\
    & \P{2}{1} & & \P{2}{2} \\
    \X{1} & & \begin{scope}[xshift=6ex] \X{2} \end{scope} & \\
  };%
  \draw %
  \contract{X1}{P1-1}{i1}{$i_1$}{out=90, in=180}%
  \contract{X1}{P2-1}{i2}{$i_2$}{out=90, in=180}%
  \contract{X2}{P2-2}{i2}{$i'_2$}{out=90, in=180};%
  \begin{scope}[index/.append style={yshift=-3.5ex}]
    \draw \contract{X2}{P1-2}{i1}{$i'_1$}{out=90, in=180};%
  \end{scope}
  \begin{scope}[opacity=1, index/.append style={text opacity=1}]
    \draw%
    \contract{P1-1}{P1-2}{o1}{$o_1$}{out=0, in=180}%
    \contract{P2-1}{P2-2}{o2}{$o_2$}{out=0, in=180}%
    (P1-1-k1) node[index, right] {$k_1$}%
    (P1-2-k1) node[index, right] {$k'_1$}%
    (P2-1-k2) node[index, right] {$k_2$}%
    (P2-2-k2) node[index, right] {$k'_2$}%
    (X1-cin) node[index, right] {$c_\text{in}$}%
    (X2-cin) node[index, right] {$c'_\text{in}$};%
  \end{scope}
\end{tikzpicture}
    }
    \caption{KFAC-expand}\label{subfig:kfc-factor}
  \end{subfigure}
  \hfill
  \begin{subfigure}[b]{0.253\linewidth}
    \centering
    \resizebox{\linewidth}{!}{%
\input{fig/tensor_networks/commands.tex}
\def\P#1#2{%
  \tensornode{P#1-#2}%
  {$\tPi^{(#1)}$}%
  {{k#1}}
  {{o#1}}
  {{i#1}}
  {{}}
}%
\def\X#1{%
  \tensornode{X#1}%
  {$\tX$}%
  {{cin}}
  {{i1, i2}}
  {{}}
  {{}}
}%
\def\I#1{%
  \tensornode{I#1}%
  {$\vone$}%
  {{}}
  {{}}
  {{}}
  {{i}}
}%
\begin{tikzpicture}[ultra thick, opacity=0.3, index/.append style={fill opacity=1, text opacity=0.3}]%
  \matrix[row sep=0.25ex,column sep=0ex]{
    & \begin{scope}[opacity=1]\I{1}\end{scope} & & \begin{scope}[opacity=1]\I{2}\end{scope} \\
    & \P{1}{1} & & \P{1}{2} \\
    & \begin{scope}[opacity=1]\I{3}\end{scope} & & \begin{scope}[opacity=1]\I{4}\end{scope} \\
    & \P{2}{1} & & \P{2}{2} \\
    \X{1} & & \begin{scope}[xshift=6ex] \X{2} \end{scope} & \\
  };%
  \draw %
  \contract{X1}{P1-1}{i1}{$i_1$}{out=90, in=180}%
  \contract{X1}{P2-1}{i2}{$i_2$}{out=90, in=180}%
  \contract{X2}{P2-2}{i2}{$i'_2$}{out=90, in=180};%
  \begin{scope}[index/.append style={yshift=-3.5ex}]
    \draw \contract{X2}{P1-2}{i1}{$i'_1$}{out=90, in=180};%
  \end{scope}
  \begin{scope}[opacity=1, index/.append style={text opacity=1}]
    \draw%
    (I1-i) to node [midway, index, right] {$o_1$} (P1-1-o1)%
    (I3-i) to node [midway, index, right] {$o_2$} (P2-1-o2)%
    (I2-i) to node [midway, index, right] {$o_1'$} (P1-2-o1)%
    (I4-i) to node [midway, index, right] {$o_2'$} (P2-2-o2)%
    (P1-1-k1) node[index, right] {$k_1$}%
    (P1-2-k1) node[index, right] {$k'_1$}%
    (P2-1-k2) node[index, right] {$k_2$}%
    (P2-2-k2) node[index, right] {$k'_2$}%
    (X1-cin) node[index, right] {$c_\text{in}$}%
    (X2-cin) node[index, right] {$c'_\text{in}$};%
  \end{scope}
\end{tikzpicture}
    }
    \caption{KFAC-reduce}\label{subfig:kfac-reduce-factor}
  \end{subfigure}
  \hfill
  \begin{minipage}[b]{0.4\linewidth}
    \caption{TNs of input-based Kronecker factors for KFAC approximations of the Fisher/GGN (no batching, no groups).
      The unfolded input is shaded, only additional contractions are highlighted.
      (\subref{subfig:kfc-factor}) $\mOmega$ (KFC/KFAC-expand) from \citet{grosse2016kroneckerfactored}, (\subref{subfig:kfac-reduce-factor}) $\hat{\mOmega}$ (KFAC-reduce) from~\citet{eschenhagen2023kroneckerfactored} (vectors of ones effectively amount to sums).}\label{fig:kfac}
  \end{minipage}
  \vspace{-2ex}
\end{figure}
\textbf{KFAC for transpose convolution:} Our approach enables us to derive KFAC for transpose convolutions.
We are not aware of previous works doing so.
This seems surprising because, similar to \Cref{subsec:convolution}, transpose convolution can be seen as matrix multiplication between the kernel and an unfolded input.
From this formulation we can immediately obtain KFAC through the weight sharing view of \citet{eschenhagen2023kroneckerfactored}.
The Kronecker factor requires unfolding the input similar to \texttt{im2col}, but for transpose convolutions.
This operation is currently not provided by ML libraries.
We can overcome this limitation, express the unfolding operation as TN, and---for the first time---establish KFAC (expand and reduce) for transpose convolutions (see \Cref{subsec:app:kfac-transpose-convolution} for details).

\section{TN Simplifications \& Implementation}\label{sec:implementation}
Many convolutions in real-world CNNs use structured connectivity patterns that allow for simplifications which we describe here along with implementation aspects.

\begin{figure}[t]
  \centering
  \begin{minipage}[b]{0.75\linewidth}
  \begin{subfigure}[b]{0.56\linewidth}
    \centering
    \resizebox{\linewidth}{!}{
      \input{fig/tensor_networks/commands.tex}
      \def\P#1#2{%
        \tensornode{P-#1}%
        {$\tPi#2$}%
        {{k}}
        {{o}}
        {{i}}
        {{}}
      }%
      \begin{tikzpicture}[ultra thick]
        \matrix [row sep=3ex,column sep=3ex,
        ampersand replacement=\&, 
        ]{%
          \P{1}{(I, K, \textcolor{maincolor}{S}, P, \textcolor{maincolor}{D})}
          \&
          \node {$\bm{=}$};
          \& \P{2}{(I, K, \textcolor{secondcolor}{D}, P, \textcolor{secondcolor}{S})}\\
        };
        \draw%
        (P-1-i) node [index, left] {$i$}%
        (P-1-o) node [index, above] {\textcolor{maincolor}{$o$}}%
        (P-1-k) node [index, right] {\textcolor{maincolor}{$k$}}%
        (P-2-i) node [index, left] {$i$}%
        (P-2-o) node [index, above] {\textcolor{secondcolor}{$k$}}%
        (P-2-k) node [index, right] {\textcolor{secondcolor}{$o$}}%
        ;%
      \end{tikzpicture}}
    \caption{Kernel-output swap}\label{subfig:tn-simplifications-swap}
  \end{subfigure}
  \hfill
  \begin{subfigure}[b]{0.41\linewidth}
    \centering
    \resizebox{\linewidth}{!}{
      \input{fig/tensor_networks/commands.tex}
      \def\P#1#2{%
        \tensornode{P-#1}%
        {$\tPi#2$}%
        {{k}}
        {{o}}
        {{i}}
        {{}}
      }%
      \begin{tikzpicture}[ultra thick]
        \matrix [row sep=3ex,column sep=3ex,
        ampersand replacement=\&, 
        ]{%
          \P{1}{(I, K, K, 0, 1)} \& \node {$\bm{=}$}; \&
          \begin{scope}[
            tensor/.append style={%
              fill opacity=0,%
              draw opacity=0,%
              minimum width=5ex,%
            }%
            ]
            \P{2}{}
          \end{scope}\\
        };
        \draw%
        (P-1-i) node [index, left] {$i$}%
        (P-1-o) node [index, above] {$o$}%
        (P-1-k) node [index, right] {$k$}%
        (P-2-i) node [index, left] {$i$}%
        (P-2-o) node [index, above] {$o$}%
        (P-2-k) node [index, right] {$k$}%
        ;%
        \coordinate (P-2-i-inner) at ($(P-2-i)+(1.3ex, 0)$);%
        \coordinate (P-2-o-inner) at ($(P-2-o)+(0, -1.3ex)$);%
        \coordinate (P-2-k-inner) at ($(P-2-k)+(-1.3ex, 0)$);%
        \draw[arrows={Triangle}-] (P-2-i-inner) to ++(0.01ex,0);%
        \draw [out=30, in=270] (P-2-i-inner) to (P-2-o-inner);%
        \draw [out=-30, in=180] (P-2-i-inner) to (P-2-k-inner);%
      \end{tikzpicture}}
    \caption{Dense convolution}\label{subfig:tn-simplifications-dense}
  \end{subfigure}
  \begin{subfigure}[t]{1.0\linewidth}
    \centering
    \resizebox{0.8\linewidth}{!}{
      \input{fig/tensor_networks/commands.tex}
      \def\P#1#2{%
        \tensornode{P-#1}%
        {$\tPi#2$}%
        {{k}}
        {{o}}
        {{i}}
        {{}}
      }%
      \begin{tikzpicture}[ultra thick]
        \matrix [row sep=3ex,column sep=3ex,
        ampersand replacement=\&, 
        ]{%
          \tensornode{V}{$\tV$}{{i}}{{}}{{}}{{}}
          \&
          \P{1}{(I, K, S > K, 0, 1)}
          \&
          \node {$\bm{=}$};
          \&
          \tensornode{Vsub}{$\tilde{\tV}$}{{i}}{{}}{{}}{{}}
          \&
          \P{2}{(\nicefrac{IK}{S}, K, K, 0, 1)}\\
        };
        \draw%
        \contract{V}{P-1}{i}{$i$}{}%
        \contract{Vsub}{P-2}{i}{$i'$}{}%
        (P-1-o) node [index, above] {$o$}%
        (P-1-k) node [index, right] {$k$}%
        (P-2-o) node [index, above] {$o$}%
        (P-2-k) node [index, right] {$k$}%
        ;%
      \end{tikzpicture}}
    \caption{Down-sampling convolution}\label{subfig:tn-simplifications-down-sampling}
  \end{subfigure}
  \end{minipage}
  \hfill
  \begin{minipage}[b]{0.2\linewidth}
    \caption{TN illustrations of index pattern simplifications and transformations.
      See \Cref{subsec:app-additional-properties} for the math formulation.}\label{fig:tn-simplifications}
    \vspace{-2ex}
  \end{minipage}
\end{figure}

\subsection{Index Pattern Structure \& Simplifications}\label{subsec:pattern-structure}
The index pattern $\tPi$ encodes the connectivity of a convolution and depends on its hyper-parameters.
Along one dimension, $\tPi = \tPi(I, K, S, P, D)$ with input size $I$, kernel size $K$, stride $S$, padding $P$, and dilation $D$.
We provide pseudo-code for computing $\tPi$ in \Cref{sec:app:implementation-details} which is easy to implement efficiently with standard functions from any numerical library (\Cref{alg:index-pattern-tensor}).
Its entries are
\begin{align}\label{equ:index-pattern-kronecker}
  \left[ \tPi(I, K, S, P, D) \right]_{i,o,k}
  &= \delta_{i, 1 + (k-1)D + (o-1)S - P}\,,
\end{align}
with $i = 1, \dots, I, o = 1, \dots, O, k = 1, \dots, K$
and output size $O(I, K, S, P, D) = 1 + \left\lfloor \nicefrac{(I + 2P - (K + (K - 1) (D - 1)))}{S} \right\rfloor$.
Since $\tPi$ is binary and has size linear in $I, O, K$, it is cheap to pre-compute and cache.
The index pattern's symmetries allow for re-wiring a TN.
For instance, the symmetry of $(k, D)$ and $(o, S)$ in \Cref{equ:index-pattern-kronecker} and $O(I,K,S,P,D)$ permits a \emph{kernel-output swap}, exchanging the role of kernel and output dimension (\Cref{subfig:tn-simplifications-swap}).
\citet{rochette2019efficient} used this to phrase the per-example gradient computation (\Cref{subfig:weight-vjp}) as convolution.

For many convolutions of real-world CNNs (see \Cref{sec:app:hyper-parameters} for a hyper-parameter study) the index pattern possesses structure that simplifies its contraction with other tensors into either smaller contractions or reshapes: \emph{Dense convolutions} use a shared kernel size and stride, and thus process non-overlapping adjacent tiles of the input.
Their index pattern's action can be expressed as a cheap reshape (\Cref{subfig:tn-simplifications-dense}).
Such convolutions are common in DenseNets~\citep{huang2017densely}, MobileNets~\citep{howard2017mobilenets,sandler2018mobilenetv2}, ResNets~\citep{he2016deep}, and ConvNeXts~\citep{liu2022convnet}.
InceptionV3~\citep{szegedy2016rethinking} has 2d \emph{mixed-dense convolutions} that are dense along one dimension.
\emph{Down-sampling convolutions} use a larger stride than kernel size, hence only process a sub-set of their input, and are used in ResNet18~\citep{he2016deep}, ResNext101~\citep{xie2017aggregated}, and WideResNet101~\citep{zagoruyko2016wide}.
Their pattern contracts with a tensor $\tV$ like that of a dense convolution with a sub-tensor $\tilde{\tV}$ (\Cref{subfig:tn-simplifications-down-sampling}).
\Cref{subsec:evaluation} shows that those simplifications accelerate computations.

\subsection{Practical Benefits of the TN Abstraction \& Limitations for Convolutions}\label{subsec:implementation-aspects}

\textbf{Contraction order optimization:} There exist various orders in which to carry out the summations in a TN and their performance can vary by orders of magnitude.
One extreme approach is to carry out all summations via nested for-loops.
This so-called Feynman path integral algorithm requires little memory, but many FLOPS since it does not re-cycle intermediate results.
The other extreme is sequential pair-wise contraction.
This builds up intermediate results and can greatly reduce FLOPS.
The schedule is represented by a binary tree, but the underlying search is in general at least \#P-hard~\citep{damm2002complexity}.
Fortunately, there exist heuristics to find high-quality contraction trees for TNs with hundreds of tensors~\citep{huang2021efficient,gray2021hyper,nvidia2023cuquantum}, implemented in packages like \texttt{opt\_einsum}~\citep{smith2018opteinsum}.

\textbf{Index slicing:} A common problem with high-quality schedules is that intermediates exceed memory.
Dynamic slicing~\citep{huang2021efficient} (e.g.\,\texttt{cotengra}~\citep{gray2021hyper}) is a simple method to decompose a contraction until it becomes feasible by breaking it up into smaller identical sub-tasks whose aggregation adds a small overhead.
This enables peak memory reduction and distribution.

\textbf{Sparsity:} $\tPi$ is sparse as only a small fraction of the input contributes to an output element.
For a convolution with stride $S<K$ and default parameters ($P=0, D=1)$, for fixed output and kernel indices $k,o$, there is exactly one non-zero entry in $[\tPi]_{:,o,k}$.
Hence $\mathtt{nnz}(\tPi) = OK$, which corresponds to a sparsity of $\nicefrac{1}{I}$.
Padding leads to more kernel elements that do not contribute to an output pixel, and therefore a sparser $\tPi$.
For down-sampling and dense convolutions, we showed how $\tPi$'s algebraic structure allows to simplify its contraction.
However, if that is not possible, $\tPi$ contains explicit zeros that add unnecessary FLOPS.
One way to circumvent this is to match a TN with that of an operation with efficient implementation (like \texttt{im2col}, (transpose) convolution) using transformations like the \emph{kernel-output swap} or by introducing identity tensors to complete a template, as done in \citet{rochette2019efficient,dangel2021unfoldnd} for per-sample gradients and \texttt{im2col}.

\textbf{Approximate contraction \& structured dropout:} TNs offer a principled approach for stochastic approximation via Monte-Carlo estimation to save memory and run time at the cost of accuracy.
The basic idea is best explained on a matrix product $\smash{\mC := \mA \mB = \sum_{n=1}^N \left[ \mA \right]_{:,n} \left[ \mB \right]_{n,:}}$ with $\smash{\mA \in \sR^{I\times N}, \mB \in \sR^{N,O}}$.
To approximate the sum, we introduce a distribution over $n$'s range, then use column-row-sampling~\citep[CRS, ][]{adelman2021faster} to form an unbiased Monte-Carlo approximation with sampled indices, which only requires the sub-matrices with active column-row pairs.
Bernoulli-CRS samples without replacement by assigning a Bernoulli random variable $\mathrm{Bernoulli}(\pi_n)$ with probability $\pi_n$ for column-row pair $n$ to be included in the contraction.
The Bernoulli estimator is $\tilde{\mC} := \sum_{n=1}^{N} \nicefrac{z_n}{\pi_n} \left[ \mA \right]_{n,:} \left[ \mB \right]_{n,:}$ with $z_n \sim \mathrm{Bernoulli}(\pi_n)$.
With a shared keep probability, $\pi_n := p$, this yields the unbiased estimator $\mC' = \nicefrac{1}{p} \sum_{n=1,\dots,N} \mA' \mB'$ where $\mA' = \mA \mK$ and $\mB' = \mK \mB$ with $\mK = \diag(z_1, \dots, z_N)$ are the sub-matrices of $\mA, \mB$ containing the active column-row pairs.
CRS applies to a single contraction.
For TNs with multiple sums, we can apply it individually, and also impose a distribution over the result indices, which computes a (scaled) sub-tensor.


\section{Experiments}\label{sec:experiments-and-evaluation}
\begin{figure}[t]
  \centering
  \begin{minipage}[b]{0.6\linewidth}
    \centering
    \pgfkeys{/pgfplots/zmystyle/.style={%
        boxplotbasicstyle,%
        height=0.58\linewidth,%
        ytick={3,1,0.3},%
        ymajorgrids,%
        minor ytick={10,9,8,7,6,5,4,3,2,0.9,0.8,0.7,0.6,0.5,0.4,0.3,0.2,0.1},%
        log ticks with fixed point,%
      }}
\begin{tikzpicture}

\definecolor{color0}{rgb}{0.12156862745098,0.466666666666667,0.705882352941177}
\definecolor{color1}{rgb}{0.0901960784313725,0.745098039215686,0.811764705882353}
\definecolor{color2}{rgb}{0.580392156862745,0.403921568627451,0.741176470588235}
\definecolor{color3}{rgb}{0.890196078431372,0.466666666666667,0.76078431372549}

\begin{axis}[
axis line style={white!80!black},
tick pos=left,
xmin=0.5, xmax=11.5,
xtick style={color=gray},
xtick={1.5,3.75,6,8.25,10.5},
xticklabels={Forward,Input VJP,Weight VJP,KFC,KFAC-red.},
ylabel={TN versus PT (logarithmic)},
ymin=0.137044871018717, ymax=8.97183280616081,
ymode=log,
zmystyle
]
\path [draw=black, ]
(axis cs:0.5,1)
--(axis cs:11.5,1);

\addplot [color0]
table {%
1 1.59927479757071
1 1.15219594102096
};
\addplot [color0]
table {%
1 3.19474902045781
1 5.12786927532713
};
\addplot [color0]
table {%
0.835 1.15219594102096
1.165 1.15219594102096
};
\addplot [color0]
table {%
0.835 5.12786927532713
1.165 5.12786927532713
};
\addplot [color0, mark=o, mark size=3, mark options={solid,fill opacity=0,draw=black}, only marks]
table {%
1 7.41882080972977
};
\addplot [color1]
table {%
2 1.05279922570089
2 0.732303578459289
};
\addplot [color1]
table {%
2 1.5262258918302
2 2.15893117557101
};
\addplot [color1]
table {%
1.835 0.732303578459289
2.165 0.732303578459289
};
\addplot [color1]
table {%
1.835 2.15893117557101
2.165 2.15893117557101
};
\addplot [color1, mark=o, mark size=3, mark options={solid,fill opacity=0,draw=black}, only marks]
table {%
2 2.32643190987456
};
\addplot [color0]
table {%
3.25 1.53578978507391
3.25 0.784287426635228
};
\addplot [color0]
table {%
3.25 2.3281565276142
3.25 3.50513351127121
};
\addplot [color0]
table {%
3.085 0.784287426635228
3.415 0.784287426635228
};
\addplot [color0]
table {%
3.085 3.50513351127121
3.415 3.50513351127121
};
\addplot [color0, mark=o, mark size=3, mark options={solid,fill opacity=0,draw=black}, only marks]
table {%
3.25 5.30929071184138
};
\addplot [color1]
table {%
4.25 0.954033703296296
4.25 0.752727745755602
};
\addplot [color1]
table {%
4.25 1.12765183745918
4.25 1.38167346569684
};
\addplot [color1]
table {%
4.085 0.752727745755602
4.415 0.752727745755602
};
\addplot [color1]
table {%
4.085 1.38167346569684
4.415 1.38167346569684
};
\addplot [color1, mark=o, mark size=3, mark options={solid,fill opacity=0,draw=black}, only marks]
table {%
4.25 0.68156981086126
4.25 0.659759465083098
4.25 0.591139716482069
4.25 1.40354960102706
4.25 1.45680566253413
};
\addplot [color0]
table {%
5.5 1.14711526075564
5.5 0.349067227706645
};
\addplot [color0]
table {%
5.5 3.0778358039215
5.5 4.96785152085861
};
\addplot [color0]
table {%
5.335 0.349067227706645
5.665 0.349067227706645
};
\addplot [color0]
table {%
5.335 4.96785152085861
5.665 4.96785152085861
};
\addplot [color0, mark=o, mark size=3, mark options={solid,fill opacity=0,draw=black}, only marks]
table {%
5.5 6.11390939877441
};
\addplot [color1]
table {%
6.5 0.681670231178204
6.5 0.364493528786432
};
\addplot [color1]
table {%
6.5 0.977493787964857
6.5 1.30057398954806
};
\addplot [color1]
table {%
6.335 0.364493528786432
6.665 0.364493528786432
};
\addplot [color1]
table {%
6.335 1.30057398954806
6.665 1.30057398954806
};
\addplot [color1, mark=o, mark size=3, mark options={solid,fill opacity=0,draw=black}, only marks]
table {%
6.5 0.165733032412545
};
\addplot [color0]
table {%
7.75 1.07229318439295
7.75 0.709932907038578
};
\addplot [color0]
table {%
7.75 1.80088130959883
7.75 2.25573036756196
};
\addplot [color0]
table {%
7.585 0.709932907038578
7.915 0.709932907038578
};
\addplot [color0]
table {%
7.585 2.25573036756196
7.915 2.25573036756196
};
\addplot [color0, mark=o, mark size=3, mark options={solid,fill opacity=0,draw=black}, only marks]
table {%
7.75 4.71179802358024
7.75 4.84636269104023
7.75 3.45449029271799
};
\addplot [color1]
table {%
8.75 0.625744324238036
8.75 0.25456683207541
};
\addplot [color1]
table {%
8.75 1.12657925302402
8.75 1.68495919058468
};
\addplot [color1]
table {%
8.585 0.25456683207541
8.915 0.25456683207541
};
\addplot [color1]
table {%
8.585 1.68495919058468
8.915 1.68495919058468
};
\addplot [color0]
table {%
10 0.905576938644048
10 0.779949708090248
};
\addplot [color0]
table {%
10 1.5705266802466
10 1.92305124729776
};
\addplot [color0]
table {%
9.835 0.779949708090248
10.165 0.779949708090248
};
\addplot [color0]
table {%
9.835 1.92305124729776
10.165 1.92305124729776
};
\addplot [color0, mark=o, mark size=3, mark options={solid,fill opacity=0,draw=black}, only marks]
table {%
10 6.12152196176341
10 3.69550472271328
};
\addplot [color1]
table {%
11 0.26768204873977
11 0.217261415937963
};
\addplot [color1]
table {%
11 0.53208823292818
11 0.912139373543072
};
\addplot [color1]
table {%
10.835 0.217261415937963
11.165 0.217261415937963
};
\addplot [color1]
table {%
10.835 0.912139373543072
11.165 0.912139373543072
};
\path [draw=color0, fill=color2]
(axis cs:0.67,1.59927479757071)
--(axis cs:1.33,1.59927479757071)
--(axis cs:1.33,3.19474902045781)
--(axis cs:0.67,3.19474902045781)
--(axis cs:0.67,1.59927479757071)
--cycle;
\path [draw=color1, fill=color3]
(axis cs:1.67,1.05279922570089)
--(axis cs:2.33,1.05279922570089)
--(axis cs:2.33,1.5262258918302)
--(axis cs:1.67,1.5262258918302)
--(axis cs:1.67,1.05279922570089)
--cycle;
\path [draw=color0, fill=color2]
(axis cs:2.92,1.53578978507391)
--(axis cs:3.58,1.53578978507391)
--(axis cs:3.58,2.3281565276142)
--(axis cs:2.92,2.3281565276142)
--(axis cs:2.92,1.53578978507391)
--cycle;
\path [draw=color1, fill=color3]
(axis cs:3.92,0.954033703296296)
--(axis cs:4.58,0.954033703296296)
--(axis cs:4.58,1.12765183745918)
--(axis cs:3.92,1.12765183745918)
--(axis cs:3.92,0.954033703296296)
--cycle;
\path [draw=color0, fill=color2]
(axis cs:5.17,1.14711526075564)
--(axis cs:5.83,1.14711526075564)
--(axis cs:5.83,3.0778358039215)
--(axis cs:5.17,3.0778358039215)
--(axis cs:5.17,1.14711526075564)
--cycle;
\path [draw=color1, fill=color3]
(axis cs:6.17,0.681670231178204)
--(axis cs:6.83,0.681670231178204)
--(axis cs:6.83,0.977493787964857)
--(axis cs:6.17,0.977493787964857)
--(axis cs:6.17,0.681670231178204)
--cycle;
\path [draw=color0, fill=color2]
(axis cs:7.42,1.07229318439295)
--(axis cs:8.08,1.07229318439295)
--(axis cs:8.08,1.80088130959883)
--(axis cs:7.42,1.80088130959883)
--(axis cs:7.42,1.07229318439295)
--cycle;
\path [draw=color1, fill=color3]
(axis cs:8.42,0.625744324238036)
--(axis cs:9.08,0.625744324238036)
--(axis cs:9.08,1.12657925302402)
--(axis cs:8.42,1.12657925302402)
--(axis cs:8.42,0.625744324238036)
--cycle;
\path [draw=color0, fill=color2]
(axis cs:9.67,0.905576938644048)
--(axis cs:10.33,0.905576938644048)
--(axis cs:10.33,1.5705266802466)
--(axis cs:9.67,1.5705266802466)
--(axis cs:9.67,0.905576938644048)
--cycle;
\path [draw=color1, fill=color3]
(axis cs:10.67,0.26768204873977)
--(axis cs:11.33,0.26768204873977)
--(axis cs:11.33,0.53208823292818)
--(axis cs:10.67,0.53208823292818)
--(axis cs:10.67,0.26768204873977)
--cycle;
\addplot [color0]
table {%
0.67 2.30653191880153
1.33 2.30653191880153
};
\addplot [color1]
table {%
1.67 1.21250271167177
2.33 1.21250271167177
};
\addplot [color0]
table {%
2.92 1.80523520444377
3.58 1.80523520444377
};
\addplot [color1]
table {%
3.92 1.04973041454523
4.58 1.04973041454523
};
\addplot [color0]
table {%
5.17 1.99632049216987
5.83 1.99632049216987
};
\addplot [color1]
table {%
6.17 0.874837699418893
6.83 0.874837699418893
};
\addplot [color0]
table {%
7.42 1.18634294428183
8.08 1.18634294428183
};
\addplot [color1]
table {%
8.42 0.730241212253938
9.08 0.730241212253938
};
\addplot [color0]
table {%
9.67 1.09266446799474
10.33 1.09266446799474
};
\addplot [color1]
table {%
10.67 0.33345179101999
11.33 0.33345179101999
};
\end{axis}

\end{tikzpicture}
  \end{minipage}
  \hfill
  \begin{minipage}[b]{0.385\linewidth}
    \caption{Run time ratios of TN (\textcolor{maincolor}{w}/\textcolor{secondcolor}{o} simplifications) vs.\,standard implementation for dense convolutions of 9 CNNs.
      With simplifications, convolution and input VJP achieve median ratios slightly above 1, and the TN implementation is faster for weight VJP, KFC \& KFAC-reduce.
      The code in \Cref{fig:visual-abstract-code} corresponds to default, \textcolor{maincolor}{TN}, and \textcolor{secondcolor}{simplified TN} KFC implementation.}\label{fig:benchmark-dense-cuda}
    \vspace{-1.5ex}
  \end{minipage}
    \vspace{-2ex}
\end{figure}

Here, we demonstrate computational benefits of TNs for less standard routines of second-order methods and showcase their flexibility to perform stochastic autodiff in novel ways.

\subsection{Run Time Evaluation}\label{subsec:evaluation}

We implement the presented TNs' contraction strings and operands\footnote{\texttt{einsum} does not yet support index un-grouping, so we must reshape manually before and after.}
in PyTorch~\citep{paszke2019pytorch}.
The simplifications from \Cref{sec:implementation} can be applied on top and yield a modified \texttt{einsum} expression.
To find a contraction schedule, we use \texttt{opt\_einsum}~\citep{smith2018opteinsum} with default settings.
We extract the unique convolutions of 9 architectures for ImageNet and smaller data sets, then compare some operations from \Cref{tab:einsum-expressions} with their standard implementation on an Nvidia Tesla T4 GPU (16 GB); see \Cref{sec:app:benchmark} for all details.
Due to space constraints, we highlight important insights here and provide references to the corresponding material in the appendix.
In general, the performance gap between standard and TN implementation decreases the less common an operation is (\Cref{fig:app:benchmark-overview}); from forward pass (inference \& training), to VJPs (training), to KFAC (training with a second-order method).
This is intuitive as more frequently used routines have been optimized more aggressively.

\textbf{Impact of simplifications:}
While general convolutions remain unaffected (\Cref{subfig:app:effect-simplifications-general}) when applying the transformations of \Cref{sec:implementation}, mixed dense, dense, and down-sampling convolutions consistently enjoy significant run time improvements (\Cref{subfig:app:effect-simplifications-mixed-dense,subfig:app:effect-simplifications-dense,subfig:app:effect-simplifications-down}).
As an example, we show the performance comparison for dense convolutions in \Cref{fig:benchmark-dense-cuda}: The performance ratio's median between TN and standard forward and input VJP is close to 1, that is both require almost the same time.
In the median, the TN even outperforms PyTorch's highly optimized weight VJP, also for down-sampling convolutions (\Cref{fig:app:weight-vjp}).
For KFC, the median performance ratios are well below 1 for dense, mixed dense \& sub-sampling convolutions (\Cref{fig:app:kfc-factor}).

\begin{figure}[t]
  \centering
  \begin{minipage}[b]{0.6\linewidth}
    \centering
    \pgfkeys{/pgfplots/zmystyle/.style={%
        mybasic,%
        width = \linewidth,
        height = 0.52\linewidth,
        every axis plot post/.append style={%
          mark size=1.5,
          mark options={fill opacity=0.5, solid, line width = 1pt}
        },
        execute at begin picture={\colorlet{color0}{maincolor}},
        ylabel={Aux.\,mem.\,TN [MiB]},
      },
    }
\begin{tikzpicture}

\definecolor{color0}{rgb}{0.580392156862745,0.403921568627451,0.741176470588235}

\begin{axis}[
axis line style={white!80!black},
log basis x={10},
tick pos=left,
xlabel={Auxiliary memory PT [MiB]},
xmin=0.47892170924971, xmax=4764.02233265664,
xmode=log,
ylabel={Auxiliary memory TN [MiB]},
ymin=0.513044481744355, ymax=1122.7248961758,
ymode=log,
zmystyle
]
\addplot [, color0, mark=*, mark size=3, mark options={solid}, only marks]
table {%
174.58984375 1
186.34765625 1
1 1
93.79296875 1
1 1
293.109375 1
286.58984375 1
71.0859375 1
142.7578125 1
1 1
70.7734375 1
1 1
1 80.2734375
73.02734375 1
758.5 86.65625
758.375 84.36328125
441.55859375 1
178.11328125 1
84.78515625 1
127.9140625 1
1 1
1 1
1 1
1 1
1 1
1 1
53.828125 1
575.671875 26.0625
431.74609375 192.00390625
647.7265625 70.3671875
161.6953125 55.66015625
215.421875 1
53.328125 1
107.484375 1
161.4921875 1
39.7421875 1
63.8125 1
293.03125 1
1151.0703125 576.0078125
575.125 256.00390625
575.15625 1
287.61328125 128.00390625
287.6328125 1
144.00390625 1
144.00390625 1
3135.16015625 1
1567.8046875 1
784.00390625 1
392.00390625 1
1 1
1 1
1 1
1 1
1 1
1 1
430.8671875 1
1 1
1 1
215.3203125 1
1 1
1 1
1 1
1 1
1 1
1 1
1 1
1 1
1 1
1 1
1 1
1 1
1 1
1 1
1 1
1 1
1 1
1 1
1 1
1 1
1 1
1 1
1 1
1 1
1 1
1 1
1 1
1 1
1 1
1 1
1 1
1 1
1 1
1 1
63.796875 1
1 1
1 1
1 1
71.4140625 1
1 1
1 1
1 1
1 1
1 1
1 1
1 1
1 1
1 1
1 1
1 1
1 1
1 1
1 1
1 1
126.91796875 1
126.7265625 1
1 1
1 1
1 1
1 1
1 1
1 1
1 1
1 1
1 1
1 1
1 1
1 1
1 1
1 1
19.2890625 1
1 1
1 1
};
\addplot [, black]
table {%
0.727743912511554 0.727743912511554
791.497946732539 791.497946732539
};
\end{axis}

\end{tikzpicture}
  \end{minipage}
  \hfill
  \begin{minipage}[b]{0.38\linewidth}
    \caption{Extra memory used by the standard versus our TN implementation (simplifications enabled) of KFAC-reduce.
      Each point represents a convolution from 9 CNNs, clipped below by 1\,MiB.
      TNs consistently use less memory than the standard implementation (one exception), and often no extra memory at all.
      We observe memory savings up to 3\,GiB.
    }\label{fig:benchmark-peakmem}
    \vspace{-1ex}
  \end{minipage}
  \vspace{-3ex}
\end{figure}

\textbf{KFAC-reduce:} For all convolution types, the TN implementation achieves its largest improvements for $\hat{\mOmega}$ and consistently outperforms the PyTorch implementation in the median when simplifications are enabled (\Cref{fig:app:kfac-reduce}).
The standard implementation unfolds the input, takes the row-average, then forms its outer product.
The TN does not need to expand $\llbracket \tX \rrbracket$ in memory and instead averages the index pattern tensors, which reduces peak memory and run time.
We observe performance ratios down to 0.22\,x (speed-ups up to $\approx4.5$\,x, \Cref{tab:app:kfac-reduce-factor-gpu}) and consistently lower memory consumption with savings up to 3\,GiB (\Cref{fig:benchmark-peakmem}).
Hence, our approach not only significantly reduces the overhead of 2nd-order optimizers based on KFAC-reduce, but also allows them to operate on larger batches without exceeding memory (\citet{eschenhagen2023kroneckerfactored} specifically mention memory as important limitation of their method).
Other examples for KFAC algorithms where computing the input-based Kronecker factor adds significant time and memory overhead are that of \citet{petersen2023isaac,benzing2022gradient} which only use $\mOmega$ (setting $\mGamma \propto \mI$), or \citet{lin2024structured,lin2023simplifying} which remove matrix inversion.

\textbf{Downstream improvements with KFAC-reduce:} To demonstrate the speed-ups of KFAC-reduce in practical algorithms, we apply our work to the SINGD optimizer~\cite{lin2024structured} and benchmark the impact of our TN implementation on its memory and run time in comparison to SGD without momentum.
Concretely, we investigate SINGD with KFAC-reduce and diagonal pre-conditioners on ResNet18 and VGG19 on ImageNet-like synthetic data $(3, 256, 256)$ using a batch size of 128.
We measured per-iteration time and peak memory on an NVIDIA A40 with 48 GiB of RAM.
For SINGD, we compare computing the Kronecker factors with the standard approach (`SINGD') via input unfolding versus our TN implementation (`SINGD+TN').
\Cref{tab:singd-performance} summarizes the results.

\begin{wraptable}[11]{r}{0.54\textwidth}
  \vspace*{-2ex}
  \begin{small}
    \caption{Impact of our TN implementation on SINGD's run time and peak memory compared to SGD.}\label{tab:singd-performance}
    \begin{tabular}{llcc}
      \toprule
      & \textbf{Optimizer} & \textbf{Per iter.\,[s]} & \textbf{Peak mem.\,[GiB]}
      \\
      \midrule
      \parbox[t]{2mm}{\multirow{3}{*}{\rotatebox[origin=c]{90}{\tiny ResNet18}}}
      &
        SGD & $0.12$ ($1.0$\,x) & $3.6$ ($1.0$\,x)
      \\
      & SINGD & $0.19$ ($1.7$\,x) & $4.5$ ($1.3$\,x)
      \\
      & SINGD+TN & $0.16$ ($1.3$\,x) & $3.6$ ($1.0$\,x)
      \\
      \midrule
      \parbox[t]{2mm}{\multirow{3}{*}{\rotatebox[origin=c]{90}{\tiny VGG19}}}
      & SGD & $0.69$ ($1.0$\,x) & $14$ ($1.0$\,x)
      \\
      & SINGD & $1.0$ ($1.5$\,x) & $32$ ($2.3$\,x)
      \\
      & SINGD+TN & $0.80$ ($1.2$\,x) & $16$ ($1.1$\,x)
      \\
      \bottomrule
    \end{tabular}
  \end{small}
  \vspace{-1ex}
\end{wraptable}

On both nets, our TN implementation halves SINGD's run time, and almost completely eliminates the memory, overhead compared to SGD.
On VGG19, it dramatically lowers the memory overhead, cutting it down by a factor of 2 from 32 GiB to 16 GiB.
This enables using larger batches or more frequently updating the pre-conditioner, underlining the utility of our approach for reducing the computational gap between approximate second-order and first-order methods.

\subsection{Randomized Autodiff via Approximate Contraction}

CRS is an alternative to checkpointing~\citep{griewank2008evaluating} to lower memory consumption of backpropagation~\citep{oktay2021randomized,chen2023dropit,adelman2021faster}.
Here, we focus on unbiased gradient approximations by applying the exact forward pass, but CRS when computing the weight VJP, which requires storing a sub-tensor of $\tX$.
For convolutions, the approaches of existing works are limited by the supported functionality of ML libraries.
\citet{adelman2021faster} restrict to sampling $\tX$ along $c_{\text{in}}$, which eliminates many gradient entries as the index is part of the gradient.
The randomized gradient would thus only train a sub-tensor of the kernel per step.
\citet{oktay2021randomized,chen2023dropit} apply unstructured dropout to $\tX$, store it in sparse form, and restore the sparsified tensor during the backward pass.
This reduces memory, but not computation.

Our TN implementation is more flexible and can, for example, tackle spatial dimensions with CRS.
This reduces memory to the same extent, but also run time due to fewer contractions.
Importantly, it does not zero out the gradient for entire filters.
In \Cref{fig:subsampling-weight-gradient} we compare the gradient approximation errors of channel and spatial sub-sampling.
For the same memory reduction, spatial sub-sampling yields a smaller approximation error on both real \& synthetic data.
E.g., instead of keeping 75\,\% of channels, we achieve the same approximation quality using only 35\,\% of pixels.

\begin{figure}[t]
  \centering
  \pgfkeys{/pgfplots/subsamplingbasicstyle/.style={
      mybasic,
      width = 1.1\linewidth,
      height = 0.6\linewidth,
      every axis plot post/.append style={
        mark size=1.5, mark options={solid, line width = 1pt}
      },
      execute at begin picture={\colorlet{color0}{maincolor}},
      execute at begin picture={\colorlet{color1}{secondcolor}},
    }
  }
  \centering
  \begin{subfigure}[t]{0.495\linewidth}
    \centering
    \caption{\scriptsize Real-world data}\label{subfig:subsampling-real-world}
    \pgfkeys{/pgfplots/zmystyle/.style={
        subsamplingbasicstyle,
      }}
\begin{tikzpicture}

\definecolor{color0}{rgb}{0.996078431372549,0.682352941176471,0.203921568627451}
\definecolor{color1}{rgb}{0.0980392156862745,0.235294117647059,0.243137254901961}

\begin{axis}[
axis line style={white!80!black},
legend style={fill opacity=0.8, draw opacity=1, text opacity=1, draw=white!80!black},
tick pos=left,
xlabel={Keep probability \(\displaystyle p\)},
xmin=0, xmax=1,
ylabel={Normalized error},
ymin=0, ymax=4.88018434442684,
zmystyle
]
\path [draw=color0, ]
(axis cs:0.050000000745058,3.58651846738)
--(axis cs:0.050000000745058,4.65018359331947);

\path [draw=color0, ]
(axis cs:0.0999999940395355,2.86256301345619)
--(axis cs:0.0999999940395355,3.37847468910424);

\path [draw=color0, ]
(axis cs:0.149999991059303,2.25252276580237)
--(axis cs:0.149999991059303,2.6506693728695);

\path [draw=color0, ]
(axis cs:0.199999988079071,1.92402518422137)
--(axis cs:0.199999988079071,2.15387045710553);

\path [draw=color0, ]
(axis cs:0.249999985098839,1.68747106304959)
--(axis cs:0.249999985098839,1.83058072814151);

\path [draw=color0, ]
(axis cs:0.299999982118607,1.49146446022617)
--(axis cs:0.299999982118607,1.58115025725735);

\path [draw=color0, ]
(axis cs:0.349999994039536,1.33622152002603)
--(axis cs:0.349999994039536,1.40232697335929);

\path [draw=color0, ]
(axis cs:0.399999976158142,1.2079648574451)
--(axis cs:0.399999976158142,1.2511810699841);

\path [draw=color0, ]
(axis cs:0.449999988079071,1.09869736895343)
--(axis cs:0.449999988079071,1.11702013745526);

\path [draw=color0, ]
(axis cs:0.5,1)
--(axis cs:0.5,1);

\path [draw=color0, ]
(axis cs:0.549999952316284,0.89516957493517)
--(axis cs:0.549999952316284,0.912612383644847);

\path [draw=color0, ]
(axis cs:0.599999964237213,0.798478264822072)
--(axis cs:0.599999964237213,0.833034436689311);

\path [draw=color0, ]
(axis cs:0.649999976158142,0.714609573736077)
--(axis cs:0.649999976158142,0.750098981008644);

\path [draw=color0, ]
(axis cs:0.699999988079071,0.626027416605179)
--(axis cs:0.699999988079071,0.668984902482803);

\path [draw=color0, ]
(axis cs:0.75,0.535999101620377)
--(axis cs:0.75,0.588850289363204);

\path [draw=color0, ]
(axis cs:0.799999952316284,0.464649180956827)
--(axis cs:0.799999952316284,0.515774633101477);

\path [draw=color0, ]
(axis cs:0.850000023841858,0.389773751723362)
--(axis cs:0.850000023841858,0.440396915448116);

\path [draw=color0, ]
(axis cs:0.899999976158142,0.307511234013611)
--(axis cs:0.899999976158142,0.354858899386352);

\path [draw=color0, ]
(axis cs:0.949999988079071,0.2058993697019)
--(axis cs:0.949999988079071,0.26416422428175);

\path [draw=color1, ]
(axis cs:0.050000000745058,0.768965090971163)
--(axis cs:0.050000000745058,1.55052914454443);

\path [draw=color1, ]
(axis cs:0.0999999940395355,0.656097389050921)
--(axis cs:0.0999999940395355,1.12364324252943);

\path [draw=color1, ]
(axis cs:0.149999991059303,0.534028571205753)
--(axis cs:0.149999991059303,1.00036701623855);

\path [draw=color1, ]
(axis cs:0.199999988079071,0.518598729715305)
--(axis cs:0.199999988079071,0.821585613145871);

\path [draw=color1, ]
(axis cs:0.249999985098839,0.442249964414203)
--(axis cs:0.249999985098839,0.812076206268704);

\path [draw=color1, ]
(axis cs:0.299999982118607,0.433659843410286)
--(axis cs:0.299999982118607,0.812718518768516);

\path [draw=color1, ]
(axis cs:0.349999994039536,0.395639252964793)
--(axis cs:0.349999994039536,0.68051302998847);

\path [draw=color1, ]
(axis cs:0.399999976158142,0.343660038397195)
--(axis cs:0.399999976158142,0.60392872722399);

\path [draw=color1, ]
(axis cs:0.449999988079071,0.405321865983605)
--(axis cs:0.449999988079071,0.561117542080283);

\path [draw=color1, ]
(axis cs:0.5,0.368737011119123)
--(axis cs:0.5,0.517751521900896);

\path [draw=color1, ]
(axis cs:0.549999952316284,0.319010071275845)
--(axis cs:0.549999952316284,0.455757458688602);

\path [draw=color1, ]
(axis cs:0.599999964237213,0.264801696654718)
--(axis cs:0.599999964237213,0.447526942971785);

\path [draw=color1, ]
(axis cs:0.649999976158142,0.252632166686791)
--(axis cs:0.649999976158142,0.401773266252738);

\path [draw=color1, ]
(axis cs:0.699999988079071,0.19114786370811)
--(axis cs:0.699999988079071,0.344590068020253);

\path [draw=color1, ]
(axis cs:0.75,0.164579556531505)
--(axis cs:0.75,0.305713706069394);

\path [draw=color1, ]
(axis cs:0.799999952316284,0.146847491815946)
--(axis cs:0.799999952316284,0.269254410907367);

\path [draw=color1, ]
(axis cs:0.850000023841858,0.113658621952258)
--(axis cs:0.850000023841858,0.241512499704636);

\path [draw=color1, ]
(axis cs:0.899999976158142,0.0920762515452076)
--(axis cs:0.899999976158142,0.184982436260922);

\path [draw=color1, ]
(axis cs:0.949999988079071,0.0501685711720384)
--(axis cs:0.949999988079071,0.141363967659195);

\addplot [, color0, mark=-, mark size=3, mark options={solid}, only marks, forget plot]
table {%
0.050000000745058 3.58651846738
0.0999999940395355 2.86256301345619
0.149999991059303 2.25252276580237
0.199999988079071 1.92402518422137
0.249999985098839 1.68747106304959
0.299999982118607 1.49146446022617
0.349999994039536 1.33622152002603
0.399999976158142 1.2079648574451
0.449999988079071 1.09869736895343
0.5 1
0.549999952316284 0.89516957493517
0.599999964237213 0.798478264822072
0.649999976158142 0.714609573736077
0.699999988079071 0.626027416605179
0.75 0.535999101620377
0.799999952316284 0.464649180956827
0.850000023841858 0.389773751723362
0.899999976158142 0.307511234013611
0.949999988079071 0.2058993697019
};
\addplot [, color0, mark=-, mark size=3, mark options={solid}, only marks, forget plot]
table {%
0.050000000745058 4.65018359331947
0.0999999940395355 3.37847468910424
0.149999991059303 2.6506693728695
0.199999988079071 2.15387045710553
0.249999985098839 1.83058072814151
0.299999982118607 1.58115025725735
0.349999994039536 1.40232697335929
0.399999976158142 1.2511810699841
0.449999988079071 1.11702013745526
0.5 1
0.549999952316284 0.912612383644847
0.599999964237213 0.833034436689311
0.649999976158142 0.750098981008644
0.699999988079071 0.668984902482803
0.75 0.588850289363204
0.799999952316284 0.515774633101477
0.850000023841858 0.440396915448116
0.899999976158142 0.354858899386352
0.949999988079071 0.26416422428175
};
\addplot [, color1, mark=-, mark size=3, mark options={solid}, only marks, forget plot]
table {%
0.050000000745058 0.768965090971163
0.0999999940395355 0.656097389050921
0.149999991059303 0.534028571205753
0.199999988079071 0.518598729715305
0.249999985098839 0.442249964414203
0.299999982118607 0.433659843410286
0.349999994039536 0.395639252964793
0.399999976158142 0.343660038397195
0.449999988079071 0.405321865983605
0.5 0.368737011119123
0.549999952316284 0.319010071275845
0.599999964237213 0.264801696654718
0.649999976158142 0.252632166686791
0.699999988079071 0.19114786370811
0.75 0.164579556531505
0.799999952316284 0.146847491815946
0.850000023841858 0.113658621952258
0.899999976158142 0.0920762515452076
0.949999988079071 0.0501685711720384
};
\addplot [, color1, mark=-, mark size=3, mark options={solid}, only marks, forget plot]
table {%
0.050000000745058 1.55052914454443
0.0999999940395355 1.12364324252943
0.149999991059303 1.00036701623855
0.199999988079071 0.821585613145871
0.249999985098839 0.812076206268704
0.299999982118607 0.812718518768516
0.349999994039536 0.68051302998847
0.399999976158142 0.60392872722399
0.449999988079071 0.561117542080283
0.5 0.517751521900896
0.549999952316284 0.455757458688602
0.599999964237213 0.447526942971785
0.649999976158142 0.401773266252738
0.699999988079071 0.344590068020253
0.75 0.305713706069394
0.799999952316284 0.269254410907367
0.850000023841858 0.241512499704636
0.899999976158142 0.184982436260922
0.949999988079071 0.141363967659195
};
\addplot [, color0, mark=*, mark size=2.5, mark options={solid}, only marks]
table {%
0.050000000745058 4.11835103034973
0.0999999940395355 3.12051885128021
0.149999991059303 2.45159606933594
0.199999988079071 2.03894782066345
0.249999985098839 1.75902589559555
0.299999982118607 1.53630735874176
0.349999994039536 1.36927424669266
0.399999976158142 1.2295729637146
0.449999988079071 1.10785875320435
0.5 1
0.549999952316284 0.903890979290008
0.599999964237213 0.815756350755692
0.649999976158142 0.73235427737236
0.699999988079071 0.647506159543991
0.75 0.562424695491791
0.799999952316284 0.490211907029152
0.850000023841858 0.415085333585739
0.899999976158142 0.331185066699982
0.949999988079071 0.235031796991825
};
\addlegendentry{channel}
\addplot [, color1, mark=*, mark size=2.5, mark options={solid}, only marks]
table {%
0.050000000745058 1.1597471177578
0.0999999940395355 0.889870315790176
0.149999991059303 0.767197793722153
0.199999988079071 0.670092171430588
0.249999985098839 0.627163085341454
0.299999982118607 0.623189181089401
0.349999994039536 0.538076141476631
0.399999976158142 0.473794382810593
0.449999988079071 0.483219704031944
0.5 0.44324426651001
0.549999952316284 0.387383764982223
0.599999964237213 0.356164319813251
0.649999976158142 0.327202716469765
0.699999988079071 0.267868965864182
0.75 0.235146631300449
0.799999952316284 0.208050951361656
0.850000023841858 0.177585560828447
0.899999976158142 0.138529343903065
0.949999988079071 0.0957662694156169
};
\addlegendentry{spatial}
\end{axis}

\end{tikzpicture}
    \vspace{-6ex}
  \end{subfigure}
  \hspace{-3.5ex}
  \begin{subfigure}[t]{0.495\linewidth}
    \centering
    \caption{\scriptsize Synthetic data}\label{subfig:subsampling-synthetic}
    \pgfkeys{/pgfplots/zmystyle/.style={
        subsamplingbasicstyle,
        every axis y label/.append style = {opacity=0},
        every y tick label/.append style = {opacity=0},
        legend style = {opacity=0, text opacity=0},
      }}
\begin{tikzpicture}

\definecolor{color0}{rgb}{0.996078431372549,0.682352941176471,0.203921568627451}
\definecolor{color1}{rgb}{0.0980392156862745,0.235294117647059,0.243137254901961}

\begin{axis}[
axis line style={white!80!black},
legend style={fill opacity=0.8, draw opacity=1, text opacity=1, draw=white!80!black},
tick pos=left,
xlabel={Keep probability \(\displaystyle p\)},
xmin=0, xmax=1,
ylabel={Normalized error},
ymin=0, ymax=5.13488728628373,
zmystyle
]
\path [draw=color0, ]
(axis cs:0.050000000745058,4.04307372215691)
--(axis cs:0.050000000745058,4.89257982131538);

\path [draw=color0, ]
(axis cs:0.0999999940395355,2.85674871569446)
--(axis cs:0.0999999940395355,3.33092189664075);

\path [draw=color0, ]
(axis cs:0.149999991059303,2.24280330610786)
--(axis cs:0.149999991059303,2.56391308831658);

\path [draw=color0, ]
(axis cs:0.199999988079071,1.92214877662392)
--(axis cs:0.199999988079071,2.09589805546074);

\path [draw=color0, ]
(axis cs:0.249999985098839,1.6711563137581)
--(axis cs:0.249999985098839,1.7960660191963);

\path [draw=color0, ]
(axis cs:0.299999982118607,1.47739532018515)
--(axis cs:0.299999982118607,1.55491883730081);

\path [draw=color0, ]
(axis cs:0.349999994039536,1.33895018248685)
--(axis cs:0.349999994039536,1.38057365269534);

\path [draw=color0, ]
(axis cs:0.399999976158142,1.20966102751921)
--(axis cs:0.399999976158142,1.23146486607362);

\path [draw=color0, ]
(axis cs:0.449999988079071,1.09769042016473)
--(axis cs:0.449999988079071,1.10801455496344);

\path [draw=color0, ]
(axis cs:0.5,1)
--(axis cs:0.5,1);

\path [draw=color0, ]
(axis cs:0.549999952316284,0.902898760335926)
--(axis cs:0.549999952316284,0.91238620708656);

\path [draw=color0, ]
(axis cs:0.599999964237213,0.813277024186429)
--(axis cs:0.599999964237213,0.828071933829013);

\path [draw=color0, ]
(axis cs:0.649999976158142,0.727434529244846)
--(axis cs:0.649999976158142,0.751496563971096);

\path [draw=color0, ]
(axis cs:0.699999988079071,0.650911524086422)
--(axis cs:0.699999988079071,0.676313624591404);

\path [draw=color0, ]
(axis cs:0.75,0.564345697489952)
--(axis cs:0.75,0.609040816696907);

\path [draw=color0, ]
(axis cs:0.799999952316284,0.491523905203039)
--(axis cs:0.799999952316284,0.536037890508479);

\path [draw=color0, ]
(axis cs:0.850000023841858,0.403589421061557)
--(axis cs:0.850000023841858,0.460894585104901);

\path [draw=color0, ]
(axis cs:0.899999976158142,0.310557628123188)
--(axis cs:0.899999976158142,0.369883554728603);

\path [draw=color0, ]
(axis cs:0.949999988079071,0.198151126132951)
--(axis cs:0.949999988079071,0.252212033046736);

\path [draw=color1, ]
(axis cs:0.050000000745058,0.837873383517057)
--(axis cs:0.050000000745058,1.59528353596231);

\path [draw=color1, ]
(axis cs:0.0999999940395355,0.682797340047721)
--(axis cs:0.0999999940395355,1.29936759124958);

\path [draw=color1, ]
(axis cs:0.149999991059303,0.595391050825933)
--(axis cs:0.149999991059303,1.25234415148654);

\path [draw=color1, ]
(axis cs:0.199999988079071,0.47238134165208)
--(axis cs:0.199999988079071,1.02663318495353);

\path [draw=color1, ]
(axis cs:0.249999985098839,0.328524551833082)
--(axis cs:0.249999985098839,0.934149255311083);

\path [draw=color1, ]
(axis cs:0.299999982118607,0.290645279770799)
--(axis cs:0.299999982118607,0.701484463328414);

\path [draw=color1, ]
(axis cs:0.349999994039536,0.346982405205426)
--(axis cs:0.349999994039536,0.60677712796146);

\path [draw=color1, ]
(axis cs:0.399999976158142,0.320394226941735)
--(axis cs:0.399999976158142,0.548545226421683);

\path [draw=color1, ]
(axis cs:0.449999988079071,0.273881319400274)
--(axis cs:0.449999988079071,0.493048253420389);

\path [draw=color1, ]
(axis cs:0.5,0.31311821369486)
--(axis cs:0.5,0.455820786673258);

\path [draw=color1, ]
(axis cs:0.549999952316284,0.302288832815099)
--(axis cs:0.549999952316284,0.414623037426066);

\path [draw=color1, ]
(axis cs:0.599999964237213,0.262241972213406)
--(axis cs:0.599999964237213,0.354419892305713);

\path [draw=color1, ]
(axis cs:0.649999976158142,0.230061026659044)
--(axis cs:0.649999976158142,0.339946807179418);

\path [draw=color1, ]
(axis cs:0.699999988079071,0.194711652830628)
--(axis cs:0.699999988079071,0.297197657391044);

\path [draw=color1, ]
(axis cs:0.75,0.173600193190912)
--(axis cs:0.75,0.262363020252844);

\path [draw=color1, ]
(axis cs:0.799999952316284,0.145844272733237)
--(axis cs:0.799999952316284,0.252893455862497);

\path [draw=color1, ]
(axis cs:0.850000023841858,0.128764308852071)
--(axis cs:0.850000023841858,0.223722563820964);

\path [draw=color1, ]
(axis cs:0.899999976158142,0.105685739198523)
--(axis cs:0.899999976158142,0.172807960232896);

\path [draw=color1, ]
(axis cs:0.949999988079071,0.046430521948359)
--(axis cs:0.949999988079071,0.115094630675294);

\addplot [, color0, mark=-, mark size=3, mark options={solid}, only marks, forget plot]
table {%
0.050000000745058 4.04307372215691
0.0999999940395355 2.85674871569446
0.149999991059303 2.24280330610786
0.199999988079071 1.92214877662392
0.249999985098839 1.6711563137581
0.299999982118607 1.47739532018515
0.349999994039536 1.33895018248685
0.399999976158142 1.20966102751921
0.449999988079071 1.09769042016473
0.5 1
0.549999952316284 0.902898760335926
0.599999964237213 0.813277024186429
0.649999976158142 0.727434529244846
0.699999988079071 0.650911524086422
0.75 0.564345697489952
0.799999952316284 0.491523905203039
0.850000023841858 0.403589421061557
0.899999976158142 0.310557628123188
0.949999988079071 0.198151126132951
};
\addplot [, color0, mark=-, mark size=3, mark options={solid}, only marks, forget plot]
table {%
0.050000000745058 4.89257982131538
0.0999999940395355 3.33092189664075
0.149999991059303 2.56391308831658
0.199999988079071 2.09589805546074
0.249999985098839 1.7960660191963
0.299999982118607 1.55491883730081
0.349999994039536 1.38057365269534
0.399999976158142 1.23146486607362
0.449999988079071 1.10801455496344
0.5 1
0.549999952316284 0.91238620708656
0.599999964237213 0.828071933829013
0.649999976158142 0.751496563971096
0.699999988079071 0.676313624591404
0.75 0.609040816696907
0.799999952316284 0.536037890508479
0.850000023841858 0.460894585104901
0.899999976158142 0.369883554728603
0.949999988079071 0.252212033046736
};
\addplot [, color1, mark=-, mark size=3, mark options={solid}, only marks, forget plot]
table {%
0.050000000745058 0.837873383517057
0.0999999940395355 0.682797340047721
0.149999991059303 0.595391050825933
0.199999988079071 0.47238134165208
0.249999985098839 0.328524551833082
0.299999982118607 0.290645279770799
0.349999994039536 0.346982405205426
0.399999976158142 0.320394226941735
0.449999988079071 0.273881319400274
0.5 0.31311821369486
0.549999952316284 0.302288832815099
0.599999964237213 0.262241972213406
0.649999976158142 0.230061026659044
0.699999988079071 0.194711652830628
0.75 0.173600193190912
0.799999952316284 0.145844272733237
0.850000023841858 0.128764308852071
0.899999976158142 0.105685739198523
0.949999988079071 0.046430521948359
};
\addplot [, color1, mark=-, mark size=3, mark options={solid}, only marks, forget plot]
table {%
0.050000000745058 1.59528353596231
0.0999999940395355 1.29936759124958
0.149999991059303 1.25234415148654
0.199999988079071 1.02663318495353
0.249999985098839 0.934149255311083
0.299999982118607 0.701484463328414
0.349999994039536 0.60677712796146
0.399999976158142 0.548545226421683
0.449999988079071 0.493048253420389
0.5 0.455820786673258
0.549999952316284 0.414623037426066
0.599999964237213 0.354419892305713
0.649999976158142 0.339946807179418
0.699999988079071 0.297197657391044
0.75 0.262363020252844
0.799999952316284 0.252893455862497
0.850000023841858 0.223722563820964
0.899999976158142 0.172807960232896
0.949999988079071 0.115094630675294
};
\addplot [, color0, mark=*, mark size=2.5, mark options={solid}, only marks]
table {%
0.050000000745058 4.46782677173614
0.0999999940395355 3.0938353061676
0.149999991059303 2.40335819721222
0.199999988079071 2.00902341604233
0.249999985098839 1.7336111664772
0.299999982118607 1.51615707874298
0.349999994039536 1.35976191759109
0.399999976158142 1.22056294679642
0.449999988079071 1.10285248756409
0.5 1
0.549999952316284 0.907642483711243
0.599999964237213 0.820674479007721
0.649999976158142 0.739465546607971
0.699999988079071 0.663612574338913
0.75 0.58669325709343
0.799999952316284 0.513780897855759
0.850000023841858 0.432242003083229
0.899999976158142 0.340220591425896
0.949999988079071 0.225181579589844
};
\addlegendentry{channel}
\addplot [, color1, mark=*, mark size=2.5, mark options={solid}, only marks]
table {%
0.050000000745058 1.21657845973969
0.0999999940395355 0.991082465648651
0.149999991059303 0.923867601156235
0.199999988079071 0.749507263302803
0.249999985098839 0.631336903572082
0.299999982118607 0.496064871549606
0.349999994039536 0.476879766583443
0.399999976158142 0.434469726681709
0.449999988079071 0.383464786410332
0.5 0.384469500184059
0.549999952316284 0.358455935120583
0.599999964237213 0.30833093225956
0.649999976158142 0.285003916919231
0.699999988079071 0.245954655110836
0.75 0.217981606721878
0.799999952316284 0.199368864297867
0.850000023841858 0.176243436336517
0.899999976158142 0.13924684971571
0.949999988079071 0.0807625763118266
};
\addlegendentry{spatial}
\end{axis}

\end{tikzpicture}
  \end{subfigure}

  \vspace*{-2ex}

  \caption{Sampling spatial axes is more effective than channels
    on both (\subref{subfig:subsampling-real-world}) real-world and
    (\subref{subfig:subsampling-synthetic}) synthetic data. We take the
    untrained All-CNN-C~\citep{springenberg2015striving}
    for CIFAR-100 with cross-entropy loss, disable dropout, and modify the
    convolutions to use a fraction $p$ of $\tX$ when computing the weight
    gradient via Bernoulli-CRS. For mini-batches of size 128, we compute the
    deterministic gradients for all kernels, then flatten and concatenate them
    into a vector $\vg$; likewise for its proxy $\hat{\vg}$. CRS is described
    by $(p_{c_{\text{in}}}, p_{i_1}, p_{i_2})$, the keep rates along the
    channel and spatial dimensions. We compare \textcolor{maincolor}{channel} and
    \textcolor{secondcolor}{spatial} sampling with same memory reduction,
    i.e.\,$\textcolor{maincolor}{(p, 1, 1)}$ and $\textcolor{secondcolor}{(1, \sqrt{p},
      \sqrt{p})}$. To measure approximation quality, we use the normalized
    residual norm $\nicefrac{\left\lVert\vg - \hat{\vg}
      \right\rVert_2}{\left\lVert \vg \right\rVert_2}$ and report mean and
    standard deviation of 10 different model and batch
    initializations.}\label{fig:subsampling-weight-gradient}
  \vspace{-2ex}
\end{figure}

\section{Related Work}
\textbf{Structured convolutions:} We use the TN formulation of convolution from \citet{hayashi2019einconv} who focus on connecting kernel factorizations to existing (depth-wise separable~\citep{howard2017mobilenets,sandler2018mobilenetv2}, factored~\citep{szegedy2016rethinking}, bottleneck~\citep{he2016deep}, flattened/CP decomposed, low-rank filter~\citep{smith1997scientist,rigamonti2013learning,tai2015convolutional}) convolutions and explore new factorizations.
Our work focuses on operations related to convolutions, diagram manipulations, the index pattern structure, and computational performance/flexibility. Structured convolutions integrate seamlessly with our framework by replacing the kernel with its factorized TN.

\textbf{Higher-order autodiff:} ML frameworks focus on differentiating scalar-valued objectives once.
Recent works \citep{laue2018computing,laue2020simple,ma2020autohoot} developed a tensor calculus to compute higher-order derivatives of tensor-valued functions and compiler optimizations through linear algebra and common sub-expression elimination.
Phrasing convolution as \texttt{einsum}, we allow it to be integrated into such frameworks, benefit from their optimizations, and complement them with our convolution-specific simplifications.

\section{Conclusion}
We used tensor networks (TNs), a diagrammatic representation of tensor multiplications, to simplify convolutions and many related operations.
We derived the diagrams of autodiff and less standard routines for curvature approximations like KFAC with support for all hyper-parameters, channel groups, batching, and generalization to arbitrary dimensions.
All amount to simple \texttt{einsum} expressions that can easily be modified---e.g.\,to perform stochastic backpropagation---and benefit from under-the-hood optimizations before evaluation.
We complemented those by convolution-specific symbolic simplifications based on structure in the connectivity pattern and showed their effectiveness to advance second-order methods.
Our TN implementation accelerates the computation of KFAC up to 4.5\,x and uses significantly less memory.
Beyond performance improvements, the simplifying perspective also allowed us to formulate KFAC for transpose convolution. More broadly, our work underlines the elegance of TNs for reasoning about tensor multiplications and function transformations (differentiation, batching, slicing, simplification) in terms of diagrams at less cognitive load without sacrificing rigour.
We believe they are a powerful tool for the ML community that will open up new algorithmic possibilities due to their simplicity \& flexibility.

\begin{ack}
  The author would like to thank Luca Thiede, Andres Fernandez Rodr\'iguez, and Kirill Neklyudov for providing feedback to the manuscript.
  Resources used in preparing this research were provided, in part, by the Province of Ontario, the Government of Canada through CIFAR, and companies sponsoring the Vector Institute.
\end{ack}


{
  \small
  \bibliographystyle{icml2024.bst}

}

\clearpage
\appendix
\renewcommand\thefigure{\thesection\arabic{figure}}
\renewcommand\thetable{\thesection\arabic{table}}
\renewcommand\thealgorithm{\thesection\arabic{algorithm}}
\renewcommand{\theequation}{\thesection\arabic{equation}}
\renewcommand{\thetransformation}{\thesection\arabic{transformation}}

\makeatletter
\vbox{%
  \hsize\textwidth
  \linewidth\hsize
  \vskip 0.1in
  \@toptitlebar
  \centering
  {\LARGE\bf \@title\,(Supplementary Material)\par}
  \@bottomtitlebar
  \vskip 0.3in \@minus 0.1in
}
\makeatother

\startcontents[sections]
\printcontents[sections]{l}{1}{\setcounter{tocdepth}{2}}
\vspace{2em}

\section{Limitations}\label{sec:app:limitations}
Here we comment on limitations on our approach.

\paragraph{No common sub-expression elimination (CSE):} Our implementation
relies on \texttt{opt\_einsum} which focuses on contraction order optimization.
This optimization is efficient when all operands are different. However, with
multiple occurrences of operands, computing shared sub-expressions might be an
advantageous optimization approach which \texttt{opt\_einsum} does not account
for. The second-order quantity TNs from \Cref{sec:app:second-order,subsec:kfac}
contain such sub-expressions, for instance $\llbracket \tX \rrbracket$ and
$\vone^{\top}_{O_1 O_2}\llbracket \tX \rrbracket$ in KFAC-expand and
KFAC-reduce, and $\tS^{(\tW)}$ in the GGN quantities from
\Cref{fig:tensor-networks-higher-order}. The efficiency of CSE depends on how
costly the shared tensor is to compute. For instance, computing $\tS^{(\tW)}$ is
expensive and therefore CSE is the more suitable optimization technique. For the
input-based Kronecker factors which require the unfolded input, either
contraction path optimization or CSE might be better. This is because the
optimal contraction order may not correspond to 2x input unfolding and exhibit
more parallelism which may lead to faster run times on a GPU. It would be
interesting to integrate CSE into the contraction path optimization and develop
a heuristic to choose a contraction path, for instance based on a weighted sum
of FLOPs and memory.

\paragraph{No index slicing:} We mention index slicing as a technique to reduce
peak memory of, and distribute, TN contractions. However, our implementation
does not use index slicing, although there are packages like
\texttt{cotengra}~\cite{gray2021hyper} with an interface similar to
\texttt{opt\_einsum}. We did not experiment with index slicing as our benchmark
uses a single GPU and did not encounter out-of-memory errors. Still, we mention
this technique, as, in combination with CSE, it could automatically reduce peak
memory of the GGN quantities from~\Cref{fig:tensor-networks-higher-order} which
suffer from high memory requirements.



\section{Visual Tour of Tensor Network Operations for Convolutions}
\label{sec:app:visual-tour}
Here, we extend the presented operations with a batch axis and allow for grouped
convolutions.

\begin{figure}[!t]
  \captionsetup[subfigure]{justification=centering}
  \centering
  \begin{subfigure}[t]{0.325\linewidth}
    \centering
\input{fig/tensor_networks/commands.tex}
\def\X{%
  \tensornode{X}%
  {$\tX$}%
  {{gcin}}
  {{i1, i2}}
  {{n}}
  {{}}
}%
\def\W{%
  \tensornode{W}%
  {$\tW$}%
  {{gcout}}
  {{k2, k1}}
  {{cin}}
  {{}}
}%
\begin{tikzpicture}[ultra thick]%
  \matrix[row sep=3ex,column sep=0ex]{
    & \P{1} & \\
    & \P{2} & \\
    \X & & \W \\
  };%
  \draw %
  \contract{X}{P1}{i1}{$i_1$}{out=90, in=180}%
  \contract{X}{P2}{i2}{$i_2$}{out=90, in=180}%
  \contract{P1}{W}{k1}{$k_1$}{out=0, in=90}%
  \contract{P2}{W}{k2}{$k_2$}{out=0, in=90}%
  (P1-o1) node[index, above] {$o_1$}%
  (P2-o2) node[index, above] {$o_2$}%
  ;%
  \node [index, left] at (X-n) {\textcolor{maincolor}{$n$}};%

  \coordinate (X-gcin-ungroup) at ($(X-gcin)+(0ex, 0)$);%
  \draw (X-gcin) to (X-gcin-ungroup);%
  \coordinate (X-cin) at ($(X-gcin-ungroup)+(3ex,-1ex)$);%
  \coordinate (X-g) at ($(X-gcin-ungroup)+(3ex,1ex)$);%
  \draw [secondcolor, out=30, in=180] (X-gcin-ungroup) to (X-g);%
  \draw [out=330, in=180] (X-gcin-ungroup) to (X-cin);%
  \draw [out=0, in=180] (X-cin) to node [index, midway] {$\tilde{c}_{\text{in}}$} (W-cin);%
  \draw[arrows=-{Triangle}] (X-gcin-ungroup) to ++(-0.1ex,0);%

  \coordinate (W-gcout-ungroup) at ($(W-gcout)+(0ex, 0)$);%
  \draw (W-gcout) to (W-gcout-ungroup);%
  \coordinate (W-cout) at ($(W-gcout-ungroup)+(3ex,-1ex)$);%
  \coordinate (W-g) at ($(W-gcout-ungroup)+(3ex,1ex)$);%
  \draw [secondcolor, out=30, in=180] (W-gcout-ungroup) to (W-g);%
  \draw [out=330, in=180] (W-gcout-ungroup) to (W-cout) node [index, right] {$\tilde{c}_{\text{out}}$};%
  \draw[arrows=-{Triangle}] (W-gcout-ungroup) to ++(-0.1ex,0);%

  \coordinate (g) at ($(W-g)+(1.5ex, 1.5ex)$);%
  \draw [secondcolor]%
  (X-g) |- (g)%
  (W-g) |- (g)%
  node [index, right] {\textcolor{secondcolor}{$g$}}%
  ;%
\end{tikzpicture}
%
    \caption{Convolution (no bias) \\[0.5ex]
      $\etY_{n, (g, \tilde{c}_{\text{out}}), o_1, o_2}$
      \\[1ex]
      \quad\Cref{eq:convolution-batch-groups}}\label{subfig:convolution-batch-groups}
  \end{subfigure}
  \hfill
  \begin{subfigure}[t]{0.325\linewidth}
    \centering
\input{fig/tensor_networks/commands.tex}
\def\I{%
  \tensornode{I}%
  {$vdelta$}%
  {{cout-right}}
  {{}}
  {{cout-left}}
  {{}}
}%
\def\X{%
  \tensornode{X}%
  {$\tX$}%
  {{gcin}}
  {{i1, i2}}
  {{n}}
  {{}}
}%
\begin{tikzpicture}[ultra thick]%
  \matrix[row sep=3ex,column sep=0ex]{
    & \P{1} &\\
    & \P{2} & \\
    \X & & \begin{scope}[opacity=0, tensor/.append style={opacity=0, xscale=0}]\I\end{scope} \\
  };%
  \draw %
  \contract{X}{P1}{i1}{$i_1$}{out=90, in=180}%
  \contract{X}{P2}{i2}{$i_2$}{out=90, in=180}%
  (P1-k1) node[index, right] {$k'_1$}%
  (P2-k2) node[index, right] {$k'_2$}%
  (P1-o1) node[index, above] {$o_1$}%
  (P2-o2) node[index, above] {$o_2$}%
  (I-cout-left) node[index, left, yshift=0.5ex] {$\tilde{c}'_{\text{out}}$} %
  to%
  (I-cout-right) node[index, right, yshift=0.5ex] {$\tilde{c}_{\text{out}}$}%
  ;%
  \node [index, left] at (X-n) {\textcolor{maincolor}{$n$}};%

  \coordinate (X-gcin-ungroup) at ($(X-gcin)+(0ex, 0)$);%
  \draw (X-gcin) to (X-gcin-ungroup);%
  \coordinate (X-cin) at ($(X-gcin-ungroup)+(3ex,-1ex)$);%
  \coordinate (X-g) at ($(X-gcin-ungroup)+(3ex,1ex)$);%
  \draw [secondcolor, out=30, in=180] (X-gcin-ungroup) to (X-g);%
  \draw [out=330, in=180] (X-gcin-ungroup) to (X-cin) node [index, right, yshift=-0.5ex] {$\tilde{c}_{\text{in}}$};%
  \draw[arrows=-{Triangle}] (X-gcin-ungroup) to ++(-0.1ex,0);%

  \coordinate (g) at ($(I-cout-right)+(0,5ex)$);%
  \coordinate (gprime) at ($(I-cout-right)+(0,3ex)$);%

  \draw[secondcolor]%
  (X-g) |- (g) node[index, right] {$g$}%
  (gprime) node[index, right] {$g'$} -| ++(-2ex,2ex)%
  ;%
\end{tikzpicture}
%
    \caption{Weight Jacobian \\
      \quad$\displaystyle\frac{\partial \etY_{n, (g, \tilde{c}_{\text{out}}), o_1, o_2}}{\partial
        \etW_{(g', \tilde{c}'_{\text{out}}), \tilde{c}'_{\text{in}}, k'_1,
          k'_2}}$}\label{subfig:weight-jacobian-batch-groups}
  \end{subfigure}
  \begin{subfigure}[t]{0.325\linewidth}
    \centering
\input{fig/tensor_networks/commands.tex}
\def\I{%
  \tensornode{I}%
  {$\vdelta$}%
  {{cin}}
  {{i1, i2}}
  {{n}}
  {{}}
}%
\def\W{%
  \tensornode{W}%
  {$\tW$}%
  {{gcout}}
  {{k2, k1}}
  {{cin}}
  {{}}
}%
\begin{tikzpicture}[ultra thick]%
  \matrix[row sep=3ex,column sep=0ex]{
    &\P{1} & \\
    &\P{2} & \\
    \begin{scope}[opacity=0, tensor/.append style={opacity=0, xscale=0}]\I\end{scope} & &\W \\
  };%
  \draw %
  \contract{P1}{W}{k1}{$k_1$}{out=0, in=90}%
  \contract{P2}{W}{k2}{$k_2$}{out=0, in=90}%
  (W-cin) node[index, left] {$\tilde{c}'_{\text{in}}$}%
  (P1-i1) node[index, left] {$i'_1$}%
  (P2-i2) node[index, left] {$i'_2$}%
  (P1-o1) node[index, above] {$o_1$}%
  (P2-o2) node[index, above] {$o_2$};%
  \draw%
  (I-n) node [index, left] {$\textcolor{maincolor}{n}\phantom{'}\!$} to (I-cin) node
  [index, right] {$\textcolor{maincolor}{n'}$}%
  ;%

  \coordinate (W-gcout-ungroup) at ($(W-gcout)+(0ex, 0)$);%
  \draw (W-gcout) to (W-gcout-ungroup);%
  \coordinate (W-cout) at ($(W-gcout-ungroup)+(3ex,-1ex)$);%
  \coordinate (W-g) at ($(W-gcout-ungroup)+(3ex,1ex)$);%
  \draw [secondcolor, out=30, in=180] (W-gcout-ungroup) to (W-g);%
  \draw [out=330, in=180] (W-gcout-ungroup) to (W-cout) node [index, right] {$\tilde{c}_{\text{out}}$};%
  \draw[arrows=-{Triangle}] (W-gcout-ungroup) to ++(-0.1ex,0);%

  \coordinate (g) at ($(W-g)+(2ex,3ex)$);%
  \coordinate (gprime) at ($(W-g)+(2ex,1ex)$);%
  \draw[secondcolor]%
  (W-g) |- (g) node[index, right] {$g$}%
  (gprime) node[index, right] {$g'$} -| ++(-2ex,2ex)%
  ;%
\end{tikzpicture}
    \caption{Input Jacobian\\ \quad$\displaystyle \frac{\partial \etY_{n, (g,
          \tilde{c}_{\text{out}}), o_1, o_2}}{\partial \etX_{n',(g',
          \tilde{c}'_{\text{in}}), i'_1,
          i'_2}}$}\label{subfig:input-jacobian-batch-groups}
  \end{subfigure}

  \vspace{2ex}

  \begin{subfigure}[t]{0.495\linewidth}
    \centering
\input{fig/tensor_networks/commands.tex}
\def\X{%
  \tensornode{X}%
  {$\tX$}%
  {{gcin}}
  {{i1, i2}}
  {{n}}
  {{}}
}%
\def\V{%
  \tensornode{V}%
  {$\tV$}%
  {{gcout}}
  {{o1, o2}}
  {{n}}
  {{}}
}%
\begin{tikzpicture}[ultra thick, opacity=0.3, index/.append style={fill opacity=1, text opacity=0.3}]%
  \matrix[row sep=3ex,column sep=0ex]{
    &\P{1} &\\
    &\P{2} & \\
    \X & & \begin{scope}[opacity=1, xshift=4ex] \V \end{scope} \\
  };%
  \draw %
  \contract{X}{P1}{i1}{$i_1$}{out=90, in=180}%
  \contract{X}{P2}{i2}{$i_2$}{out=90, in=180}%
  (P1-k1) node[index, right] {$k'_1$}%
  (P2-k2) node[index, right] {$k'_2$};%
  \begin{scope}[opacity=1, index/.append style={text opacity=1}]
    \draw %
    (V-o2) |- node[index, xshift=-3.5ex] {$o_1$} (P1-o1)%
    (V-o1) |- node[index, xshift=-2ex] {$o_2$} (P2-o2)%
    ;%
    \draw[maincolor]%
    (X-n) to ++(0, +3ex) -| node [index, midway] {$n$} (V-n)
    ;%

    \coordinate (X-gcin-ungroup) at ($(X-gcin)+(0ex, 0)$);%
    \draw (X-gcin) to (X-gcin-ungroup);%
    \coordinate (X-cin) at ($(X-gcin-ungroup)+(3ex,-1ex)$);%
    \coordinate (X-g) at ($(X-gcin-ungroup)+(3ex,1ex)$);%
    \draw [secondcolor, out=30, in=180] (X-gcin-ungroup) to (X-g);%
    \draw [out=330, in=180] (X-gcin-ungroup) to (X-cin)
    node [index, right] {$\tilde{c}'_{\text{in}}$}
    ;%
    \draw[arrows=-{Triangle}] (X-gcin-ungroup) to ++(-0.1ex,0);%

    \coordinate (V-gcout-ungroup) at ($(V-gcout)+(0ex, 0)$);%
    \draw (V-gcout) to (V-gcout-ungroup);%
    \coordinate (V-cout) at ($(V-gcout-ungroup)+(3ex,-1ex)$);%
    \coordinate (V-g) at ($(V-gcout-ungroup)+(3ex,1ex)$);%
    \draw [secondcolor, out=30, in=180] (V-gcout-ungroup) to (V-g);%
    \draw [out=330, in=180] (V-gcout-ungroup) to (V-cout) node [index, right] {$\tilde{c}'_{\text{out}}$};%
    \draw[arrows=-{Triangle}] (V-gcout-ungroup) to ++(-0.1ex,0);%

    \coordinate (g) at ($(V-g)+(1.5ex, 3.5ex)$);%
    \draw [secondcolor]%
    (V-g) |- (g)%
    (X-g) |- (g)%
    node [index, right] {\textcolor{secondcolor}{$g'$}}%
    ;%
  \end{scope}
\end{tikzpicture}
    \caption{Weight VJP
      \\
      $\displaystyle
      \etV^{(\tW)}_{(g', \tilde{c}'_{\text{out}}), \tilde{c}'_{\text{in}},
        k'_1, k'_2}$
      \\[1ex]
      =
      \\
      $\displaystyle
      \sum_{\substack{n, g, \tilde{c}_{\text{out}},\\ o_1,
          o_2}} \!\!\etV_{n, (g, \tilde{c}_{\text{out}}), o_1, o_2}
      \frac{\partial \etY_{n, (g, \tilde{c}_{\text{out}}), o_1,
          o_2}}{\partial \etW_{(g', \tilde{c}'_{\text{out}}),
          \tilde{c}'_{\text{in}}, k'_1, k'_2}}$}\label{subfig:weight-vjp-batch-groups}
  \end{subfigure}
  \hfill
  \begin{subfigure}[t]{0.495\linewidth}
    \centering
\input{fig/tensor_networks/commands.tex}
\def\V{%
  \tensornode{V}%
  {$\tV^{(\tY)}$}%
  {{gcout}}
  {{o1, o2}}
  {{n}}
  {{}}
}%
\def\W{%
  \tensornode{W}%
  {$\tW$}%
  {{gcout}}
  {{k2, k1}}
  {{cin}}
  {{}}
}%
\begin{tikzpicture}[ultra thick, opacity=0.3, index/.append style={fill opacity=1, text opacity=0.3}]%
  \matrix[row sep=3ex,column sep=0ex]{
    &\P{1} &\\
    &\P{2} & \\
    \begin{scope}[opacity=1]
      \V
    \end{scope}
    & & \W \\
  };
  \draw %
  \contract{P1}{W}{k1}{$k_1$}{out=0, in=90}%
  \contract{P2}{W}{k2}{$k_2$}{out=0, in=90}%
  (W-cin) node[index, left] {$\tilde{c}'_{\text{in}}$}%
  (P1-i1) node[index, left] {$i'_1$}%
  (P2-i2) node[index, left] {$i'_2$};%
  \begin{scope}[opacity=1, index/.append style={text opacity=1}]
    \draw %
    (V-o1) |- node[index, xshift=3.5ex] {$o_1$} (P1-o1)%
    (V-o2) |- node[index, xshift=2ex] {$o_2$} (P2-o2)%
    ;%
    \node [index, left] at (V-n) {$\textcolor{maincolor}{n'}$};%

    \coordinate (V-gcout-ungroup) at ($(V-gcout)+(0ex, 0)$);%
    \draw (V-gcout) to (V-gcout-ungroup);%
    \coordinate (V-cout) at ($(V-gcout-ungroup)+(3ex,-1ex)$);%
    \coordinate (V-g) at ($(V-gcout-ungroup)+(3ex,1ex)$);%
    \draw [secondcolor, out=30, in=180] (V-gcout-ungroup) to (V-g);%
    \draw [out=330, in=180] (V-gcout-ungroup) to (V-cout);%
    \draw[arrows=-{Triangle}] (V-gcout-ungroup) to ++(-0.1ex,0);%

    \coordinate (W-gcout-ungroup) at ($(W-gcout)+(0ex, 0)$);%
    \draw (W-gcout) to (W-gcout-ungroup);%
    \coordinate (W-cout) at ($(W-gcout-ungroup)+(3ex,-1ex)$);%
    \coordinate (W-g) at ($(W-gcout-ungroup)+(3ex,1ex)$);%
    \draw [secondcolor, out=30, in=180] (W-gcout-ungroup) to (W-g);%
    \draw [out=330, in=180] (W-gcout-ungroup) to (W-cout);%
    \draw[arrows=-{Triangle}] (W-gcout-ungroup) to ++(-0.1ex,0);%

    \coordinate (g) at ($(W-g)+(1.5ex, 1.5ex)$);%
    \draw [secondcolor]%
    (V-g) |- (g)%
    (W-g) |- (g)%
    node [index, right] {\textcolor{secondcolor}{$g'$}}%
    ;%
    \draw (V-cout) to ++(0ex,-1.5ex) node [index, xshift=3ex] {$\tilde{c}_{\text{out}}$} -| (W-cout);%
  \end{scope}
\end{tikzpicture}
    \caption{Input VJP
      \\
      $\displaystyle
      \etV^{(\tX)}_{n', (g', \tilde{c}'_{\text{in}}), i'_1, i'_2}$
      \\[1ex]
      =
      \\
      $\displaystyle
      \sum_{\substack{n, g, \tilde{c}_{\text{out}},\\ o_1,
          o_2}} \!\!\etV_{n, (g, \tilde{c}_{\text{out}}), o_1, o_2}
      \frac{\partial \etY_{n, (g, \tilde{c}_{\text{out}}), o_1,
          o_2}}{\partial \etX_{n', (g', \tilde{c}'_{\text{in}}),
          , i'_1, i'_2}}$}\label{subfig:input-vjp-batch-groups}
  \end{subfigure}

  \vspace{1ex}

  \caption{TNs of the (\subref{subfig:convolution-batch-groups}) forward pass,
    (\subref{subfig:weight-jacobian-batch-groups},
    \subref{subfig:input-jacobian-batch-groups}) Jacobians, and
    (\subref{subfig:weight-vjp-batch-groups},
    \subref{subfig:input-vjp-batch-groups}) VJPs with \textcolor{maincolor}{batch
      axis} and \textcolor{secondcolor}{channel groups}. They generalize
    \Cref{fig:visual-abstract,fig:example-tensor-network-derivative}
    from the main text. For the VJPs, the Jacobians are shaded.}
  \label{fig:tensor-networks-batch-groups}
\end{figure}

\subsection{Convolution \& First-order Derivatives}

\paragraph{Adding a batch dimension (\texttt{vmap}-ing):} Adding a batch axis to
all presented operations is trivial. We only need to add an additional leg to
the batched tensors, and connect these legs via element-wise or inner multiplication,
depending on whether the result tensor is batched or not.

\paragraph{Grouped convolutions:} Grouped convolutions were originally proposed
by \citet{krizhevsky2012imagenet} and allow for parallelizing, distributing, and
reducing the parameters of the convolution operation. They split $C_{\text{in}}$
input channels into $G$ groups of size $\tilde{C}_{\text{in}} :=
\nicefrac{C_{\text{in}}}{G}$, then perform independent convolutions per group,
each producing $\tilde{C}_{\text{out}} := \nicefrac{C_{\text{out}}}{G}$ output
channels which are concatenated in the output. Each group uses a kernel
$\tW_g$of size $\tilde{C}_{\text{out}} \times \tilde{C}_{\text{in}} \times K_1
\times K_2$. These kernels are stacked into a single tensor $\tW \in
\sR^{C_{\text{out}}, \tilde{C}_{\text{in}}, K_1, K_2}$ such that
$[\tW]_{(g,:),:,:,:} = \tW_g$. To support groups, we thus decompose the channel
indices into $c_{\text{in}} := (\tilde{c}_{\text{in}}, g)$ and $c_{\text{out}}
:= (\tilde{c}_{\text{out}}, g)$. For the forward pass this yields the grouped
convolution (without bias)
\begin{align}
  \label{eq:convolution-batch-groups}
  \textstyle
  \etY_{(g, \tilde{c}_{\text{out}}), o_1, o_2}%
  =%
  \sum_{i_1, i_2, \tilde{c}_{\text{in}}, k_1, k_2}%
  \etX_{(g, \tilde{c}_{\text{in}}), i_1, i_2}%
  \etPi^{(1)}_{i_1, o_1, k_1}%
  \etPi^{(2)}_{i_2, o_2, k_2}%
  \etW_{(g, \tilde{c}_{\text{out}}), c_{\text{in}}, k_1, k_2}\,.%
\end{align}
\Cref{subfig:convolution-batch-groups} shows the
batched version of \Cref{eq:convolution-batch-groups} as TN. Applying the
differentiation rule from \Cref{sec:operations} leads to the Jacobians and VJPs
shown in the remaining panels of \Cref{fig:tensor-networks-batch-groups}.

\subsection{Exact Second-order Information}

\begin{figure}[!t]
  \begin{subfigure}[t]{0.495\linewidth}
    \centering
    \input{fig/tensor_networks/commands.tex}
\def\P#1#2{%
  \tensornode{P#1-#2}%
  {$\tPi^{(#1)}$}%
  {{k#1}}
  {{o#1}}
  {{i#1}}
  {{}}
}%
\def\X#1{%
  \tensornode{X#1}%
  {$\tX$}%
  {{gcin}}
  {{i1, i2}}
  {{n}}
  {{}}
}%
\def\S#1{%
  \tensornode{S#1}%
  {$\tS^{(\tY)}$}%
  {{gcout}}
  {{o1, o2}}
  {{n}}
  {{c}}
}%
\begin{tikzpicture}[ultra thick, opacity=0.3, index/.append style={fill opacity=1, text opacity=0.3}]%
  \matrix[row sep=2.5ex,column sep=0ex]{
    & & \P{1}{1} & \\%
    & & \P{2}{1} & \\%
    & \X{1} & & \\%
    \S{1} & & \\%
  };%
  \begin{scope}[xshift=19ex]
    \matrix[row sep=2.5ex,column sep=0ex]{
      & & \P{1}{2} & \coordinate (k1-low); \\
      & & \P{2}{2} & \coordinate (k2-low); \\
      &\X{2} & & \coordinate (cin-low); \\
      \S{2} & & & \coordinate (cout-low); \\
    };%
  \end{scope}
  \begin{scope}[index/.append style={xshift=-1ex, yshift=0.25ex}]
    \draw %
    \contract{S1}{P1-1}{o1}{$o_1$}{out=90, in=180}%
    \contract{S1}{P2-1}{o2}{$o_2$}{out=90, in=180}%
    \contract{S2}{P1-2}{o1}{$o'_1$}{out=90, in=180}%
    \contract{S2}{P2-2}{o2}{$o'_2$}{out=90, in=180}%
    ;%
  \end{scope}
  \begin{scope}[index/.append style={yshift=0.4ex}]
    \draw %
    \contract{X1}{P1-1}{i1}{$i_1$}{out=90, in=180}%
    \contract{X1}{P2-1}{i2}{$i_2$}{out=90, in=180}%
    \contract{X2}{P1-2}{i1}{$i'_1$}{out=90, in=180}%
    \contract{X2}{P2-2}{i2}{$i'_2$}{out=90, in=180}%
    ;%
  \end{scope}
  \begin{scope}[opacity=1, index/.append style={text opacity=1}]
    \coordinate (n) at ($(X1-n)+(0, -2.5ex)$);%
    \coordinate (g) at ($(cout-low)+(0, -2.4ex)$);%
    \coordinate (k1) at ($(k1-low)+(0, 2.5ex)$);%
    \coordinate (k2) at ($(k2-low)+(0, 2.5ex)$);%
    \coordinate (cout) at ($(cout-low)+(0, 2.5ex)$);%
    \draw %
    (P1-2-k1) |- (k1)%
    (P2-2-k2) |- (k2)%
    ;%
    \draw %
    (P1-1-k1) |- (k1) to ++(2ex, 0) node[index, right] {$k_1$}%
    (P2-1-k2) |- (k2) to ++(2ex, 0) node[index, right] {$k_2$}%
    \contract{S1}{S2}{c}{$c$}{out=0, in=180}%
    ;%
    \draw[maincolor] %
    (X1-n) to (n)%
    (X2-n) |- (n)%
    (S1-n) |- (n)%
    (S2-n) |- (n) node[index, left, xshift=-0.5ex] {$n$}%
    ;%

    \coordinate (X1-gcin-ungroup) at ($(X1-gcin)+(0ex, 0)$);%
    \draw (X1-gcin) to (X1-gcin-ungroup);%
    \coordinate (X1-cin) at ($(X1-gcin-ungroup)+(3ex,-1ex)$);%
    \coordinate (X1-g) at ($(X1-gcin-ungroup)+(3ex,1ex)$);%
    \draw [secondcolor, out=30, in=180] (X1-gcin-ungroup) to (X1-g);%
    \draw [out=330, in=180] (X1-gcin-ungroup) to (X1-cin);%
    \draw[arrows=-{Triangle}] (X1-gcin-ungroup) to ++(-0.1ex,0);%

    \coordinate (X2-gcin-ungroup) at ($(X2-gcin)+(0ex, 0)$);%
    \draw (X2-gcin) to (X2-gcin-ungroup);%
    \coordinate (X2-cin) at ($(X2-gcin-ungroup)+(3ex,-1ex)$);%
    \coordinate (X2-g) at ($(X2-gcin-ungroup)+(3ex,1ex)$);%
    \draw [secondcolor, out=30, in=180] (X2-gcin-ungroup) to (X2-g);%
    \draw [out=330, in=180] (X2-gcin-ungroup) to (X2-cin);%
    \draw[arrows=-{Triangle}] (X2-gcin-ungroup) to ++(-0.1ex,0);%

    \coordinate (S1-gcout-ungroup) at ($(S1-gcout)+(0ex, 0)$);%
    \draw (S1-gcout) to (S1-gcout-ungroup);%
    \coordinate (S1-cout) at ($(S1-gcout-ungroup)+(3ex,-1ex)$);%
    \coordinate (S1-g) at ($(S1-gcout-ungroup)+(3ex,1ex)$);%
    \draw [secondcolor, out=30, in=180] (S1-gcout-ungroup) to (S1-g);%
    \draw [out=330, in=180] (S1-gcout-ungroup) to (S1-cout);%
    \draw[arrows=-{Triangle}] (S1-gcout-ungroup) to ++(-0.1ex,0);%

    \coordinate (S2-gcout-ungroup) at ($(S2-gcout)+(0ex, 0)$);%
    \draw (S2-gcout) to (S2-gcout-ungroup);%
    \coordinate (S2-cout) at ($(S2-gcout-ungroup)+(3ex,-1ex)$);%
    \coordinate (S2-g) at ($(S2-gcout-ungroup)+(3ex,1ex)$);%
    \draw [secondcolor, out=30, in=180] (S2-gcout-ungroup) to (S2-g);%
    \draw [out=330, in=180] (S2-gcout-ungroup) to (S2-cout);%
    \draw[arrows=-{Triangle}] (S2-gcout-ungroup) to ++(-0.1ex,0);%

    \coordinate (g) at ($(X2-g)+(1.5ex, 3ex)$);%
    \draw [secondcolor]%
    (X1-g) |- (g) to ++(2ex,0) node [index, right] {\textcolor{secondcolor}{$g$}}%
    (S1-g) to ++(0,3ex) -| (g)%
    (S2-g) to ++(0,3ex) -| (g)%
    (X2-g) -| (g)%
    ;%

    \coordinate (cin) at ($(X2-cin)+(4ex, 3.5ex)$);%
    \draw%
    (X1-cin) to ++(2ex, 0) |- (cin) to ++(2ex, 0) node [index, right] {$\tilde{c}_{\text{in}}$}%
    (X2-cin) -| (cin)%
    ;%

    \coordinate (cout) at ($(S2-cout)+(11.5ex, 3.5ex)$);%
    \draw%
    (S1-cout) to ++(2ex, 0) |- (cout) to ++(2ex, 0) node [index, right] {$\tilde{c}_{\text{out}}$}%
    (S2-cout) -| (cout)%
    ;%
  \end{scope}
\end{tikzpicture}
    \caption{GGN diagonal}\label{subfig:ggn-diagonal-batch-groups}
  \end{subfigure}
  \hfill
  \begin{subfigure}[t]{0.495\linewidth}
    \centering
\input{fig/tensor_networks/commands.tex}
\def\P#1#2{%
  \tensornode{P#1-#2}%
  {$\tPi^{(#1)}$}%
  {{k#1}}
  {{o#1}}
  {{i#1}}
  {{}}
}%
\def\X#1{%
  \tensornode{X#1}%
  {$\tX$}%
  {{gcin}}
  {{i1, i2}}
  {{n}}
  {{}}
}%
\def\S#1{%
  \tensornode{S#1}%
  {$\tS^{(\tY)}$}%
  {{gcout}}
  {{o1, o2}}
  {{n}}
  {{c}}
}%
\begin{tikzpicture}[ultra thick, opacity=0.3, index/.append style={fill opacity=1, text opacity=0.3}]%
  \matrix[row sep=2.5ex,column sep=0ex]{
    & & \P{1}{1} & \\%
    & & \P{2}{1} & \\%
    & \X{1} & & \\%
    \S{1} & & \\%
  };%
  \begin{scope}[xshift=20ex]
    \matrix[row sep=2.5ex,column sep=0ex]{
      & & \P{1}{2} & \coordinate (k1-low); \\
      & & \P{2}{2} & \coordinate (k2-low); \\
      &\X{2} & & \coordinate (cin-low); \\
      \S{2} & & & \coordinate (cout-low); \\
    };%
  \end{scope}
  \begin{scope}[index/.append style={xshift=-1ex, yshift=0.25ex}]
    \draw %
    \contract{S1}{P1-1}{o1}{$o_1$}{out=90, in=180}%
    \contract{S1}{P2-1}{o2}{$o_2$}{out=90, in=180}%
    \contract{S2}{P1-2}{o1}{$o'_1$}{out=90, in=180}%
    \contract{S2}{P2-2}{o2}{$o'_2$}{out=90, in=180}%
    ;%
  \end{scope}
  \begin{scope}[index/.append style={yshift=0.4ex}]
    \draw %
    \contract{X1}{P1-1}{i1}{$i_1$}{out=90, in=180}%
    \contract{X1}{P2-1}{i2}{$i_2$}{out=90, in=180}%
    \contract{X2}{P1-2}{i1}{$i'_1$}{out=90, in=180}%
    \contract{X2}{P2-2}{i2}{$i'_2$}{out=90, in=180}%
    ;%
  \end{scope}
  \begin{scope}[opacity=1, index/.append style={text opacity=1, yshift = 2.5ex, xshift=1ex}]
    \draw %
    (P1-1-k1) to ++ (0, +2.5ex) -| (P1-2-k1) node [index] {$k_1$}%
    (P2-1-k2) to ++ (0, +2.5ex) -| (P2-2-k2) node [index] {$k_2$}%
    ;%
  \end{scope}
  \begin{scope}[opacity=1, index/.append style={text opacity=1}]
    \draw %
    (S1-c) to ++(-2ex, 0) node [index, left] {$c$}%
    (S2-c) to ++(-2ex, 0) node [index, left] {$c'$}%
    ;%
    \draw[maincolor] %
    (X1-n) to [out=180, in=90] (S1-n) to ++(-2ex,0) node [index, left] {$n$}%
    (X2-n) to [out=180, in=90] (S2-n) to ++(-2ex,0) node [index, left] {$n'$}%
    ;%

    \coordinate (X1-gcin-ungroup) at ($(X1-gcin)+(0ex, 0)$);%
    \draw (X1-gcin) to (X1-gcin-ungroup);%
    \coordinate (X1-cin) at ($(X1-gcin-ungroup)+(3ex,-1ex)$);%
    \coordinate (X1-g) at ($(X1-gcin-ungroup)+(3ex,1ex)$);%
    \draw [secondcolor, out=30, in=180] (X1-gcin-ungroup) to (X1-g);%
    \draw [out=330, in=180] (X1-gcin-ungroup) to (X1-cin);%
    \draw[arrows=-{Triangle}] (X1-gcin-ungroup) to ++(-0.1ex,0);%

    \coordinate (X2-gcin-ungroup) at ($(X2-gcin)+(0ex, 0)$);%
    \draw (X2-gcin) to (X2-gcin-ungroup);%
    \coordinate (X2-cin) at ($(X2-gcin-ungroup)+(3ex,-1ex)$);%
    \coordinate (X2-g) at ($(X2-gcin-ungroup)+(3ex,1ex)$);%
    \draw [secondcolor, out=30, in=180] (X2-gcin-ungroup) to (X2-g);%
    \draw [out=330, in=180] (X2-gcin-ungroup) to (X2-cin);%
    \draw[arrows=-{Triangle}] (X2-gcin-ungroup) to ++(-0.1ex,0);%

    \coordinate (S1-gcout-ungroup) at ($(S1-gcout)+(0ex, 0)$);%
    \draw (S1-gcout) to (S1-gcout-ungroup);%
    \coordinate (S1-cout) at ($(S1-gcout-ungroup)+(3ex,-1ex)$);%
    \coordinate (S1-g) at ($(S1-gcout-ungroup)+(3ex,1ex)$);%
    \draw [secondcolor, out=30, in=180] (S1-gcout-ungroup) to (S1-g);%
    \draw [out=330, in=180] (S1-gcout-ungroup) to (S1-cout);%
    \draw[arrows=-{Triangle}] (S1-gcout-ungroup) to ++(-0.1ex,0);%

    \coordinate (S2-gcout-ungroup) at ($(S2-gcout)+(0ex, 0)$);%
    \draw (S2-gcout) to (S2-gcout-ungroup);%
    \coordinate (S2-cout) at ($(S2-gcout-ungroup)+(3ex,-1ex)$);%
    \coordinate (S2-g) at ($(S2-gcout-ungroup)+(3ex,1ex)$);%
    \draw [secondcolor, out=30, in=180] (S2-gcout-ungroup) to (S2-g);%
    \draw [out=330, in=180] (S2-gcout-ungroup) to (S2-cout);%
    \draw[arrows=-{Triangle}] (S2-gcout-ungroup) to ++(-0.1ex,0);%

    \coordinate (g) at ($(X2-g)+(0, 3ex)$);%
    \draw [secondcolor]%
    (X1-g) |- (g) node [index, left, xshift=-1ex] {\textcolor{secondcolor}{$g$}}%
    (S1-g) to ++(0,3ex) -| (g)%
    (S2-g) to ++(0,3ex) -| (g)%
    (X2-g) -| (g)%
    ;%

    \coordinate (cin) at ($(X1-cin)+(4ex, 3.5ex)$);%
    \draw%
    (X1-cin) to ++(2ex, 0) |- (cin) to ++(2ex, 0)%
    (X2-cin) to ++(3.5ex, 0) |- (cin) %
    node [index, right] {$\tilde{c}_{\text{in}}$}%
    ;%

    \coordinate (cout) at ($(S1-cout)+(4ex, 3.5ex)$);%
    \draw%
    (S1-cout) to ++(2ex, 0) |- (cout) to ++(2ex, 0)%
    (S2-cout) to ++(2ex, 0) |- (cout) %
    node [index, right] {$\tilde{c}_{\text{out}}$}%
    ;%
  \end{scope}
\end{tikzpicture}
    \caption{GGN Gram matrix (empirical
      NTK)}\label{subfig:ggn-gram-batch-groups}
  \end{subfigure}

  \caption{TNs of (\subref{subfig:ggn-diagonal-batch-groups}) the
    GGN diagonal and (\subref{subfig:ggn-gram-batch-groups}) the GGN Gram matrix
    with \textcolor{maincolor}{batching} and \textcolor{secondcolor}{channel groups}.
    They extend \Cref{subfig:ggn-diagonal,subfig:ggn-gram} from the main text.
  }\label{fig:higher-order-batch-groups}
\end{figure}

In \Cref{fig:higher-order-batch-groups} we show the TNs for the GGN diagonal and
the GGN Gram matrix (empirical NTK matrix)
from \Cref{fig:tensor-networks-higher-order} extended by channel groups and a batch
axis.

\begin{figure}
  \centering 
\input{fig/tensor_networks/commands.tex}
\def\P#1#2{%
  \tensornode{P#1-#2}%
  {$\tPi^{(#1)}$}%
  {{gk#1}}
  {{o#1}}
  {{i#1}}
  {{}}
}%
\def\X#1{%
  \tensornode{X#1}%
  {$\tX$}%
  {{gcin}}
  {{i1, i2}}
  {{}}
  {{}}
}%
\def\S#1{%
  \tensornode{S#1}%
  {$\tS^{(\tY)}$}%
  {{gcout}}
  {{o1, o2}}
  {{}}
  {{c}}
}%
\begin{tikzpicture}[ultra thick, opacity=0.3, index/.append style={fill opacity=1, text opacity=0.3}]%
  \matrix[row sep=1ex,column sep=0ex]{
    & & \P{1}{1} & \\%
    & & \P{2}{1} & \\%
    & \X{1} & & \\%
    \begin{scope}[opacity=1]\S{1}\end{scope} & & \\%
  };%
  \begin{scope}[xshift=26ex]
    \matrix[row sep=1ex,column sep=0ex]{
      & & \P{1}{2} & \coordinate (k1-low); \\
      & & \P{2}{2} & \coordinate (k2-low); \\
      &\X{2} & & \coordinate (cin-low); \\
      \begin{scope}[opacity=1]\S{2}\end{scope} & & & \coordinate (cout-low); \\
    };%
  \end{scope}
  \begin{scope}[index/.append style={xshift=-1ex, yshift=0.25ex}]
    \draw %
    \contract{S1}{P1-1}{o1}{$o_1$}{out=90, in=180}%
    \contract{S1}{P2-1}{o2}{$o_2$}{out=90, in=180}%
    \contract{S2}{P1-2}{o1}{$o'_1$}{out=90, in=180}%
    \contract{S2}{P2-2}{o2}{$o'_2$}{out=90, in=180}%
    ;%
  \end{scope}
  \begin{scope}[index/.append style={yshift=0.4ex}]
    \draw %
    \contract{X1}{P1-1}{i1}{$i_1$}{out=90, in=180}%
    \contract{X1}{P2-1}{i2}{$i_2$}{out=90, in=180}%
    \contract{X2}{P1-2}{i1}{$i'_1$}{out=90, in=180}%
    \contract{X2}{P2-2}{i2}{$i'_2$}{out=90, in=180}%
    ;%
  \end{scope}
  \begin{scope}[opacity=1, index/.append style={text opacity=1}]
    \draw %
    \contract{S1}{S2}{c}{$c$}{out=0, in=180}%
    ;%
    \coordinate (gk1-ungroup) at ($(P1-1-gk1)+(2ex, 0)$);
    \draw[arrows=-{Triangle}] (gk1-ungroup) to ++(-0.1ex,0);%
    \draw (P1-1-gk1) to (gk1-ungroup);
    \coordinate (k1) at ($(gk1-ungroup)+(3ex,-1ex)$);
    \coordinate (gk1) at ($(gk1-ungroup)+(3ex,1ex)$);
    \draw [out=30, in=180] (gk1-ungroup) to (gk1);
    \draw [out=330, in=180] (gk1-ungroup) to (k1) node [index, right] {$\tilde{k}_1$};

    \coordinate (gk1prime-ungroup) at ($(P1-2-gk1)+(2ex, 0)$);
    \draw[arrows=-{Triangle}] (gk1prime-ungroup) to ++(-0.1ex,0);%
    \draw (P1-2-gk1) to (gk1prime-ungroup);
    \coordinate (k1prime) at ($(gk1prime-ungroup)+(3ex,-1ex)$);
    \coordinate (gk1prime) at ($(gk1prime-ungroup)+(3ex,1ex)$);
    \draw [out=30, in=180] (gk1prime-ungroup) to (gk1prime);
    \draw [out=330, in=180] (gk1prime-ungroup) to (k1prime) node [index, right] {$\tilde{k}'_1$};

    \coordinate (gk1-join) at ($(gk1prime-ungroup)+(7ex,2.5ex)$);
    \draw (gk1) |- (gk1-join) node [index, right] {$g_{K_1}$};
    \draw (gk1prime) |- (gk1-join);

    \coordinate (gk2-ungroup) at ($(P2-1-gk2)+(2ex, 0)$);
    \draw[arrows=-{Triangle}] (gk2-ungroup) to ++(-0.1ex,0);%
    \draw (P2-1-gk2) to (gk2-ungroup);
    \coordinate (k2) at ($(gk2-ungroup)+(3ex,-1ex)$);
    \coordinate (gk2) at ($(gk2-ungroup)+(3ex,1ex)$);
    \draw [out=30, in=180] (gk2-ungroup) to (gk2);
    \draw [out=330, in=180] (gk2-ungroup) to (k2) node [index, right] {$\tilde{k}_2$};

    \coordinate (gk2prime-ungroup) at ($(P2-2-gk2)+(2ex, 0)$);
    \draw[arrows=-{Triangle}] (gk2prime-ungroup) to ++(-0.1ex,0);%
    \draw (P2-2-gk2) to (gk2prime-ungroup);
    \coordinate (k2prime) at ($(gk2prime-ungroup)+(3ex,-1ex)$);
    \coordinate (gk2prime) at ($(gk2prime-ungroup)+(3ex,1ex)$);
    \draw [out=30, in=180] (gk2prime-ungroup) to (gk2prime);
    \draw [out=330, in=180] (gk2prime-ungroup) to (k2prime) node [index, right] {$\tilde{k}'_2$};

    \coordinate (gk2-join) at ($(gk2prime-ungroup)+(7ex,2.5ex)$);
    \draw (gk2) |- (gk2-join) node [index, right] {$g_{K_2}$};
    \draw (gk2prime) |- (gk2-join);

    \coordinate (gcin-ungroup) at ($(X1-gcin)+(2ex, 0)$);
    \draw[arrows=-{Triangle}] (gcin-ungroup) to ++(-0.1ex,0);%
    \draw (X1-gcin) to (gcin-ungroup);
    \coordinate (cin) at ($(gcin-ungroup)+(3ex,-1ex)$);
    \coordinate (gcin) at ($(gcin-ungroup)+(3ex,1ex)$);
    \draw [out=30, in=180] (gcin-ungroup) to (gcin);
    \draw [out=330, in=180] (gcin-ungroup) to (cin) node [index, right] {$\tilde{c}_{\text{in}}$};

    \coordinate (gcinprime-ungroup) at ($(X2-gcin)+(2ex, 0)$);
    \draw[arrows=-{Triangle}] (gcinprime-ungroup) to ++(-0.1ex,0);%
    \draw (X2-gcin) to (gcinprime-ungroup);
    \coordinate (cinprime) at ($(gcinprime-ungroup)+(3ex,-1ex)$);
    \coordinate (gcinprime) at ($(gcinprime-ungroup)+(3ex,1ex)$);
    \draw [out=30, in=180] (gcinprime-ungroup) to (gcinprime);
    \draw [out=330, in=180] (gcinprime-ungroup) to (cinprime) node [index, right] {$\tilde{c}_{\text{in}}'$};

    \coordinate (gcin-join) at ($(gcinprime-ungroup)+(7ex,2.5ex)$);
    \draw (gcin) |- (gcin-join) node [index, right] {$g_{C_{\text{in}}}$};
    \draw (gcinprime) |- (gcin-join);

    \coordinate (gcout-ungroup) at ($(S1-gcout)+(2ex, 0)$);
    \draw[arrows=-{Triangle}] (gcout-ungroup) to ++(-0.1ex,0);%
    \draw (S1-gcout) to (gcout-ungroup);
    \coordinate (cin) at ($(gcout-ungroup)+(3ex,-1ex)$);
    \coordinate (gcout) at ($(gcout-ungroup)+(3ex,1ex)$);
    \draw [out=30, in=180] (gcout-ungroup) to (gcout);
    \draw [out=330, in=180] (gcout-ungroup) to (cin) node [index, right] {$\tilde{c}_{\text{out}}$};

    \coordinate (gcoutprime-ungroup) at ($(S2-gcout)+(2ex, 0)$);
    \draw[arrows=-{Triangle}] (gcoutprime-ungroup) to ++(-0.1ex,0);%
    \draw (S2-gcout) to (gcoutprime-ungroup);
    \coordinate (cinprime) at ($(gcoutprime-ungroup)+(3ex,-1ex)$);
    \coordinate (gcoutprime) at ($(gcoutprime-ungroup)+(3ex,1ex)$);
    \draw [out=30, in=180] (gcoutprime-ungroup) to (gcoutprime);
    \draw [out=330, in=180] (gcoutprime-ungroup) to (cinprime) node [index, right] {$\tilde{c}_{\text{out}}'$};

    \coordinate (gcout-join) at ($(gcoutprime-ungroup)+(7ex,2.5ex)$);
    \draw (gcout) |- (gcout-join) node [index, right] {$g_{C_{\text{out}}}$};
    \draw (gcoutprime) |- (gcout-join);
  \end{scope}
\end{tikzpicture}
  \caption{TN of a GGN mini-block diagonal without batching and channel
    groups.}\label{subfig:ggn-mini-block}
\end{figure}

\paragraph{Diagonal block extraction:} Combined with index un-grouping, diagonal
extraction generalizes to larger blocks: Let $\mA\in \sR^{KI\times KJ}$ be a
matrix of $K$ horizontally and vertically concatenated blocks
$\mA^{(k_1,k_2)}\in \sR^{I\times J}, k_i=1\dots,K$. We can extract the diagonal
blocks by restoring the sub-structure, then taking the diagonal along the
$K$-dimensional index,
\begin{equation*}
  \resizebox{!}{2.5ex}{%
    \tikz[baseline=-0.5ex,%
    index/.append style={fill=white, fill opacity = 1},%
    very thick,%
    ]{%
      \tensornode{A}%
      {$\left\{ \mA^{(k,k)} \right\}$}%
      {{j}}
      {{}}
      {{k, i}}
      {{}};
      \node[left, index] at (A-k) {$k$};%
      \node[left, index, fill opacity=0, text opacity=1] at (A-i) {$i$};%
      \node[right, index] at (A-j) {$j$};%
    }%
  } = \resizebox{!}{3.5ex}{%
    \tikz[baseline=-0.5ex,%
    index/.append style={fill=white, fill opacity = 1},%
    very thick,%
    ]{%
      \tensornode{A}%
      {$\mA$}%
      {{k2j}}
      {{}}
      {{k1i}}
      {{}};
      \coordinate (k1i) at ($(A)+(-11ex,0)$);%
      \coordinate (k2j) at ($(A)+(11ex,0)$);%
      \draw[arrows=-{Triangle[reversed]}] (k1i) to ++(-0.1ex,0);%
      \draw[arrows=-{Triangle[reversed]}] (k2j) to ++(0.1ex,0);%
      \draw (A-k2j) to node [midway, index, xshift=-1.5ex] {$(k, j)$} (k2j);%
      \draw (A-k1i) to node [midway, index, xshift=1.5ex] {$(k, i)$} (k1i);%

      \coordinate (k1) at ($(k1i)+(-2ex,2.5ex)$);%
      \draw[out=150, in=270] (k1i) to (k1);%
      \draw (k1) to ++(-2ex, 0ex) node [index, left] {$k$};%
      \coordinate (i) at ($(k1i)+(-4ex,-0.5ex)$);%
      \draw[out=210, in=0] (k1i) to (i) node [index, left] {$i$};%

      \coordinate (k2) at ($(k2j)+(2ex,2.5ex)$);%
      \draw[out=30, in=270] (k2j) to (k2);%
      \draw (k2) to (k1);%
      \coordinate (j) at ($(k2j)+(4ex,-0.5ex)$);%
      \draw[out=330, in=180] (k2j) to (j) node [index, right] {$j$};%
    }%
  }. %
\end{equation*}
We can apply this procedure to the GGN from \Cref{subfig:ggn}. Assume we want to
divide the output channel, input channel, and spatial dimensions into
$G_{C_{\text{out}}}, G_{C_{\text{in}}}, G_{K_1}, G_{K_2}$ groups. A group will
thus be indexed with a tuple $(g_{C_{\text{out}}}, g_{C_{\text{in}}}, g_{K_1},
g_{K_2})$ and the corresponding GGN block will be of dimension $C_{\text{out}} /
G_{C_{\text{out}}} \times C_{\text{in}} / G_{C_{\text{in}}} \times K_1 / G_{K_1}
\times K_2 / G_{K_2} \times C_{\text{out}} / G_{C_{\text{out}}} \times
C_{\text{in}} / G_{C_{\text{in}}} \times K_1 / G_{K_1} \times K_2 / G_{K_2}$ and
correspond to the GGN for $[\tW]_{(g_{C_{\text{out}}},:), (g_{C_{\text{in}}},:),
  (g_{K_1},:), (g_{K_2},:)}$. This process of un-grouping the output dimensions,
then taking the diagonal along the group indices, is illustrated in
\Cref{subfig:ggn-mini-block}. Note that if we choose $G_{C_{\text{out}}} =
C_{\text{out}}, G_{C_{\text{in}}} = C_{\text{in}}, G_{K_1} = K_1, G_{K_2} =
K_2$, each block will be a single number and hence we recover the GGN diagonal
from \Cref{subfig:ggn-diagonal}. If instead we $G_{C_{\text{out}}} =
G_{C_{\text{in}}} G_{K_1} G_{K_2} = 1$, we obtain the full GGN from
\Cref{subfig:ggn}. The outlined schemes allows to extract mini-blocks of
arbitrary size along the diagonal (subject to the total dimension).

\subsection{Kronecker-factored Approximate Curvature (KFAC) for Grouped Convolutions}\label{app:kfac-grouped-convolution}

We were unable to find a definition of KFAC for grouped convolutions. Hence, we
derive it here and present the TN diagrams. We use the perspective that grouped
convolutions are independent convolutions over channel groups which are then
concatenated. For each of those convolutions, we can then apply established the
KFAC approximation for convolutions without groups. For a group $g$ we have the
kernel $\tW_g = [\tW]_{(g,:),:,:,:}$ and the unfolded input of its associated
input channels, $\llbracket \tX_g \rrbracket = \llbracket \tX
\rrbracket_{(g,:),:,:}= \llbracket [\tX]_{(g,:),:,:} \rrbracket$ (or $\llbracket
\tX_{n,g} \rrbracket = \llbracket \tX_n \rrbracket_{(g,:),:,:}= \llbracket
[\tX]_{n,(g,:),:,:} \rrbracket$ in the batched setting).

\begin{figure}[!t]
  \begin{subfigure}[t]{0.49\linewidth}
    \centering
\input{fig/tensor_networks/commands.tex}
\def\P#1#2{%
  \tensornode{P#1-#2}%
  {$\tPi^{(#1)}$}%
  {{k#1}}
  {{o#1}}
  {{i#1}}
  {{}}
}%
\def\X#1{%
  \tensornode{X#1}%
  {$\tX$}%
  {{gcin}}
  {{i1, i2}}
  {{n}}
  {{}}
}%
\begin{tikzpicture}[ultra thick, opacity=0.3, index/.append style={fill opacity=1, text opacity=0.3}]%
  \matrix[row sep=3ex,column sep=0ex]{
    & \P{1}{1} & & \P{1}{2} \\
    & \P{2}{1} & & \P{2}{2} \\
    \X{1} & & \begin{scope}[xshift=6ex] \X{2} \end{scope} & \\
  };%
  \draw %
  \contract{X1}{P1-1}{i1}{$i_1$}{out=90, in=180}%
  \contract{X1}{P2-1}{i2}{$i_2$}{out=90, in=180}%
  \contract{X2}{P2-2}{i2}{$i'_2$}{out=90, in=180};%
  \begin{scope}[index/.append style={yshift=-3.5ex}]
    \draw \contract{X2}{P1-2}{i1}{$i'_1$}{out=90, in=180};%
  \end{scope}
  \begin{scope}[opacity=1, index/.append style={text opacity=1}]
    \draw%
    \contract{P1-1}{P1-2}{o1}{$o_1$}{out=0, in=180}%
    \contract{P2-1}{P2-2}{o2}{$o_2$}{out=0, in=180}%
    (P1-1-k1) node[index, right] {$k_1$}%
    (P1-2-k1) node[index, right] {$k'_1$}%
    (P2-1-k2) node[index, right] {$k_2$}%
    (P2-2-k2) node[index, right] {$k'_2$}%
    ;%
    \coordinate (n) at ($(X2-n)+(-2.5ex, 2.5ex)$);%
    \draw[maincolor]%
    (X1-n) |- (n)%
    (X2-n) |- (n) node [index] {$n$}%
    ;%

    \coordinate (X1-gcin-ungroup) at ($(X1-gcin)+(0ex, 0)$);%
    \draw (X1-gcin) to (X1-gcin-ungroup);%
    \coordinate (X1-cin) at ($(X1-gcin-ungroup)+(3ex,-1ex)$);%
    \coordinate (X1-g) at ($(X1-gcin-ungroup)+(3ex,1ex)$);%
    \draw [secondcolor, out=30, in=180] (X1-gcin-ungroup) to (X1-g);%
    \draw [out=330, in=180] (X1-gcin-ungroup) to (X1-cin) node [index, right] {$\tilde{c}_{\text{in}}$};%
    \draw[arrows=-{Triangle}] (X1-gcin-ungroup) to ++(-0.1ex,0);%

    \coordinate (X2-gcin-ungroup) at ($(X2-gcin)+(0ex, 0)$);%
    \draw (X2-gcin) to (X2-gcin-ungroup);%
    \coordinate (X2-cin) at ($(X2-gcin-ungroup)+(3ex,-1ex)$);%
    \coordinate (X2-g) at ($(X2-gcin-ungroup)+(3ex,1ex)$);%
    \draw [secondcolor, out=30, in=180] (X2-gcin-ungroup) to (X2-g);%
    \draw [out=330, in=180] (X2-gcin-ungroup) to (X2-cin) node [index, right] {$\tilde{c}'_{\text{in}}$};%
    \draw[arrows=-{Triangle}] (X2-gcin-ungroup) to ++(-0.1ex,0);%

    \coordinate (g) at ($(X2-g)+(1.5ex, 3ex)$);%
    \draw [secondcolor]%
    (X1-g) |- (g) node [index, right] {\textcolor{secondcolor}{$g$}}%
    (X2-g) |- (g)%
    ;%
  \end{scope}
\end{tikzpicture}
    \caption{KFC/KFAC-expand factor}\label{subfig:kfc-factor-batch-groups}
  \end{subfigure}
 \hfill
 \begin{subfigure}[t]{0.49\linewidth}
   \centering
\input{fig/tensor_networks/commands.tex}
\def\P#1#2{%
  \tensornode{P#1-#2}%
  {$\tPi^{(#1)}$}%
  {{k#1}}
  {{o#1}}
  {{i#1}}
  {{}}
}%
\def\X#1{%
  \tensornode{X#1}%
  {$\tX$}%
  {{gcin}}
  {{i1, i2}}
  {{n}}
  {{}}
}%
\def\I#1{%
  \tensornode{I#1}%
  {$\vone$}%
  {{}}
  {{}}
  {{}}
  {{i}}
}%
\begin{tikzpicture}[ultra thick, opacity=0.3, index/.append style={fill opacity=1, text opacity=0.3}]%
  \matrix[row sep=0.25ex,column sep=0ex]{
    & \begin{scope}[opacity=1]\I{1}\end{scope} & & \begin{scope}[opacity=1]\I{2}\end{scope} \\
    & \P{1}{1} & & \P{1}{2} \\
    & \begin{scope}[opacity=1]\I{3}\end{scope} & & \begin{scope}[opacity=1]\I{4}\end{scope} \\
    & \P{2}{1} & & \P{2}{2} \\
    \X{1} & & \begin{scope}[xshift=6ex] \X{2} \end{scope} & \\
  };%
  \draw %
  \contract{X1}{P1-1}{i1}{$i_1$}{out=90, in=180}%
  \contract{X1}{P2-1}{i2}{$i_2$}{out=90, in=180}%
  \contract{X2}{P2-2}{i2}{$i'_2$}{out=90, in=180};%
  \begin{scope}[index/.append style={yshift=-3.5ex}]
    \draw \contract{X2}{P1-2}{i1}{$i'_1$}{out=90, in=180};%
  \end{scope}
  \begin{scope}[opacity=1, index/.append style={text opacity=1}]
    \draw%
    (I1-i) to node [midway, index, right] {$o_1$} (P1-1-o1)%
    (I3-i) to node [midway, index, right] {$o_2$} (P2-1-o2)%
    (I2-i) to node [midway, index, right] {$o_1'$} (P1-2-o1)%
    (I4-i) to node [midway, index, right] {$o_2'$} (P2-2-o2)%
    (P1-1-k1) node[index, right] {$k_1$}%
    (P1-2-k1) node[index, right] {$k'_1$}%
    (P2-1-k2) node[index, right] {$k_2$}%
    (P2-2-k2) node[index, right] {$k'_2$}%
    ;%

    \coordinate (X1-gcin-ungroup) at ($(X1-gcin)+(0ex, 0)$);%
    \draw (X1-gcin) to (X1-gcin-ungroup);%
    \coordinate (X1-cin) at ($(X1-gcin-ungroup)+(3ex,-1ex)$);%
    \coordinate (X1-g) at ($(X1-gcin-ungroup)+(3ex,1ex)$);%
    \draw [secondcolor, out=30, in=180] (X1-gcin-ungroup) to (X1-g);%
    \draw [out=330, in=180] (X1-gcin-ungroup) to (X1-cin) node [index, right] {$\tilde{c}_{\text{in}}$};%
    \draw[arrows=-{Triangle}] (X1-gcin-ungroup) to ++(-0.1ex,0);%

    \coordinate (X2-gcin-ungroup) at ($(X2-gcin)+(0ex, 0)$);%
    \draw (X2-gcin) to (X2-gcin-ungroup);%
    \coordinate (X2-cin) at ($(X2-gcin-ungroup)+(3ex,-1ex)$);%
    \coordinate (X2-g) at ($(X2-gcin-ungroup)+(3ex,1ex)$);%
    \draw [secondcolor, out=30, in=180] (X2-gcin-ungroup) to (X2-g);%
    \draw [out=330, in=180] (X2-gcin-ungroup) to (X2-cin) node [index, right] {$\tilde{c}'_{\text{in}}$};%
    \draw[arrows=-{Triangle}] (X2-gcin-ungroup) to ++(-0.1ex,0);%

    \coordinate (g) at ($(X2-g)+(1.5ex, 1.5ex)$);%
    \draw [secondcolor]%
    (X1-g) |- (g) node [index, right] {\textcolor{secondcolor}{$g$}}%
    (X2-g) |- (g)%
    ;%

    \draw[maincolor]%
    (X1-n) to ++(0, 3.25ex) node [index, right] {$n$} -| (X2-n)%
    ;%
  \end{scope}
\end{tikzpicture}
   \caption{KFAC-reduce factor}\label{subfig:kfac-reduce-factor-batch-groups}
 \end{subfigure}

 \caption{TN diagrams of input-based factors in Kronecker approximations of the
   GGN for convolutions with \textcolor{maincolor}{batching} and
   \textcolor{secondcolor}{channel groups}. They extend \Cref{fig:kfac} from the
   main text.}\label{fig:higher-order-batch-groups}
\end{figure}

\paragraph{KFC/KFAC-expand for grouped convolutions:}
Applying the regular KFC approximation to the kernel of group $g$, this yields
the Fisher approximation $\mOmega_g \otimes \mGamma_g$ with $\mGamma_g \in
\sR^{\tilde{C}_{\text{out}} \times \tilde{C}_{\text{out}}}$ and $\mOmega_g =
\nicefrac{1}{N} \sum_{n=1}^N \llbracket \tX_{n, g} \rrbracket \llbracket
\tX_{n,g} \rrbracket^\top \in \sR^{\tilde{C}_{\text{in}} K_1 K_2 \times
  \tilde{C}_{\text{in}} K_1 K_2}$ where $\tX_{n, g}$ is the input tensor for
sample $n$ and group $g$ (remember the index structure $\tX_{n, (g,
  \tilde{c}_{\text{in}}), i_1, i_2}$). \Cref{subfig:kfc-factor-batch-groups}
shows the diagram for $\{ N \mOmega_g \}_{g=1}^G$.

\paragraph{KFAC-reduce for grouped convolutions:} Proceeding in the same way,
but using the unfolded input averaged over output locations, we obtain the
Fisher approximation $\hat{\mOmega}_g \otimes \hat{\mGamma}_g$ with
$\hat{\mGamma}_g \in \sR^{\tilde{C}_{\text{out}} \times
  \tilde{C}_{\text{out}}}$ and $\hat{\mOmega}_g = \nicefrac{1}{N (O_1 O_2)^2}
\sum_{n=1}^N \vone_{O_1 O_2}^{\top}\llbracket \tX_{n, g} \rrbracket
(\vone_{O_1 O_2}^{\top}\llbracket \tX_{n,g} \rrbracket)^{\top} \in
\sR^{\tilde{C}_{\text{in}} K_1 K_2 \times \tilde{C}_{\text{in}} K_1 K_2}$ for
the kernel of group $g$. \Cref{subfig:kfac-reduce-factor-batch-groups} shows
the diagram for $\{ N (O_1 O_2)^2 \hat{\mOmega}_g \}_{g=1}^G$.

\subsection{Kronecker-factored Approximate Curvature (KFAC) for Transpose Convolution}\label{subsec:app:kfac-transpose-convolution}
Here we derive the KFAC approximation for transpose convolutions.

We describe transpose convolution in terms of its associated convolution from an input space $\gX = \sR^{C_{\text{in}}\times I_1 \times I_2}$ to an output space $\gY = \sR^{C_{\text{out}} \times O_1 \times O_2}$.
The convolution has hyper-parameters $K_{1,2}, S_{1,2}, P_{1,2}, D_{1,2}$ with index patterns $\tPi^{(1)} = \tPi(I_1, K_1, S_1, P_1, D_1) \in \sR^{I_1 \times O_1 \times K_1}$ and $\tPi^{(2)} = \tPi(I_2, K_2, S_2, P_2, D_2) \in \sR^{I_2 \times O_2 \times K_2}$.

\paragraph{Transpose convolution as matrix multiplication:}
Transpose convolution maps a $\tY \in \gY$ into an $\tX \in \gX$.
In ML frameworks like PyTorch, its kernel $\tilde{\tW}$ is stored as $C_{\text{out}}\times C_{\text{in}} \times K_1 \times K_2$ tensor.
The relation $\tX = \tilde{\tW} \star_{\text{T}} \tY$ where $\star_{\text{T}}$ denotes transpose convolution is given by \Cref{subfig:input-vjp},
\begin{align}\label{eq:app:transposed-convolution}
  \etX_{c_{\text{in}}, i_1, i_2}
  =
  \sum_{c_{\text{out}}=1}^{C_{\text{out}}}
  \sum_{k_1=1}^{K_1}
  \sum_{k_2=1}^{K_2}
  \sum_{o_1=1}^{O_1}
  \sum_{o_2=1}^{O_2}
  \etPi^{(1)}_{i_1, o_1, k_1}
  \etPi^{(2)}_{i_2, o_2, k_2}
  \etY_{c_{\text{out}}, k_1, k_2}
  \tilde{\etW}_{c_{\text{out}}, c_{\text{in}}, k_1, k_2}
\end{align}
Our goal is to turn the express the above as matrix multiplication.
To do that, we first define the matrix reshape $\mX$ of $\tX$ via $\mX \in \sR^{C_{\text{in}} \times I_1 I_2}$ such that $[\mX]_{c_{\text{in}}, (i_1, i_2)} = \etX_{c_{\text{in}}, i_1, i_2}$.
Next, we consider a transposed kernel $\tW$ of $\tilde{\tW}$ with changed order of the first two indices, i.e.\,$\tW \in \sR^{C_{\text{in}} \times C_{\text{out}} \times K_1 \times K_2}$ such that
\begin{align}
  \etW_{c_{\text{in}}, c_{\text{out}}, k_1, k_2}
  =
  \tilde{\etW}_{c_{\text{out}}, c_{\text{in}}, k_1, k_2}\,.
\end{align}
This transposition is necessary to convert the kernel's layout in the ML framework to a layout that admits \Cref{eq:app:transposed-convolution} to be expressed as matrix multiplication.
Using a matrix reshape $\mW$ of $\tW$ via $\mW \in \sR^{C_{\text{in}} \times C_{\text{out}} K_1 K_2}$ such that $[\mW]_{c_{\text{in}}, (c_{\text{out}}, k_1, k_2)} = \etW_{c_{\text{in}}, c_{\text{out}}, k_1, k_2}$, we can express \Cref{eq:app:transposed-convolution} as matrix multiplication
\begin{align}
  \label{eq:app:conv-transpose-as-matrix-multiplication}
  \mX &= \mW \llbracket \tY \rrbracket_{\text{T}}
        \shortintertext{where $\llbracket \tY \rrbracket_{\text{T}} \in \sR^{C_{\text{out}} K_1 K_2 \times I_1 I_2}$ is the \emph{transpose-unfolded input} to the transpose convolution
        (note that $\llbracket \cdot \rrbracket \neq \llbracket \cdot \rrbracket_{\text{T}}$!)}
        \label{eq:app:transpose-unfolded-input}
        [\llbracket \tY \rrbracket_{\text{T}}]_{(c_{\text{out}}, k_1, k_2), (i_1, i_2)}
  &=
    \sum_{o_1=1}^{O_1}
    \sum_{o_2=1}^{O_2}
    \etPi^{(1)}_{i_1, o_1, k_1}
    \etPi^{(2)}_{i_2, o_2, k_2}
    \etY_{c_{\text{out}}, o_1, o_2}\,.
\end{align}
To the best of our knowledge there is no API for $\llbracket \cdot \rrbracket_{\text{T}}$ in existing ML frameworks.
Our approach can provide a simple and efficient implementation of $\llbracket \cdot \rrbracket$ through the TN shown in \Cref{subfig:app:kfac-transpose-convolutions-unfold} which corresponds to \Cref{eq:app:transpose-unfolded-input}.
As \Cref{eq:app:conv-transpose-as-matrix-multiplication} is of the same form as \Cref{eq:convolution-with-unfolded-input}, it is now straightforward to write down the KFAC approximations for transpose convolution.

\begin{figure}
  \centering
  \begin{subfigure}[b]{0.2\linewidth}
    \centering
    \resizebox{\linewidth}{!}{
\input{fig/tensor_networks/commands.tex}
\begin{tikzpicture}[ultra thick]%
  \matrix[row sep=3ex,column sep=3.5ex]{
    \\
    & \P{1}
    \\
    & \P{2} \\
    \tensornode{Y}%
    {$\tY$}%
    {{cout}}
    {{o1,o2}}
    {{}}
    {{}};
    \\
  };%
  \draw %
  \contract{Y}{P1}{o1}{$o_1$}{out=90, in=170}%
  \contract{Y}{P2}{o2}{$o_2$}{out=90, in=170}%
  ;%
  \node [index, right] at (Y-cout) {$c_{\text{out}}$};%
  \node [index, right] at (P1-k1) {$k_1$};%
  \node [index, right] at (P2-k2) {$k_2$};%
  \node [index, left] at (P1-i1) {$i_1$};%
  \node [index, left] at (P2-i2) {$i_2$};%
\end{tikzpicture}
    }
    \caption{Transpose-unfolded input}\label{subfig:app:kfac-transpose-convolutions-unfold}
  \end{subfigure}
  \hfill
  \begin{subfigure}[b]{0.4\linewidth}
    \centering
    \resizebox{\linewidth}{!}{
\input{fig/tensor_networks/commands.tex}
\def\P#1#2{%
  \tensornode{P#1-#2}%
  {$\tPi^{(#1)}$}%
  {{k#1}}
  {{o#1}}
  {{i#1}}
  {{}}
}%
\def\Y#1{%
  \tensornode{Y#1}%
  {$\tY$}%
  {{cout}}
  {{o1, o2}}
  {{}}
  {{}}
}%
\begin{tikzpicture}[ultra thick, opacity=0.3, index/.append style={fill opacity=1, text opacity=0.3}]%
  \matrix[row sep=3ex,column sep=2ex]{
    & \P{1}{1} & & \P{1}{2} \\
    & \P{2}{1} & & \P{2}{2} \\
    \Y{1} & & \begin{scope}[xshift=6ex] \Y{2} \end{scope} & \\
  };%
  \draw %
  \contract{Y1}{P1-1}{o1}{$o_1$}{out=90, in=170}%
  \contract{Y1}{P2-1}{o2}{$o_2$}{out=90, in=170}%
  \contract{Y2}{P1-2}{o1}{$o'_1$}{out=90, in=170}
  \contract{Y2}{P2-2}{o2}{$o'_2$}{out=90, in=170}
  ;%
  \begin{scope}[opacity=1, index/.append style={text opacity=1}]
    \draw%
    (P1-1-i1) to ++(0, -3.5ex) -| (P1-2-i1) node[midway, index, xshift=-10ex] {$i_1$}%
    (P2-1-i2) to ++(0, -3.5ex) -| (P2-2-i2) node[midway, index, xshift=-10ex] {$i_2$}%
    (P1-1-k1) node[index, right] {$k_1$}%
    (P1-2-k1) node[index, right] {$k'_1$}%
    (P2-1-k2) node[index, right] {$k_2$}%
    (P2-2-k2) node[index, right] {$k'_2$}%
    (Y1-cout) node[index, right] {$c_\text{out}$}%
    (Y2-cout) node[index, right] {$c'_\text{out}$}%
    ;%
  \end{scope}
\end{tikzpicture}
    }
    \caption{KFAC-expand}\label{subfig:app:kfac-transpose-convolutions-expand}
  \end{subfigure}
  \hfill
  \begin{subfigure}[b]{0.35\linewidth}
    \centering
    \resizebox{\linewidth}{!}{
\input{fig/tensor_networks/commands.tex}
\def\P#1#2{%
  \tensornode{P#1-#2}%
  {$\tPi^{(#1)}$}%
  {{k#1}}
  {{o#1}}
  {{i#1}}
  {{}}
}%
\def\Y#1{%
  \tensornode{Y#1}%
  {$\tY$}%
  {{cout}}
  {{o1, o2}}
  {{}}
  {{}}
}%
\def\I#1#2{%
  \tensornode{I#1-#2}%
  {$\vone$}%
  {{}}
  {{}}
  {{i#1}}
  {{}}
}%
\begin{tikzpicture}[ultra thick, opacity=0.3, index/.append style={fill opacity=1, text opacity=0.3}]%
  \matrix[row sep=1ex,column sep=1ex]{
    & \P{1}{1} & & \P{1}{2} \\
    &
    \begin{scope}[opacity=1]
      \I{1}{1}
    \end{scope}
    & &
    \begin{scope}[opacity=1]
      \I{1}{2}
    \end{scope}
    \\
    & \P{2}{1} & & \P{2}{2} \\
    &
    \begin{scope}[opacity=1]
      \I{2}{1}
    \end{scope}
    & &
    \begin{scope}[opacity=1]
      \I{2}{2}
    \end{scope}
    \\
    \Y{1} & & \begin{scope}[xshift=6ex] \Y{2} \end{scope} & \\
  };%
  \draw %
  \contract{Y1}{P1-1}{o1}{$o_1$}{out=90, in=170}%
  \contract{Y1}{P2-1}{o2}{$o_2$}{out=90, in=170}%
  \contract{Y2}{P1-2}{o1}{$o'_1$}{out=90, in=170}
  \contract{Y2}{P2-2}{o2}{$o'_2$}{out=90, in=170}
  ;%
  \begin{scope}[opacity=1, index/.append style={text opacity=1}]
    \draw%
    \contract{I1-1}{P1-1}{i1}{$i_1$}{}%
    \contract{I1-2}{P1-2}{i1}{$i_1'$}{}%
    \contract{I2-1}{P2-1}{i2}{$i_2$}{}%
    \contract{I2-2}{P2-2}{i2}{$i_2'$}{}%
    (P1-1-k1) node[index, right] {$k_1$}%
    (P1-2-k1) node[index, right] {$k'_1$}%
    (P2-1-k2) node[index, right] {$k_2$}%
    (P2-2-k2) node[index, right] {$k'_2$}%
    (Y1-cout) node[index, right] {$c_\text{out}$}%
    (Y2-cout) node[index, right] {$c'_\text{out}$}%
    ;%
  \end{scope}
\end{tikzpicture}

    }
    \caption{KFAC-reduce}\label{subfig:app:kfac-transpose-convolutions-reduce}
  \end{subfigure}
  \caption{TNs for extending KFAC to transpose convolutions (no batching and groups).}\label{fig:app:kfac-transpose-convolutions}
\end{figure}

\paragraph{KFAC-expand:}
We will define the KFAC-expand approximation for the GGN w.r.t.\,the flattened kernel $\vw$ of $\mW$.
Note that, in practise, this approximation must be properly transformed back to the layout $\tilde{\tW}$ of the ML framework.
We have $\mG(\vw) \approx \mOmega \otimes \mGamma$, with $\mGamma \in \sR^{C_{\text{in}}\times C_{\text{in}}}$ computed from backpropagated gradients, and the input-based Kronecker factor
\begin{align}
  \mOmega
  =
  \llbracket \tY \rrbracket_{\text{T}}
  \llbracket \tY \rrbracket_{\text{T}}^{\top}
  \in \sR^{C_{\text{out}} K_1 K_2 \times C_{\text{out}} K_1 K_2}\,.
\end{align}
See \Cref{subfig:app:kfac-transpose-convolutions-expand} for the corresponding TN.

\paragraph{KFAC-reduce:}
For KFAC-reduce, we have $\mG(\vw) \approx \hat{\mOmega} \otimes \hat{\mGamma}$, with $\hat{\mGamma} \in \sR^{C_{\text{in}}\times C_{\text{in}}}$ computed from backpropagated gradients, and the input-based Kronecker factor
\begin{align}
  \hat{\mOmega}
  =
  \frac{1}{(I_1 I_2)^2}
  \left(
  \vone_{I_1 I_2}^{\top}
  \llbracket \tY \rrbracket_{\text{T}}
  \right)
  \left(
  \vone_{I_1 I_2}^{\top}
  \llbracket \tY \rrbracket_{\text{T}}
  \right)^{\top}
  \in \sR^{C_{\text{out}} K_1 K_2 \times C_{\text{out}} K_1 K_2}\,.
\end{align}
See \Cref{subfig:app:kfac-transpose-convolutions-reduce} for the corresponding TN.

\paragraph{With batching and groups:}
In the presence of $G$ groups, we have per-group kernels $\tilde{\tW}_g = [\tilde{\tW}]_{(g,:),:,:,:} \in \sR^{\nicefrac{C_{\text{out}}}{G} \times \nicefrac{C_{\text{in}}}{G} \times K_1 \times K_2}$ and $\tW_g \in \sR^{\nicefrac{C_{\text{in}}}{G} \times \nicefrac{C_{\text{out}}}{G} \times K_1 \times K_2}$, as well as per-group transpose-unfolded inputs $\llbracket \tY_g \rrbracket_{\text{T}} = {\llbracket \tY \rrbracket_{\text{T}}}_{(g,:),:,:}= \llbracket [\tY]_{(g,:),:,:} \rrbracket_{\text{T}} \in \sR^{\nicefrac{C_{\text{out}}}{G} K_1 K_2 \times I_1 I_2}$.
Each group corresponds to a transpose convolution in itself.
With batching, we have an additional leading batch dimension, i.e.\,$\llbracket \tY_{n,g} \rrbracket_{\text{T}}$.
Applying the same steps from above, we can define the KFAC approximation for the GGN w.r.t.\,the flattened per-group kernel $\vw_g$ of $\mW_g$.

For KFAC-expand, we have $\mG(\vw_g) \approx \mOmega_g \otimes \mGamma_g$, with $\mGamma_g \in \sR^{\nicefrac{C_{\text{in}}}{G}\times \nicefrac{C_{\text{in}}}{G}}$ computed from backpropagated gradients, and the input-based Kronecker factor
\begin{align*}
  \mOmega_g
  =
  \frac{1}{N}
  \sum_{n=1}^{N}
  \llbracket \tY_{n,g} \rrbracket_{\text{T}}
  \llbracket \tY_{n,g} \rrbracket_{\text{T}}^{\top}
  \in \sR^{\nicefrac{C_{\text{out}}}{G} K_1 K_2 \times \nicefrac{C_{\text{out}}}{G} K_1 K_2}\,.
\end{align*}
For KFAC-reduce, we have $\mG(\vw_g) \approx \hat{\mOmega}_g \otimes \hat{\mGamma}_g$, with $\hat{\mGamma}_g \in \sR^{\nicefrac{C_{\text{in}}}{G}\times \nicefrac{C_{\text{in}}}{G}}$ computed from backpropagated gradients, and the input-based Kronecker factor
\begin{align*}
  \hat{\mOmega}_g
  =
  \frac{1}{N (O_1 O_2)^2}
  \sum_{n=1}^{N}
  \left(
  \vone_{I_1 I_2}^{\top}
  \llbracket \tY_{n,g} \rrbracket_{\text{T}}
  \right)
  \left(
  \vone_{I_1 I_2}^{\top}
  \llbracket \tY_{n,g} \rrbracket_{\text{T}}
  \right)^{\top}
  \in \sR^{\nicefrac{C_{\text{out}}}{G} K_1 K_2 \times \nicefrac{C_{\text{out}}}{G} K_1 K_2}\,.
\end{align*}

\subsection{Further Operations \& Extensive Overview}

\paragraph{Consecutive convolutions:} We can chain two, or more, convolutions
into a single TN diagram (\Cref{fig:app:consecutive-convolutions}) to obtain a
deep linear CNN~\cite{singh2023hessian} similar to deep linear networks which
are popular for analytical studies.

\begin{figure}[t]
  \centering
  \input{fig/tensor_networks/commands.tex}
\def\Pprime#1{%
  \tensornode{Pprime#1}%
  {${\tPi\mathbf{'}}^{(#1)}$}%
  {{k#1}}
  {{o#1}}
  {{i#1}}
  {{}}
}%
\def\Wprime{%
  \tensornode{Wprime}%
  {$\tW'$}%
  {{cout}}
  {{k2, k1}}
  {{cin}}
  {{}}
}%
\begin{tikzpicture}[ultra thick, opacity=0.3, index/.append style={fill opacity=1, text opacity=0.3}]%
  \matrix[row sep=3ex,column sep=0ex]{
    & \P{1} & & \begin{scope}[opacity=1]\Pprime{1}\end{scope} \\
    & \P{2} & & \begin{scope}[opacity=1]\Pprime{2}\end{scope} \\
    \X & & \W & & \begin{scope}[opacity=1]\Wprime\end{scope} \\
  };%
  \draw %
  \contract{X}{W}{cin}{$c_{\text{in}}$}{}%
  \contract{X}{P1}{i1}{$i_1$}{out=90, in=180}%
  \contract{X}{P2}{i2}{$i_2$}{out=90, in=180}%
  \contract{P1}{W}{k1}{$k_1$}{out=0, in=90}%
  \contract{P2}{W}{k2}{$k_2$}{out=0, in=90};%
  \begin{scope}[opacity=1, index/.append style={text opacity=1}]
    \draw%
    \contract{Pprime1}{Wprime}{k1}{$k'_1$}{out=0, in=90}%
    \contract{Pprime2}{Wprime}{k2}{$k'_2$}{out=0, in=90}%
    (W-cout) -- node[index] {$c_{\text{out}}$} (Wprime-cin)%
    (P1-o1) -- node[index] {$o_1$} (Pprime1-i1)%
    (P2-o2) -- node[index] {$o_2$} (Pprime2-i2)%
    (Wprime-cout) node[index, right] {$c'_{\text{out}}$}%
    (Pprime1-o1) node[index, above] {$o'_1$}%
    (Pprime2-o2) node[index, above] {$o'_2$};%
  \end{scope}
\end{tikzpicture}
  \caption{TN of two consecutive convolutions without groups and without batch
    axis.} \label{fig:app:consecutive-convolutions}
\end{figure}

\paragraph{Convolution weight/input JVPs:} In the main text, we derived the
Jacobians of convolution (\Cref{subsec:differentiation}) which can be used to
derive the JVPs. A JVP propagates perturbations $\tV^{(\tW)} \in
\sR^{C_{\text{out}}\times C_{\text{in}} \times K_1 \times K_2}$ and $\tV^{(\tX)}
\in \sR^{C_{\text{in}} \times I_1 \times I_2}$ in the input space to
perturbations in the output space by contracting the perturbation with the
Jacobian. See \Cref{tab:app:einsum-expressions-extensive} for the general
\texttt{einsum} expressions.

\paragraph{Batched convolution weight VJP:} To obtain per-sample gradients, the
weight VJP must be carried out without summing over the batch axis which amounts
to keeping the batch index in the output index tuple.

\paragraph{VJPs and JVPs of \texttt{im2col}:} With the TN differentiation
technique described in \Cref{subsec:differentiation} we can compute the Jacobian
of the unfolding operation, then contract it with perturbations $V^{(\tX)} \in
\sR^{C_{\text{in}}\times K_1 \times K_2}$ in input space to obtain the JVP, or
with perturbations $V^{(\llbracket \tX \rrbracket)} \in \sR^{O_1 O_2
  \times C_{\text{in}} K_1 K_2}$ to obtain the VJP.

\paragraph{Approximate Hessian diagonals (HesScale/BL89):}
\citet{becker1989improving,elsayed2024revisiting} proposed approximate procedures
for the Hessian diagonal which cost roughly a gradient. They can be understood
as modifications of the Hessian backpropagation equations from
\citet{dangel2020modular}.

Consider a layer with input $\vx$, output $\vy$, and weights $\vw$ inside a
sequential feedforward neural network (for a convolutional layer, these
correspond to the flattened input, output, and kernel). To compute per-layer
Hessians of a loss $\ell$, each layer backpropagates its incoming Hessian
$\nabla_{\vy}^2\ell$ according to~\cite{dangel2020modular}
\begin{align}\label{equ:app:hbp}
  \begin{split}
    \nabla^2_{\vx}\ell
    &=
      (\mJ_{\vx}\vy)^{\top} \nabla_{\vy}^2\ell (\mJ_{\vx}\vy)
      +
      \sum_i \frac{\partial \ell}{\partial \evy_i} \nabla^2_{\vx}\evy_i\,,
    \\
    \nabla^2_{\vw}\ell
    &=
      (\mJ_{\vw}\vy)^{\top} \nabla_{\vy}^2\ell (\mJ_{\vw}\vy)
      +
      \sum_i \frac{\partial \ell}{\partial \evy_i} \nabla^2_{\vw}\evy_i\,.
  \end{split}
\end{align}
The scheme of \cite{becker1989improving,elsayed2024revisiting} imposes
diagonal structure on the backpropagated quantity. A layer receives a
backpropagated diagonal $\vd^{(\vy)}$ such that $\diag(\vd^{(\vy)}) \approx
\nabla_{\vy}^2\ell$, and backpropagates it according to \Cref{equ:app:hbp}, but
with a post-processing step to obtain a diagonal backpropagated quantity,
\begin{align}\label{equ:app:hesscale}
  \begin{split}
    \vd^{(\vx)}
    &=
      \diag \left(
      (\mJ_{\vx}\vy)^{\top} \diag(\vd^{(\vy)}) (\mJ_{\vx}\vy)
      \right)
      +
      \diag \left(
      \sum_i \frac{\partial \ell}{\partial \evy_i} \nabla^2_{\vx}\evy_i
      \right)\,,
    \\
    \vd^{(\vw)}
    &=
      \diag \left(
      (\mJ_{\vw}\vy)^{\top} \diag(\vd^{(\vw)}) (\mJ_{\vw}\vy)
      \right)
      +
      \diag
      \left(
      \sum_i \frac{\partial \ell}{\partial \evy_i} \nabla^2_{\vw}\evy_i
      \right)\,,
  \end{split}
\end{align}
where $\diag(\vd^{(\vx)}) \approx \nabla_{\vx}^2\ell$ and $\diag(\vd^{(\vw)})
\approx \nabla_{\vw}^2\ell$ is an approximation to the Hessian diagonal.

For convolutional layers, which are linear in the input and weight, the second
summands are zero due to $\nabla_{\vx}^2\evy_i = \vzero = \nabla_{\vw}^2\evy_i$.
The first terms of \Cref{equ:app:hesscale} require (i) embedding a diagonal
vector into a matrix, (ii) applying MJPs and JMPs, and (iii) extracting the
result's diagonal. Those can be expressed as a single TN. We show the diagrams
in \Cref{fig:app:hesscale}, using tensors rather than their flattened versions,
that is $(\vx, \vy, \vw, \vd^{(\vx)}, \vd^{(\vy)}, \vd^{(\vw)}) \to (\tX, \tY,
\tW, \tD^{(\tX)}, \tD^{(\tY)}, \tD^{(\tW)})$.

\begin{figure}[!t]
  \centering
  \begin{subfigure}[t]{0.495\linewidth}
    \centering
    \resizebox{\linewidth}{!}{
\input{fig/tensor_networks/commands.tex}
\def\P#1#2{%
  \tensornode{P#1-#2}%
  {$\tPi^{(#1)}$}%
  {{k#1}}
  {{o#1}}
  {{i#1}}
  {{}}
}%
\def\W#1{%
  \tensornode{W#1}%
  {$\tW$}%
  {{cout}}
  {{k2, k1}}
  {{cin}}
  {{}}
}%
\def\D{%
  \tensornode{D}%
  {$\tD^{(\tY)}$}%
  {{cout}}
  {{o1, o2}}
  {{}}
  {{}}
}%
\begin{tikzpicture}[ultra thick, opacity=0.3, index/.append style={fill opacity=1, text opacity=0.3}]%
  \matrix[row sep=3.5ex,column sep=0ex]{
    & \P{1}{1} \P{1}{2} \\
    & \P{2}{1} \P{2}{2} \\
    \begin{scope}[opacity=1, xshift=-8ex]\D\end{scope} & & \W{1}\\
  };%
  \begin{scope}[xshift=20ex]
    \matrix[row sep=3.5ex,column sep=0ex]{
      \P{1}{2} \\
      \P{2}{2} \\
      & \W{2}\\
    };%
  \end{scope}
  \draw %
  \contract{P1-1}{W1}{k1}{$k_1$}{out=0, in=90}%
  \contract{P2-1}{W1}{k2}{$k_2$}{out=0, in=90}%
  \contract{P1-2}{W2}{k1}{$k'_1$}{out=0, in=90}%
  \contract{P2-2}{W2}{k2}{$k'_2$}{out=0, in=90}%
  ;
   \begin{scope}[opacity=1, index/.append style={text opacity=1}]
     \coordinate (o1) at ($(P1-1-o1)+(-2ex, 0.5ex)$);%
     \draw (D-o1) to [out=90, in=180] (o1)%
     (P1-1-o1) |- (o1)%
     (P1-2-o1) |- (o1) node [index, left] {$o_1$}%
     ;%
     \coordinate (o2) at ($(P2-1-o2)+(-2ex, 0.5ex)$);%
     \draw (D-o2) to [out=90, in=180] (o2)%
     (P2-1-o2) |- (o2)%
     (P2-2-o2) |- (o2) node [index, left] {$o_2$}%
     ;%
     \coordinate (i1) at ($(P1-1-i1)+(-2ex, -2.75ex)$);%
     \draw%
     (P1-1-i1) |- (i1)%
     (P1-2-i1) |- (i1) node [index, left] {$i_1$}%
     ;%
     \coordinate (i2) at ($(P2-1-i2)+(-2ex, -2.75ex)$);%
     \draw%
     (P2-1-i2) |- (i2)%
     (P2-2-i2) |- (i2) node [index, left] {$i_2$}%
     ;%
     \coordinate (cout) at ($(D-cout)!0.5!(W1-cout)$);%
     \draw%
     (D-cout) to (cout)%
     (W1-cout) to ++(0, 2.75ex) -| (cout)%
     (W2-cout) to ++(0, 2.75ex) -| (cout)%
     node [index, left, xshift=-1.5ex] {$c_{\text{out}}$}%
     ;%
     \coordinate (cin) at ($(W2-cin)+(7ex, -2.75ex)$);%
     \draw%
     (W1-cin) |- (cin)%
     (W2-cin) |- (cin) node [index, right] {$c_{\text{in}}$}%
     ;%
  \end{scope}
\end{tikzpicture}
%
    }
    \caption{HesScale/BL89 input backpropagation}\label{subfig:app:hesscale-input-simple}
  \end{subfigure}
  \hfill
  \begin{subfigure}[t]{0.495\linewidth}
    \centering
    \resizebox{\linewidth}{!}{
\input{fig/tensor_networks/commands.tex}
\def\P#1#2{%
  \tensornode{P#1-#2}%
  {$\tPi^{(#1)}$}%
  {{k#1}}
  {{o#1}}
  {{i#1}}
  {{}}
}%
\def\X#1{%
  \tensornode{X#1}%
  {$\tX$}%
  {{cin}}
  {{i1, i2}}
  {{}}
  {{}}
}%
\def\D{%
  \tensornode{D}%
  {$\tD^{(\tY)}$}%
  {{cout}}
  {{o1, o2}}
  {{}}
  {{}}
}%
\begin{tikzpicture}[ultra thick, opacity=0.3, index/.append style={fill opacity=1, text opacity=0.3}]%
  \matrix[row sep=3.5ex,column sep=0ex]{
    & & \P{1}{1} \P{1}{2} \\
    & & \P{2}{1} \P{2}{2} \\
    \begin{scope}[opacity=1, xshift=-9ex]\D\end{scope} & \X{1} & \\
  };%
  \begin{scope}[xshift=22ex]
    \matrix[row sep=3.5ex,column sep=0ex]{
      &\P{1}{2} \\
      &\P{2}{2} \\
      \X{2} & \\
    };%
  \end{scope}
  \draw %
  \contract{X1}{P1-1}{i1}{$i_1$}{out=90, in=180}%
  \contract{X2}{P1-2}{i1}{$i'_1$}{out=90, in=180}%
  \contract{X1}{P2-1}{i2}{$i_2$}{out=90, in=180}%
  \contract{X2}{P2-2}{i2}{$i'_2$}{out=90, in=180}%
  ;
   \begin{scope}[opacity=1, index/.append style={text opacity=1}]
     \coordinate (o1) at ($(P1-1-o1)+(-2ex, 0.5ex)$);%
     \draw (D-o1) to [out=90, in=180] (o1)%
     (P1-1-o1) |- (o1)%
     (P1-2-o1) |- (o1) node [index, left] {$o_1$}%
     ;%
     \coordinate (o2) at ($(P2-1-o2)+(-2ex, 0.5ex)$);%
     \draw (D-o2) to [out=90, in=180] (o2)%
     (P2-1-o2) |- (o2)%
     (P2-2-o2) |- (o2) node [index, left] {$o_2$}%
     ;%
     \coordinate (k1) at ($(P1-2-k1)+(2ex, -2.75ex)$);%
     \draw%
     (P1-1-k1) |- (k1)%
     (P1-2-k1) |- (k1) node [index, right] {$k_1$}%
     ;%
     \coordinate (k2) at ($(P2-2-k2)+(2ex, -2.75ex)$);%
     \draw%
     (P2-1-k2) |- (k2)%
     (P2-2-k2) |- (k2) node [index, right] {$k_2$}%
     ;%
     \node [index, right] at (D-cout) {$c_{\text{out}}$};%
     \coordinate (cin) at ($(X2-cin)+(2ex, 2.75ex)$);%
     \draw%
     (X1-cin) |- (cin)%
     (X2-cin) |- (cin) node [index, right] {$c_{\text{in}}$}%
     ;%
  \end{scope}
\end{tikzpicture}
    }
    \caption{HesScale/BL89 weight backpropagation}\label{subfig:app:hesscale-weight-simple}
  \end{subfigure}

  \vspace{1ex}

  \begin{subfigure}[t]{0.495\linewidth}
    \centering
    \resizebox{\linewidth}{!}{
\input{fig/tensor_networks/commands.tex}
\def\P#1#2{%
  \tensornode{P#1-#2}%
  {$\tPi^{(#1)}$}%
  {{k#1}}
  {{o#1}}
  {{i#1}}
  {{}}
}%
\def\W#1{%
  \tensornode{W#1}%
  {$\tW$}%
  {{gcout}}
  {{k2, k1}}
  {{cin}}
  {{}}
}%
\def\D{%
  \tensornode{D}%
  {$\tD^{(\tY)}$}%
  {{gcout}}
  {{o1, o2}}
  {{n}}
  {{}}
}%
\begin{tikzpicture}[ultra thick, opacity=0.3, index/.append style={fill opacity=1, text opacity=0.3}]%
  \matrix[row sep=3.5ex,column sep=0ex]{
    & \P{1}{1} \P{1}{2} \\
    & \P{2}{1} \P{2}{2} \\
    \begin{scope}[opacity=1, xshift=-8ex]\D\end{scope} & & \W{1}\\
  };%
  \begin{scope}[xshift=20ex]
    \matrix[row sep=3.5ex,column sep=0ex]{
      \P{1}{2} \\
      \P{2}{2} \\
      & \W{2}\\
    };%
  \end{scope}
  \draw %
  \contract{P1-1}{W1}{k1}{$k_1$}{out=0, in=90}%
  \contract{P2-1}{W1}{k2}{$k_2$}{out=0, in=90}%
  \contract{P1-2}{W2}{k1}{$k'_1$}{out=0, in=90}%
  \contract{P2-2}{W2}{k2}{$k'_2$}{out=0, in=90}%
  ;
   \begin{scope}[opacity=1, index/.append style={text opacity=1}]
     \coordinate (o1) at ($(P1-1-o1)+(-2ex, 0.5ex)$);%
     \draw (D-o1) to [out=90, in=180] (o1)%
     (P1-1-o1) |- (o1)%
     (P1-2-o1) |- (o1) node [index, left] {$o_1$}%
     ;%
     \coordinate (o2) at ($(P2-1-o2)+(-2ex, 0.5ex)$);%
     \draw (D-o2) to [out=90, in=180] (o2)%
     (P2-1-o2) |- (o2)%
     (P2-2-o2) |- (o2) node [index, left] {$o_2$}%
     ;%
     \coordinate (i1) at ($(P1-1-i1)+(-2ex, -2.75ex)$);%
     \draw%
     (P1-1-i1) |- (i1)%
     (P1-2-i1) |- (i1) node [index, left] {$i_1$}%
     ;%
     \coordinate (i2) at ($(P2-1-i2)+(-2ex, -2.75ex)$);%
     \draw%
     (P2-1-i2) |- (i2)%
     (P2-2-i2) |- (i2) node [index, left] {$i_2$}%
     ;%
     \node [index, left] at (D-n) {$\textcolor{maincolor}{n}$};%

     \coordinate (W1-gcout-ungroup) at ($(W1-gcout)+(0ex, 0)$);%
     \draw (W1-gcout) to (W1-gcout-ungroup);%
     \coordinate (W1-cout) at ($(W1-gcout-ungroup)+(3ex,-1ex)$);%
     \coordinate (W1-g) at ($(W1-gcout-ungroup)+(3ex,1ex)$);%
     \draw [secondcolor, out=30, in=180] (W1-gcout-ungroup) to (W1-g);%
     \draw [out=330, in=180] (W1-gcout-ungroup) to (W1-cout);%
     \draw[arrows=-{Triangle}] (W1-gcout-ungroup) to ++(-0.1ex,0);%

     \coordinate (W2-gcout-ungroup) at ($(W2-gcout)+(0ex, 0)$);%
     \draw (W2-gcout) to (W2-gcout-ungroup);%
     \coordinate (W2-cout) at ($(W2-gcout-ungroup)+(3ex,-1ex)$);%
     \coordinate (W2-g) at ($(W2-gcout-ungroup)+(3ex,1ex)$);%
     \draw [secondcolor, out=30, in=180] (W2-gcout-ungroup) to (W2-g);%
     \draw [out=330, in=180] (W2-gcout-ungroup) to (W2-cout);%
     \draw[arrows=-{Triangle}] (W2-gcout-ungroup) to ++(-0.1ex,0);%

     \coordinate (D-gcout-ungroup) at ($(D-gcout)+(0ex, 0)$);%
     \draw (D-gcout) to (D-gcout-ungroup);%
     \coordinate (D-cout) at ($(D-gcout-ungroup)+(3ex,-1ex)$);%
     \coordinate (D-g) at ($(D-gcout-ungroup)+(3ex,1ex)$);%
     \draw [secondcolor, out=30, in=180] (D-gcout-ungroup) to (D-g);%
     \draw [out=330, in=180] (D-gcout-ungroup) to (D-cout);%
     \draw[arrows=-{Triangle}] (D-gcout-ungroup) to ++(-0.1ex,0);%

     \coordinate (g) at ($(W2-g)+(1.5ex, 3.5ex)$);%
     \draw [secondcolor]%
     (W1-g) |- (g)%
     (W2-g) |- (g)%
     (D-g) |- (g)%
     node [index, right] {\textcolor{secondcolor}{$g$}}%
     ;%

     \coordinate (cout) at ($(D-cout)+(0, -1.75ex)$);%
     \draw %
     (D-cout) |- (cout)%
     (W1-cout) |- (cout)%
     (W2-cout) |- (cout) node [index, right, xshift=1ex] {$\tilde{c}_{\text{out}}$}%
     ;%

     \coordinate (cin) at ($(W2-cin)+(12.5ex, 2.5ex)$);%
     \draw %
     (W1-cin) |- (cin)%
     (W2-cin) |- (cin)%
     node [index, right] {$\tilde{c}_{\text{in}}$}%
     ;%
  \end{scope}
\end{tikzpicture}
%
    }
    \caption{HesScale/BL89 input backpropagation (+ batch, groups)}\label{subfig:app:hesscale-input}
  \end{subfigure}
  \hfill
  \begin{subfigure}[t]{0.495\linewidth}
    \centering
    \resizebox{\linewidth}{!}{
\input{fig/tensor_networks/commands.tex}
\def\P#1#2{%
  \tensornode{P#1-#2}%
  {$\tPi^{(#1)}$}%
  {{k#1}}
  {{o#1}}
  {{i#1}}
  {{}}
}%
\def\X#1{%
  \tensornode{X#1}%
  {$\tX$}%
  {{gcin}}
  {{i1, i2}}
  {{n}}
  {{}}
}%
\def\D{%
  \tensornode{D}%
  {$\tD^{(\tY)}$}%
  {{gcout}}
  {{o1, o2}}
  {{n}}
  {{}}
}%
\begin{tikzpicture}[ultra thick, opacity=0.3, index/.append style={fill opacity=1, text opacity=0.3}]%
  \matrix[row sep=3.5ex,column sep=0ex]{
    & & \P{1}{1} \P{1}{2} \\
    & & \P{2}{1} \P{2}{2} \\
    \begin{scope}[opacity=1, xshift=-11ex]\D\end{scope} & \X{1} & \\
  };%
  \begin{scope}[xshift=22ex]
    \matrix[row sep=3.5ex,column sep=0ex]{
      &\P{1}{2} \\
      &\P{2}{2} \\
      \X{2} & \\
    };%
  \end{scope}
  \draw %
  \contract{X1}{P1-1}{i1}{$i_1$}{out=90, in=180}%
  \contract{X2}{P1-2}{i1}{$i'_1$}{out=90, in=180}%
  \contract{X1}{P2-1}{i2}{$i_2$}{out=90, in=180}%
  \contract{X2}{P2-2}{i2}{$i'_2$}{out=90, in=180}%
  ;
   \begin{scope}[opacity=1, index/.append style={text opacity=1}]
     \coordinate (o1) at ($(P1-1-o1)+(-2ex, 0.5ex)$);%
     \draw (D-o1) to [out=90, in=180] (o1)%
     (P1-1-o1) |- (o1)%
     (P1-2-o1) |- (o1) node [index, left] {$o_1$}%
     ;%
     \coordinate (o2) at ($(P2-1-o2)+(-2ex, 0.5ex)$);%
     \draw (D-o2) to [out=90, in=180] (o2)%
     (P2-1-o2) |- (o2)%
     (P2-2-o2) |- (o2) node [index, left] {$o_2$}%
     ;%
     \coordinate (k1) at ($(P1-2-k1)+(2ex, -2.75ex)$);%
     \draw%
     (P1-1-k1) |- (k1)%
     (P1-2-k1) |- (k1) node [index, right] {$k_1$}%
     ;%
     \coordinate (k2) at ($(P2-2-k2)+(2ex, -2.75ex)$);%
     \draw%
     (P2-1-k2) |- (k2)%
     (P2-2-k2) |- (k2) node [index, right] {$k_2$}%
     ;%
     \coordinate (n) at ($(X1-n)+(-2ex, 4.5ex)$);%
     \draw[maincolor]%
     (X1-n) |- (n)%
     (D-n) |- (n)%
     (X2-n) |- (n) node [index, left, xshift=-1.5ex] {$n$}%
     ;%

     \coordinate (X1-gcin-ungroup) at ($(X1-gcin)+(0ex, 0)$);%
     \draw (X1-gcin) to (X1-gcin-ungroup);%
     \coordinate (X1-cin) at ($(X1-gcin-ungroup)+(3ex,-1ex)$);%
     \coordinate (X1-g) at ($(X1-gcin-ungroup)+(3ex,1ex)$);%
     \draw [secondcolor, out=30, in=180] (X1-gcin-ungroup) to (X1-g);%
     \draw [out=330, in=180] (X1-gcin-ungroup) to (X1-cin);%
     \draw[arrows=-{Triangle}] (X1-gcin-ungroup) to ++(-0.1ex,0);%

     \coordinate (X2-gcin-ungroup) at ($(X2-gcin)+(0ex, 0)$);%
     \draw (X2-gcin) to (X2-gcin-ungroup);%
     \coordinate (X2-cin) at ($(X2-gcin-ungroup)+(3ex,-1ex)$);%
     \coordinate (X2-g) at ($(X2-gcin-ungroup)+(3ex,1ex)$);%
     \draw [secondcolor, out=30, in=180] (X2-gcin-ungroup) to (X2-g);%
     \draw [out=330, in=180] (X2-gcin-ungroup) to (X2-cin);%
     \draw[arrows=-{Triangle}] (X2-gcin-ungroup) to ++(-0.1ex,0);%

     \coordinate (D-gcout-ungroup) at ($(D-gcout)+(0ex, 0)$);%
     \draw (D-gcout) to (D-gcout-ungroup);%
     \coordinate (D-cout) at ($(D-gcout-ungroup)+(3ex,-1ex)$);%
     \coordinate (D-g) at ($(D-gcout-ungroup)+(3ex,1ex)$);%
     \draw [secondcolor, out=30, in=180] (D-gcout-ungroup) to (D-g);%
     \draw [out=330, in=180] (D-gcout-ungroup) to (D-cout)  node [index, right] {$\tilde{c}_{\text{out}}$};%
     \draw[arrows=-{Triangle}] (D-gcout-ungroup) to ++(-0.1ex,0);%

     \coordinate (g) at ($(X2-g)+(1.5ex, 1.5ex)$);%
     \draw [secondcolor]%
     (X1-g) |- (g)%
     (X2-g) |- (g)%
     (D-g) |- (g)%
     node [index, right] {\textcolor{secondcolor}{$g$}}%
     ;%

     \coordinate (cin) at ($(X2-cin)+(1.5ex, -1.5ex)$);%
     \draw %
     (X1-cin) |- (cin)%
     (X2-cin) |- (cin)%
     node [index, right] {$\tilde{c}_{\text{in}}$}%
     ;%
  \end{scope}
\end{tikzpicture}
%
    }
    \caption{HesScale/BL89 weight backpropagation (+ batch, groups)}\label{subfig:app:hesscale-weight}
  \end{subfigure}

  \caption{TN diagrams for
    HesScale/BL89~\cite{becker1989improving,elsayed2024revisiting}
    backpropagations through convolutional layers to approximate the Hessian
    diagonals $\tD^{(\tX)}, \tD^{(\tW)}$. JMPs and MJPs are shaded.
    (\subref{subfig:app:hesscale-input-simple},
    \subref{subfig:app:hesscale-weight-simple}) show the simple versions without
    batching and without channel groups. (\subref{subfig:app:hesscale-input},
    \subref{subfig:app:hesscale-weight}) include \textcolor{maincolor}{batching} and
    \textcolor{secondcolor}{channel groups}.}\label{fig:app:hesscale}
\end{figure}


\clearpage
\begin{landscape}
\newcolumntype{b}{X} 
\newcolumntype{s}{>{\hsize=.4\hsize}X} 
\begin{table}
  \centering%
  \tikzset{%
    einsum/.style={%
      inner sep=0pt,%
      font=\ttfamily\bfseries\color{secondcolor},%
      align=left,%
      scale=0.8,%
    }%
  }%
  \caption{Extensive list of convolution and related operations (extension from
    \Cref{tab:einsum-expressions} in the main text). All operations consider two
    spatial dimensions and support batching and channel groups. Generalization
    to other dimensions follow by introducing more spatial indices \texttt{i\_3,
      o\_3, \dots} and kernel indices \texttt{k\_3,
      \dots}.}\label{tab:app:einsum-expressions-extensive}
  \begin{scriptsize}
    \begin{tabularx}{1.6\textheight}{ssb}
      \toprule
      \textbf{Operation}
      & \textbf{Operands}
      & \textbf{Contraction string (\texttt{einops}}~\cite{rogozhnikov2022einops} \textbf{convention)}
      \\ \midrule
      Convolution (no bias)~\cite{hayashi2019einconv}
      & $\tX, \tPi^{(1)}, \tPi^{(2)}, \tW$
      & \tikz[baseline=-0.5ex]{%
        \node[einsum]{%
        "n (g c\_in) i1 i2, i1 o1 k1, i2 o2 k2, (g c\_out) c\_in k1 k2
        -> n (g c\_out) o1 o2"%
        }}%
      \\
      Unfolded input (\texttt{im2col}, $\llbracket \tX \rrbracket$)
      & $\tX, \tPi^{(1)}, \tPi^{(2)}$
      & \tikz[baseline=-0.5ex]{%
        \node[einsum]{%
        "n c\_in i1 i2, i1 o1 k1, i2 o2 k2
        -> n (c\_in k1 k2) (o1 o2)"%
        }}%
      \\
      Unfolded kernel (Toeplitz)
      & $\tPi^{(1)}, \tPi^{(2)}, \tW$
      & \tikz[baseline=-0.5ex]{%
        \node[einsum]{%
        "i1 o1 k1, i2 o2 k2, c\_out c\_in k1 k2
        -> (c\_out o1 o2) (c\_in i1 i2)"%
        }}%
      \\
      Folded output (\texttt{col2im})
      & $\tY, \tPi^{(1)}, \tPi^{(2)}$
      & \tikz[baseline=-0.5ex]{%
        \node[einsum]{%
        "n (g c\_out) o1 o2, i1 o1 k1, i2 o2 k2
        -> n (g c\_in) i1 i2"%
        }}%
      \\
      Transpose-unfolded input ($\llbracket \tY \rrbracket_{\text{T}}$)
      & $\tY, \tPi^{(1)}, \tPi^{(2)}$
      & \tikz[baseline=-0.5ex]{%
        \node[einsum]{%
        "n (g c\_out) o1 o2, i1 o1 k1, i2 o2 k2
        -> n (g c\_in k1 k2) i1 i2"%
        }}%
      \\
      \midrule
      Convolution weight VJP
      & $\tX, \tPi^{(1)}, \tPi^{(2)}, \tV^{(\tY)}$
      & \tikz[baseline=-0.5ex]{%
        \node[einsum]{%
        "n (g c\_in) i1 i2, i1 o1 k1, i2 o2 k2, n (g c\_out) o1 o2
        -> c\_out c\_in k1 k2"%
        }}%
      \\
      Convolution input VJP (transpose convolution)
      & $\tW, \tPi^{(1)}, \tPi^{(2)}, \tV^{(\tY)}$
      & \tikz[baseline=-0.5ex]{%
        \node[einsum]{%
        "(g c\_out) c\_in k1 k2, i1 o1 k1, i2 o2 k2, n (g c\_out) o1 o2
        -> n (g c\_in) i1 i2"%
        }}%
      \\
      Convolution weight VJP (per-sample/batched)~\cite{rochette2019efficient}
      & $\tX, \tPi^{(1)}, \tPi^{(2)}, \tV^{(\tY)}$
      & \tikz[baseline=-0.5ex]{%
        \node[einsum]{%
        "n (g c\_in) i1 i2, i1 o1 k1, i2 o2 k2, n (g c\_out) o1 o2 %
        -> n (g c\_out) c\_in k1 k2"%
        }}%
      \\
      \midrule
      Convolution weight JVP
      & $\tX, \tPi^{(1)}, \tPi^{(2)}, \tV^{(\tW)}$
      & \tikz[baseline=-0.5ex]{%
        \node[einsum]{%
        "n (g c\_in) i1 i2, i1 o1 k1, i2 o2 k2, (g c\_out) c\_in k1 k2 %
        -> n (g c\_out) o1 o2"%
        }}%
      \\
      Convolution input JVP
      & $\tW, \tPi^{(1)}, \tPi^{(2)}, \tV^{(\tX)}$
      & \tikz[baseline=-0.5ex]{%
        \node[einsum]{%
        "(g c\_out) c\_in i1 i2, i1 o1 k1, i2 o2 k2, n (g c\_in) i1 i2 %
        -> n (g c\_out) o1 o2"%
        }}%
      \\
      \midrule
      \texttt{im2col} VJP
      & $\tPi^{(1)}, \tPi^{(2)}, \tV^{(\llbracket \tX \rrbracket)}$
      & \tikz[baseline=-0.5ex]{%
        \node[einsum]{%
        "i1 o1 k1, i2 o2 k2, n (c\_in k1 k2) (o1 o2) %
        -> n c\_in i1 i2"%
        }}%
      \\
      \texttt{im2col} JVP
      & $\tPi^{(1)}, \tPi^{(2)}, \tV^{(\tX)}$
      & \tikz[baseline=-0.5ex]{%
        \node[einsum]{%
        "i1 o1 k1, i2 o2 k2, n c\_in i1 i2 %
        -> n (c\_in k1 k2) (o1 o2)"%
        }}%
      \\
      \midrule
      KFC/KFAC-expand~\cite{grosse2016kroneckerfactored,eschenhagen2023kroneckerfactored}
      & $\tX, \tPi^{(1)}, \tPi^{(2)}, \tX, \tPi^{(1)}, \tPi^{(2)}$
      & \tikz[baseline=-0.5ex]{%
        \node[einsum]{%
        "n (g c\_in) i1 i2, i1 o1 k1, i2 o2 k2,
        n (g c\_in\_) i1\_ i2\_, i1\_ o1 k1\_, i2\_ o2 k2\_\\
      \ -> g (c\_in k1 k2) (c\_in\_ k1\_ k2\_)"%
      }}%
      \\
      KFAC-reduce~\cite{eschenhagen2023kroneckerfactored}
      & $\tX, \tPi^{(1)}, \tPi^{(2)}, \tX, \tPi^{(1)}, \tPi^{(2)}$
      & \tikz[baseline=-0.5ex]{%
        \node[einsum]{%
        "n (g c\_in) i1 i2, i1 o1 k1, i2 o2 k2,
        n (g c\_in\_) i1\_ i2\_, i1\_ o1\_ k1\_, i2\_ o2\_ k2\_\\
      \ -> g (c k1 k2) (c\_ k1\_ k2\_)"%
      }}%
      \\
      KFC/KFAC-expand for transpose convolution
      & $\tY, \tPi^{(1)}, \tPi^{(2)}, \tY, \tPi^{(1)}, \tPi^{(2)}$
      & \tikz[baseline=-0.5ex]{%
        \node[einsum]{%
        "n (g c\_out) o1 o2, i1 o1 k1, i2 o2 k2,
        n (g c\_out\_) o1\_ o2\_, i1\_ o1 k1\_, i2\_ o2 k2\_\\
      \ -> g (c\_out k1 k2) (c\_out\_ k1\_ k2\_)"%
      }}%
      \\
      KFAC-reduce for transpose convolution
      & $\tY, \tPi^{(1)}, \tPi^{(2)}, \tY, \tPi^{(1)}, \tPi^{(2)}$
      & \tikz[baseline=-0.5ex]{%
        \node[einsum]{%
        "n (g c\_out) o1 o2, i1 o1 k1, i2 o2 k2,
        n (g c\_out\_) o1\_ o2\_, i1\_ o1\_ k1\_, i2\_ o2\_ k2\_\\
      \ -> g (c\_out k1 k2) (c\_out\_ k1\_ k2\_)"%
      }}%
      \\
      \midrule
      GGN Gram/empirical NTK matrix~\cite{dangel2022vivit,osawa2023asdl,novak2022fast}
      & $\tX, \tPi^{(1)}, \tPi^{(2)}, \tS^{(\tY)}, \tX, \tPi^{(1)}, \tPi^{(2)}, \tS^{(\tY)}$
      & \tikz[baseline=-0.5ex]{%
        \node[einsum]{%
        "n (g c\_in) i1 i2, i1 o1 k1, i2 o2 k2,
        c n (g c\_out) o1 o2, n\_ (g c\_in) i1\_ i2\_, i1\_ o1\_ k1,\\
      \ i2\_ o2\_ k2, c\_ n\_ (g c\_out) o1\_ o2\_
      -> (c n) (c\_ n\_)"%
      }}%
      \\
      GGN/Fisher diagonal~\cite{dangel2020backpack,osawa2023asdl}
      & $\tX, \tPi^{(1)}, \tPi^{(2)}, \tS^{(\tY)}, \tX, \tPi^{(1)}, \tPi^{(2)}, \tS^{(\tY)}$
      & \tikz[baseline=-0.5ex]{%
        \node[einsum]{%
        "n (g c\_in) i1 i2, i1 o1 k1, i2 o2 k2, c n (g c\_out) o1 o2,
        n (g c\_in) i1\_ i2\_, i1\_ o1\_ k1, \\
      \ i2\_ o2\_ k2,
      c n (g c\_out) o1\_ o2\_ -> (g c\_out) c\_in k1 k2"%
      }}%
      \\
      GGN/Fisher diagonal (per-sample/batched)
      & $\tX, \tPi^{(1)}, \tPi^{(2)}, \tS^{(\tY)}, \tX, \tPi^{(1)}, \tPi^{(2)}, \tS^{(\tY)}$
      & \tikz[baseline=-0.5ex]{%
        \node[einsum]{%
        "n (g c\_in) i1 i2, i1 o1 k1, i2 o2 k2, c n (g c\_out) o1 o2,
        n (g c\_in) i1\_ i2\_, i1\_ o1\_ k1,\\
      \ i2\_ o2\_ k2, c n (g c\_out) o1\_ o2\_ -> n (g c\_out) c\_in k1 k2"%
      }}%
      \\
      \midrule
      Approximate weight Hessian diagonal~\cite{becker1989improving,elsayed2024revisiting}
      & $\tX, \tPi^{(1)}, \tPi^{(2)}, \tD^{(\tY)}, \tX, \tPi^{(1)}, \tPi^{(2)}$
      & \tikz[baseline=-0.5ex]{%
        \node[einsum]{%
        "n (g c\_in) i1 i2, i1 o1 k1, i2 o2 k2,
        n (g c\_out) o1 o2,
        n (g c\_in) i1\_ i2\_, i1\_ o1 k1, i2\_ o2 k2\\
      \ -> (g c\_out) c\_in k1 k2"%
      }}%
      \\
      Approximate input Hessian diagonal~\cite{becker1989improving,elsayed2024revisiting}
      & $\tW, \tPi^{(1)}, \tPi^{(2)}, \tD^{(\tY)}, \tW, \tPi^{(1)}, \tPi^{(2)}$
      & \tikz[baseline=-0.5ex]{%
        \node[einsum]{%
        "(g c\_out) c\_in k1 k2, i1 o1 k1, i2 o2 k2,
        n (g c\_out) o1 o2,
        (g c\_out) c\_in k1\_ k2\_, i1 o1 k1\_,\\
      \ i2 o2 k2\_ -> n (g c\_in) i1 i2"%
      }}%
      \\
      Approximate weight Hessian diagonal (per-sample/batched)
      & $\tX, \tPi^{(1)}, \tPi^{(2)}, \tD^{(\tY)}, \tX, \tPi^{(1)}, \tPi^{(2)}$
      & \tikz[baseline=-0.5ex]{%
        \node[einsum]{%
        "n (g c\_in) i1 i2, i1 o1 k1, i2 o2 k2,
        n (g c\_out) o1 o2,
        n (g c\_in) i1\_ i2\_, i1\_ o1 k1, i2\_ o2 k2\\
      \ -> n (g c\_out) c\_in k1 k2"%
      }}%
      \\
      \bottomrule
    \end{tabularx}
  \end{scriptsize}
\end{table}


\end{landscape}
\clearpage

\clearpage

\section{Exact Second-Order Information}\label{sec:app:second-order}
Here, we look at computing second-order information of a loss w.r.t.\, to the kernel of a convolution.
Its computation can be phrased as backpropagation with a final extraction step~\cite{dangel2023backpropagation} which contains less standard operations like Jacobian-matrix products (JMPs) and sub-tensor extraction.
TNs can express this extraction step in a single diagram.

Consider a datum $(\vx, \vt)$ and its loss $\ell(\vw) = \ell(\vf, \vt)$ where
$\vf := f_{\vw}(\vx) \in \sR^C$ is the prediction of a CNN with a convolution
with flattened kernel $\vw$ and flattened output $\vy$ (derivations carry over
to a batch loss). The kernel's generalized Gauss-Newton (GGN)
matrix~\cite{schraudolph2002fast} $\mG(\vw) = ( \mJ_{\vw}\vf )^{\top}
\nabla_{\vf}^2 \ell ( \mJ_{\vw}\vf ) \in \sR^{C_{\text{out}} C_{\text{in}} K_1
  K_2 \times C_{\text{out}} C_{\text{in}} K_1 K_2 }$ is a positive semi-definite
Hessian proxy preferred by many
applications~\citep[e.g.][]{daxberger2021laplace,martens2010deep} and coincides
with the Fisher information matrix for many common losses~\cite{martens2020new}.
It is the self-outer product of a backpropagated symmetric factorization
$\mS^{(\vy)} = (\mJ_{\vy} \vf)^{\top} \mS^{(\vf)} \in \sR^{C_{\text{out}} O_1
  O_2 \times C}$ of the loss Hessian, $\nabla_{\vf}^2 \ell(\vf, \vy) =
\mS^{(\vf)} (\mS^{(\vf)})^{\top}$. During backpropagation, the convolution
extracts information about $\mG(\vw) = (\mJ_{\vw} \vy)^{\top} \mS^{(\vy)}
(\mS^{(\vy)})^{\top} \mJ_{\vw}\vy$.

In TN notation, this is easy to express without flattening: We simply compose
two VJP diagrams from \Cref{subfig:weight-vjp} with an extra leg (MJP) and add
the outer-product contraction to obtain the tensor version $\tG(\tW)\in
\sR^{C_{\text{out}} \times C_{\text{in}} \times K_1 \times K_2 \times
  C_{\text{out}} \times C_{\text{in}} \times K_1 \times K_2}$ of $\mG(\vw)$
(\Cref{subfig:ggn}). The GGN is often further approximated by sub-tensors as it
is too large. These slicing operations are also easy to integrate into the
diagrams, e.g.\,to extract diagonal elements
(\Cref{subfig:ggn-diagonal}~\cite{dangel2020backpack,osawa2023asdl}), or
mini-block diagonals
(\Cref{subfig:ggn-mini-block}~\citep[][]{dangel2020modular,bahamou2023mini}).
This also removes redundant computations compared to computing, then slicing,
the matrix. The same ideas apply to the GGN Gram matrix
$(\mS^{(\vw)})^{\top}\mS^{(\vw)} \in \sR^{C \times C}$ (\Cref{subfig:ggn-gram}).
It contains the GGN spectrum~\cite{dangel2022vivit} and is related to the
empirical NTK for square loss \cite{novak2022fast}.

\begin{figure}[t]
  \centering
  \begin{subfigure}[b]{0.33\linewidth}
    \centering
    \resizebox{\linewidth}{!}{%
\input{fig/tensor_networks/commands.tex}
\def\P#1#2{%
  \tensornode{P#1-#2}%
  {$\tPi^{(#1)}$}%
  {{k#1}}
  {{o#1}}
  {{i#1}}
  {{}}
}%
\def\X#1{%
  \tensornode{X#1}%
  {$\tX$}%
  {{cin}}
  {{i1, i2}}
  {{}}
  {{}}
}%
\def\S#1{%
  \tensornode{S#1}%
  {$\tS^{(\tY)}$}%
  {{cout}}
  {{o1, o2}}
  {{}}
  {{c}}
}%
\begin{tikzpicture}[ultra thick, opacity=0.3, index/.append style={fill opacity=1, text opacity=0.3}]%
  \matrix[row sep=1ex,column sep=0ex]{
    & & \P{1}{1} & \\%
    & & \P{2}{1} & \\%
    & \X{1} & & \\%
    \begin{scope}[opacity=1]\S{1}\end{scope} & & \\%
  };%
  \begin{scope}[xshift=22ex]
    \matrix[row sep=1ex,column sep=0ex]{
      & & \P{1}{2} & \coordinate (k1-low); \\
      & & \P{2}{2} & \coordinate (k2-low); \\
      &\X{2} & & \coordinate (cin-low); \\
      \begin{scope}[opacity=1]\S{2}\end{scope} & & & \coordinate (cout-low); \\
    };%
  \end{scope}
  \begin{scope}[index/.append style={xshift=-1ex, yshift=0.25ex}]
    \draw %
    \contract{S1}{P1-1}{o1}{$o_1$}{out=90, in=180}%
    \contract{S1}{P2-1}{o2}{$o_2$}{out=90, in=180}%
    \contract{S2}{P1-2}{o1}{$o'_1$}{out=90, in=180}%
    \contract{S2}{P2-2}{o2}{$o'_2$}{out=90, in=180}%
    ;%
  \end{scope}
  \begin{scope}[index/.append style={yshift=0.4ex}]
    \draw %
    \contract{X1}{P1-1}{i1}{$i_1$}{out=90, in=180}%
    \contract{X1}{P2-1}{i2}{$i_2$}{out=90, in=180}%
    \contract{X2}{P1-2}{i1}{$i'_1$}{out=90, in=180}%
    \contract{X2}{P2-2}{i2}{$i'_2$}{out=90, in=180}%
    ;%
  \end{scope}
  \begin{scope}[opacity=1, index/.append style={text opacity=1}]
    \draw %
    (S2-cout) node [index, right] {$c'_{\text{out}}$}%
    (X2-cin) node [index, right] {$c'_{\text{in}}$}%
    (P1-2-k1) node [index, right] {$k_1'$}%
    (P2-2-k2) node [index, right] {$k_2'$}%
    ;%
    \draw %
    (S1-cout) node[index, right] {$c_{\text{out}}$}%
    (X1-cin) node[index, right] {$c_{\text{in}}$}%
    (P1-1-k1) node[index, right] {$k_1$}%
    (P2-1-k2) node[index, right] {$k_2$}%
    \contract{S1}{S2}{c}{$c$}{out=0, in=180}%
    ;%
  \end{scope}
\end{tikzpicture}
    }
    \caption{GGN/Fisher}\label{subfig:ggn}
  \end{subfigure}
  \hfill
  \begin{subfigure}[b]{0.33\linewidth}
    \centering
    \resizebox{\linewidth}{!}{%
\input{fig/tensor_networks/commands.tex}
\def\P#1#2{%
  \tensornode{P#1-#2}%
  {$\tPi^{(#1)}$}%
  {{k#1}}
  {{o#1}}
  {{i#1}}
  {{}}
}%
\def\X#1{%
  \tensornode{X#1}%
  {$\tX$}%
  {{cin}}
  {{i1, i2}}
  {{}}
  {{}}
}%
\def\S#1{%
  \tensornode{S#1}%
  {$\tS^{(\tY)}$}%
  {{cout}}
  {{o1, o2}}
  {{}}
  {{c}}
}%
\begin{tikzpicture}[ultra thick, opacity=0.3, index/.append style={fill opacity=1, text opacity=0.3}]%
  \matrix[row sep=1ex,column sep=0ex]{
    & & \P{1}{1} & \\%
    & & \P{2}{1} & \\%
    & \X{1} & & \\%
    \begin{scope}[opacity=1]\S{1}\end{scope} & & \\%
  };%
  \begin{scope}[xshift=18ex]
    \matrix[row sep=1ex,column sep=0ex]{
      & & \P{1}{2} & \coordinate (k1-low); \\
      & & \P{2}{2} & \coordinate (k2-low); \\
      &\X{2} & & \coordinate (cin-low); \\
      \begin{scope}[opacity=1]\S{2}\end{scope} & & & \coordinate (cout-low); \\
    };%
  \end{scope}
  \begin{scope}[index/.append style={xshift=-1ex, yshift=0.25ex}]
    \draw %
    \contract{S1}{P1-1}{o1}{$o_1$}{out=90, in=180}%
    \contract{S1}{P2-1}{o2}{$o_2$}{out=90, in=180}%
    \contract{S2}{P1-2}{o1}{$o'_1$}{out=90, in=180}%
    \contract{S2}{P2-2}{o2}{$o'_2$}{out=90, in=180}%
    ;%
  \end{scope}
  \begin{scope}[index/.append style={yshift=0.4ex}]
    \draw %
    \contract{X1}{P1-1}{i1}{$i_1$}{out=90, in=180}%
    \contract{X1}{P2-1}{i2}{$i_2$}{out=90, in=180}%
    \contract{X2}{P1-2}{i1}{$i'_1$}{out=90, in=180}%
    \contract{X2}{P2-2}{i2}{$i'_2$}{out=90, in=180}%
    ;%
  \end{scope}
  \begin{scope}[opacity=1, index/.append style={text opacity=1}]
    \coordinate (k1) at ($(k1-low)+(0, 2.5ex)$);%
    \coordinate (k2) at ($(k2-low)+(0, 2.5ex)$);%
    \coordinate (cin) at ($(cin-low)+(0, 2.5ex)$);%
    \coordinate (cout) at ($(cout-low)+(0, 2.5ex)$);%
    \draw %
    (S2-cout) to ++(0, 2.5ex)%
    (X2-cin) to ++(0, 2.5ex)%
    (P1-2-k1) to ++(0, 2.5ex)%
    (P2-2-k2) to ++(0, 2.5ex)%
    ;%
    \draw %
    (S1-cout) |- (cout) to ++(2ex, 0) node[index, right] {$c_{\text{out}}$}%
    (X1-cin) |- (cin) to ++(2ex, 0) node[index, right] {$c_{\text{in}}$}%
    (P1-1-k1) |- (k1) to ++(2ex, 0) node[index, right] {$k_1$}%
    (P2-1-k2) |- (k2) to ++(2ex, 0) node[index, right] {$k_2$}%
    \contract{S1}{S2}{c}{$c$}{out=0, in=180}%
    ;%
  \end{scope}
\end{tikzpicture}
    }
    \caption{GGN/Fisher diagonal}\label{subfig:ggn-diagonal}
  \end{subfigure}
  \hfill
  \begin{subfigure}[b]{0.3\linewidth}
    \centering
    \resizebox{\linewidth}{!}{%
\input{fig/tensor_networks/commands.tex}
\def\P#1#2{%
  \tensornode{P#1-#2}%
  {$\tPi^{(#1)}$}%
  {{k#1}}
  {{o#1}}
  {{i#1}}
  {{}}
}%
\def\X#1{%
  \tensornode{X#1}%
  {$\tX$}%
  {{cin}}
  {{i1, i2}}
  {{}}
  {{}}
}%
\def\S#1{%
  \tensornode{S#1}%
  {$\tS^{(\tY)}$}%
  {{cout}}
  {{o1, o2}}
  {{}}
  {{c}}
}%
\begin{tikzpicture}[ultra thick, opacity=0.3, index/.append style={fill opacity=1, text opacity=0.3}]%
  \matrix[row sep=1ex,column sep=0ex]{
    & & \P{1}{1} & \\%
    & & \P{2}{1} & \\%
    & \X{1} & & \\%
    \begin{scope}[opacity=1]\S{1}\end{scope} & & \\%
  };%
  \begin{scope}[xshift=19ex]
    \matrix[row sep=1ex,column sep=0ex]{
      & & \P{1}{2} & \coordinate (k1-low); \\
      & & \P{2}{2} & \coordinate (k2-low); \\
      &\X{2} & & \coordinate (cin-low); \\
      \begin{scope}[opacity=1]\S{2}\end{scope} & & & \coordinate (cout-low); \\
    };%
  \end{scope}
  \begin{scope}[index/.append style={xshift=-1ex, yshift=0.25ex}]
    \draw %
    \contract{S1}{P1-1}{o1}{$o_1$}{out=90, in=180}%
    \contract{S1}{P2-1}{o2}{$o_2$}{out=90, in=180}%
    \contract{S2}{P1-2}{o1}{$o'_1$}{out=90, in=180}%
    \contract{S2}{P2-2}{o2}{$o'_2$}{out=90, in=180}%
    ;%
  \end{scope}
  \begin{scope}[index/.append style={yshift=0.4ex}]
    \draw %
    \contract{X1}{P1-1}{i1}{$i_1$}{out=90, in=180}%
    \contract{X1}{P2-1}{i2}{$i_2$}{out=90, in=180}%
    \contract{X2}{P1-2}{i1}{$i'_1$}{out=90, in=180}%
    \contract{X2}{P2-2}{i2}{$i'_2$}{out=90, in=180}%
    ;%
  \end{scope}
  \begin{scope}[opacity=1, index/.append style={text opacity=1, yshift = 2.5ex, xshift=1ex}]
    \draw %
    (P1-1-k1) to ++ (0, +2.5ex) -| (P1-2-k1) node [index] {$k_1$}%
    (P2-1-k2) to ++ (0, +2.5ex) -| (P2-2-k2) node [index] {$k_2$}%
    (X1-cin) to ++ (0, +2.5ex) -| (X2-cin) node [index] {$c_{\text{in}}$}%
    (S1-cout) to ++ (0, +2.5ex) -| (S2-cout) node [index] {$c_{\text{out}}$}%
    ;%
  \end{scope}
  \begin{scope}[opacity=1, index/.append style={text opacity=1}]
    \draw %
    (S1-c) to ++(-2ex, 0) node [index, left] {$c$}%
    (S2-c) to ++(-2ex, 0) node [index, left] {$c'$}%
    ;%
  \end{scope}
\end{tikzpicture}
    }
    \caption{Gram (empirical NTK)}\label{subfig:ggn-gram}
  \end{subfigure}

  \vspace{-1ex}
  \caption{TN composition and sub-tensor extraction for second-order
    information. Weight MJPs from \Cref{subfig:weight-vjp} are shaded.
    (\subref{subfig:ggn}) exact and (\subref{subfig:ggn-diagonal}) diagonal of
    the kernel's GGN (the same applies to structurally similar matrices like the
    gradient covariance~\cite{jastrzebski2020break}). (\subref{subfig:ggn-gram})
    TN of the GGN Gram matrix.}\label{fig:tensor-networks-higher-order}

  \vspace{-3ex}
\end{figure}



\clearpage

\section{Implementation Details}
\label{sec:app:implementation-details}
Here we present details on the index pattern computation, and additional
transformations.

\subsection{Index Pattern Tensor Computation for Convolutions}
\Cref{alg:index-pattern-tensor} lists pseudo-code for the index pattern
computation from the convolution hyper-parameters $K, S, P, D$, and the spatial
input dimension $I$, that is $\tPi(I, K, S, P, D)$. \emph{Unlike in the main
  text, we use 0-based indexing which is more common in numerical libraries.}
For self-consistency, we re-state the relation of the hyper-parameters to output
dimension from \citep[Relationship 15]{dumoulin2016guide},
\begin{align}\label{equ:spatial-out-dimension}
  O(I, K, S, P, D)
  =  1 +
  \left\lfloor
  \frac{I + 2P - K - (K - 1) (D - 1) }{S}
  \right\rfloor\,.
\end{align}

\begin{algorithm}
  \caption{Computing the convolution index pattern tensor $\tPi$ for a spatial
    dimension.}\label{alg:index-pattern-tensor}
  \begin{small}
    \begin{algorithmic}
      \Require Input size $I \in \sN^{+}$, kernel size $K \in \sN^{+}$, stride %
      $S\in \sN^{+}$, padding $P \in \sN^+_{0}$,
      dilation $D\in \sN^{+}$%
      \State $O \gets 1 + \left\lfloor \frac{I + 2 P - K - (K - 1) (D - 1) }{S} \right\rfloor$%
      \Comment{\texttt{Compute output dimension \citep[Relationship 15]{dumoulin2016guide}}}%
      \State $\tPi \gets \vzero_{I \times O \times K}$%
      \Comment{\texttt{Initialize index pattern tensor}}%
      \For{$o = 0, \dots, O-1$, $k = 0, \dots, K-1$}%
        \Comment{\texttt{Use 0-based indexing!}}%
        \State $i \gets k D + oS - P$%
        \Comment{\texttt{Reconstruct contributing input element}}%
        \If{$0 \le i \le I-1$}%
          \Comment{\texttt{Check in bounds}}%
          \State $\etPi_{i,o,k} \gets 1$%
        \EndIf%
      \EndFor\\%
      \Return{Index pattern tensor $\tPi \in \{0, 1\}^{I \times O \times K}$}%
    \end{algorithmic}
  \end{small}
\end{algorithm}

\subsection{Index Pattern Tensor for Standalone Transpose Convolution}
Although a transpose convolution is defined w.r.t.\,a reference convolution with
hyper-parameters $K, S, P, D$, most libraries offer standalone implementations
of transpose convolution. We describe the transpose convolution by its
associated convolution, that is as a mapping from $\sR^{C_{\text{out}} \times
  O_1 \times O_2}$ (the convolution's output space) to $\sR^{C_{\text{in}}
  \times I_1 \times I_2}$ (the convolution's input space). For convolution
with $S>1$, we cannot infer $I$ from $O, K, S, P, D$, as multiple $I$s map to
the same $O$ if $(I + 2P - K - (K-1)(D-1)) \mod S \neq 0$ (see the floor
operation in \Cref{alg:index-pattern-tensor}). We need to either supply $I$
directly, or the remainder
\begin{align*}
  A
  =
  I + 2P - K - (K - 1) (D - 1)
  - S (O - 1)
\end{align*}
(often called \texttt{output\_padding}) to make $I$ unambiguous. Then, we compute
\begin{align}
  \label{eq:unambiguous-input-size}
  I = (O - 1) S  - 2 P + K + (K - 1) (D - 1)+ A\,.
\end{align}
to get $I(O, A)$ and call \Cref{alg:index-pattern-tensor} to obtain $\tPi(I(O,
A), K, S, P, D)$.

\subsection{Details on Index Pattern Simplifications}\label{subsec:app-additional-properties}
In the following, we will assume the absence of boundary pixels that don't overlap with the kernel, that is
\begin{align}\label{equ:no-boundary-pixels}
  I + 2P - (K + (K - 1) (D - 1)) \mod S = 0\,,
\end{align}
where the floor operation in $O(I, K, S, P, D)$ is obsolete. This can always be
assured by narrowing $\tX$ before a convolution. Based on our hyper-parameter
analysis of real-world CNNs (\Cref{sec:app:hyper-parameters}), we identify:
\begin{transformation}[Dense convolutions]\label{trans:dense}
  Assume \Cref{equ:no-boundary-pixels}. For $K=S$ with default padding and
  dilation ($P=0$, $D=1$), patches are adjacent non-overlapping tiles,
  accessible by un-grouping the input index $i$ into a tuple index $(\tilde{i},
  \tilde{k})$ of size $\nicefrac{I}{K} \times K$:
  \begin{align*}
    \left[ \tPi(I, K, K, 0, 1) \right]_{i,o,k}
    =
    \left[\tPi(I, K, K, 0, 1) \right]_{(\tilde{i}, \tilde{k}), o, k}
    =
    \delta_{\tilde{i}, o} \delta_{\tilde{k},k}\,.
  \end{align*}
  Point-wise convolutions ($K=S=1$) are a special case with
  pattern $[\tPi(I, 1, 1, 0, 1)]_{i,o,k} = \delta_{i,o}$.
\end{transformation}
Point-wise convolutions with $K=S=1$ are common in DenseNets~\cite{huang2017densely},
MobileNets~\cite{howard2017mobilenets,sandler2018mobilenetv2} and
ResNets~\cite{he2016deep}. InceptionV3~\cite{szegedy2016rethinking} has 2d
`mixed dense' convolutions that are point-wise along one spatial dimension.
ConvNeXt~\cite{liu2022convnet} uses dense convolutions with $K=S \in \{2, 4\}$.

\begin{transformation}[Down-sampling convolutions]\label{trans:down-sampling}
  For $S>K$ with default padding and dilation ($P=0$, $D=1$), some elements do
  not overlap with the kernel. If the input dimension $i$ is summed, all
  participating tensors can be pruned to remove the explicit zeros. Assume $I
  \mod S = 0$. Then, pruning amounts to un-grouping $i$ into $(i', s)$ of size
  $\nicefrac{I}{S} \times S$, narrowing $s$ to $K$ entries, and grouping back
  into an index $\tilde{i}$ of size $\nicefrac{K I}{S}$. After pruning, the
  index pattern represents a dense convolution with input size
  $\nicefrac{KI}{S}$, kernel size $K$, and stride $K$. In a contraction with
  some tensor $\tV$,
  \begin{align*}
    \textstyle
    \sum_{i=1}^I
    \left[ \tV \right]_{\dots, i, \dots}
    \left[ \tPi(I, K, S>K, 0, 1)\right]_{i,o,k}
    =
    \sum_{\tilde{i}=1}^{\nicefrac{I}{S}}
    [ \tilde{\tV} ]_{\dots, \tilde{i},\dots}
    \left[ \tPi(\nicefrac{KI}{S}, K, K, 0, 1)\right]_{\tilde{i},o,k}
  \end{align*}
  with sub-tensor $ [ \tilde{\tV} ]_{\dots,\tilde{i},\dots} = [\left[ \tV
  \right]_{\dots,(i', s),\dots}]_{\dots, (:, :K),\dots}$ where $:\!K$ means
  narrowing to $K$ elements.
\end{transformation}
\Cref{trans:down-sampling} converts down-sampling convolutions to dense
convolutions, which can be further simplified with \Cref{trans:dense}. We find
down-sampling convolutions with $S=2 > K=1$ in ResNet18~\cite{he2016deep},
ResNext101~\cite{xie2017aggregated}, and WideResNet101~\cite{zagoruyko2016wide}.
Those convolutions discard 75\,\% of their input! Knowledge that an operation
only consumes a fraction of its input could be used to eliminate those `dead'
computations in preceding operations, reducing FLOPS and memory.

\begin{transformation}[Kernel-output dimension swap]\label{trans:swap-k-o-axis}
  Assume \Cref{equ:no-boundary-pixels}. Transposing kernel and output
  dimensions in an index pattern yields another index pattern with same input
  size, kernel size $O(I, K, S, P, D)$, and swapped stride and dilation:
  \begin{align*}
    \left[
    \tPi(I, K, \textcolor{maincolor}{S}, P, \textcolor{maincolor}{D})
    \right]_{i,\textcolor{maincolor}{o},\textcolor{maincolor}{k}}
    =
    \left[
    \tPi(I, O,
    \textcolor{secondcolor}{D}, P, \textcolor{secondcolor}{S})
    \right]_{i,\textcolor{secondcolor}{k},\textcolor{secondcolor}{o}}\,.
  \end{align*}
\end{transformation}
This transformation is easy to see from the symmetry of $(k, D)$ and $(o, S)$ in
\Cref{equ:index-pattern-kronecker} and $O(I,K,S,P,D)$. It converts index pattern
contractions over output into kernel dimensions, like in convolutions. An
example is the weight VJP from \Cref{subfig:weight-vjp}, which---after swapping
kernel and output dimensions---resembles the TN for convolution from
\Cref{fig:visual-abstract} with kernel $\tV$. \citet{rochette2019efficient} use
this to phrase the computation of per-example gradients as convolution.

\Cref{subsec:app-additional-properties} presents more properties of $\tPi$ based
on the sub-sampling interpretation of stride and dilation along the output and
kernel dimensions. We also provide a transformation for swapping input and
output dimensions, relating convolution and transpose convolution as described
in \cite{dumoulin2016guide}.

For completeness, we state additional index pattern tensor properties here
(using 1-based indexing):

\begin{transformation}[Sub-sampling interpretation of stride]\label{trans:stride}
  Strided convolutions ($S>1$) sub-sample non-strided convolutions along the
  output dimension, ignoring all but every $S$th
  output~\cite{dumoulin2016guide}. In other words, $\left[ \tPi(I, K,
    \textcolor{maincolor}{S}, P, D) \right]_{i,o,k} = \left[ \tPi(I, K,
    \textcolor{maincolor}{1}, P, D) \right]_{i,1 + \textcolor{maincolor}{S} (o-1),k}$ or,
  in tensor notation ($[\cdot]_{::S}$ denotes slicing with steps of $S$),
  \begin{align*}
    \tPi(I, K, \textcolor{maincolor}{S}, P, D)
    =
    \left[
    \tPi(I, K, \textcolor{maincolor}{1}, P, D)
    \right]_{:,::\textcolor{maincolor}{S},:}\,.
  \end{align*}
\end{transformation}

\begin{transformation}[Sub-sampling interpretation of
  dilation]\label{trans:dilation}
  Dilated convolutions ($D>1$) with kernel size $K$ sub-sample the kernel of a
  non-dilated convolution of kernel size $K + (D-1)(K-1)$, ignoring all but
  every $D$th kernel element. In other words, $\left[ \tPi(I, K, S, P,
    \textcolor{maincolor}{D}) \right]_{i,o,k} = \left[ \tPi(I, K + (K-1)(D-1), S, P,
    \textcolor{maincolor}{1}) \right]_{i,o, 1 + \textcolor{maincolor}{D} (k-1)}$ or, in
  tensor notation,
  \begin{align*}
    \tPi(I, K, S, P, \textcolor{maincolor}{D})
    =
    \left[
    \tPi(I, K + (K-1)(D-1), S, P, \textcolor{maincolor}{1})
    \right]_{:,:, ::\textcolor{maincolor}{D}}\,.
  \end{align*}
\end{transformation}

\begin{transformation}[Transpose convolution as
  convolution]\label{trans:swap-i-o-axis} Assume \Cref{equ:no-boundary-pixels}.
  Consider a non-strided ($S=1$), non-dilated ($D=1$) convolution with index
  pattern $\tPi(I, K, 1, P, 1)$ and output dimension $O(I, K, 1, P, 1)$.
  Transposing the spatial dimensions and flipping the kernel dimension yields
  another index pattern with modified padding $P' = K - P - 1$. In other words,
  for all $i = 1, \dots, I$, $k = 1, \dots, K$, $o = 1, \dots, O$
  \begin{align*}
    \left[ \tPi(I, K, 1, P, 1) \right]_{i, o, k}
    =
    \left[ \tPi(O, K, 1, P', 1) \right]_{o, i, K+1 - k}\,.
  \end{align*}
\end{transformation}


\section{Convolution Layer Hyper-parameter Analysis}\label{sec:app:hyper-parameters}
Here we give an overview of and characterize convolutions in popular
architectures (see \Cref{tab:app:hyperparameters}). We include moderately deep
CNNs on Fashion MNIST, CIFAR-10, and CIFAR-100 from the DeepOBS
benchmark~\cite{schneider2019deepobs}, and deep CNNs on ImageNet (AlexNet,
ResNet18, InceptionV3, MobileNetV2, ResNext101). Regarding the hyper-parameters,
we make the following observations:
\begin{itemize}
\item Many CNNs do not use a bias term. This is because the output of those
  layers feeds directly into a batch normalization layer, which is invariant
  under the addition of a bias term.
\item All investigated convolutions use default dilation.
\item Group convolutions are rarely used.
  MobileNetV2 and ConvNeXt-base (\Cref{tab:app:hyperparameters-mobilenet-v2,tab:app:hyperparameters-convnext-base}) use group convolutions that interpret each individual channel as a group.
  ResNext101 (\Cref{tab:app:hyperparameters-resnext101}) uses group convolutions that interpret a collection of channels as a group.
  ConvNeXt-base (\Cref{tab:app:hyperparameters-convnext-base}) uses dense convolutions with $P=0$ and $S = K \in \{2, 4\}$.
\item Many networks use dense convolutions, that is convolutions with unit
  kernel size $(K=1)$, unit stride $(S=1)$, and no padding $(P=0)$. These
  convolutions have a trivial index pattern and can therefore be simplified.
\item InceptionV3 (\Cref{tab:app:hyperparameters-inception-v3}) uses
  two-dimensional convolutions with one trivial dimension (`mixed dense') with
  unit kernel size, unit stride, and no padding along one direction. For this
  spatial dimension, the index pattern can be simplified.
\item ResNet18 (\Cref{tab:app:hyperparameters-resnet18}) and ResNext101
  (\Cref{tab:app:hyperparameters-resnext101}) use convolutions with $S>K$ for
  down-sampling whose kernel only overlaps with a fraction of the input. The
  index pattern can be simplified.
\end{itemize}

\clearpage
\def\hyperparameterTable#1{%
  \scalebox{0.9}{%
    \begin{tiny}
      \csvloop{%
        file=#1,%
        separator=semicolon,%
        tabular={c|ccccccccc},%
        table head={%
          \toprule \textbf{Name (count)}%
          & \textbf{Input shape}%
          & \textbf{Output shape}%
          & \textbf{Kernel}%
          & \textbf{Stride}%
          & \textbf{Padding}%
          & \textbf{Dilation}%
          & \textbf{Groups}%
          & \textbf{Bias}%
          & \textbf{Type}%
          \\\midrule%
        },%
        command={%
          \csvcoli\ (\csvcolx)%
          & \csvcolii%
          & \csvcoliii%
          & \csvcoliv%
          & \csvcolv%
          & \csvcolvi%
          & \csvcolvii%
          & \csvcolviii%
          & \csvcolix%
          & \csvcolxi%
        },%
        table foot=\bottomrule}%
    \end{tiny}
  }
}%
\def\convTypeTable#1{%
  \scalebox{0.9}[]{%
    \begin{tiny}
      \csvloop{%
        file=#1,%
        separator=semicolon,%
        tabular={c},%
        table head={%
          \toprule%
          \textbf{Type}%
          \\\midrule%
        },%
        command={%
          \csvcolxi\scalebox{0}[0.862]{\phantom{$10^{-1}$}}%
        },%
        table foot=\bottomrule}%
    \end{tiny}
  }
}%
\def\convTypeTableRebuttal#1{%
  \scalebox{0.9}{%
    \begin{tiny}
      \csvloop{%
        file=#1,%
        separator=semicolon,%
        tabular={c},%
        table head={%
          \toprule%
          \textbf{Type}%
          \\\midrule%
        },%
        command={%
          \csvcolxi%
        },%
        table foot=\bottomrule}%
    \end{tiny}
  }
}%

\def\datapath{exp/exp01_conv2d_vs_einconv2d/results/layer_details/}
\begin{table}[p]
  \centering
  \vspace{-3.5ex}
  \caption{Hyper-parameters of convolutions in different CNNs. For convolutions
    with identical hyper-parameters, we only show one instance and its
    multiplicity.}\label{tab:app:hyperparameters}
  \begin{subtable}[t]{1.0\linewidth}
    \centering
    \caption{3c3d, CIFAR-10 (3, 32, 32)}\label{tab:app:hyperparameters-cifar10-3c3d}
    \hyperparameterTable{\datapath cifar10_3c3d_in_3_32_32_out_10.csv}
  \end{subtable}
  \begin{subtable}[t]{1.0\linewidth}
    \centering
    \caption{2c2d, Fashion MNIST (1, 28, 28)}\label{tab:app:hyperparameters-fmnist-2c2d}
    \hyperparameterTable{\datapath fmnist_2c2d_in_1_28_28_out_10.csv}
  \end{subtable}
  \begin{subtable}[t]{1.0\linewidth}
    \centering
    \caption{All-CNN-C, CIFAR-100 (3, 32, 32)}\label{tab:app:hyperparameters-cifar100-allcnnc}
    \hyperparameterTable{\datapath cifar100_allcnnc_in_3_32_32_out_100.csv}
  \end{subtable}
  \begin{subtable}[t]{1.0\linewidth}
    \centering
    \caption{AlexNet, ImageNet (3, 256, 256)}\label{tab:app:hyperparameters-alexnet}
    \hyperparameterTable{\datapath alexnet_in_3_256_256_out_1000.csv}
  \end{subtable}
  \begin{subtable}[t]{1.0\linewidth}
    \centering
    \caption{ResNet18, ImageNet (3, 256, 256)}\label{tab:app:hyperparameters-resnet18}
    \hyperparameterTable{\datapath resnet18_in_3_256_256_out_1000.csv}
  \end{subtable}
  \begin{subtable}[t]{1.0\linewidth}
    \centering
    \caption{ResNext101\_32x8d, ImageNet (3, 256, 256)}\label{tab:app:hyperparameters-resnext101}
    \hyperparameterTable{\datapath resnext101_32x8d_in_3_256_256_out_1000.csv}
  \end{subtable}
  \begin{subtable}[t]{1.0\linewidth}
    \centering
    \caption{ConvNeXt-base, ImageNet (3, 256, 256)}\label{tab:app:hyperparameters-convnext-base}
    \hyperparameterTable{\datapath convnext_base_in_3_256_256_out_1000.csv}
  \end{subtable}
\end{table}
\begin{table}[p]\ContinuedFloat
    \begin{subtable}[t]{1.0\linewidth}
    \centering
    \caption{InceptionV3, ImageNet (3, 299, 299)}\label{tab:app:hyperparameters-inception-v3}
    \hyperparameterTable{\datapath inception_v3_in_3_299_299_out_1000.csv}
  \end{subtable}
  \begin{subtable}[t]{1.0\linewidth}
    \centering
    \caption{MobileNetV2, ImageNet (3, 256, 256)}\label{tab:app:hyperparameters-mobilenet-v2}
    \hyperparameterTable{\datapath mobilenet_v2_in_3_256_256_out_1000.csv}
  \end{subtable}
\end{table}

\clearpage

\section{Run Time Evaluation Details (GPU)}\label{sec:app:benchmark}

Here we provide all details on the run time evaluation from the main text. We
consider the convolutions from the CNNs from \Cref{sec:app:hyper-parameters}.
Experiments were carried out on an Nvidia Tesla T4 (16\,GB memory). We use a
batch size of 32 for the ImageNet architectures, and 128 for the others.

\subsection{Protocol \& Overview}
We compare different implementations of the same operations in PyTorch. The base
line (referenced by `PT') uses PyTorch's built-in functionalities for
convolutions and related operations, such as \texttt{torch.nn.functional.conv2d}
(forward), \texttt{torch.nn.functional.unfold} (KFC, KFAC-reduce), and PyTorch's
built-in automatic differentiation \texttt{torch.autograd.grad} (VJPs).

Our TN implementation (referenced by `TN') sets up operands and the
string-valued equation for each routine. Optionally, we can apply the
simplifications from \Cref{sec:implementation} as a post-processing step before
contraction, which yields a modified equation and operand list (`TN + opt').
Finally, we determine the contraction path using
\texttt{opt\_einsum.contract\_path} and perform the contraction with its PyTorch
back-end (\texttt{opt\_einsum.contract}). We only measure the contraction time
as in practical settings, the contraction path search would be executed once,
then cached. We also exclude final operations to obtain the correct shape or
scale (flattening, reshaping, scaling by constant) in all implementations
(including the base line).

For each operation and each convolution layer, we perform 50 independent
repetitions and report the minimum time in tables. To summarize those tables, we
extract the performance ratios, that is the TN implementation's run time divided
by the base line's. Ratios larger than 1 mean that the TN implementation is
slower, ratios smaller than 1 indicate that it is faster than the base line. We
collect those ratios for the different convolution types (general, mixed dense,
dense, sub-sampling) and display them separately using box plots. Each operation
has two boxes, corresponding to the un-simplified (TN), and the simplified (TN +
opt) implementation. For the box plots, we use \texttt{matplotlib}'s default
settings (a box extends from the first quartile to the third quartile of the
data, with a line at the median; whiskers extend from the box by 1.5x the
inter-quartile range; flier points are those past the end of the whiskers).
\Cref{fig:app:benchmark-overview} summarizes the entire GPU benchmark.
\Cref{fig:app:effect-simplifications} shows the same information with each
convolution type as an individual plot.

\begin{figure}[!t]
  \centering
  \pgfkeys{/pgfplots/zmystyle/.style={boxplotbasicstyle, grid=none, height=0.55\linewidth}}
  \input{exp/exp11_summarize_benchmark/fig/operations_conv_types/cuda_0.tex}
  \caption{Benchmark overview. We measure the performance ratios of our TN
    implementation w.r.t.\,a base line in PyTorch (PT). \textcolor{maincolor}{Blue}
    boxes show the performance ratios of TN versus PT,
    \textcolor{secondcolor}{secondcolor} boxes show the performance ratios of TN+opt
    versus PT.}
  \label{fig:app:benchmark-overview}
\end{figure}

\begin{figure}[!h]
  \pgfkeys{/pgfplots/zmystyle/.style={%
      boxplotbasicstyle,%
      height=0.8\linewidth,%
      log ticks with fixed point
    }}
  \centering
  \begin{subfigure}[t]{0.495\linewidth}
    \centering
    \caption{Mixed dense}\label{subfig:app:effect-simplifications-mixed-dense}
\begin{tikzpicture}

\definecolor{color0}{rgb}{0.12156862745098,0.466666666666667,0.705882352941177}
\definecolor{color1}{rgb}{0.0901960784313725,0.745098039215686,0.811764705882353}
\definecolor{color2}{rgb}{0.580392156862745,0.403921568627451,0.741176470588235}
\definecolor{color3}{rgb}{0.890196078431372,0.466666666666667,0.76078431372549}

\begin{axis}[
axis line style={white!80!black},
tick pos=left,
xmin=0.5, xmax=11.5,
xtick style={color=gray},
xtick={1.5,3.75,6,8.25,10.5},
xticklabels={Forward,Input VJP,Weight VJP,KFC,KFAC-red.},
ylabel={TN versus PT (logarithmic)},
ymin=0.344581849596042, ymax=3.27951210174897,
ymode=log,
zmystyle
]
\path [draw=black, ]
(axis cs:0.5,1)
--(axis cs:11.5,1);

\addplot [color0]
table {%
1 2.46023692065264
1 2.4505726759626
};
\addplot [color0]
table {%
1 2.68523201814311
1 2.9602691527339
};
\addplot [color0]
table {%
0.835 2.4505726759626
1.165 2.4505726759626
};
\addplot [color0]
table {%
0.835 2.9602691527339
1.165 2.9602691527339
};
\addplot [color0, mark=o, mark size=3, mark options={solid,fill opacity=0,draw=black}, only marks]
table {%
1 1.59138793085551
1 1.55902931467978
};
\addplot [color1]
table {%
2 2.11941964143051
2 2.11299128561488
};
\addplot [color1]
table {%
2 2.24359226442037
2 2.42270096537717
};
\addplot [color1]
table {%
1.835 2.11299128561488
2.165 2.11299128561488
};
\addplot [color1]
table {%
1.835 2.42270096537717
2.165 2.42270096537717
};
\addplot [color1, mark=o, mark size=3, mark options={solid,fill opacity=0,draw=black}, only marks]
table {%
2 1.34187038156887
2 1.36634167604259
2 2.50778708904802
};
\addplot [color0]
table {%
3.25 2.21507580188499
3.25 2.20854703470289
};
\addplot [color0]
table {%
3.25 2.57006747044357
3.25 2.79165785357732
};
\addplot [color0]
table {%
3.085 2.20854703470289
3.415 2.20854703470289
};
\addplot [color0]
table {%
3.085 2.79165785357732
3.415 2.79165785357732
};
\addplot [color0, mark=o, mark size=3, mark options={solid,fill opacity=0,draw=black}, only marks]
table {%
3.25 1.54759047558896
3.25 1.5407083891019
};
\addplot [color1]
table {%
4.25 1.95084083469693
4.25 1.91820809710367
};
\addplot [color1]
table {%
4.25 2.21912496307565
4.25 2.47712335498054
};
\addplot [color1]
table {%
4.085 1.91820809710367
4.415 1.91820809710367
};
\addplot [color1]
table {%
4.085 2.47712335498054
4.415 2.47712335498054
};
\addplot [color1, mark=o, mark size=3, mark options={solid,fill opacity=0,draw=black}, only marks]
table {%
4.25 1.23647095403953
4.25 1.29759375066617
};
\addplot [color0]
table {%
5.5 1.14323001754823
5.5 1.06391407199899
};
\addplot [color0]
table {%
5.5 1.94265472891544
5.5 2.47824944822851
};
\addplot [color0]
table {%
5.335 1.06391407199899
5.665 1.06391407199899
};
\addplot [color0]
table {%
5.335 2.47824944822851
5.665 2.47824944822851
};
\addplot [color1]
table {%
6.5 1.02553324034011
6.5 0.815138975157994
};
\addplot [color1]
table {%
6.5 1.201294802175
6.5 1.24145248621953
};
\addplot [color1]
table {%
6.335 0.815138975157994
6.665 0.815138975157994
};
\addplot [color1]
table {%
6.335 1.24145248621953
6.665 1.24145248621953
};
\addplot [color0]
table {%
7.75 0.481001751597008
7.75 0.428947696001238
};
\addplot [color0]
table {%
7.75 0.558160201068993
7.75 0.562950905252086
};
\addplot [color0]
table {%
7.585 0.428947696001238
7.915 0.428947696001238
};
\addplot [color0]
table {%
7.585 0.562950905252086
7.915 0.562950905252086
};
\addplot [color0, mark=o, mark size=3, mark options={solid,fill opacity=0,draw=black}, only marks]
table {%
7.75 1.07269443005032
7.75 1.08734621108146
};
\addplot [color1]
table {%
8.75 0.395634576133383
8.75 0.381742432018256
};
\addplot [color1]
table {%
8.75 0.419588191320375
8.75 0.419782733825825
};
\addplot [color1]
table {%
8.585 0.381742432018256
8.915 0.381742432018256
};
\addplot [color1]
table {%
8.585 0.419782733825825
8.915 0.419782733825825
};
\addplot [color1, mark=o, mark size=3, mark options={solid,fill opacity=0,draw=black}, only marks]
table {%
8.75 1.30612826047223
8.75 1.38161374883581
};
\addplot [color0]
table {%
10 0.672063608405375
10 0.514678490551516
};
\addplot [color0]
table {%
10 0.964153769253581
10 1.25178763944338
};
\addplot [color0]
table {%
9.835 0.514678490551516
10.165 0.514678490551516
};
\addplot [color0]
table {%
9.835 1.25178763944338
10.165 1.25178763944338
};
\addplot [color0, mark=o, mark size=3, mark options={solid,fill opacity=0,draw=black}, only marks]
table {%
10 1.42887647675525
};
\addplot [color1]
table {%
11 0.500394870626054
11 0.412697481777046
};
\addplot [color1]
table {%
11 0.689893985157816
11 0.89263511853432
};
\addplot [color1]
table {%
10.835 0.412697481777046
11.165 0.412697481777046
};
\addplot [color1]
table {%
10.835 0.89263511853432
11.165 0.89263511853432
};
\path [draw=color0, fill=color2]
(axis cs:0.67,2.46023692065264)
--(axis cs:1.33,2.46023692065264)
--(axis cs:1.33,2.68523201814311)
--(axis cs:0.67,2.68523201814311)
--(axis cs:0.67,2.46023692065264)
--cycle;
\path [draw=color1, fill=color3]
(axis cs:1.67,2.11941964143051)
--(axis cs:2.33,2.11941964143051)
--(axis cs:2.33,2.24359226442037)
--(axis cs:1.67,2.24359226442037)
--(axis cs:1.67,2.11941964143051)
--cycle;
\path [draw=color0, fill=color2]
(axis cs:2.92,2.21507580188499)
--(axis cs:3.58,2.21507580188499)
--(axis cs:3.58,2.57006747044357)
--(axis cs:2.92,2.57006747044357)
--(axis cs:2.92,2.21507580188499)
--cycle;
\path [draw=color1, fill=color3]
(axis cs:3.92,1.95084083469693)
--(axis cs:4.58,1.95084083469693)
--(axis cs:4.58,2.21912496307565)
--(axis cs:3.92,2.21912496307565)
--(axis cs:3.92,1.95084083469693)
--cycle;
\path [draw=color0, fill=color2]
(axis cs:5.17,1.14323001754823)
--(axis cs:5.83,1.14323001754823)
--(axis cs:5.83,1.94265472891544)
--(axis cs:5.17,1.94265472891544)
--(axis cs:5.17,1.14323001754823)
--cycle;
\path [draw=color1, fill=color3]
(axis cs:6.17,1.02553324034011)
--(axis cs:6.83,1.02553324034011)
--(axis cs:6.83,1.201294802175)
--(axis cs:6.17,1.201294802175)
--(axis cs:6.17,1.02553324034011)
--cycle;
\path [draw=color0, fill=color2]
(axis cs:7.42,0.481001751597008)
--(axis cs:8.08,0.481001751597008)
--(axis cs:8.08,0.558160201068993)
--(axis cs:7.42,0.558160201068993)
--(axis cs:7.42,0.481001751597008)
--cycle;
\path [draw=color1, fill=color3]
(axis cs:8.42,0.395634576133383)
--(axis cs:9.08,0.395634576133383)
--(axis cs:9.08,0.419588191320375)
--(axis cs:8.42,0.419588191320375)
--(axis cs:8.42,0.395634576133383)
--cycle;
\path [draw=color0, fill=color2]
(axis cs:9.67,0.672063608405375)
--(axis cs:10.33,0.672063608405375)
--(axis cs:10.33,0.964153769253581)
--(axis cs:9.67,0.964153769253581)
--(axis cs:9.67,0.672063608405375)
--cycle;
\path [draw=color1, fill=color3]
(axis cs:10.67,0.500394870626054)
--(axis cs:11.33,0.500394870626054)
--(axis cs:11.33,0.689893985157816)
--(axis cs:10.67,0.689893985157816)
--(axis cs:10.67,0.500394870626054)
--cycle;
\addplot [color0]
table {%
0.67 2.52218902995192
1.33 2.52218902995192
};
\addplot [color1]
table {%
1.67 2.17698162959808
2.33 2.17698162959808
};
\addplot [color0]
table {%
2.92 2.40880657361853
3.58 2.40880657361853
};
\addplot [color1]
table {%
3.92 2.14390734744617
4.58 2.14390734744617
};
\addplot [color0]
table {%
5.17 1.75219237449184
5.83 1.75219237449184
};
\addplot [color1]
table {%
6.17 1.06467403362364
6.83 1.06467403362364
};
\addplot [color0]
table {%
7.42 0.525203203060172
8.08 0.525203203060172
};
\addplot [color1]
table {%
8.42 0.408550537415579
9.08 0.408550537415579
};
\addplot [color0]
table {%
9.67 0.76247669149464
10.33 0.76247669149464
};
\addplot [color1]
table {%
10.67 0.592760601855747
11.33 0.592760601855747
};
\end{axis}

\end{tikzpicture}
  \end{subfigure}
  \hfill
  \begin{subfigure}[t]{0.495\linewidth}
    \centering
    \caption{Dense}\label{subfig:app:effect-simplifications-dense}
\begin{tikzpicture}

\definecolor{color0}{rgb}{0.12156862745098,0.466666666666667,0.705882352941177}
\definecolor{color1}{rgb}{0.0901960784313725,0.745098039215686,0.811764705882353}
\definecolor{color2}{rgb}{0.580392156862745,0.403921568627451,0.741176470588235}
\definecolor{color3}{rgb}{0.890196078431372,0.466666666666667,0.76078431372549}

\begin{axis}[
axis line style={white!80!black},
tick pos=left,
xmin=0.5, xmax=11.5,
xtick style={color=gray},
xtick={1.5,3.75,6,8.25,10.5},
xticklabels={Forward,Input VJP,Weight VJP,KFC,KFAC-red.},
ylabel={TN versus PT (logarithmic)},
ymin=0.137044871018717, ymax=8.97183280616081,
ymode=log,
zmystyle
]
\path [draw=black, ]
(axis cs:0.5,1)
--(axis cs:11.5,1);

\addplot [color0]
table {%
1 1.59927479757071
1 1.15219594102096
};
\addplot [color0]
table {%
1 3.19474902045781
1 5.12786927532713
};
\addplot [color0]
table {%
0.835 1.15219594102096
1.165 1.15219594102096
};
\addplot [color0]
table {%
0.835 5.12786927532713
1.165 5.12786927532713
};
\addplot [color0, mark=o, mark size=3, mark options={solid,fill opacity=0,draw=black}, only marks]
table {%
1 7.41882080972977
};
\addplot [color1]
table {%
2 1.05279922570089
2 0.732303578459289
};
\addplot [color1]
table {%
2 1.5262258918302
2 2.15893117557101
};
\addplot [color1]
table {%
1.835 0.732303578459289
2.165 0.732303578459289
};
\addplot [color1]
table {%
1.835 2.15893117557101
2.165 2.15893117557101
};
\addplot [color1, mark=o, mark size=3, mark options={solid,fill opacity=0,draw=black}, only marks]
table {%
2 2.32643190987456
};
\addplot [color0]
table {%
3.25 1.53578978507391
3.25 0.784287426635228
};
\addplot [color0]
table {%
3.25 2.3281565276142
3.25 3.50513351127121
};
\addplot [color0]
table {%
3.085 0.784287426635228
3.415 0.784287426635228
};
\addplot [color0]
table {%
3.085 3.50513351127121
3.415 3.50513351127121
};
\addplot [color0, mark=o, mark size=3, mark options={solid,fill opacity=0,draw=black}, only marks]
table {%
3.25 5.30929071184138
};
\addplot [color1]
table {%
4.25 0.954033703296296
4.25 0.752727745755602
};
\addplot [color1]
table {%
4.25 1.12765183745918
4.25 1.38167346569684
};
\addplot [color1]
table {%
4.085 0.752727745755602
4.415 0.752727745755602
};
\addplot [color1]
table {%
4.085 1.38167346569684
4.415 1.38167346569684
};
\addplot [color1, mark=o, mark size=3, mark options={solid,fill opacity=0,draw=black}, only marks]
table {%
4.25 0.68156981086126
4.25 0.659759465083098
4.25 0.591139716482069
4.25 1.40354960102706
4.25 1.45680566253413
};
\addplot [color0]
table {%
5.5 1.14711526075564
5.5 0.349067227706645
};
\addplot [color0]
table {%
5.5 3.0778358039215
5.5 4.96785152085861
};
\addplot [color0]
table {%
5.335 0.349067227706645
5.665 0.349067227706645
};
\addplot [color0]
table {%
5.335 4.96785152085861
5.665 4.96785152085861
};
\addplot [color0, mark=o, mark size=3, mark options={solid,fill opacity=0,draw=black}, only marks]
table {%
5.5 6.11390939877441
};
\addplot [color1]
table {%
6.5 0.681670231178204
6.5 0.364493528786432
};
\addplot [color1]
table {%
6.5 0.977493787964857
6.5 1.30057398954806
};
\addplot [color1]
table {%
6.335 0.364493528786432
6.665 0.364493528786432
};
\addplot [color1]
table {%
6.335 1.30057398954806
6.665 1.30057398954806
};
\addplot [color1, mark=o, mark size=3, mark options={solid,fill opacity=0,draw=black}, only marks]
table {%
6.5 0.165733032412545
};
\addplot [color0]
table {%
7.75 1.07229318439295
7.75 0.709932907038578
};
\addplot [color0]
table {%
7.75 1.80088130959883
7.75 2.25573036756196
};
\addplot [color0]
table {%
7.585 0.709932907038578
7.915 0.709932907038578
};
\addplot [color0]
table {%
7.585 2.25573036756196
7.915 2.25573036756196
};
\addplot [color0, mark=o, mark size=3, mark options={solid,fill opacity=0,draw=black}, only marks]
table {%
7.75 4.71179802358024
7.75 4.84636269104023
7.75 3.45449029271799
};
\addplot [color1]
table {%
8.75 0.625744324238036
8.75 0.25456683207541
};
\addplot [color1]
table {%
8.75 1.12657925302402
8.75 1.68495919058468
};
\addplot [color1]
table {%
8.585 0.25456683207541
8.915 0.25456683207541
};
\addplot [color1]
table {%
8.585 1.68495919058468
8.915 1.68495919058468
};
\addplot [color0]
table {%
10 0.905576938644048
10 0.779949708090248
};
\addplot [color0]
table {%
10 1.5705266802466
10 1.92305124729776
};
\addplot [color0]
table {%
9.835 0.779949708090248
10.165 0.779949708090248
};
\addplot [color0]
table {%
9.835 1.92305124729776
10.165 1.92305124729776
};
\addplot [color0, mark=o, mark size=3, mark options={solid,fill opacity=0,draw=black}, only marks]
table {%
10 6.12152196176341
10 3.69550472271328
};
\addplot [color1]
table {%
11 0.26768204873977
11 0.217261415937963
};
\addplot [color1]
table {%
11 0.53208823292818
11 0.912139373543072
};
\addplot [color1]
table {%
10.835 0.217261415937963
11.165 0.217261415937963
};
\addplot [color1]
table {%
10.835 0.912139373543072
11.165 0.912139373543072
};
\path [draw=color0, fill=color2]
(axis cs:0.67,1.59927479757071)
--(axis cs:1.33,1.59927479757071)
--(axis cs:1.33,3.19474902045781)
--(axis cs:0.67,3.19474902045781)
--(axis cs:0.67,1.59927479757071)
--cycle;
\path [draw=color1, fill=color3]
(axis cs:1.67,1.05279922570089)
--(axis cs:2.33,1.05279922570089)
--(axis cs:2.33,1.5262258918302)
--(axis cs:1.67,1.5262258918302)
--(axis cs:1.67,1.05279922570089)
--cycle;
\path [draw=color0, fill=color2]
(axis cs:2.92,1.53578978507391)
--(axis cs:3.58,1.53578978507391)
--(axis cs:3.58,2.3281565276142)
--(axis cs:2.92,2.3281565276142)
--(axis cs:2.92,1.53578978507391)
--cycle;
\path [draw=color1, fill=color3]
(axis cs:3.92,0.954033703296296)
--(axis cs:4.58,0.954033703296296)
--(axis cs:4.58,1.12765183745918)
--(axis cs:3.92,1.12765183745918)
--(axis cs:3.92,0.954033703296296)
--cycle;
\path [draw=color0, fill=color2]
(axis cs:5.17,1.14711526075564)
--(axis cs:5.83,1.14711526075564)
--(axis cs:5.83,3.0778358039215)
--(axis cs:5.17,3.0778358039215)
--(axis cs:5.17,1.14711526075564)
--cycle;
\path [draw=color1, fill=color3]
(axis cs:6.17,0.681670231178204)
--(axis cs:6.83,0.681670231178204)
--(axis cs:6.83,0.977493787964857)
--(axis cs:6.17,0.977493787964857)
--(axis cs:6.17,0.681670231178204)
--cycle;
\path [draw=color0, fill=color2]
(axis cs:7.42,1.07229318439295)
--(axis cs:8.08,1.07229318439295)
--(axis cs:8.08,1.80088130959883)
--(axis cs:7.42,1.80088130959883)
--(axis cs:7.42,1.07229318439295)
--cycle;
\path [draw=color1, fill=color3]
(axis cs:8.42,0.625744324238036)
--(axis cs:9.08,0.625744324238036)
--(axis cs:9.08,1.12657925302402)
--(axis cs:8.42,1.12657925302402)
--(axis cs:8.42,0.625744324238036)
--cycle;
\path [draw=color0, fill=color2]
(axis cs:9.67,0.905576938644048)
--(axis cs:10.33,0.905576938644048)
--(axis cs:10.33,1.5705266802466)
--(axis cs:9.67,1.5705266802466)
--(axis cs:9.67,0.905576938644048)
--cycle;
\path [draw=color1, fill=color3]
(axis cs:10.67,0.26768204873977)
--(axis cs:11.33,0.26768204873977)
--(axis cs:11.33,0.53208823292818)
--(axis cs:10.67,0.53208823292818)
--(axis cs:10.67,0.26768204873977)
--cycle;
\addplot [color0]
table {%
0.67 2.30653191880153
1.33 2.30653191880153
};
\addplot [color1]
table {%
1.67 1.21250271167177
2.33 1.21250271167177
};
\addplot [color0]
table {%
2.92 1.80523520444377
3.58 1.80523520444377
};
\addplot [color1]
table {%
3.92 1.04973041454523
4.58 1.04973041454523
};
\addplot [color0]
table {%
5.17 1.99632049216987
5.83 1.99632049216987
};
\addplot [color1]
table {%
6.17 0.874837699418893
6.83 0.874837699418893
};
\addplot [color0]
table {%
7.42 1.18634294428183
8.08 1.18634294428183
};
\addplot [color1]
table {%
8.42 0.730241212253938
9.08 0.730241212253938
};
\addplot [color0]
table {%
9.67 1.09266446799474
10.33 1.09266446799474
};
\addplot [color1]
table {%
10.67 0.33345179101999
11.33 0.33345179101999
};
\end{axis}

\end{tikzpicture}
  \end{subfigure}

  \begin{subfigure}[t]{0.495\linewidth}
    \centering
    \caption{Down-sampling}\label{subfig:app:effect-simplifications-down}
\begin{tikzpicture}

\definecolor{color0}{rgb}{0.12156862745098,0.466666666666667,0.705882352941177}
\definecolor{color1}{rgb}{0.0901960784313725,0.745098039215686,0.811764705882353}
\definecolor{color2}{rgb}{0.580392156862745,0.403921568627451,0.741176470588235}
\definecolor{color3}{rgb}{0.890196078431372,0.466666666666667,0.76078431372549}

\begin{axis}[
axis line style={white!80!black},
tick pos=left,
xmin=0.5, xmax=11.5,
xtick style={color=gray},
xtick={1.5,3.75,6,8.25,10.5},
xticklabels={Forward,Input VJP,Weight VJP,KFC,KFAC-red.},
ylabel={TN versus PT (logarithmic)},
ymin=0.356866060920604, ymax=5.06455535478399,
ymode=log,
zmystyle
]
\path [draw=black, ]
(axis cs:0.5,1)
--(axis cs:11.5,1);

\addplot [color0]
table {%
1 1.88605563239138
1 1.37392250430349
};
\addplot [color0]
table {%
1 2.83253124542281
1 4.22673130466074
};
\addplot [color0]
table {%
0.835 1.37392250430349
1.165 1.37392250430349
};
\addplot [color0]
table {%
0.835 4.22673130466074
1.165 4.22673130466074
};
\addplot [color1]
table {%
2 1.08163795848431
2 0.968470984302919
};
\addplot [color1]
table {%
2 1.36294334773553
2 1.48138841366899
};
\addplot [color1]
table {%
1.835 0.968470984302919
2.165 0.968470984302919
};
\addplot [color1]
table {%
1.835 1.48138841366899
2.165 1.48138841366899
};
\addplot [color0]
table {%
3.25 1.3340871308979
3.25 0.844344178309862
};
\addplot [color0]
table {%
3.25 1.92323726384394
3.25 2.51228605183415
};
\addplot [color0]
table {%
3.085 0.844344178309862
3.415 0.844344178309862
};
\addplot [color0]
table {%
3.085 2.51228605183415
3.415 2.51228605183415
};
\addplot [color1]
table {%
4.25 1.1655912213179
4.25 0.789541123017264
};
\addplot [color1]
table {%
4.25 1.64896924538288
4.25 2.24055722762165
};
\addplot [color1]
table {%
4.085 0.789541123017264
4.415 0.789541123017264
};
\addplot [color1]
table {%
4.085 2.24055722762165
4.415 2.24055722762165
};
\addplot [color0]
table {%
5.5 0.993701833460124
5.5 0.835399200975706
};
\addplot [color0]
table {%
5.5 2.28866774524304
5.5 2.98567624008154
};
\addplot [color0]
table {%
5.335 0.835399200975706
5.665 0.835399200975706
};
\addplot [color0]
table {%
5.335 2.98567624008154
5.665 2.98567624008154
};
\addplot [color1]
table {%
6.5 0.610222206314606
6.5 0.496035849824804
};
\addplot [color1]
table {%
6.5 0.994613327744404
6.5 1.12365043827573
};
\addplot [color1]
table {%
6.335 0.496035849824804
6.665 0.496035849824804
};
\addplot [color1]
table {%
6.335 1.12365043827573
6.665 1.12365043827573
};
\addplot [color0]
table {%
7.75 2.25150354641202
7.75 1.79970587350613
};
\addplot [color0]
table {%
7.75 3.18452685016055
7.75 4.48927330767106
};
\addplot [color0]
table {%
7.585 1.79970587350613
7.915 1.79970587350613
};
\addplot [color0]
table {%
7.585 4.48927330767106
7.915 4.48927330767106
};
\addplot [color1]
table {%
8.75 0.701584827640437
8.75 0.688925344673295
};
\addplot [color1]
table {%
8.75 0.802040784076284
8.75 0.81291370392081
};
\addplot [color1]
table {%
8.585 0.688925344673295
8.915 0.688925344673295
};
\addplot [color1]
table {%
8.585 0.81291370392081
8.915 0.81291370392081
};
\addplot [color1, mark=o, mark size=3, mark options={solid,fill opacity=0,draw=black}, only marks]
table {%
8.75 0.528109093333629
};
\addplot [color0]
table {%
10 1.95336734083687
10 1.81540267598385
};
\addplot [color0]
table {%
10 3.03443622840916
10 3.43523309658439
};
\addplot [color0]
table {%
9.835 1.81540267598385
10.165 1.81540267598385
};
\addplot [color0]
table {%
9.835 3.43523309658439
10.165 3.43523309658439
};
\addplot [color1]
table {%
11 0.423344739067368
11 0.402596989737241
};
\addplot [color1]
table {%
11 0.541241931307674
11 0.626449967671783
};
\addplot [color1]
table {%
10.835 0.402596989737241
11.165 0.402596989737241
};
\addplot [color1]
table {%
10.835 0.626449967671783
11.165 0.626449967671783
};
\path [draw=color0, fill=color2]
(axis cs:0.67,1.88605563239138)
--(axis cs:1.33,1.88605563239138)
--(axis cs:1.33,2.83253124542281)
--(axis cs:0.67,2.83253124542281)
--(axis cs:0.67,1.88605563239138)
--cycle;
\path [draw=color1, fill=color3]
(axis cs:1.67,1.08163795848431)
--(axis cs:2.33,1.08163795848431)
--(axis cs:2.33,1.36294334773553)
--(axis cs:1.67,1.36294334773553)
--(axis cs:1.67,1.08163795848431)
--cycle;
\path [draw=color0, fill=color2]
(axis cs:2.92,1.3340871308979)
--(axis cs:3.58,1.3340871308979)
--(axis cs:3.58,1.92323726384394)
--(axis cs:2.92,1.92323726384394)
--(axis cs:2.92,1.3340871308979)
--cycle;
\path [draw=color1, fill=color3]
(axis cs:3.92,1.1655912213179)
--(axis cs:4.58,1.1655912213179)
--(axis cs:4.58,1.64896924538288)
--(axis cs:3.92,1.64896924538288)
--(axis cs:3.92,1.1655912213179)
--cycle;
\path [draw=color0, fill=color2]
(axis cs:5.17,0.993701833460124)
--(axis cs:5.83,0.993701833460124)
--(axis cs:5.83,2.28866774524304)
--(axis cs:5.17,2.28866774524304)
--(axis cs:5.17,0.993701833460124)
--cycle;
\path [draw=color1, fill=color3]
(axis cs:6.17,0.610222206314606)
--(axis cs:6.83,0.610222206314606)
--(axis cs:6.83,0.994613327744404)
--(axis cs:6.17,0.994613327744404)
--(axis cs:6.17,0.610222206314606)
--cycle;
\path [draw=color0, fill=color2]
(axis cs:7.42,2.25150354641202)
--(axis cs:8.08,2.25150354641202)
--(axis cs:8.08,3.18452685016055)
--(axis cs:7.42,3.18452685016055)
--(axis cs:7.42,2.25150354641202)
--cycle;
\path [draw=color1, fill=color3]
(axis cs:8.42,0.701584827640437)
--(axis cs:9.08,0.701584827640437)
--(axis cs:9.08,0.802040784076284)
--(axis cs:8.42,0.802040784076284)
--(axis cs:8.42,0.701584827640437)
--cycle;
\path [draw=color0, fill=color2]
(axis cs:9.67,1.95336734083687)
--(axis cs:10.33,1.95336734083687)
--(axis cs:10.33,3.03443622840916)
--(axis cs:9.67,3.03443622840916)
--(axis cs:9.67,1.95336734083687)
--cycle;
\path [draw=color1, fill=color3]
(axis cs:10.67,0.423344739067368)
--(axis cs:11.33,0.423344739067368)
--(axis cs:11.33,0.541241931307674)
--(axis cs:10.67,0.541241931307674)
--(axis cs:10.67,0.423344739067368)
--cycle;
\addplot [color0]
table {%
0.67 2.45708832072198
1.33 2.45708832072198
};
\addplot [color1]
table {%
1.67 1.23494468558542
2.33 1.23494468558542
};
\addplot [color0]
table {%
2.92 1.78863774096403
3.58 1.78863774096403
};
\addplot [color1]
table {%
3.92 1.41758749431511
4.58 1.41758749431511
};
\addplot [color0]
table {%
5.17 1.33889812787125
5.83 1.33889812787125
};
\addplot [color1]
table {%
6.17 0.754098964215456
6.83 0.754098964215456
};
\addplot [color0]
table {%
7.42 2.64838507657877
8.08 2.64838507657877
};
\addplot [color1]
table {%
8.42 0.770436654590951
9.08 0.770436654590951
};
\addplot [color0]
table {%
9.67 2.35851665865151
10.33 2.35851665865151
};
\addplot [color1]
table {%
10.67 0.46295134875552
11.33 0.46295134875552
};
\end{axis}

\end{tikzpicture}
  \end{subfigure}
  \hfill
  \begin{subfigure}[t]{0.495\linewidth}
    \centering
    \caption{General}\label{subfig:app:effect-simplifications-general}
\begin{tikzpicture}

\definecolor{color0}{rgb}{0.12156862745098,0.466666666666667,0.705882352941177}
\definecolor{color1}{rgb}{0.0901960784313725,0.745098039215686,0.811764705882353}
\definecolor{color2}{rgb}{0.580392156862745,0.403921568627451,0.741176470588235}
\definecolor{color3}{rgb}{0.890196078431372,0.466666666666667,0.76078431372549}

\begin{axis}[
axis line style={white!80!black},
tick pos=left,
xmin=0.5, xmax=11.5,
xtick style={color=gray},
xtick={1.5,3.75,6,8.25,10.5},
xticklabels={Forward,Input VJP,Weight VJP,KFC,KFAC-red.},
ylabel={TN versus PT (logarithmic)},
ymin=0.0460800396890979, ymax=26.0932222695582,
ymode=log,
zmystyle
]
\path [draw=black, ]
(axis cs:0.5,1)
--(axis cs:11.5,1);

\addplot [color0]
table {%
1 3.07466396705552
1 0.525983824154647
};
\addplot [color0]
table {%
1 6.26184972902956
1 10.3484465930735
};
\addplot [color0]
table {%
0.835 0.525983824154647
1.165 0.525983824154647
};
\addplot [color0]
table {%
0.835 10.3484465930735
1.165 10.3484465930735
};
\addplot [color0, mark=o, mark size=3, mark options={solid,fill opacity=0,draw=black}, only marks]
table {%
1 11.1143814142224
1 19.5518044271115
1 15.7566575364248
};
\addplot [color1]
table {%
2 3.07376647049064
2 0.459094522743973
};
\addplot [color1]
table {%
2 6.17159229039609
2 10.2964472466807
};
\addplot [color1]
table {%
1.835 0.459094522743973
2.165 0.459094522743973
};
\addplot [color1]
table {%
1.835 10.2964472466807
2.165 10.2964472466807
};
\addplot [color1, mark=o, mark size=3, mark options={solid,fill opacity=0,draw=black}, only marks]
table {%
2 11.0863931157091
2 19.5609829358057
2 15.7213644059416
};
\addplot [color0]
table {%
3.25 1.97478609071369
3.25 0.306602075002283
};
\addplot [color0]
table {%
3.25 3.96504370157623
3.25 6.16135260154379
};
\addplot [color0]
table {%
3.085 0.306602075002283
3.415 0.306602075002283
};
\addplot [color0]
table {%
3.085 6.16135260154379
3.415 6.16135260154379
};
\addplot [color0, mark=o, mark size=3, mark options={solid,fill opacity=0,draw=black}, only marks]
table {%
3.25 8.63622568496884
3.25 7.06675293761222
};
\addplot [color1]
table {%
4.25 1.99811642016714
4.25 0.245227364821287
};
\addplot [color1]
table {%
4.25 3.97914368561097
4.25 6.10397790090546
};
\addplot [color1]
table {%
4.085 0.245227364821287
4.415 0.245227364821287
};
\addplot [color1]
table {%
4.085 6.10397790090546
4.415 6.10397790090546
};
\addplot [color1, mark=o, mark size=3, mark options={solid,fill opacity=0,draw=black}, only marks]
table {%
4.25 8.70824958227125
4.25 7.04794867872801
};
\addplot [color0]
table {%
5.5 2.39142009365388
5.5 0.258041678022451
};
\addplot [color0]
table {%
5.5 5.29867334053462
5.5 8.90278993274964
};
\addplot [color0]
table {%
5.335 0.258041678022451
5.665 0.258041678022451
};
\addplot [color0]
table {%
5.335 8.90278993274964
5.665 8.90278993274964
};
\addplot [color1]
table {%
6.5 2.36502848759696
6.5 0.257599781055045
};
\addplot [color1]
table {%
6.5 4.27533354952236
6.5 7.12534215957134
};
\addplot [color1]
table {%
6.335 0.257599781055045
6.665 0.257599781055045
};
\addplot [color1]
table {%
6.335 7.12534215957134
6.665 7.12534215957134
};
\addplot [color1, mark=o, mark size=3, mark options={solid,fill opacity=0,draw=black}, only marks]
table {%
6.5 8.89323548795748
6.5 7.55897936745994
};
\addplot [color0]
table {%
7.75 0.928731571806936
7.75 0.0634347307333202
};
\addplot [color0]
table {%
7.75 1.90692494649583
7.75 3.14867046315771
};
\addplot [color0]
table {%
7.585 0.0634347307333202
7.915 0.0634347307333202
};
\addplot [color0]
table {%
7.585 3.14867046315771
7.915 3.14867046315771
};
\addplot [color0, mark=o, mark size=3, mark options={solid,fill opacity=0,draw=black}, only marks]
table {%
7.75 4.02898611572527
7.75 4.40517843410241
7.75 8.3644225413074
7.75 3.68611989439665
7.75 6.34292107596406
7.75 4.16952301215333
};
\addplot [color1]
table {%
8.75 1.02978702842775
8.75 0.0614681134247496
};
\addplot [color1]
table {%
8.75 1.99227266655037
8.75 3.30005321467678
};
\addplot [color1]
table {%
8.585 0.0614681134247496
8.915 0.0614681134247496
};
\addplot [color1]
table {%
8.585 3.30005321467678
8.915 3.30005321467678
};
\addplot [color1, mark=o, mark size=3, mark options={solid,fill opacity=0,draw=black}, only marks]
table {%
8.75 4.54585546760952
8.75 8.32895358139788
8.75 6.13317231629844
};
\addplot [color0]
table {%
10 0.41533571104182
10 0.216579164271876
};
\addplot [color0]
table {%
10 0.882388464186273
10 1.22722307407157
};
\addplot [color0]
table {%
9.835 0.216579164271876
10.165 0.216579164271876
};
\addplot [color0]
table {%
9.835 1.22722307407157
10.165 1.22722307407157
};
\addplot [color0, mark=o, mark size=3, mark options={solid,fill opacity=0,draw=black}, only marks]
table {%
10 1.67964558324244
10 2.29400845356964
};
\addplot [color1]
table {%
11 0.394608980513521
11 0.212432244251576
};
\addplot [color1]
table {%
11 0.868971930741341
11 1.22948240025032
};
\addplot [color1]
table {%
10.835 0.212432244251576
11.165 0.212432244251576
};
\addplot [color1]
table {%
10.835 1.22948240025032
11.165 1.22948240025032
};
\addplot [color1, mark=o, mark size=3, mark options={solid,fill opacity=0,draw=black}, only marks]
table {%
11 1.64322153441689
11 2.28716763848796
};
\path [draw=color0, fill=color2]
(axis cs:0.67,3.07466396705552)
--(axis cs:1.33,3.07466396705552)
--(axis cs:1.33,6.26184972902956)
--(axis cs:0.67,6.26184972902956)
--(axis cs:0.67,3.07466396705552)
--cycle;
\path [draw=color1, fill=color3]
(axis cs:1.67,3.07376647049064)
--(axis cs:2.33,3.07376647049064)
--(axis cs:2.33,6.17159229039609)
--(axis cs:1.67,6.17159229039609)
--(axis cs:1.67,3.07376647049064)
--cycle;
\path [draw=color0, fill=color2]
(axis cs:2.92,1.97478609071369)
--(axis cs:3.58,1.97478609071369)
--(axis cs:3.58,3.96504370157623)
--(axis cs:2.92,3.96504370157623)
--(axis cs:2.92,1.97478609071369)
--cycle;
\path [draw=color1, fill=color3]
(axis cs:3.92,1.99811642016714)
--(axis cs:4.58,1.99811642016714)
--(axis cs:4.58,3.97914368561097)
--(axis cs:3.92,3.97914368561097)
--(axis cs:3.92,1.99811642016714)
--cycle;
\path [draw=color0, fill=color2]
(axis cs:5.17,2.39142009365388)
--(axis cs:5.83,2.39142009365388)
--(axis cs:5.83,5.29867334053462)
--(axis cs:5.17,5.29867334053462)
--(axis cs:5.17,2.39142009365388)
--cycle;
\path [draw=color1, fill=color3]
(axis cs:6.17,2.36502848759696)
--(axis cs:6.83,2.36502848759696)
--(axis cs:6.83,4.27533354952236)
--(axis cs:6.17,4.27533354952236)
--(axis cs:6.17,2.36502848759696)
--cycle;
\path [draw=color0, fill=color2]
(axis cs:7.42,0.928731571806936)
--(axis cs:8.08,0.928731571806936)
--(axis cs:8.08,1.90692494649583)
--(axis cs:7.42,1.90692494649583)
--(axis cs:7.42,0.928731571806936)
--cycle;
\path [draw=color1, fill=color3]
(axis cs:8.42,1.02978702842775)
--(axis cs:9.08,1.02978702842775)
--(axis cs:9.08,1.99227266655037)
--(axis cs:8.42,1.99227266655037)
--(axis cs:8.42,1.02978702842775)
--cycle;
\path [draw=color0, fill=color2]
(axis cs:9.67,0.41533571104182)
--(axis cs:10.33,0.41533571104182)
--(axis cs:10.33,0.882388464186273)
--(axis cs:9.67,0.882388464186273)
--(axis cs:9.67,0.41533571104182)
--cycle;
\path [draw=color1, fill=color3]
(axis cs:10.67,0.394608980513521)
--(axis cs:11.33,0.394608980513521)
--(axis cs:11.33,0.868971930741341)
--(axis cs:10.67,0.868971930741341)
--(axis cs:10.67,0.394608980513521)
--cycle;
\addplot [color0]
table {%
0.67 3.9877849767546
1.33 3.9877849767546
};
\addplot [color1]
table {%
1.67 3.86810353659764
2.33 3.86810353659764
};
\addplot [color0]
table {%
2.92 3.35932866286477
3.58 3.35932866286477
};
\addplot [color1]
table {%
3.92 3.36509500099497
4.58 3.36509500099497
};
\addplot [color0]
table {%
5.17 3.9418953049528
5.83 3.9418953049528
};
\addplot [color1]
table {%
6.17 3.1092860383707
6.83 3.1092860383707
};
\addplot [color0]
table {%
7.42 1.34248519975462
8.08 1.34248519975462
};
\addplot [color1]
table {%
8.42 1.42460631331979
9.08 1.42460631331979
};
\addplot [color0]
table {%
9.67 0.602045987203911
10.33 0.602045987203911
};
\addplot [color1]
table {%
10.67 0.56462942392321
11.33 0.56462942392321
};
\end{axis}

\end{tikzpicture}
  \end{subfigure}

  \caption{Impact of \textcolor{secondcolor}{TN simplifications} (non-simplified
    performance ratios shown in \textcolor{maincolor}{blue}). TN simplifications
    improve performance on
    (\subref{subfig:app:effect-simplifications-mixed-dense}) mixed dense,
    (\subref{subfig:app:effect-simplifications-dense}) dense, and
    (\subref{subfig:app:effect-simplifications-down}) down-sampling
    convolutions. (\subref{subfig:app:effect-simplifications-general}) General
    convolutions are not affected by TN
    simplifications.}\label{fig:app:effect-simplifications}
\end{figure}

\clearpage

\subsection{Forward Pass}
We compare TN and TN+opt with PyTorch's \texttt{torch.nn.functional.conv2d}.
\Cref{fig:app:forward-pass} visualizes the performance ratios for different
convolution categories. \Cref{tab:app:forward-pass-gpu} contains the detailed
run times and performance factors.

\begin{figure}[!h]
  \centering
  \pgfkeys{/pgfplots/zmystyle/.style={boxplotbasicstyle}}
\begin{tikzpicture}

\definecolor{color0}{rgb}{0.12156862745098,0.466666666666667,0.705882352941177}
\definecolor{color1}{rgb}{0.0901960784313725,0.745098039215686,0.811764705882353}
\definecolor{color2}{rgb}{0.580392156862745,0.403921568627451,0.741176470588235}
\definecolor{color3}{rgb}{0.890196078431372,0.466666666666667,0.76078431372549}

\begin{axis}[
axis line style={white!80!black},
tick pos=left,
xmin=0.5, xmax=9.25,
xtick style={color=gray},
xtick={1.5,3.75,6,8.25},
xticklabels={General,Dense mix,Dense,Down},
ylabel={TN versus PT (logarithmic)},
ymin=0.380563982560047, ymax=23.5974515110603,
ymode=log,
zmystyle
]
\path [draw=black, ]
(axis cs:0.5,1)
--(axis cs:9.25,1);

\addplot [color0]
table {%
1 3.07466396705552
1 0.525983824154647
};
\addplot [color0]
table {%
1 6.26184972902956
1 10.3484465930735
};
\addplot [color0]
table {%
0.835 0.525983824154647
1.165 0.525983824154647
};
\addplot [color0]
table {%
0.835 10.3484465930735
1.165 10.3484465930735
};
\addplot [color0, mark=o, mark size=3, mark options={solid,fill opacity=0,draw=black}, only marks]
table {%
1 11.1143814142224
1 19.5518044271115
1 15.7566575364248
};
\addplot [color1]
table {%
2 3.07376647049064
2 0.459094522743973
};
\addplot [color1]
table {%
2 6.17159229039609
2 10.2964472466807
};
\addplot [color1]
table {%
1.835 0.459094522743973
2.165 0.459094522743973
};
\addplot [color1]
table {%
1.835 10.2964472466807
2.165 10.2964472466807
};
\addplot [color1, mark=o, mark size=3, mark options={solid,fill opacity=0,draw=black}, only marks]
table {%
2 11.0863931157091
2 19.5609829358057
2 15.7213644059416
};
\addplot [color0]
table {%
3.25 2.46023692065264
3.25 2.4505726759626
};
\addplot [color0]
table {%
3.25 2.68523201814311
3.25 2.9602691527339
};
\addplot [color0]
table {%
3.085 2.4505726759626
3.415 2.4505726759626
};
\addplot [color0]
table {%
3.085 2.9602691527339
3.415 2.9602691527339
};
\addplot [color0, mark=o, mark size=3, mark options={solid,fill opacity=0,draw=black}, only marks]
table {%
3.25 1.59138793085551
3.25 1.55902931467978
};
\addplot [color1]
table {%
4.25 2.11941964143051
4.25 2.11299128561488
};
\addplot [color1]
table {%
4.25 2.24359226442037
4.25 2.42270096537717
};
\addplot [color1]
table {%
4.085 2.11299128561488
4.415 2.11299128561488
};
\addplot [color1]
table {%
4.085 2.42270096537717
4.415 2.42270096537717
};
\addplot [color1, mark=o, mark size=3, mark options={solid,fill opacity=0,draw=black}, only marks]
table {%
4.25 1.34187038156887
4.25 1.36634167604259
4.25 2.50778708904802
};
\addplot [color0]
table {%
5.5 1.59927479757071
5.5 1.15219594102096
};
\addplot [color0]
table {%
5.5 3.19474902045781
5.5 5.12786927532713
};
\addplot [color0]
table {%
5.335 1.15219594102096
5.665 1.15219594102096
};
\addplot [color0]
table {%
5.335 5.12786927532713
5.665 5.12786927532713
};
\addplot [color0, mark=o, mark size=3, mark options={solid,fill opacity=0,draw=black}, only marks]
table {%
5.5 7.41882080972977
};
\addplot [color1]
table {%
6.5 1.05279922570089
6.5 0.732303578459289
};
\addplot [color1]
table {%
6.5 1.5262258918302
6.5 2.15893117557101
};
\addplot [color1]
table {%
6.335 0.732303578459289
6.665 0.732303578459289
};
\addplot [color1]
table {%
6.335 2.15893117557101
6.665 2.15893117557101
};
\addplot [color1, mark=o, mark size=3, mark options={solid,fill opacity=0,draw=black}, only marks]
table {%
6.5 2.32643190987456
};
\addplot [color0]
table {%
7.75 1.88605563239138
7.75 1.37392250430349
};
\addplot [color0]
table {%
7.75 2.83253124542281
7.75 4.22673130466074
};
\addplot [color0]
table {%
7.585 1.37392250430349
7.915 1.37392250430349
};
\addplot [color0]
table {%
7.585 4.22673130466074
7.915 4.22673130466074
};
\addplot [color1]
table {%
8.75 1.08163795848431
8.75 0.968470984302919
};
\addplot [color1]
table {%
8.75 1.36294334773553
8.75 1.48138841366899
};
\addplot [color1]
table {%
8.585 0.968470984302919
8.915 0.968470984302919
};
\addplot [color1]
table {%
8.585 1.48138841366899
8.915 1.48138841366899
};
\path [draw=color0, fill=color2]
(axis cs:0.67,3.07466396705552)
--(axis cs:1.33,3.07466396705552)
--(axis cs:1.33,6.26184972902956)
--(axis cs:0.67,6.26184972902956)
--(axis cs:0.67,3.07466396705552)
--cycle;
\path [draw=color1, fill=color3]
(axis cs:1.67,3.07376647049064)
--(axis cs:2.33,3.07376647049064)
--(axis cs:2.33,6.17159229039609)
--(axis cs:1.67,6.17159229039609)
--(axis cs:1.67,3.07376647049064)
--cycle;
\path [draw=color0, fill=color2]
(axis cs:2.92,2.46023692065264)
--(axis cs:3.58,2.46023692065264)
--(axis cs:3.58,2.68523201814311)
--(axis cs:2.92,2.68523201814311)
--(axis cs:2.92,2.46023692065264)
--cycle;
\path [draw=color1, fill=color3]
(axis cs:3.92,2.11941964143051)
--(axis cs:4.58,2.11941964143051)
--(axis cs:4.58,2.24359226442037)
--(axis cs:3.92,2.24359226442037)
--(axis cs:3.92,2.11941964143051)
--cycle;
\path [draw=color0, fill=color2]
(axis cs:5.17,1.59927479757071)
--(axis cs:5.83,1.59927479757071)
--(axis cs:5.83,3.19474902045781)
--(axis cs:5.17,3.19474902045781)
--(axis cs:5.17,1.59927479757071)
--cycle;
\path [draw=color1, fill=color3]
(axis cs:6.17,1.05279922570089)
--(axis cs:6.83,1.05279922570089)
--(axis cs:6.83,1.5262258918302)
--(axis cs:6.17,1.5262258918302)
--(axis cs:6.17,1.05279922570089)
--cycle;
\path [draw=color0, fill=color2]
(axis cs:7.42,1.88605563239138)
--(axis cs:8.08,1.88605563239138)
--(axis cs:8.08,2.83253124542281)
--(axis cs:7.42,2.83253124542281)
--(axis cs:7.42,1.88605563239138)
--cycle;
\path [draw=color1, fill=color3]
(axis cs:8.42,1.08163795848431)
--(axis cs:9.08,1.08163795848431)
--(axis cs:9.08,1.36294334773553)
--(axis cs:8.42,1.36294334773553)
--(axis cs:8.42,1.08163795848431)
--cycle;
\addplot [color0]
table {%
0.67 3.9877849767546
1.33 3.9877849767546
};
\addplot [color1]
table {%
1.67 3.86810353659764
2.33 3.86810353659764
};
\addplot [color0]
table {%
2.92 2.52218902995192
3.58 2.52218902995192
};
\addplot [color1]
table {%
3.92 2.17698162959808
4.58 2.17698162959808
};
\addplot [color0]
table {%
5.17 2.30653191880153
5.83 2.30653191880153
};
\addplot [color1]
table {%
6.17 1.21250271167177
6.83 1.21250271167177
};
\addplot [color0]
table {%
7.42 2.45708832072198
8.08 2.45708832072198
};
\addplot [color1]
table {%
8.42 1.23494468558542
9.08 1.23494468558542
};
\end{axis}

\end{tikzpicture}
  \caption{Forward pass performance ratios of \textcolor{maincolor}{TN versus PT} and
    \textcolor{secondcolor}{TN+opt versus PT} for different convolution types on
    GPU.}\label{fig:app:forward-pass}
\end{figure}

\clearpage

\def\memoryFormat#1{%
  \num[%
  round-mode=figures,%
  round-precision=3,%
  scientific-notation=false,%
  text-series-to-math = true,%
  ]{#1}%
}%
\def\runtimeFormat#1{%
  \num[%
  round-mode=places,%
  round-precision=2,%
  scientific-notation=true,%
  exponent-product=\ensuremath{\cdot},%
  ]{#1}%
}%
\def\factorFormat#1{%
  \num[%
  round-mode=places,%
  round-precision=2,%
  text-series-to-math = true,%
  ]{#1}\,x%
}%
\def\highlightSpeedup#1{%
  \def\speedupValue{#1}%
  \ifthenelse{\lengthtest{\speedupValue pt < 1.0pt}}{%
    \bfseries%
  }{}%
}%
\def\performanceTableCPU#1{%
  \scalebox{0.9}{
    \begin{tiny}
      \csvloop{%
        file=#1,%
        separator=semicolon,%
        tabular={c|ccc|ccc|c},%
        table head={%
          \toprule%
          \multirow{2}{*}{\textbf{Name}}%
          & \multicolumn{3}{|c}{\textbf{PyTorch}}%
          & \multicolumn{3}{|c|}{\textbf{TACO}}%
          & \multirow{2}{*}{\textbf{Combined}}%
          \\%
          & \textbf{TN [s]}%
          & \textbf{PT [s]}%
          & \textbf{Factor}%
          & \textbf{Dense [s]}%
          & \textbf{Sparse [s]}%
          & \textbf{Factor}%
          \\\midrule},%
        command={%
          \csvcoli%
          & \runtimeFormat{\csvcoliii}%
          & \runtimeFormat{\csvcolii}%
          & {\highlightSpeedup{\csvcolvi}\factorFormat{\csvcolvi}}%
          & \runtimeFormat{\csvcoliv}%
          & \runtimeFormat{\csvcolv}%
          & \factorFormat{\csvcolvii}%
          & {\highlightSpeedup{\csvcolviii}\factorFormat{\csvcolviii}}%
        },%
        table foot=\bottomrule}%
    \end{tiny}
  }
}
\def\performanceTableGPU#1{%
  \scalebox{0.9}{
    \begin{tiny}
      \csvloop{%
        file=#1,%
        separator=semicolon,%
        tabular={c|ccc},%
        table head={%
          \toprule%
          \textbf{Name}
          & \textbf{TN [s]}%
          & \textbf{PT [s]}%
          & \textbf{Factor}%
          \\\midrule},%
        command={%
          \csvcoli%
          & \runtimeFormat{\csvcoliii}%
          & \runtimeFormat{\csvcolii}%
          & {\highlightSpeedup{\csvcolvi}\factorFormat{\csvcolvi}}%
        },%
        table foot=\bottomrule}%
    \end{tiny}
  }
}
\def\performanceTableGPUSimplified#1{%
  \scalebox{0.9}{%
    \begin{tiny}
      \csvloop{%
        file=#1,%
        separator=semicolon,%
        tabular={ccc},%
        table head={%
          \toprule%
          \textbf{TN + opt [s]}%
          & \textbf{PT [s]}%
          & \textbf{Factor}%
          \\\midrule},%
        command={%
          \runtimeFormat{\csvcoliii}%
          & \runtimeFormat{\csvcolii}%
          & {\highlightSpeedup{\csvcolvi}\factorFormat{\csvcolvi}}%
        },%
        table foot=\bottomrule}%
    \end{tiny}
  }
}
\def\performanceTableTACO#1{%
  \scalebox{0.9}{
    \begin{tiny}
      \csvloop{%
        file=#1,%
        separator=semicolon,%
        tabular={c|ccc},%
        table head={%
          \toprule%
          \textbf{Name}
          & \textbf{Dense [s]}%
          & \textbf{Sparse [s]}%
          & \textbf{Factor}%
          \\\midrule},%
        command={%
          \csvcoli%
          & \runtimeFormat{\csvcoliv}\scalebox{0}[0.8]{\phantom{$10^{-1}$}}%
          & \runtimeFormat{\csvcolv}%
          & \factorFormat{\csvcolvii}%
        },%
        table foot=\bottomrule}%
    \end{tiny}
  }
}
\def\peakmemTable#1{%
  \scalebox{0.9}{
    \begin{tiny}
      \csvloop{%
        file=#1,%
        separator=semicolon,%
        tabular={c|c},
        table head={%
          \toprule%
          \textbf{Name}
          & \textbf{TN [MiB]}%
          \\\midrule},%
        command={%
          \csvcoli%
          & {%
            \memoryFormat{\csvcoliii}%
          }%
        },%
        table foot=\bottomrule}%
    \end{tiny}
  }
}
\def\peakmemTableSimplified#1{%
  \scalebox{0.9}{
    \begin{tiny}
      \csvloop{%
        file=#1,%
        separator=semicolon,%
        tabular={cc},%
        table head={%
          \toprule%
          \textbf{TN + opt [MiB]}%
          & \textbf{PT [MiB]}%
          \\\midrule},%
        command={%
          {%
            \memoryFormat{\csvcoliii}%
          }%
          &
          {%
            \memoryFormat{\csvcolii}%
          }%
        },%
        table foot=\bottomrule}%
    \end{tiny}
  }
}

\def\datapath{exp/exp02_performance_forward/results/}
\def\datapathHyper{exp/exp01_conv2d_vs_einconv2d/results/layer_details/}
\begin{table}[p]
  \centering
  \caption{Forward pass performance comparison on GPU.}\label{tab:app:forward-pass-gpu}
  \vspace{-2ex}
  \begin{subtable}[t]{1.0\linewidth}
    \centering
    \caption{3c3d, CIFAR-10, input shape (128, 3, 32, 32)}\label{tab:app:forward-pass-gpu-cifar10-3c3d}
    \vspace{-0.5ex}
    \performanceTableGPU{\datapath cifar10_3c3d_input_shape_3_32_32_cuda_0_batch_size_128/times.csv}
    \performanceTableGPUSimplified{\datapath cifar10_3c3d_input_shape_3_32_32_cuda_0_batch_size_128_simplify/times.csv}
    \convTypeTable{\datapathHyper cifar10_3c3d_in_3_32_32_out_10.csv}
  \end{subtable}
  \begin{subtable}[t]{1.0\linewidth}
    \centering
    \caption{F-MNIST 2c2d, input shape (128, 1, 28, 28)}\label{tab:app:forward-pass-gpu-fmnist-2c2d}
    \vspace{-0.5ex}
    \performanceTableGPU{\datapath fmnist_2c2d_input_shape_1_28_28_cuda_0_batch_size_128/times.csv}
    \performanceTableGPUSimplified{\datapath fmnist_2c2d_input_shape_1_28_28_cuda_0_batch_size_128_simplify/times.csv}
    \convTypeTable{\datapathHyper fmnist_2c2d_in_1_28_28_out_10.csv}
  \end{subtable}
  \begin{subtable}[t]{1.0\linewidth}
    \centering
    \caption{CIFAR-100 All-CNN-C, input shape (128, 3, 32, 32)}\label{tab:app:forward-pass-gpu-cifar100-allcnnc}
    \vspace{-0.5ex}
    \performanceTableGPU{\datapath cifar100_allcnnc_input_shape_3_32_32_cuda_0_batch_size_128/times.csv}
    \performanceTableGPUSimplified{\datapath cifar100_allcnnc_input_shape_3_32_32_cuda_0_batch_size_128_simplify/times.csv}
    \convTypeTable{\datapathHyper cifar100_allcnnc_in_3_32_32_out_100.csv}
  \end{subtable}
  \begin{subtable}[t]{1.0\linewidth}
    \centering
    \caption{Alexnet, input shape (32, 3, 256, 256)}\label{tab:app:forward-pass-gpu-alexnet}
    \vspace{-0.5ex}
    \performanceTableGPU{\datapath alexnet_input_shape_3_256_256_cuda_0_batch_size_32/times.csv}
    \performanceTableGPUSimplified{\datapath alexnet_input_shape_3_256_256_cuda_0_batch_size_32_simplify/times.csv}
    \convTypeTable{\datapathHyper alexnet_in_3_256_256_out_1000.csv}
  \end{subtable}
  \begin{subtable}[t]{1.0\linewidth}
    \centering
    \caption{ResNet18, input shape (32, 3, 256, 256)}\label{tab:app:forward-pass-gpu-resnet18}
    \vspace{-0.5ex}
    \performanceTableGPU{\datapath resnet18_input_shape_3_256_256_cuda_0_batch_size_32/times.csv}
    \performanceTableGPUSimplified{\datapath resnet18_input_shape_3_256_256_cuda_0_batch_size_32_simplify/times.csv}
    \convTypeTable{\datapathHyper resnet18_in_3_256_256_out_1000.csv}
  \end{subtable}
  \begin{subtable}[t]{1.0\linewidth}
    \centering
    \caption{ResNext101, input shape (32, 3, 256, 256)}\label{tab:app:forward-pass-gpu-resnext101}
    \vspace{-0.5ex}
    \performanceTableGPU{\datapath resnext101_32x8d_input_shape_3_256_256_cuda_0_batch_size_32/times.csv}
    \performanceTableGPUSimplified{\datapath resnext101_32x8d_input_shape_3_256_256_cuda_0_batch_size_32_simplify/times.csv}
    \convTypeTable{\datapathHyper resnext101_32x8d_in_3_256_256_out_1000.csv}
  \end{subtable}
  \begin{subtable}[t]{1.0\linewidth}
    \centering
    \caption{ConvNeXt-base, input shape (32, 3, 256,
      256)}\label{tab:app:forward-pass-gpu-convnext-base}
    \vspace{-0.5ex}
    \performanceTableGPU{\datapath convnext_base_input_shape_3_256_256_cuda_0_batch_size_32/times.csv}
    \performanceTableGPUSimplified{\datapath convnext_base_input_shape_3_256_256_cuda_0_batch_size_32_simplify/times.csv}
    \convTypeTable{\datapathHyper convnext_base_in_3_256_256_out_1000.csv}
  \end{subtable}
\end{table}
\begin{table}[p]\ContinuedFloat
  \centering
  \vspace*{-10ex}
  \begin{subtable}[t]{1.0\linewidth}
    \centering
    \caption{InceptionV3, input shape (32, 3, 299, 299)}\label{tab:app:forward-pass-gpu-inception-v3}
    \performanceTableGPU{\datapath inception_v3_input_shape_3_299_299_cuda_0_batch_size_32/times.csv}
    \performanceTableGPUSimplified{\datapath inception_v3_input_shape_3_299_299_cuda_0_batch_size_32_simplify/times.csv}
    \convTypeTable{\datapathHyper inception_v3_in_3_299_299_out_1000.csv}
  \end{subtable}
  \begin{subtable}[t]{1.0\linewidth}
    \centering
    \caption{MobileNetV2, input shape (32, 3, 256, 256)}\label{tab:app:forward-pass-gpu-mobilenet-v2}
    \performanceTableGPU{\datapath mobilenet_v2_input_shape_3_256_256_cuda_0_batch_size_32/times.csv}
    \performanceTableGPUSimplified{\datapath mobilenet_v2_input_shape_3_256_256_cuda_0_batch_size_32_simplify/times.csv}
    \convTypeTable{\datapathHyper mobilenet_v2_in_3_256_256_out_1000.csv}
  \end{subtable}
\end{table}

\clearpage

\subsection{Input VJP}
We compare TN and TN+opt with a PyTorch implementation of the input VJP via
\texttt{torch.autograd.grad}. \Cref{fig:app:input-vjp} visualizes the
performance ratios for different convolution categories.
\Cref{tab:app:input-vjp-gpu} contains the detailed run times and performance
factors.

\begin{figure}[!h]
  \centering
  \pgfkeys{/pgfplots/zmystyle/.style={boxplotbasicstyle}}
\begin{tikzpicture}

\definecolor{color0}{rgb}{0.12156862745098,0.466666666666667,0.705882352941177}
\definecolor{color1}{rgb}{0.0901960784313725,0.745098039215686,0.811764705882353}
\definecolor{color2}{rgb}{0.580392156862745,0.403921568627451,0.741176470588235}
\definecolor{color3}{rgb}{0.890196078431372,0.466666666666667,0.76078431372549}

\begin{axis}[
axis line style={white!80!black},
tick pos=left,
xmin=0.5, xmax=9.25,
xtick style={color=gray},
xtick={1.5,3.75,6,8.25},
xticklabels={General,Dense mix,Dense,Down},
ylabel={TN versus PT (logarithmic)},
ymin=0.205140228286858, ymax=10.4099576913811,
ymode=log,
zmystyle
]
\path [draw=black, ]
(axis cs:0.5,1)
--(axis cs:9.25,1);

\addplot [color0]
table {%
1 1.97478609071369
1 0.306602075002283
};
\addplot [color0]
table {%
1 3.96504370157623
1 6.16135260154379
};
\addplot [color0]
table {%
0.835 0.306602075002283
1.165 0.306602075002283
};
\addplot [color0]
table {%
0.835 6.16135260154379
1.165 6.16135260154379
};
\addplot [color0, mark=o, mark size=3, mark options={solid,fill opacity=0,draw=black}, only marks]
table {%
1 8.63622568496884
1 7.06675293761222
};
\addplot [color1]
table {%
2 1.99811642016714
2 0.245227364821287
};
\addplot [color1]
table {%
2 3.97914368561097
2 6.10397790090546
};
\addplot [color1]
table {%
1.835 0.245227364821287
2.165 0.245227364821287
};
\addplot [color1]
table {%
1.835 6.10397790090546
2.165 6.10397790090546
};
\addplot [color1, mark=o, mark size=3, mark options={solid,fill opacity=0,draw=black}, only marks]
table {%
2 8.70824958227125
2 7.04794867872801
};
\addplot [color0]
table {%
3.25 2.21507580188499
3.25 2.20854703470289
};
\addplot [color0]
table {%
3.25 2.57006747044357
3.25 2.79165785357732
};
\addplot [color0]
table {%
3.085 2.20854703470289
3.415 2.20854703470289
};
\addplot [color0]
table {%
3.085 2.79165785357732
3.415 2.79165785357732
};
\addplot [color0, mark=o, mark size=3, mark options={solid,fill opacity=0,draw=black}, only marks]
table {%
3.25 1.54759047558896
3.25 1.5407083891019
};
\addplot [color1]
table {%
4.25 1.95084083469693
4.25 1.91820809710367
};
\addplot [color1]
table {%
4.25 2.21912496307565
4.25 2.47712335498054
};
\addplot [color1]
table {%
4.085 1.91820809710367
4.415 1.91820809710367
};
\addplot [color1]
table {%
4.085 2.47712335498054
4.415 2.47712335498054
};
\addplot [color1, mark=o, mark size=3, mark options={solid,fill opacity=0,draw=black}, only marks]
table {%
4.25 1.23647095403953
4.25 1.29759375066617
};
\addplot [color0]
table {%
5.5 1.53578978507391
5.5 0.784287426635228
};
\addplot [color0]
table {%
5.5 2.3281565276142
5.5 3.50513351127121
};
\addplot [color0]
table {%
5.335 0.784287426635228
5.665 0.784287426635228
};
\addplot [color0]
table {%
5.335 3.50513351127121
5.665 3.50513351127121
};
\addplot [color0, mark=o, mark size=3, mark options={solid,fill opacity=0,draw=black}, only marks]
table {%
5.5 5.30929071184138
};
\addplot [color1]
table {%
6.5 0.954033703296296
6.5 0.752727745755602
};
\addplot [color1]
table {%
6.5 1.12765183745918
6.5 1.38167346569684
};
\addplot [color1]
table {%
6.335 0.752727745755602
6.665 0.752727745755602
};
\addplot [color1]
table {%
6.335 1.38167346569684
6.665 1.38167346569684
};
\addplot [color1, mark=o, mark size=3, mark options={solid,fill opacity=0,draw=black}, only marks]
table {%
6.5 0.68156981086126
6.5 0.659759465083098
6.5 0.591139716482069
6.5 1.40354960102706
6.5 1.45680566253413
};
\addplot [color0]
table {%
7.75 1.3340871308979
7.75 0.844344178309862
};
\addplot [color0]
table {%
7.75 1.92323726384394
7.75 2.51228605183415
};
\addplot [color0]
table {%
7.585 0.844344178309862
7.915 0.844344178309862
};
\addplot [color0]
table {%
7.585 2.51228605183415
7.915 2.51228605183415
};
\addplot [color1]
table {%
8.75 1.1655912213179
8.75 0.789541123017264
};
\addplot [color1]
table {%
8.75 1.64896924538288
8.75 2.24055722762165
};
\addplot [color1]
table {%
8.585 0.789541123017264
8.915 0.789541123017264
};
\addplot [color1]
table {%
8.585 2.24055722762165
8.915 2.24055722762165
};
\path [draw=color0, fill=color2]
(axis cs:0.67,1.97478609071369)
--(axis cs:1.33,1.97478609071369)
--(axis cs:1.33,3.96504370157623)
--(axis cs:0.67,3.96504370157623)
--(axis cs:0.67,1.97478609071369)
--cycle;
\path [draw=color1, fill=color3]
(axis cs:1.67,1.99811642016714)
--(axis cs:2.33,1.99811642016714)
--(axis cs:2.33,3.97914368561097)
--(axis cs:1.67,3.97914368561097)
--(axis cs:1.67,1.99811642016714)
--cycle;
\path [draw=color0, fill=color2]
(axis cs:2.92,2.21507580188499)
--(axis cs:3.58,2.21507580188499)
--(axis cs:3.58,2.57006747044357)
--(axis cs:2.92,2.57006747044357)
--(axis cs:2.92,2.21507580188499)
--cycle;
\path [draw=color1, fill=color3]
(axis cs:3.92,1.95084083469693)
--(axis cs:4.58,1.95084083469693)
--(axis cs:4.58,2.21912496307565)
--(axis cs:3.92,2.21912496307565)
--(axis cs:3.92,1.95084083469693)
--cycle;
\path [draw=color0, fill=color2]
(axis cs:5.17,1.53578978507391)
--(axis cs:5.83,1.53578978507391)
--(axis cs:5.83,2.3281565276142)
--(axis cs:5.17,2.3281565276142)
--(axis cs:5.17,1.53578978507391)
--cycle;
\path [draw=color1, fill=color3]
(axis cs:6.17,0.954033703296296)
--(axis cs:6.83,0.954033703296296)
--(axis cs:6.83,1.12765183745918)
--(axis cs:6.17,1.12765183745918)
--(axis cs:6.17,0.954033703296296)
--cycle;
\path [draw=color0, fill=color2]
(axis cs:7.42,1.3340871308979)
--(axis cs:8.08,1.3340871308979)
--(axis cs:8.08,1.92323726384394)
--(axis cs:7.42,1.92323726384394)
--(axis cs:7.42,1.3340871308979)
--cycle;
\path [draw=color1, fill=color3]
(axis cs:8.42,1.1655912213179)
--(axis cs:9.08,1.1655912213179)
--(axis cs:9.08,1.64896924538288)
--(axis cs:8.42,1.64896924538288)
--(axis cs:8.42,1.1655912213179)
--cycle;
\addplot [color0]
table {%
0.67 3.35932866286477
1.33 3.35932866286477
};
\addplot [color1]
table {%
1.67 3.36509500099497
2.33 3.36509500099497
};
\addplot [color0]
table {%
2.92 2.40880657361853
3.58 2.40880657361853
};
\addplot [color1]
table {%
3.92 2.14390734744617
4.58 2.14390734744617
};
\addplot [color0]
table {%
5.17 1.80523520444377
5.83 1.80523520444377
};
\addplot [color1]
table {%
6.17 1.04973041454523
6.83 1.04973041454523
};
\addplot [color0]
table {%
7.42 1.78863774096403
8.08 1.78863774096403
};
\addplot [color1]
table {%
8.42 1.41758749431511
9.08 1.41758749431511
};
\end{axis}

\end{tikzpicture}
  \caption{Input VJP performance ratios of \textcolor{maincolor}{TN versus PT} and
    \textcolor{secondcolor}{TN+opt versus PT} for different convolution types on
    GPU.}\label{fig:app:input-vjp}
\end{figure}

\clearpage

\def\datapath{exp/exp04_performance_input_vjp/results/input_vjp/}
\def\datapathHyper{exp/exp01_conv2d_vs_einconv2d/results/layer_details/}
\begin{table}[p]
  \centering
  \caption{Input VJP performance comparison on GPU.}\label{tab:app:input-vjp-gpu}
  \vspace{-1ex}
  \begin{subtable}[t]{1.0\linewidth}
    \centering
    \caption{3c3d, CIFAR-10, input shape (128, 3, 32, 32)}\label{tab:app:input-vjp-gpu-cifar10-3c3d}
    \vspace{-0.5ex}
    \performanceTableGPU{\datapath cifar10_3c3d_input_shape_3_32_32_cuda_0_batch_size_128/times.csv}
    \performanceTableGPUSimplified{\datapath cifar10_3c3d_input_shape_3_32_32_cuda_0_batch_size_128_simplify/times.csv}
    \convTypeTable{\datapathHyper cifar10_3c3d_in_3_32_32_out_10.csv}
  \end{subtable}
  \begin{subtable}[t]{1.0\linewidth}
    \centering
    \caption{F-MNIST 2c2d, input shape (128, 1, 28, 28)}\label{tab:app:input-vjp-gpu-fmnist-2c2d}
    \vspace{-0.5ex}
    \performanceTableGPU{\datapath fmnist_2c2d_input_shape_1_28_28_cuda_0_batch_size_128/times.csv}
    \performanceTableGPUSimplified{\datapath fmnist_2c2d_input_shape_1_28_28_cuda_0_batch_size_128_simplify/times.csv}
    \convTypeTable{\datapathHyper fmnist_2c2d_in_1_28_28_out_10.csv}
  \end{subtable}
  \begin{subtable}[t]{1.0\linewidth}
    \centering
    \caption{CIFAR-100 All-CNN-C, input shape (128, 3, 32, 32)}\label{tab:app:input-vjp-gpu-cifar100-allcnnc}
    \vspace{-0.5ex}
    \performanceTableGPU{\datapath cifar100_allcnnc_input_shape_3_32_32_cuda_0_batch_size_128/times.csv}
    \performanceTableGPUSimplified{\datapath cifar100_allcnnc_input_shape_3_32_32_cuda_0_batch_size_128_simplify/times.csv}
    \convTypeTable{\datapathHyper cifar100_allcnnc_in_3_32_32_out_100.csv}
  \end{subtable}
  \begin{subtable}[t]{1.0\linewidth}
    \centering
    \caption{Alexnet, input shape (32, 3, 256, 256)}\label{tab:app:input-vjp-gpu-alexnet}
    \vspace{-0.5ex}
    \performanceTableGPU{\datapath alexnet_input_shape_3_256_256_cuda_0_batch_size_32/times.csv}
    \performanceTableGPUSimplified{\datapath alexnet_input_shape_3_256_256_cuda_0_batch_size_32_simplify/times.csv}
    \convTypeTable{\datapathHyper alexnet_in_3_256_256_out_1000.csv}
  \end{subtable}
  \begin{subtable}[t]{1.0\linewidth}
    \centering
    \caption{ResNet18, input shape (32, 3, 256, 256)}\label{tab:app:input-vjp-gpu-resnet18}
    \vspace{-0.5ex}
    \performanceTableGPU{\datapath resnet18_input_shape_3_256_256_cuda_0_batch_size_32/times.csv}
    \performanceTableGPUSimplified{\datapath resnet18_input_shape_3_256_256_cuda_0_batch_size_32_simplify/times.csv}
    \convTypeTable{\datapathHyper resnet18_in_3_256_256_out_1000.csv}
  \end{subtable}
  \begin{subtable}[t]{1.0\linewidth}
    \centering
    \caption{ResNext101, input shape (32, 3, 256, 256)}\label{tab:app:input-vjp-gpu-resnext101} \vspace{-0.5ex}\performanceTableGPU{\datapath resnext101_32x8d_input_shape_3_256_256_cuda_0_batch_size_32/times.csv}
    \performanceTableGPUSimplified{\datapath resnext101_32x8d_input_shape_3_256_256_cuda_0_batch_size_32_simplify/times.csv}
    \convTypeTable{\datapathHyper resnext101_32x8d_in_3_256_256_out_1000.csv}
  \end{subtable}
  \begin{subtable}[t]{1.0\linewidth}
    \centering
    \caption{ConvNeXt-base, input shape (32, 3, 256, 256)}\label{tab:app:input-vjp-gpu-convnext-base}\vspace{-0.5ex} \performanceTableGPU{\datapath convnext_base_input_shape_3_256_256_cuda_0_batch_size_32/times.csv}
    \performanceTableGPUSimplified{\datapath convnext_base_input_shape_3_256_256_cuda_0_batch_size_32_simplify/times.csv}
    \convTypeTable{\datapathHyper convnext_base_in_3_256_256_out_1000.csv}
  \end{subtable}

\end{table}
\begin{table}[p]\ContinuedFloat
  \centering
  \vspace*{-10ex}
  \begin{subtable}[t]{1.0\linewidth}
    \centering
    \caption{InceptionV3, input shape (32, 3, 299, 299)}\label{tab:app:input-vjp-gpu-inception-v3}
    \performanceTableGPU{\datapath inception_v3_input_shape_3_299_299_cuda_0_batch_size_32/times.csv}
    \performanceTableGPUSimplified{\datapath inception_v3_input_shape_3_299_299_cuda_0_batch_size_32_simplify/times.csv}
    \convTypeTable{\datapathHyper inception_v3_in_3_299_299_out_1000.csv}
  \end{subtable}
  \begin{subtable}[t]{1.0\linewidth}
    \centering
    \caption{MobileNetV2, input shape (32, 3, 256, 256)}\label{tab:app:input-vjp-gpu-mobilenet-v2}
    \performanceTableGPU{\datapath mobilenet_v2_input_shape_3_256_256_cuda_0_batch_size_32/times.csv}
    \performanceTableGPUSimplified{\datapath mobilenet_v2_input_shape_3_256_256_cuda_0_batch_size_32_simplify/times.csv}
    \convTypeTable{\datapathHyper mobilenet_v2_in_3_256_256_out_1000.csv}
  \end{subtable}
\end{table}

\clearpage

\subsection{Weight VJP}
We compare TN and TN+opt with a PyTorch implementation of the weight VJP via
\texttt{torch.autograd.grad}. \Cref{fig:app:weight-vjp} visualizes the
performance ratios for different convolution categories.
\Cref{tab:app:weight-vjp-gpu} contains the detailed run times and performance
factors.

\begin{figure}[!h]
  \centering
  \pgfkeys{/pgfplots/zmystyle/.style={boxplotbasicstyle}}
\begin{tikzpicture}

\definecolor{color0}{rgb}{0.12156862745098,0.466666666666667,0.705882352941177}
\definecolor{color1}{rgb}{0.0901960784313725,0.745098039215686,0.811764705882353}
\definecolor{color2}{rgb}{0.580392156862745,0.403921568627451,0.741176470588235}
\definecolor{color3}{rgb}{0.890196078431372,0.466666666666667,0.76078431372549}

\begin{axis}[
axis line style={white!80!black},
tick pos=left,
xmin=0.5, xmax=9.25,
xtick style={color=gray},
xtick={1.5,3.75,6,8.25},
xticklabels={General,Dense mix,Dense,Down},
ylabel={TN versus PT (logarithmic)},
ymin=0.135801080130873, ymax=10.8650562356686,
ymode=log,
zmystyle
]
\path [draw=black, ]
(axis cs:0.5,1)
--(axis cs:9.25,1);

\addplot [color0]
table {%
1 2.39142009365388
1 0.258041678022451
};
\addplot [color0]
table {%
1 5.29867334053462
1 8.90278993274964
};
\addplot [color0]
table {%
0.835 0.258041678022451
1.165 0.258041678022451
};
\addplot [color0]
table {%
0.835 8.90278993274964
1.165 8.90278993274964
};
\addplot [color1]
table {%
2 2.36502848759696
2 0.257599781055045
};
\addplot [color1]
table {%
2 4.27533354952236
2 7.12534215957134
};
\addplot [color1]
table {%
1.835 0.257599781055045
2.165 0.257599781055045
};
\addplot [color1]
table {%
1.835 7.12534215957134
2.165 7.12534215957134
};
\addplot [color1, mark=o, mark size=3, mark options={solid,fill opacity=0,draw=black}, only marks]
table {%
2 8.89323548795748
2 7.55897936745994
};
\addplot [color0]
table {%
3.25 1.14323001754823
3.25 1.06391407199899
};
\addplot [color0]
table {%
3.25 1.94265472891544
3.25 2.47824944822851
};
\addplot [color0]
table {%
3.085 1.06391407199899
3.415 1.06391407199899
};
\addplot [color0]
table {%
3.085 2.47824944822851
3.415 2.47824944822851
};
\addplot [color1]
table {%
4.25 1.02553324034011
4.25 0.815138975157994
};
\addplot [color1]
table {%
4.25 1.201294802175
4.25 1.24145248621953
};
\addplot [color1]
table {%
4.085 0.815138975157994
4.415 0.815138975157994
};
\addplot [color1]
table {%
4.085 1.24145248621953
4.415 1.24145248621953
};
\addplot [color0]
table {%
5.5 1.14711526075564
5.5 0.349067227706645
};
\addplot [color0]
table {%
5.5 3.0778358039215
5.5 4.96785152085861
};
\addplot [color0]
table {%
5.335 0.349067227706645
5.665 0.349067227706645
};
\addplot [color0]
table {%
5.335 4.96785152085861
5.665 4.96785152085861
};
\addplot [color0, mark=o, mark size=3, mark options={solid,fill opacity=0,draw=black}, only marks]
table {%
5.5 6.11390939877441
};
\addplot [color1]
table {%
6.5 0.681670231178204
6.5 0.364493528786432
};
\addplot [color1]
table {%
6.5 0.977493787964857
6.5 1.30057398954806
};
\addplot [color1]
table {%
6.335 0.364493528786432
6.665 0.364493528786432
};
\addplot [color1]
table {%
6.335 1.30057398954806
6.665 1.30057398954806
};
\addplot [color1, mark=o, mark size=3, mark options={solid,fill opacity=0,draw=black}, only marks]
table {%
6.5 0.165733032412545
};
\addplot [color0]
table {%
7.75 0.993701833460124
7.75 0.835399200975706
};
\addplot [color0]
table {%
7.75 2.28866774524304
7.75 2.98567624008154
};
\addplot [color0]
table {%
7.585 0.835399200975706
7.915 0.835399200975706
};
\addplot [color0]
table {%
7.585 2.98567624008154
7.915 2.98567624008154
};
\addplot [color1]
table {%
8.75 0.610222206314606
8.75 0.496035849824804
};
\addplot [color1]
table {%
8.75 0.994613327744404
8.75 1.12365043827573
};
\addplot [color1]
table {%
8.585 0.496035849824804
8.915 0.496035849824804
};
\addplot [color1]
table {%
8.585 1.12365043827573
8.915 1.12365043827573
};
\path [draw=color0, fill=color2]
(axis cs:0.67,2.39142009365388)
--(axis cs:1.33,2.39142009365388)
--(axis cs:1.33,5.29867334053462)
--(axis cs:0.67,5.29867334053462)
--(axis cs:0.67,2.39142009365388)
--cycle;
\path [draw=color1, fill=color3]
(axis cs:1.67,2.36502848759696)
--(axis cs:2.33,2.36502848759696)
--(axis cs:2.33,4.27533354952236)
--(axis cs:1.67,4.27533354952236)
--(axis cs:1.67,2.36502848759696)
--cycle;
\path [draw=color0, fill=color2]
(axis cs:2.92,1.14323001754823)
--(axis cs:3.58,1.14323001754823)
--(axis cs:3.58,1.94265472891544)
--(axis cs:2.92,1.94265472891544)
--(axis cs:2.92,1.14323001754823)
--cycle;
\path [draw=color1, fill=color3]
(axis cs:3.92,1.02553324034011)
--(axis cs:4.58,1.02553324034011)
--(axis cs:4.58,1.201294802175)
--(axis cs:3.92,1.201294802175)
--(axis cs:3.92,1.02553324034011)
--cycle;
\path [draw=color0, fill=color2]
(axis cs:5.17,1.14711526075564)
--(axis cs:5.83,1.14711526075564)
--(axis cs:5.83,3.0778358039215)
--(axis cs:5.17,3.0778358039215)
--(axis cs:5.17,1.14711526075564)
--cycle;
\path [draw=color1, fill=color3]
(axis cs:6.17,0.681670231178204)
--(axis cs:6.83,0.681670231178204)
--(axis cs:6.83,0.977493787964857)
--(axis cs:6.17,0.977493787964857)
--(axis cs:6.17,0.681670231178204)
--cycle;
\path [draw=color0, fill=color2]
(axis cs:7.42,0.993701833460124)
--(axis cs:8.08,0.993701833460124)
--(axis cs:8.08,2.28866774524304)
--(axis cs:7.42,2.28866774524304)
--(axis cs:7.42,0.993701833460124)
--cycle;
\path [draw=color1, fill=color3]
(axis cs:8.42,0.610222206314606)
--(axis cs:9.08,0.610222206314606)
--(axis cs:9.08,0.994613327744404)
--(axis cs:8.42,0.994613327744404)
--(axis cs:8.42,0.610222206314606)
--cycle;
\addplot [color0]
table {%
0.67 3.9418953049528
1.33 3.9418953049528
};
\addplot [color1]
table {%
1.67 3.1092860383707
2.33 3.1092860383707
};
\addplot [color0]
table {%
2.92 1.75219237449184
3.58 1.75219237449184
};
\addplot [color1]
table {%
3.92 1.06467403362364
4.58 1.06467403362364
};
\addplot [color0]
table {%
5.17 1.99632049216987
5.83 1.99632049216987
};
\addplot [color1]
table {%
6.17 0.874837699418893
6.83 0.874837699418893
};
\addplot [color0]
table {%
7.42 1.33889812787125
8.08 1.33889812787125
};
\addplot [color1]
table {%
8.42 0.754098964215456
9.08 0.754098964215456
};
\end{axis}

\end{tikzpicture}
  \caption{Weight VJP performance ratios of \textcolor{maincolor}{TN versus PT} and
    \textcolor{secondcolor}{TN+opt versus PT} for different convolution types on
    GPU.}\label{fig:app:weight-vjp}
\end{figure}

\clearpage

\def\datapath{exp/exp03_performance_weight_vjp/results/weight_vjp/}
\def\datapathHyper{exp/exp01_conv2d_vs_einconv2d/results/layer_details/}
\begin{table}[p]
  \centering
  \caption{Weight VJP performance comparison on GPU.}\label{tab:app:weight-vjp-gpu}
  \vspace{-1ex}
  \begin{subtable}[t]{1.0\linewidth}
    \centering
    \caption{3c3d, CIFAR-10, input shape (128, 3, 32, 32)}\label{tab:app:weight-vjp-gpu-cifar10-3c3d}
    \vspace{-0.5ex}
    \performanceTableGPU{\datapath cifar10_3c3d_input_shape_3_32_32_cuda_0_batch_size_128/times.csv}
    \performanceTableGPUSimplified{\datapath cifar10_3c3d_input_shape_3_32_32_cuda_0_batch_size_128_simplify/times.csv}
    \convTypeTable{\datapathHyper cifar10_3c3d_in_3_32_32_out_10.csv}
  \end{subtable}
  \begin{subtable}[t]{1.0\linewidth}
    \centering
    \caption{F-MNIST 2c2d, input shape (128, 1, 28, 28)}\label{tab:app:weight-vjp-gpu-fmnist-2c2d}
    \vspace{-0.5ex}
    \performanceTableGPU{\datapath fmnist_2c2d_input_shape_1_28_28_cuda_0_batch_size_128/times.csv}
    \performanceTableGPUSimplified{\datapath fmnist_2c2d_input_shape_1_28_28_cuda_0_batch_size_128_simplify/times.csv}
    \convTypeTable{\datapathHyper fmnist_2c2d_in_1_28_28_out_10.csv}
  \end{subtable}
  \begin{subtable}[t]{1.0\linewidth}
    \centering
    \caption{CIFAR-100 All-CNN-C, input shape (128, 3, 32, 32)}\label{tab:app:weight-vjp-gpu-cifar100-allcnnc}
    \vspace{-0.5ex}
    \performanceTableGPU{\datapath cifar100_allcnnc_input_shape_3_32_32_cuda_0_batch_size_128/times.csv}
    \performanceTableGPUSimplified{\datapath cifar100_allcnnc_input_shape_3_32_32_cuda_0_batch_size_128_simplify/times.csv}
    \convTypeTable{\datapathHyper cifar100_allcnnc_in_3_32_32_out_100.csv}
  \end{subtable}
  \begin{subtable}[t]{1.0\linewidth}
    \centering
    \caption{Alexnet, input shape (32, 3, 256, 256)}\label{tab:app:weight-vjp-gpu-alexnet}
    \vspace{-0.5ex}
    \performanceTableGPU{\datapath alexnet_input_shape_3_256_256_cuda_0_batch_size_32/times.csv}
    \performanceTableGPUSimplified{\datapath alexnet_input_shape_3_256_256_cuda_0_batch_size_32_simplify/times.csv}
    \convTypeTable{\datapathHyper alexnet_in_3_256_256_out_1000.csv}
  \end{subtable}
  \begin{subtable}[t]{1.0\linewidth}
    \centering
    \caption{ResNet18, input shape (32, 3, 256, 256)}\label{tab:app:weight-vjp-gpu-resnet18}
    \vspace{-0.5ex}
    \performanceTableGPU{\datapath resnet18_input_shape_3_256_256_cuda_0_batch_size_32/times.csv}
    \performanceTableGPUSimplified{\datapath resnet18_input_shape_3_256_256_cuda_0_batch_size_32_simplify/times.csv}
    \convTypeTable{\datapathHyper resnet18_in_3_256_256_out_1000.csv}
  \end{subtable}
  \begin{subtable}[t]{1.0\linewidth}
    \centering
    \caption{ResNext101, input shape (32, 3, 256, 256)}\label{tab:app:weight-vjp-gpu-resnext101}\vspace{-0.5ex} \performanceTableGPU{\datapath resnext101_32x8d_input_shape_3_256_256_cuda_0_batch_size_32/times.csv}
    \performanceTableGPUSimplified{\datapath resnext101_32x8d_input_shape_3_256_256_cuda_0_batch_size_32_simplify/times.csv}
    \convTypeTable{\datapathHyper resnext101_32x8d_in_3_256_256_out_1000.csv}
  \end{subtable}
  \begin{subtable}[t]{1.0\linewidth}
    \centering
    \caption{ConvNeXt-base, input shape (32, 3, 256, 256)}\label{tab:app:weight-vjp-gpu-convnext-base} \vspace{-0.5ex}\performanceTableGPU{\datapath convnext_base_input_shape_3_256_256_cuda_0_batch_size_32/times.csv}
    \performanceTableGPUSimplified{\datapath convnext_base_input_shape_3_256_256_cuda_0_batch_size_32_simplify/times.csv}
    \convTypeTable{\datapathHyper convnext_base_in_3_256_256_out_1000.csv}
  \end{subtable}

\end{table}
\begin{table}[p]\ContinuedFloat
  \centering
  \vspace*{-10ex}
  \begin{subtable}[t]{1.0\linewidth}
    \centering
    \caption{InceptionV3, input shape (32, 3, 299, 299)}\label{tab:app:weight-vjp-gpu-inception-v3}
    \performanceTableGPU{\datapath inception_v3_input_shape_3_299_299_cuda_0_batch_size_32/times.csv}
    \performanceTableGPUSimplified{\datapath inception_v3_input_shape_3_299_299_cuda_0_batch_size_32_simplify/times.csv}
    \convTypeTable{\datapathHyper inception_v3_in_3_299_299_out_1000.csv}
  \end{subtable}
  \begin{subtable}[t]{1.0\linewidth}
    \centering
    \caption{MobileNetV2, input shape (32, 3, 256, 256)}\label{tab:app:weight-vjp-gpu-mobilenet-v2}
    \performanceTableGPU{\datapath mobilenet_v2_input_shape_3_256_256_cuda_0_batch_size_32/times.csv}
    \performanceTableGPUSimplified{\datapath mobilenet_v2_input_shape_3_256_256_cuda_0_batch_size_32_simplify/times.csv}
    \convTypeTable{\datapathHyper mobilenet_v2_in_3_256_256_out_1000.csv}
  \end{subtable}
\end{table}

\clearpage

\subsection{KFC Factor (KFAC-expand)}
We compare TN and TN+opt with a PyTorch implementation of the input-based KFC
factor based on \texttt{torch.nn.functional.unfold}. \Cref{fig:app:kfc-factor}
visualizes the performance ratios for different convolution categories.
\Cref{tab:app:kfc-factor-gpu} contains the detailed run times and performance
factors.

\begin{figure}[!h]
  \centering
  \pgfkeys{/pgfplots/zmystyle/.style={boxplotbasicstyle}}
\begin{tikzpicture}

\definecolor{color0}{rgb}{0.12156862745098,0.466666666666667,0.705882352941177}
\definecolor{color1}{rgb}{0.0901960784313725,0.745098039215686,0.811764705882353}
\definecolor{color2}{rgb}{0.580392156862745,0.403921568627451,0.741176470588235}
\definecolor{color3}{rgb}{0.890196078431372,0.466666666666667,0.76078431372549}

\begin{axis}[
axis line style={white!80!black},
tick pos=left,
xmin=0.5, xmax=9.25,
xtick style={color=gray},
xtick={1.5,3.75,6,8.25},
xticklabels={General,Dense mix,Dense,Down},
ylabel={TN versus PT (logarithmic)},
ymin=0.0480795705651229, ymax=10.6936328144865,
ymode=log,
zmystyle
]
\path [draw=black, ]
(axis cs:0.5,1)
--(axis cs:9.25,1);

\addplot [color0]
table {%
1 0.928731571806936
1 0.0634347307333202
};
\addplot [color0]
table {%
1 1.90692494649583
1 3.14867046315771
};
\addplot [color0]
table {%
0.835 0.0634347307333202
1.165 0.0634347307333202
};
\addplot [color0]
table {%
0.835 3.14867046315771
1.165 3.14867046315771
};
\addplot [color0, mark=o, mark size=3, mark options={solid,fill opacity=0,draw=black}, only marks]
table {%
1 4.02898611572527
1 4.40517843410241
1 8.3644225413074
1 3.68611989439665
1 6.34292107596406
1 4.16952301215333
};
\addplot [color1]
table {%
2 1.02978702842775
2 0.0614681134247496
};
\addplot [color1]
table {%
2 1.99227266655037
2 3.30005321467678
};
\addplot [color1]
table {%
1.835 0.0614681134247496
2.165 0.0614681134247496
};
\addplot [color1]
table {%
1.835 3.30005321467678
2.165 3.30005321467678
};
\addplot [color1, mark=o, mark size=3, mark options={solid,fill opacity=0,draw=black}, only marks]
table {%
2 4.54585546760952
2 8.32895358139788
2 6.13317231629844
};
\addplot [color0]
table {%
3.25 0.481001751597008
3.25 0.428947696001238
};
\addplot [color0]
table {%
3.25 0.558160201068993
3.25 0.562950905252086
};
\addplot [color0]
table {%
3.085 0.428947696001238
3.415 0.428947696001238
};
\addplot [color0]
table {%
3.085 0.562950905252086
3.415 0.562950905252086
};
\addplot [color0, mark=o, mark size=3, mark options={solid,fill opacity=0,draw=black}, only marks]
table {%
3.25 1.07269443005032
3.25 1.08734621108146
};
\addplot [color1]
table {%
4.25 0.395634576133383
4.25 0.381742432018256
};
\addplot [color1]
table {%
4.25 0.419588191320375
4.25 0.419782733825825
};
\addplot [color1]
table {%
4.085 0.381742432018256
4.415 0.381742432018256
};
\addplot [color1]
table {%
4.085 0.419782733825825
4.415 0.419782733825825
};
\addplot [color1, mark=o, mark size=3, mark options={solid,fill opacity=0,draw=black}, only marks]
table {%
4.25 1.30612826047223
4.25 1.38161374883581
};
\addplot [color0]
table {%
5.5 1.07229318439295
5.5 0.709932907038578
};
\addplot [color0]
table {%
5.5 1.80088130959883
5.5 2.25573036756196
};
\addplot [color0]
table {%
5.335 0.709932907038578
5.665 0.709932907038578
};
\addplot [color0]
table {%
5.335 2.25573036756196
5.665 2.25573036756196
};
\addplot [color0, mark=o, mark size=3, mark options={solid,fill opacity=0,draw=black}, only marks]
table {%
5.5 4.71179802358024
5.5 4.84636269104023
5.5 3.45449029271799
};
\addplot [color1]
table {%
6.5 0.625744324238036
6.5 0.25456683207541
};
\addplot [color1]
table {%
6.5 1.12657925302402
6.5 1.68495919058468
};
\addplot [color1]
table {%
6.335 0.25456683207541
6.665 0.25456683207541
};
\addplot [color1]
table {%
6.335 1.68495919058468
6.665 1.68495919058468
};
\addplot [color0]
table {%
7.75 2.25150354641202
7.75 1.79970587350613
};
\addplot [color0]
table {%
7.75 3.18452685016055
7.75 4.48927330767106
};
\addplot [color0]
table {%
7.585 1.79970587350613
7.915 1.79970587350613
};
\addplot [color0]
table {%
7.585 4.48927330767106
7.915 4.48927330767106
};
\addplot [color1]
table {%
8.75 0.701584827640437
8.75 0.688925344673295
};
\addplot [color1]
table {%
8.75 0.802040784076284
8.75 0.81291370392081
};
\addplot [color1]
table {%
8.585 0.688925344673295
8.915 0.688925344673295
};
\addplot [color1]
table {%
8.585 0.81291370392081
8.915 0.81291370392081
};
\addplot [color1, mark=o, mark size=3, mark options={solid,fill opacity=0,draw=black}, only marks]
table {%
8.75 0.528109093333629
};
\path [draw=color0, fill=color2]
(axis cs:0.67,0.928731571806936)
--(axis cs:1.33,0.928731571806936)
--(axis cs:1.33,1.90692494649583)
--(axis cs:0.67,1.90692494649583)
--(axis cs:0.67,0.928731571806936)
--cycle;
\path [draw=color1, fill=color3]
(axis cs:1.67,1.02978702842775)
--(axis cs:2.33,1.02978702842775)
--(axis cs:2.33,1.99227266655037)
--(axis cs:1.67,1.99227266655037)
--(axis cs:1.67,1.02978702842775)
--cycle;
\path [draw=color0, fill=color2]
(axis cs:2.92,0.481001751597008)
--(axis cs:3.58,0.481001751597008)
--(axis cs:3.58,0.558160201068993)
--(axis cs:2.92,0.558160201068993)
--(axis cs:2.92,0.481001751597008)
--cycle;
\path [draw=color1, fill=color3]
(axis cs:3.92,0.395634576133383)
--(axis cs:4.58,0.395634576133383)
--(axis cs:4.58,0.419588191320375)
--(axis cs:3.92,0.419588191320375)
--(axis cs:3.92,0.395634576133383)
--cycle;
\path [draw=color0, fill=color2]
(axis cs:5.17,1.07229318439295)
--(axis cs:5.83,1.07229318439295)
--(axis cs:5.83,1.80088130959883)
--(axis cs:5.17,1.80088130959883)
--(axis cs:5.17,1.07229318439295)
--cycle;
\path [draw=color1, fill=color3]
(axis cs:6.17,0.625744324238036)
--(axis cs:6.83,0.625744324238036)
--(axis cs:6.83,1.12657925302402)
--(axis cs:6.17,1.12657925302402)
--(axis cs:6.17,0.625744324238036)
--cycle;
\path [draw=color0, fill=color2]
(axis cs:7.42,2.25150354641202)
--(axis cs:8.08,2.25150354641202)
--(axis cs:8.08,3.18452685016055)
--(axis cs:7.42,3.18452685016055)
--(axis cs:7.42,2.25150354641202)
--cycle;
\path [draw=color1, fill=color3]
(axis cs:8.42,0.701584827640437)
--(axis cs:9.08,0.701584827640437)
--(axis cs:9.08,0.802040784076284)
--(axis cs:8.42,0.802040784076284)
--(axis cs:8.42,0.701584827640437)
--cycle;
\addplot [color0]
table {%
0.67 1.34248519975462
1.33 1.34248519975462
};
\addplot [color1]
table {%
1.67 1.42460631331979
2.33 1.42460631331979
};
\addplot [color0]
table {%
2.92 0.525203203060172
3.58 0.525203203060172
};
\addplot [color1]
table {%
3.92 0.408550537415579
4.58 0.408550537415579
};
\addplot [color0]
table {%
5.17 1.18634294428183
5.83 1.18634294428183
};
\addplot [color1]
table {%
6.17 0.730241212253938
6.83 0.730241212253938
};
\addplot [color0]
table {%
7.42 2.64838507657877
8.08 2.64838507657877
};
\addplot [color1]
table {%
8.42 0.770436654590951
9.08 0.770436654590951
};
\end{axis}

\end{tikzpicture}
  \caption{KFC/KFAC-expand factor performance ratios of \textcolor{maincolor}{TN
      versus PT} and \textcolor{secondcolor}{TN+opt versus PT} for different
    convolution types on GPU.}\label{fig:app:kfc-factor}
\end{figure}

\clearpage

\def\datapath{exp/exp05_performance_kfc_factor/results/kfc_factor/}
\def\datapathHyper{exp/exp01_conv2d_vs_einconv2d/results/layer_details/}
\begin{table}[p]
  \centering
  \caption{KFC (KFAC-expand) factor performance comparison on GPU.}\label{tab:app:kfc-factor-gpu}
  \vspace{-1ex}
  \begin{subtable}[t]{1.0\linewidth}
    \centering
    \caption{3c3d, CIFAR-10, input shape (128, 3, 32, 32)}\label{tab:app:kfc-factor-gpu-cifar10-3c3d}
    \vspace{-0.5ex}
    \performanceTableGPU{\datapath cifar10_3c3d_input_shape_3_32_32_cuda_0_batch_size_128/times.csv}
    \performanceTableGPUSimplified{\datapath cifar10_3c3d_input_shape_3_32_32_cuda_0_batch_size_128_simplify/times.csv}
    \convTypeTable{\datapathHyper cifar10_3c3d_in_3_32_32_out_10.csv}
  \end{subtable}
  \begin{subtable}[t]{1.0\linewidth}
    \centering
    \caption{F-MNIST 2c2d, input shape (128, 1, 28, 28)}\label{tab:app:kfc-factor-gpu-fmnist-2c2d}
    \vspace{-0.5ex}
    \performanceTableGPU{\datapath fmnist_2c2d_input_shape_1_28_28_cuda_0_batch_size_128/times.csv}
    \performanceTableGPUSimplified{\datapath fmnist_2c2d_input_shape_1_28_28_cuda_0_batch_size_128_simplify/times.csv}
    \convTypeTable{\datapathHyper fmnist_2c2d_in_1_28_28_out_10.csv}
  \end{subtable}
  \begin{subtable}[t]{1.0\linewidth}
    \centering
    \caption{CIFAR-100 All-CNN-C, input shape (128, 3, 32, 32)}\label{tab:app:kfc-factor-gpu-cifar100-allcnnc}
    \vspace{-0.5ex}
    \performanceTableGPU{\datapath cifar100_allcnnc_input_shape_3_32_32_cuda_0_batch_size_128/times.csv}
    \performanceTableGPUSimplified{\datapath cifar100_allcnnc_input_shape_3_32_32_cuda_0_batch_size_128_simplify/times.csv}
    \convTypeTable{\datapathHyper cifar100_allcnnc_in_3_32_32_out_100.csv}
  \end{subtable}
  \begin{subtable}[t]{1.0\linewidth}
    \centering
    \caption{Alexnet, input shape (32, 3, 256, 256)}\label{tab:app:kfc-factor-gpu-alexnet}
    \vspace{-0.5ex}
    \performanceTableGPU{\datapath alexnet_input_shape_3_256_256_cuda_0_batch_size_32/times.csv}
    \performanceTableGPUSimplified{\datapath alexnet_input_shape_3_256_256_cuda_0_batch_size_32_simplify/times.csv}
    \convTypeTable{\datapathHyper alexnet_in_3_256_256_out_1000.csv}
  \end{subtable}
  \begin{subtable}[t]{1.0\linewidth}
    \centering
    \caption{ResNet18, input shape (32, 3, 256, 256)}\label{tab:app:kfc-factor-gpu-resnet18}
    \vspace{-0.5ex}
    \performanceTableGPU{\datapath resnet18_input_shape_3_256_256_cuda_0_batch_size_32/times.csv}
    \performanceTableGPUSimplified{\datapath resnet18_input_shape_3_256_256_cuda_0_batch_size_32_simplify/times.csv}
    \convTypeTable{\datapathHyper resnet18_in_3_256_256_out_1000.csv}
  \end{subtable}
  \begin{subtable}[t]{1.0\linewidth}
    \centering
    \caption{ResNext101, input shape (32, 3, 256, 256)}\label{tab:app:kfc-factor-gpu-resnext101}\vspace{-0.5ex} \performanceTableGPU{\datapath resnext101_32x8d_input_shape_3_256_256_cuda_0_batch_size_32/times.csv}
    \performanceTableGPUSimplified{\datapath resnext101_32x8d_input_shape_3_256_256_cuda_0_batch_size_32_simplify/times.csv}
    \convTypeTable{\datapathHyper resnext101_32x8d_in_3_256_256_out_1000.csv}
  \end{subtable}
  \begin{subtable}[t]{1.0\linewidth}
    \centering
    \caption{ConvNeXt-base, input shape (32, 3, 256, 256)}\label{tab:app:kfc-factor-gpu-convnext-base} \vspace{-0.5ex}\performanceTableGPU{\datapath convnext_base_input_shape_3_256_256_cuda_0_batch_size_32/times.csv}
    \performanceTableGPUSimplified{\datapath convnext_base_input_shape_3_256_256_cuda_0_batch_size_32_simplify/times.csv}
    \convTypeTable{\datapathHyper convnext_base_in_3_256_256_out_1000.csv}
  \end{subtable}

\end{table}
\begin{table}[p]\ContinuedFloat
  \centering
  \vspace*{-10ex}
  \begin{subtable}[t]{1.0\linewidth}
    \centering
    \caption{InceptionV3, input shape (32, 3, 299, 299)}\label{tab:app:kfc-factor-gpu-inception-v3}
    \performanceTableGPU{\datapath inception_v3_input_shape_3_299_299_cuda_0_batch_size_32/times.csv}
    \performanceTableGPUSimplified{\datapath inception_v3_input_shape_3_299_299_cuda_0_batch_size_32_simplify/times.csv}
    \convTypeTable{\datapathHyper inception_v3_in_3_299_299_out_1000.csv}
  \end{subtable}
  \begin{subtable}[t]{1.0\linewidth}
    \centering
    \caption{MobileNetV2, input shape (32, 3, 256, 256)}\label{tab:app:kfc-factor-gpu-mobilenet-v2}
    \performanceTableGPU{\datapath mobilenet_v2_input_shape_3_256_256_cuda_0_batch_size_32/times.csv}
    \performanceTableGPUSimplified{\datapath mobilenet_v2_input_shape_3_256_256_cuda_0_batch_size_32_simplify/times.csv}
    \convTypeTable{\datapathHyper mobilenet_v2_in_3_256_256_out_1000.csv}
  \end{subtable}
\end{table}
\clearpage

\subsection{KFAC-reduce Factor}
We compare TN and TN+opt with a PyTorch implementation of the input-based
KFAC-reduce factor based on \texttt{torch.nn.functional.unfold}.
\Cref{fig:app:kfac-reduce} visualizes the performance ratios for
different convolution categories. \Cref{tab:app:kfac-reduce-factor-gpu} contains
the detailed run times and performance factors.

\begin{figure}[!h]
  \centering
  \pgfkeys{/pgfplots/zmystyle/.style={boxplotbasicstyle}}
\begin{tikzpicture}

\definecolor{color0}{rgb}{0.12156862745098,0.466666666666667,0.705882352941177}
\definecolor{color1}{rgb}{0.0901960784313725,0.745098039215686,0.811764705882353}
\definecolor{color2}{rgb}{0.580392156862745,0.403921568627451,0.741176470588235}
\definecolor{color3}{rgb}{0.890196078431372,0.466666666666667,0.76078431372549}

\begin{axis}[
axis line style={white!80!black},
tick pos=left,
xmin=0.5, xmax=9.25,
xtick style={color=gray},
xtick={1.5,3.75,6,8.25},
xticklabels={General,Dense mix,Dense,Down},
ylabel={TN versus PT (logarithmic)},
ymin=0.179571945772811, ymax=7.24171386001435,
ymode=log,
zmystyle
]
\path [draw=black, ]
(axis cs:0.5,1)
--(axis cs:9.25,1);

\addplot [color0]
table {%
1 0.41533571104182
1 0.216579164271876
};
\addplot [color0]
table {%
1 0.882388464186273
1 1.22722307407157
};
\addplot [color0]
table {%
0.835 0.216579164271876
1.165 0.216579164271876
};
\addplot [color0]
table {%
0.835 1.22722307407157
1.165 1.22722307407157
};
\addplot [color0, mark=o, mark size=3, mark options={solid,fill opacity=0,draw=black}, only marks]
table {%
1 1.67964558324244
1 2.29400845356964
};
\addplot [color1]
table {%
2 0.394608980513521
2 0.212432244251576
};
\addplot [color1]
table {%
2 0.868971930741341
2 1.22948240025032
};
\addplot [color1]
table {%
1.835 0.212432244251576
2.165 0.212432244251576
};
\addplot [color1]
table {%
1.835 1.22948240025032
2.165 1.22948240025032
};
\addplot [color1, mark=o, mark size=3, mark options={solid,fill opacity=0,draw=black}, only marks]
table {%
2 1.64322153441689
2 2.28716763848796
};
\addplot [color0]
table {%
3.25 0.672063608405375
3.25 0.514678490551516
};
\addplot [color0]
table {%
3.25 0.964153769253581
3.25 1.25178763944338
};
\addplot [color0]
table {%
3.085 0.514678490551516
3.415 0.514678490551516
};
\addplot [color0]
table {%
3.085 1.25178763944338
3.415 1.25178763944338
};
\addplot [color0, mark=o, mark size=3, mark options={solid,fill opacity=0,draw=black}, only marks]
table {%
3.25 1.42887647675525
};
\addplot [color1]
table {%
4.25 0.500394870626054
4.25 0.412697481777046
};
\addplot [color1]
table {%
4.25 0.689893985157816
4.25 0.89263511853432
};
\addplot [color1]
table {%
4.085 0.412697481777046
4.415 0.412697481777046
};
\addplot [color1]
table {%
4.085 0.89263511853432
4.415 0.89263511853432
};
\addplot [color0]
table {%
5.5 0.905576938644048
5.5 0.779949708090248
};
\addplot [color0]
table {%
5.5 1.5705266802466
5.5 1.92305124729776
};
\addplot [color0]
table {%
5.335 0.779949708090248
5.665 0.779949708090248
};
\addplot [color0]
table {%
5.335 1.92305124729776
5.665 1.92305124729776
};
\addplot [color0, mark=o, mark size=3, mark options={solid,fill opacity=0,draw=black}, only marks]
table {%
5.5 6.12152196176341
5.5 3.69550472271328
};
\addplot [color1]
table {%
6.5 0.26768204873977
6.5 0.217261415937963
};
\addplot [color1]
table {%
6.5 0.53208823292818
6.5 0.912139373543072
};
\addplot [color1]
table {%
6.335 0.217261415937963
6.665 0.217261415937963
};
\addplot [color1]
table {%
6.335 0.912139373543072
6.665 0.912139373543072
};
\addplot [color0]
table {%
7.75 1.95336734083687
7.75 1.81540267598385
};
\addplot [color0]
table {%
7.75 3.03443622840916
7.75 3.43523309658439
};
\addplot [color0]
table {%
7.585 1.81540267598385
7.915 1.81540267598385
};
\addplot [color0]
table {%
7.585 3.43523309658439
7.915 3.43523309658439
};
\addplot [color1]
table {%
8.75 0.423344739067368
8.75 0.402596989737241
};
\addplot [color1]
table {%
8.75 0.541241931307674
8.75 0.626449967671783
};
\addplot [color1]
table {%
8.585 0.402596989737241
8.915 0.402596989737241
};
\addplot [color1]
table {%
8.585 0.626449967671783
8.915 0.626449967671783
};
\path [draw=color0, fill=color2]
(axis cs:0.67,0.41533571104182)
--(axis cs:1.33,0.41533571104182)
--(axis cs:1.33,0.882388464186273)
--(axis cs:0.67,0.882388464186273)
--(axis cs:0.67,0.41533571104182)
--cycle;
\path [draw=color1, fill=color3]
(axis cs:1.67,0.394608980513521)
--(axis cs:2.33,0.394608980513521)
--(axis cs:2.33,0.868971930741341)
--(axis cs:1.67,0.868971930741341)
--(axis cs:1.67,0.394608980513521)
--cycle;
\path [draw=color0, fill=color2]
(axis cs:2.92,0.672063608405375)
--(axis cs:3.58,0.672063608405375)
--(axis cs:3.58,0.964153769253581)
--(axis cs:2.92,0.964153769253581)
--(axis cs:2.92,0.672063608405375)
--cycle;
\path [draw=color1, fill=color3]
(axis cs:3.92,0.500394870626054)
--(axis cs:4.58,0.500394870626054)
--(axis cs:4.58,0.689893985157816)
--(axis cs:3.92,0.689893985157816)
--(axis cs:3.92,0.500394870626054)
--cycle;
\path [draw=color0, fill=color2]
(axis cs:5.17,0.905576938644048)
--(axis cs:5.83,0.905576938644048)
--(axis cs:5.83,1.5705266802466)
--(axis cs:5.17,1.5705266802466)
--(axis cs:5.17,0.905576938644048)
--cycle;
\path [draw=color1, fill=color3]
(axis cs:6.17,0.26768204873977)
--(axis cs:6.83,0.26768204873977)
--(axis cs:6.83,0.53208823292818)
--(axis cs:6.17,0.53208823292818)
--(axis cs:6.17,0.26768204873977)
--cycle;
\path [draw=color0, fill=color2]
(axis cs:7.42,1.95336734083687)
--(axis cs:8.08,1.95336734083687)
--(axis cs:8.08,3.03443622840916)
--(axis cs:7.42,3.03443622840916)
--(axis cs:7.42,1.95336734083687)
--cycle;
\path [draw=color1, fill=color3]
(axis cs:8.42,0.423344739067368)
--(axis cs:9.08,0.423344739067368)
--(axis cs:9.08,0.541241931307674)
--(axis cs:8.42,0.541241931307674)
--(axis cs:8.42,0.423344739067368)
--cycle;
\addplot [color0]
table {%
0.67 0.602045987203911
1.33 0.602045987203911
};
\addplot [color1]
table {%
1.67 0.56462942392321
2.33 0.56462942392321
};
\addplot [color0]
table {%
2.92 0.76247669149464
3.58 0.76247669149464
};
\addplot [color1]
table {%
3.92 0.592760601855747
4.58 0.592760601855747
};
\addplot [color0]
table {%
5.17 1.09266446799474
5.83 1.09266446799474
};
\addplot [color1]
table {%
6.17 0.33345179101999
6.83 0.33345179101999
};
\addplot [color0]
table {%
7.42 2.35851665865151
8.08 2.35851665865151
};
\addplot [color1]
table {%
8.42 0.46295134875552
9.08 0.46295134875552
};
\end{axis}

\end{tikzpicture}
  \caption{KFAC-reduce factor performance ratios of \textcolor{maincolor}{TN versus
      PT} and \textcolor{secondcolor}{TN+opt versus PT} for different convolution
    types on GPU.}\label{fig:app:kfac-reduce}
\end{figure}

\clearpage

\def\datapath{exp/exp10_performance_kfac_reduce_factor/results/kfac_reduce_factor/}
\def\datapathHyper{exp/exp01_conv2d_vs_einconv2d/results/layer_details/}
\begin{table}[p]
  \centering
  \caption{KFAC-reduce factor performance comparison on GPU.}\label{tab:app:kfac-reduce-factor-gpu}
  \vspace{-1ex}
  \begin{subtable}[t]{1.0\linewidth}
    \centering
    \caption{3c3d, CIFAR-10, input shape (128, 3, 32, 32)}\label{tab:app:kfac-reduce-factor-gpu-cifar10-3c3d}
    \vspace{-0.5ex}
    \performanceTableGPU{\datapath cifar10_3c3d_input_shape_3_32_32_cuda_0_batch_size_128/times.csv}
    \performanceTableGPUSimplified{\datapath cifar10_3c3d_input_shape_3_32_32_cuda_0_batch_size_128_simplify/times.csv}
    \convTypeTable{\datapathHyper cifar10_3c3d_in_3_32_32_out_10.csv}
  \end{subtable}
  \begin{subtable}[t]{1.0\linewidth}
    \centering
    \caption{F-MNIST 2c2d, input shape (128, 1, 28, 28)}\label{tab:app:kfac-reduce-factor-gpu-fmnist-2c2d}
    \vspace{-0.5ex}
    \performanceTableGPU{\datapath fmnist_2c2d_input_shape_1_28_28_cuda_0_batch_size_128/times.csv}
    \performanceTableGPUSimplified{\datapath fmnist_2c2d_input_shape_1_28_28_cuda_0_batch_size_128_simplify/times.csv}
    \convTypeTable{\datapathHyper fmnist_2c2d_in_1_28_28_out_10.csv}
  \end{subtable}
  \begin{subtable}[t]{1.0\linewidth}
    \centering
    \caption{CIFAR-100 All-CNN-C, input shape (128, 3, 32, 32)}\label{tab:app:kfac-reduce-factor-gpu-cifar100-allcnnc}
    \vspace{-0.5ex}
    \performanceTableGPU{\datapath cifar100_allcnnc_input_shape_3_32_32_cuda_0_batch_size_128/times.csv}
    \performanceTableGPUSimplified{\datapath cifar100_allcnnc_input_shape_3_32_32_cuda_0_batch_size_128_simplify/times.csv}
    \convTypeTable{\datapathHyper cifar100_allcnnc_in_3_32_32_out_100.csv}
  \end{subtable}
  \begin{subtable}[t]{1.0\linewidth}
    \centering
    \caption{Alexnet, input shape (32, 3, 256, 256)}\label{tab:app:kfac-reduce-factor-gpu-alexnet}
    \vspace{-0.5ex}
    \performanceTableGPU{\datapath alexnet_input_shape_3_256_256_cuda_0_batch_size_32/times.csv}
    \performanceTableGPUSimplified{\datapath alexnet_input_shape_3_256_256_cuda_0_batch_size_32_simplify/times.csv}
    \convTypeTable{\datapathHyper alexnet_in_3_256_256_out_1000.csv}
  \end{subtable}
  \begin{subtable}[t]{1.0\linewidth}
    \centering
    \caption{ResNet18, input shape (32, 3, 256, 256)}\label{tab:app:kfac-reduce-factor-gpu-resnet18}
    \vspace{-0.5ex}
    \performanceTableGPU{\datapath resnet18_input_shape_3_256_256_cuda_0_batch_size_32/times.csv}
    \performanceTableGPUSimplified{\datapath resnet18_input_shape_3_256_256_cuda_0_batch_size_32_simplify/times.csv}
    \convTypeTable{\datapathHyper resnet18_in_3_256_256_out_1000.csv}
  \end{subtable}
  \begin{subtable}[t]{1.0\linewidth}
    \centering
    \caption{ResNext101, input shape (32, 3, 256, 256)}\label{tab:app:kfac-reduce-factor-gpu-resnext101} \vspace{-0.5ex}\performanceTableGPU{\datapath resnext101_32x8d_input_shape_3_256_256_cuda_0_batch_size_32/times.csv}
    \performanceTableGPUSimplified{\datapath resnext101_32x8d_input_shape_3_256_256_cuda_0_batch_size_32_simplify/times.csv}
    \convTypeTable{\datapathHyper resnext101_32x8d_in_3_256_256_out_1000.csv}
  \end{subtable}
  \begin{subtable}[t]{1.0\linewidth}
    \centering
    \caption{ConvNeXt-base, input shape (32, 3, 256, 256)}\label{tab:app:kfac-reduce-factor-gpu-convnext-base} \vspace{-0.5ex}\performanceTableGPU{\datapath convnext_base_input_shape_3_256_256_cuda_0_batch_size_32/times.csv}
    \performanceTableGPUSimplified{\datapath convnext_base_input_shape_3_256_256_cuda_0_batch_size_32_simplify/times.csv}
    \convTypeTable{\datapathHyper convnext_base_in_3_256_256_out_1000.csv}
  \end{subtable}

\end{table}
\begin{table}[p]\ContinuedFloat
  \centering
  \vspace*{-10ex}
  \begin{subtable}[t]{1.0\linewidth}
    \centering
    \caption{InceptionV3, input shape (32, 3, 299, 299)}\label{tab:app:kfac-reduce-factor-gpu-inception-v3}
    \performanceTableGPU{\datapath inception_v3_input_shape_3_299_299_cuda_0_batch_size_32/times.csv}
    \performanceTableGPUSimplified{\datapath inception_v3_input_shape_3_299_299_cuda_0_batch_size_32_simplify/times.csv}
    \convTypeTable{\datapathHyper inception_v3_in_3_299_299_out_1000.csv}
  \end{subtable}
  \begin{subtable}[t]{1.0\linewidth}
    \centering
    \caption{MobileNetV2, input shape (32, 3, 256, 256)}\label{tab:app:kfac-reduce-factor-gpu-mobilenet-v2}
    \performanceTableGPU{\datapath mobilenet_v2_input_shape_3_256_256_cuda_0_batch_size_32/times.csv}
    \performanceTableGPUSimplified{\datapath mobilenet_v2_input_shape_3_256_256_cuda_0_batch_size_32_simplify/times.csv}
    \convTypeTable{\datapathHyper mobilenet_v2_in_3_256_256_out_1000.csv}
  \end{subtable}
\end{table}
\clearpage

\section{Memory Evaluation Details (CPU)}\label{sec:app:benchmark-memory}

Here, we investigate the peak memory consumption of our proposed TN implementations.

\subsection{Theoretical \& Empirical Analysis for KFAC-reduce Factor}\label{subsec:memory-kfac-reduce}

We assume a two-dimensional convolution with input $\tX$ of shape $(C_{\text{in}}, I_1, I_2)$, output of shape $(C_{\text{out}}, O_1, O_2)$ and kernel of shape $(C_{\text{out}}, C_{\text{in}}, K_1, K_2)$.
The analysis with a batch dimension is analogous; hence we suppress it here to de-clutter the notation.

The main difference between the default and our proposed TN implementation of $\hat{\mOmega}$ from \Cref{subsec:kfac} lies in the computation of the averaged unfolded input $\llbracket \tX \rrbracket^{(\text{avg})} := \nicefrac{1}{(O_1O_2)} \vone_{O_1 O_2}^{\top} \llbracket \tX \rrbracket$ which consists of $C_{\text{in}} K_1 K_2$ numbers.
In the following, we will look at the extra memory on top of storing the input $\tX$, the averaged unfolded input $\llbracket \tX \rrbracket^{(\text{avg})}$, and the result $\hat{\mOmega}$.

\paragraph{Default implementation:} The standard implementation computes $\llbracket \tX \rrbracket^{(\text{avg})}$ via the unfolded input $\llbracket X \rrbracket$ and thus requires extra storage of $C_{\text{in}} K_1 K_2 O_1 O_2$ numbers.

\paragraph{TN implementation (general case):} The TN implementation requires storing the averaged index patterns $\tPi^{(i, \text{avg})} := \nicefrac{1}{O_i} \sum_{o=1}^{O_i} [\tPi^{(i)}]_{:,o,:}$ for $i=1,2$.
These are directly computed via a slight modification of \Cref{alg:index-pattern-tensor} and require storing $I_1 K_1 + I_2 K_2$ numbers.
In contrast to the default implementation, spatial dimensions are de-coupled and there is no dependency on $C_{\text{in}}$.

\paragraph{TN implementation (structured case):} For structured convolutions (\Cref{fig:tn-simplifications}) we can describe the action of the index pattern tensor through reshape and narrowing operations.
ML libraries usually perform these without allocating additional memory.
Hence, our symbolic simplifications completely eliminate the allocation of
temporary intermediates to compute $\llbracket \tX \rrbracket^{(\text{avg})}$.

\paragraph{Empirical results:}
To demonstrate the memory reduction inside the computation of $\hat{\mOmega}$ we measure its peak memory with the \href{https://pypi.org/project/memory-profiler/}{\texttt{memory-profiler}} library and subtract the memory required to store $\tX$ and $\hat{\mOmega}$.
This approximates the extra internal memory requirement of an implementation.
With the setup of \Cref{sec:app:benchmark} we report the minimum additional memory over 50 independent runs in \Cref{tab:app:peakmem-kfac-reduce-cpu}.
We consistently observe that the TN implementation has lower peak memory, which is further reduced by our symbolic simplifications (see for example the effect on ResNext101's dense and down-sampling convolutions in \Cref{tab:app:peakmem-kfac-reduce-cpu-resnext101}).

Our theoretical analysis from above suggests that the peak memory difference becomes most visible for many channels with large kernel and output sizes.
One example are ConxNeXt-base's \texttt{features.1.0.block.0} convolutions with $K_1 = K_2 = 7$, $O_1 = O_2 = 64$, and $C_{\text{in}} = 128$ (\Cref{tab:app:hyperparameters-convnext-base}).
For those convolutions, we observe that the default implementation requires an additional 3,140\,MiB ($\approx 3\,\text{GiB}$!)
of memory, whereas the TN implementation has zero extra memory demand (\Cref{tab:app:peakmem-kfac-reduce-cpu-convnext-base}).
This is consistent with our theoretical analysis in that the overhead is storing the unfolded input, which has $(N=32) \cdot (C_{\text{in}} = 128) \cdot (O_1 = 64) \cdot (O_2 = 64) \cdot (K_1 = 7) \cdot (K_2=7) = 822,083,584$ \texttt{float32} entries, corresponding to 3,136\,MiB.

\clearpage

\def\datapath{exp/exp10_performance_kfac_reduce_factor/results/kfac_reduce_factor/}
\def\datapathHyper{exp/exp01_conv2d_vs_einconv2d/results/layer_details/}
\begin{table}[p]
  \vspace*{-10ex}
  \centering
  \caption{Additional internally required memory to compute the KFAC-reduce factor (measured on CPU).
    The value 0 indicates that an implementation's peak memory matches the memory consumption of its input $\tX$ and result $\smash{\hat{\mOmega}}$.
  }\label{tab:app:peakmem-kfac-reduce-cpu}
  \begin{subtable}[t]{1.0\linewidth}
    \centering
    \caption{3c3d, CIFAR-10, input shape (128, 3, 32, 32)}\label{tab:app:peakmem-kfac-reduce-cpu-cifar10-3c3d}
    \peakmemTable{\datapath cifar10_3c3d_input_shape_3_32_32_cpu_batch_size_128/peakmems.csv}
    \peakmemTableSimplified{\datapath cifar10_3c3d_input_shape_3_32_32_cpu_batch_size_128_simplify/peakmems.csv}
    \convTypeTableRebuttal{\datapathHyper cifar10_3c3d_in_3_32_32_out_10.csv}
  \end{subtable}
  \begin{subtable}[t]{1.0\linewidth}
    \centering
    \caption{F-MNIST 2c2d, input shape (128, 1, 28, 28)}\label{tab:app:peakmem-kfac-reduce-cpu-fmnist-2c2d}
    \peakmemTable{\datapath fmnist_2c2d_input_shape_1_28_28_cpu_batch_size_128/peakmems.csv}
    \peakmemTableSimplified{\datapath fmnist_2c2d_input_shape_1_28_28_cpu_batch_size_128_simplify/peakmems.csv}
    \convTypeTableRebuttal{\datapathHyper fmnist_2c2d_in_1_28_28_out_10.csv}
  \end{subtable}
  \begin{subtable}[t]{1.0\linewidth}
    \centering
    \caption{CIFAR-100 All-CNN-C, input shape (128, 3, 32, 32)}\label{tab:app:peakmem-kfac-reduce-cpu-cifar100-allcnnc}
    \peakmemTable{\datapath cifar100_allcnnc_input_shape_3_32_32_cpu_batch_size_128/peakmems.csv}
    \peakmemTableSimplified{\datapath cifar100_allcnnc_input_shape_3_32_32_cpu_batch_size_128_simplify/peakmems.csv}
    \convTypeTableRebuttal{\datapathHyper cifar100_allcnnc_in_3_32_32_out_100.csv}
  \end{subtable}
  \begin{subtable}[t]{1.0\linewidth}
    \centering
    \caption{Alexnet, input shape (32, 3, 256, 256)}\label{tab:app:peakmem-kfac-reduce-cpu-alexnet}
    \peakmemTable{\datapath alexnet_input_shape_3_256_256_cpu_batch_size_32/peakmems.csv}
    \peakmemTableSimplified{\datapath alexnet_input_shape_3_256_256_cpu_batch_size_32_simplify/peakmems.csv}
    \convTypeTableRebuttal{\datapathHyper alexnet_in_3_256_256_out_1000.csv}
  \end{subtable}
  \begin{subtable}[t]{1.0\linewidth}
    \centering
    \caption{ResNet18, input shape (32, 3, 256, 256)}\label{tab:app:peakmem-kfac-reduce-cpu-resnet18}
    \peakmemTable{\datapath resnet18_input_shape_3_256_256_cpu_batch_size_32/peakmems.csv}
    \peakmemTableSimplified{\datapath resnet18_input_shape_3_256_256_cpu_batch_size_32_simplify/peakmems.csv}
    \convTypeTableRebuttal{\datapathHyper resnet18_in_3_256_256_out_1000.csv}
  \end{subtable}
  \begin{subtable}[t]{1.0\linewidth}
    \centering
    \caption{ResNext101, input shape (32, 3, 256, 256)}\label{tab:app:peakmem-kfac-reduce-cpu-resnext101} \peakmemTable{\datapath resnext101_32x8d_input_shape_3_256_256_cpu_batch_size_32/peakmems.csv}
    \peakmemTableSimplified{\datapath resnext101_32x8d_input_shape_3_256_256_cpu_batch_size_32_simplify/peakmems.csv}
    \convTypeTableRebuttal{\datapathHyper resnext101_32x8d_in_3_256_256_out_1000.csv}
  \end{subtable}
  \begin{subtable}[t]{1.0\linewidth}
    \centering
    \caption{ConvNeXt-base, input shape (32, 3, 256, 256)}\label{tab:app:peakmem-kfac-reduce-cpu-convnext-base} \peakmemTable{\datapath convnext_base_input_shape_3_256_256_cpu_batch_size_32/peakmems.csv}
    \peakmemTableSimplified{\datapath convnext_base_input_shape_3_256_256_cpu_batch_size_32_simplify/peakmems.csv}
    \convTypeTableRebuttal{\datapathHyper convnext_base_in_3_256_256_out_1000.csv}
  \end{subtable}
  \vspace{-4ex}
\end{table}
\begin{table}[p]\ContinuedFloat
  \centering
  \vspace*{-10ex}
  \begin{subtable}[t]{1.0\linewidth}
    \centering
    \caption{InceptionV3, input shape (32, 3, 299, 299)}\label{tab:app:peakmem-kfac-reduce-cpu-inception-v3}
    \peakmemTable{\datapath inception_v3_input_shape_3_299_299_cpu_batch_size_32/peakmems.csv}
    \peakmemTableSimplified{\datapath inception_v3_input_shape_3_299_299_cpu_batch_size_32_simplify/peakmems.csv}
    \convTypeTableRebuttal{\datapathHyper inception_v3_in_3_299_299_out_1000.csv}
  \end{subtable}
  \begin{subtable}[t]{1.0\linewidth}
    \centering
    \caption{MobileNetV2, input shape (32, 3, 256, 256)}\label{tab:app:peakmem-kfac-reduce-cpu-mobilenet-v2}
    \peakmemTable{\datapath mobilenet_v2_input_shape_3_256_256_cpu_batch_size_32/peakmems.csv}
    \peakmemTableSimplified{\datapath mobilenet_v2_input_shape_3_256_256_cpu_batch_size_32_simplify/peakmems.csv}
    \convTypeTableRebuttal{\datapathHyper mobilenet_v2_in_3_256_256_out_1000.csv}
  \end{subtable}
\end{table}
\clearpage











\section{Miscellaneous}
\subsection{Example: Associativity of Tensor
  Multiplication}\label{subsec:associativity-example}
Here, we demonstrate associativity of tensor multiplication through an example.
The technical challenge is that an index can only be summed once there are no
remaining tensors sharing it. Therefore, we must carry indices that are summed
in later multiplications in the intermediate results, which requires some set
arithmetic on the index sets.

Let $S_1, S_2, S_3$ be index tuples of the input tensors $\tA, \tB, \tC$, and
$S_4 \subseteq (S_1 \cup S_2 \cup S_3)$ a valid output index tuple of their
tensor multiplication $\tD = *_{(S_1, S_2, S_3, S_4)}(\tA, \tB, \tC)$. We can
either first multiply $\tA$ with $\tB$ to obtain an intermediate tensor of index
structure $S_{1,2}$, or $\tB$ with $\tC$ to obtain an intermediate tensor of
index structure $S_{2,3}$, before carrying out the remaining multiplications. To
construct the intermediate index structures, we divide the indices $\tilde{S} =
(S_1 \cup S_2 \cup S_3) \setminus S_4$ that are summed over into those only
shared between $\tA, \tB$ given by $\tilde{S}_{1,2} = (S_1 \cup S_2) \setminus
(S_4 \cup S_3)$, and those only shared among $\tB, \tC$ given by
$\tilde{S}_{2,3} = (S_2 \cup S_3) \setminus (S_4 \cup S_1)$. This yields the
intermediate indices $S_{1,2} = (S_1 \cup S_2) \setminus \tilde{S}_{1,2}$ and
$S_{2,3} = (S_2 \cup S_3) \setminus \tilde{S}_{2,3}$, and the parenthesizations
\begin{align}\label{eq:associativity}
  \begin{split}
    [\tD]_{S_4}
    &=
      \left(
      \textstyle
      \sum_{\tilde{S} \setminus \tilde{S}_{1,2}}
      \left(
      \sum_{\tilde{S}_{1,2}}
      [\tA]_{S_1}
      [\tB]_{S_2}
      \right)
      [\tC]_{S_3}
      \right)
      =
      \left(
      \textstyle
      \sum_{\tilde{S} \setminus \tilde{S}_{2,3}}
      [\tA]_{S_1}
      \left(
      \sum_{\tilde{S}_{2,3}}
      [\tB]_{S_2}
      [\tC]_{S_3}
      \right)
      \right)
    \\
    \Leftrightarrow \
    \tD & = *_{(S_{1,2}, S_3, S_4)}
      \left(
      *_{(S_2, S_3, S_{2,3})}
      (\tA, \tB),
      \tC
      \right)
      =
      *_{(S_1, S_{2,3}, S_4)}
      \left(
      \tA,
      *_{(S_1, S_2, S_{1,2})}
      (\tB, \tC)
      \right)\,.
  \end{split}
\end{align}
This generalizes to $n$-ary multiplication, allowing to break it down into
smaller multiplications. However, the index notation and set arithmetic from
\Cref{eq:associativity} quickly becomes impractical.


\subsection{Example: Matrix-matrix Multiplication as Tensor Multiplication}\label{subsec:app:example:matrix-multiplication}
Here we provide a small self-contained example that demonstrates \Cref{eq:tensor-multiplication} for matrix-matrix multiplication.

Consider two matrices $\mA, \mB$ which are compatible for multiplication and let $\mC = \mA \mB$.
In index notation, we have
\begin{equation*}
  [\mC]_{i,k} = \sum_j [\mA]_{i,j} [\mB]_{j,k}\,.
\end{equation*}
The index tuples are $S_{\mA} = (i, j)$, $S_{\mB} = (j, k)$, and $S_{\mC} = (i,k)$.
Next, we evaluate which indices are summed over.
Since the order of those indices does not matter, we can interpret the tuples as sets and use set arithmetic:
\begin{equation*}
  \left( S_{\mA} \cup S_{\mB} \right) \setminus S_{\mC}
  =
  \left((i, j) \cup (j, k) \right) \setminus (i, k)
  =
  (j) \setminus (i, k)
  =
  (j)\,.
\end{equation*}
Now we see that matrix-matrix multiplication is a case of tensor multiplication (\Cref{eq:tensor-multiplication}),
\begin{equation*}
  [\mC]_{S_{\mC}} = \sum_{(S_{\mA} \cup S_{\mB}) \setminus S_{\mC}} [\mA]_{S_{\mA}} [\mB]_{S_{\mB}}
  = *_{(S_{\mA}, S_{\mB}, S_{\mC})}(\mA, \mB)\,.
\end{equation*}



\clearpage
\section*{NeurIPS Paper Checklist}


\begin{enumerate}

\item {\bf Claims}
\item[] Question: Do the main claims made in the abstract and introduction accurately reflect the paper's contributions and scope?
\item[] Answer: \answerYes{}
\item[] Justification: We provide a bullet point list of our contributions in \Cref{sec:introduction}, each of which references the part of the paper that outlines the contribution.

\item {\bf Limitations}
\item[] Question: Does the paper discuss the limitations of the work performed by the authors?
\item[] Answer: \answerYes{}
\item[] Justification: See \Cref{subsec:implementation-aspects,sec:app:limitations}.

\item {\bf Theory Assumptions and Proofs}
\item[] Question: For each theoretical result, does the paper provide the full set of assumptions and a complete (and correct) proof?
\item[] Answer: \answerYes{}
\item[] Justification: The TN simplifications we provide in \Cref{subsec:pattern-structure} follow straightforward from the index pattern's structure \Cref{equ:index-pattern-kronecker} and are stated rigorously in \Cref{subsec:app-additional-properties}, including their assumptions.

\item {\bf Experimental Result Reproducibility}
\item[] Question: Does the paper fully disclose all the information needed to reproduce the main experimental results of the paper to the extent that it affects the main claims and/or conclusions of the paper (regardless of whether the code and data are provided or not)?
\item[] Answer: \answerYes{}
\item[] Justification: We provide implementation details in \Cref{sec:app:implementation-details}, experimental and hardware details in \Cref{sec:app:benchmark,sec:app:benchmark-memory}.

\item {\bf Open access to data and code}
\item[] Question: Does the paper provide open access to the data and code, with sufficient instructions to faithfully reproduce the main experimental results, as described in supplemental material?
\item[] Answer: \answerYes{}
\item[] Justification: We will open-source the code to reproduce all our experiments, as well as the raw data containing the results shown in the manuscript.

\item {\bf Experimental Setting/Details}
\item[] Question: Does the paper specify all the training and test details (e.g., data splits, hyperparameters, how they were chosen, type of optimizer, etc.) necessary to understand the results?
\item[] Answer: \answerYes{}
\item[] Justification: See \Cref{sec:app:benchmark,sec:app:benchmark-memory} for details on the experimental setting.

\item {\bf Experiment Statistical Significance}
\item[] Question: Does the paper report error bars suitably and correctly defined or other appropriate information about the statistical significance of the experiments?
\item[] Answer: \answerYes{}
\item[] Justification: Our run time plots contain box plots with medians and quartiles reported over different convolutions, and the randomized backpropagation results show mean and standard deviations for 10 different model and batch initializations.

\item {\bf Experiments Compute Resources}
\item[] Question: For each experiment, does the paper provide sufficient information on the computer resources (type of compute workers, memory, time of execution) needed to reproduce the experiments?
\item[] Answer: \answerYes{}
\item[] Justification: All results were obtained on a single GPU to be comparable in terms of run time. See \Cref{sec:app:benchmark} for the details.

\item {\bf Code Of Ethics}
\item[] Question: Does the research conducted in the paper conform, in every respect, with the NeurIPS Code of Ethics \url{https://neurips.cc/public/EthicsGuidelines}?
\item[] Answer: \answerYes{}
\item[] Justification:  We have read the Code of Ethics and believe that our work conforms to it.

\item {\bf Broader Impacts}
\item[] Question: Does the paper discuss both potential positive societal impacts and negative societal impacts of the work performed?
\item[] Answer: \answerYes{}
\item[] Justification: This work aims to provide a simplifying perspective and implementations of otherwise hard-to-access operations for convolutions to facilitate the exploration of algorithmic ideas and advance existing second-order methods.
  We don't see any direct negative societal impacts.

\item {\bf Safeguards}
\item[] Question: Does the paper describe safeguards that have been put in place for responsible release of data or models that have a high risk for misuse (e.g., pretrained language models, image generators, or scraped datasets)?
\item[] Answer: \answerNA{}
\item[] Justification: The paper does not release any data or models that have a high risk for misuse.

\item {\bf Licenses for existing assets}
\item[] Question: Are the creators or original owners of assets (e.g., code, data, models), used in the paper, properly credited and are the license and terms of use explicitly mentioned and properly respected?
\item[] Answer: \answerYes{}
\item[] Justification: We cite the papers introducing the neural network architectures and data sets used in our experiments.

\item {\bf New Assets}
\item[] Question: Are new assets introduced in the paper well documented and is the documentation provided alongside the assets?
\item[] Answer: \answerNA{}
\item[] Justification: The paper does not release new assets.

\item {\bf Crowdsourcing and Research with Human Subjects}
\item[] Question: For crowdsourcing experiments and research with human subjects, does the paper include the full text of instructions given to participants and screenshots, if applicable, as well as details about compensation (if any)?
\item[] Answer: \answerNA{}
\item[] Justification: The paper does not involve crowdsourcing nor research with human subjects.

\item {\bf Institutional Review Board (IRB) Approvals or Equivalent for Research with Human Subjects}
\item[] Question: Does the paper describe potential risks incurred by study participants, whether such risks were disclosed to the subjects, and whether Institutional Review Board (IRB) approvals (or an equivalent approval/review based on the requirements of your country or institution) were obtained?
\item[] Answer: \answerNA{}
\item[] Justification: The paper does not involve crowdsourcing nor research with human subjects.
\end{enumerate}

\end{document}